# Machine Learning-Assisted Sustainable Remanufacturing, Reusing and Recycling for Lithium-ion Batteries

Dissertation submitted to

**Tsinghua University**

in partial fulfillment of the requirement

for the degree of

**Doctor of Philosophy**

in

**Electrical Engineering**

by

**Shengyu Tao**

Dissertation Supervisor : Associate Professor Xuan Zhang

**September, 2025**

# ABSTRACT

Sustainable utilization of lithium-ion batteries (LIBs) has become key technologies in addressing energy crisis. However, challenges of data scarcity and data heterogeneity in battery management remain significant, especially after the battery retirement. Based on machine learning, this dissertation focuses on solving the common scientific problems of data scarcity and heterogeneity in the quality verification of new prototype batteries, residual value estimation of retired batteries, material sorting of retired batteries, and adaptive management of the entire life cycle. The core work of this paper includes:

A battery prototype quality control method based on fusion of physical information and machine learning is proposed. This method utilizes early-life long-cycle data to predict long-term degradation trajectory, enabling early quality control in manufacturing process. By employing the Arrhenius equation for multi-temperature domain adaptation, the method addresses the data scarcity issue in the quality control phase. It accurately predicts the degradation trajectory by combining ample source domain data with limited target domain data, thereby improving the efficiency of quality feedback during battery development and facilitating the identification of defective batteries for sustainable waste recycling.

A rapid residual value assessment method based on fast pulse test method and data generation is introduced. This method expands the dataset through a generative machine learning model under random retirement conditions, thus providing the necessary sample size for fast assessment of the health state and residual value of retired batteries. Using the expanded dataset, the method, combined with a random forest regression model, can quickly and accurately assess the battery health state and residual value under random retirement conditions. It significantly reduces the time, testing costs, energy consumption, and carbon emissions compared to traditional data collection methods.

A privacy-preserving collaborative material sorting method is proposed. Based on a federated learning framework, this method enables high-precision and stable material sorting using only long-cycle data from the battery end-of-life stage, without exposing sensitive information. The effectiveness of this method is validated under homogeneous and heterogeneous data distributions. This approach allows battery data to be shared in a "usable while invisible" manner between different stakeholders, optimizing material recycling strategies, improving recycling efficiency, and reducing carbon emissions.

A unified diagnostics and prognostics method is introduced. This method, based on correlation alignment (CORAL)-aided feature engineering and neural network loss functions, achieves high accuracy and stability in various state estimation tasks, such as state-of-health estimation, state-of-charge estimation, and remaining useful life prediction. This method also shows high compatibility with different battery testing methods, such as long cycling test and fast pulse test, and achieves adaptability and consistency in the general process from the battery data curation, data processing, model construction to diagnostics and prognostics.

In summary, this dissertation proposes a comprehensive machine learning-assisted battery management framework, covering early quality control, residual value assessment of retired batteries, material sorting, adaptive diagnostics and prognostics. This research advances technologies for sustainable battery utilization and provides new methodologies for the sustainable development of the battery industry. It is expected to play a significant role in achieving the global carbon neutrality goals.

# LIST OF FIGURES AND TABLES

# LIST OF SYMBOLS AND ACRONYMS

| | |
|---|---|
| LIB | Lithium-ion Battery |
| LFP | Lithium Iron Phosphate |
| LMO | Lithium Manganese Oxide |
| NCM | Nickel Manganese Cobalt Oxide |
| NCA | Nickel Cobalt Aluminum Oxide |
| EV | Electric Vehicle |
| R&D | Research and Development |
| IMV | Initial Manufacturing Variability |
| CCCV | Constant Current Constant Voltage |
| SOC | State of Charge |
| SOH | State of Health |
| RUL | Remaining Useful Life |
| RPT | Reference Performance Test |
| LAM | Loss of Active Material |
| LLI | Loss of Lithium Inventory |
| OOD | Out of Distribution |
| CORAL | Correlation Alignment |
| MV | Majority Voting |
| WDV | Wasserstein Distance Voting |
| FL | Federated Learning |
| IL | Independent Learning |
| PB | Privacy Budget |
| ReLU | Rectified Linear Unit |
| MLP | Multi-Layer Perceptron |
| LSTM | Long Short-Term Memory |
| SAGE | Shapley Additive Global Importance |
| MAPE | Mean Absolute Percentage Error |
| STD | Standard Deviation |
| FEA | Finite Element Analysis |

# CHAPTER 1 INTRODUCTION

## 1.1 Background and Motivation

The Nobel Prize in Chemistry in 2019 was awarded for the revolutionary invention of lithium-ion batteries (LIBs), which have provided enormous value across consumer electronics, electric vehicles (EVs), and energy infrastructures over the last 30 years [1]. LIBs are already making substantial contributions to decarbonization by enabling EVs, and they will be an important part of the solution allowing energy storage for intermittent sources such as wind and solar in smart grid [2-4]. However, they are typically retired from EVs while still retaining up to 80% of their initial capacity, due to concerns about safety, range, and cold starts [5]. This massive underutilization poses considerable economic and environmental difficulties to materials suppliers, battery manufacturers, and end users [6-10]. Therefore, addressing the sustainable use of retired batteries is crucial for securing the intended clean technology ambitions.

Despite great promise of remanufacturing, reusing, and recycling of retired batteries, these reutilization methods are fraught with significant confusion and inconsistencies [11-16]. Key questions such as whether, when, and how these processes should be carried out to ensure sustainable reutilization have not yet received definitive answers [17]. This uncertainty largely stems from lack of reliable field data conditions on battery status, crucial for their informed decision-making. Important information like state of charge (SOC), remaining capacity, remaining energy, material types, and degradation mechanisms is often not readily accessible at the test site [3, 18-28]. This noticeable information gap makes it challenging for decision-makers to determine the optimal strategy for handling retired batteries, leading to inefficiencies and underutilization.

Tremendous efforts have been made by both industry and academia to address these information challenges [29-32]. The European Union Battery Directive, effective from July 2023, intends to promote sustainability through battery lifecycle management, and the Battery Passport is envisioned as a digital record of battery attributes to secure traceability and integrity [33-35]. However, the data recorded in the Battery Passport is static and does not accommodate the frequent changes in battery ownership and usage post-retirement. This limitation underscores the need for dynamic methods to acquire and make available this battery information [36]. Machine learning methods have demonstrated significant potential for assessing the field state of batteries at the point of retirement, leveraging available data. Compared to their first life, retired batteries exhibit unique challenges in data scarcity and heterogeneity. Data scarcity often arises due to the restrictions on use of the battery management system or to privacy concerns related to historical usage [18, 37-39]. Data heterogeneity is another issue, as retired batteries are usually collected in batches, with each batch containing batteries of different material types, physical formats, capacity designs, and degradation mechanisms shaped by varying usage histories [17, 18, 40]. Indeed, machine learning methods are inherently data-hungry, and the issues of data scarcity and heterogeneity severely prohibit their potential application in decision-making for reuse, recycling, and remanufacturing retired batteries. Finally, promoting retired batteries into second life use will require warranties, whose cost will be determined by the expected failure distribution [41, 42]. Therefore, machine learning methodologies tailored for the data that is available from retired batteries, are indispensable for devising economically and environmentally viable reutilization decision-making.

## 1.2 Literature Review

In this section, fundamentals of LIBs, concept of remanufacturing, reusing, recycling,

adaptive lifetime management, diagnostics and prognostics tasks are systematically reviewed.

### 1.2.1 Lithium-ion Batteries

Figure 1. 1 illustrates the schematics of LIB operation, showing that the general structure of a LIB cell, comprising an anode, a cathode, an electrolyte, a separator and a current collector. An ion-conducting electrolyte (containing a dissociated lithium conducting salt) is situated between two electrodes. The separator, a porous membrane to electrically isolate the two electrodes from each other, is also in that position. Single lithium ions migrate back and forth between the electrodes of LIBs during charging and discharging and are intercalated into the active materials. During discharging, when lithium is deintercalated from the negative electrode (copper functions as current collector), electrons are released. The active materials of the positive electrode are mixed oxides. Those of the negative electrode mainly are graphite and amorphous carbon compounds.

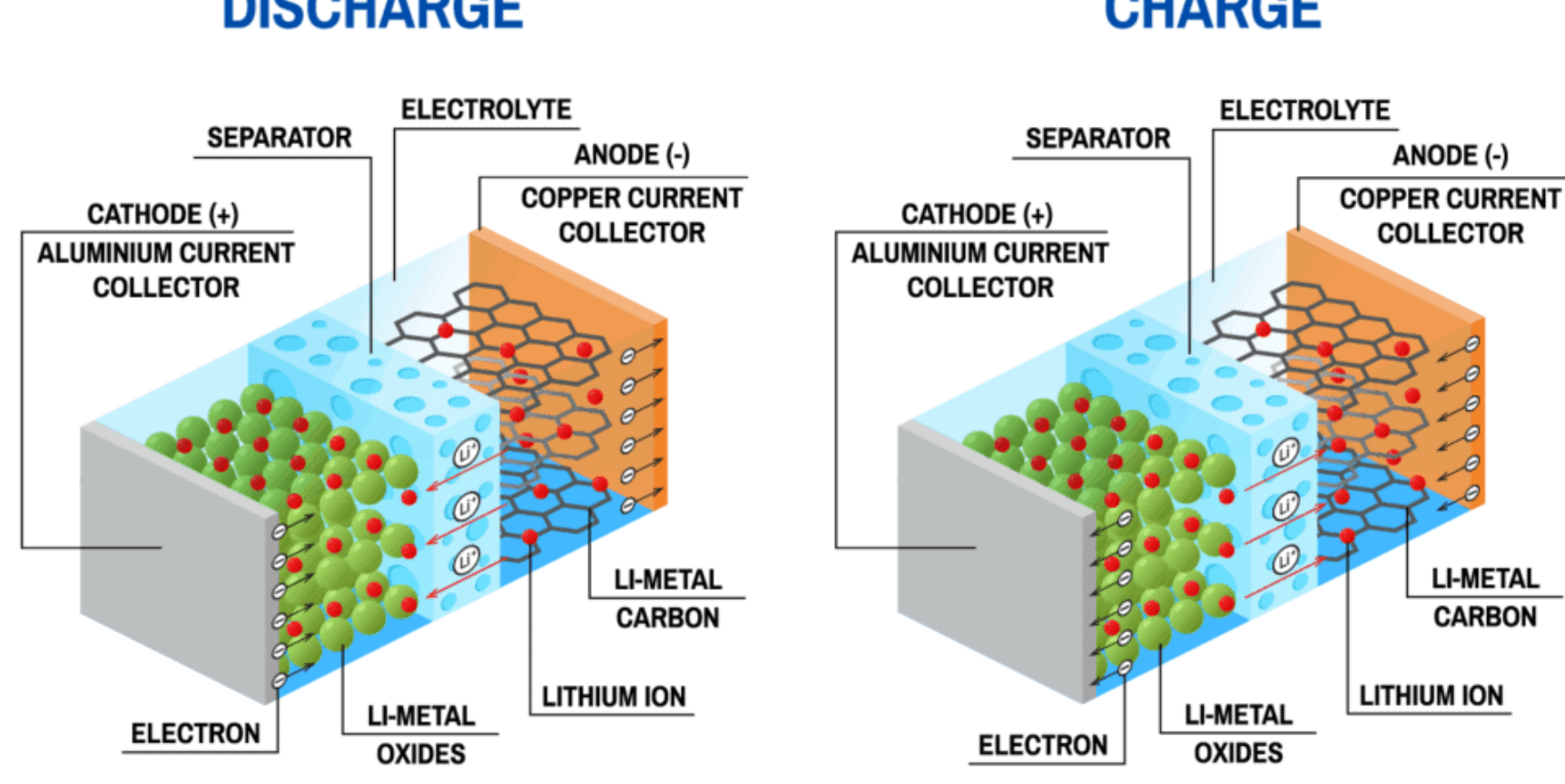

Figure 1. 1 The schematics of lithium-ion battery operation.

During the charging process, lithium atoms in cathode are ionized to Li+ ions and migrate towards the anode through the electrolyte. These Li+ ions combine with electrons to revert to lithium atoms, which are then intercalated between carbon layers in the anode.

This process is reversed during discharging. The complete chemical reaction in charging direction taking place in the anode and cathode can be given by：

$$\mathrm{C} + n\mathrm{Li}^+ + n\mathrm{e}^- \rightleftharpoons \mathrm{Li}_n\mathrm{C} \tag{1.1}$$

and

$$\mathrm{LiXXO_2} \rightleftharpoons \mathrm{Li}_{1-n}\mathrm{XXO_2} + n\mathrm{Li}^+ + n\mathrm{e}^- \tag{1.2}$$

where Li+ denotes lithium ion and $\mathrm{e}^-$ denotes electron. The symbol $\mathrm{C}$ and $\mathrm{LiXXO_2}$ stand for graphite and lithium metal oxide, respectively, which are used as the active materials for anode and cathode.

Table 1. 1 Main commercial implementations of lithium metal oxide.

| Compound | Abbreviation | Chemical structure |
|---|---|---|
| Manganese oxide | LMO | $\mathrm{LiMn_2O_4}$ |
| Nickel manganese cobalt oxide | NCM | $\mathrm{LiNi_{1/3}Mn_{1/3}Co_{1/3}O_2}$ |
| Nickel cobalt aluminum oxide | NCA | $\mathrm{LiNi_{0.8}Co_{0.15}Al_{0.05}O_2}$ |
| Iron phosphate | LFP | $\mathrm{LiFePO_4}$ |

The main commercial implementations of $\mathrm{LiXXO_2}$ are listed in Table 1. 1. Each of the implementations solve some of the technical problems regarding safety, energy density, durability and cost but they are all compromises. The detailed selection of lithium metal oxide can vary upon specific application requirements.

### 1.2.2 Remanufacturing

In novel battery prototypes manufacturing, capacity calibration tests are crucial for verifying as-manufactured battery prototypes [43]. Unlike reusing, which repurposes malfunctioned batteries for less demanding applications with minimal intervention, remanufacturing focuses on returning the battery to its original performance for high-demand uses like EVs and fresh products [12, 44]. Those batteries with undesired performance should be remanufactured to ensure the application safety and the product quality. The conventional methodologies involve extensive durability testing on selected samples to statistically validate batch quality [45, 46]. Although accelerated tests are conducted under harsh conditions like higher current densities and temperatures, they still necessitate extensive cycles per sample up to the end-of-life (EOL) criteria, such as 80% nominal capacity. However, translating outcomes from these accelerated tests to standard scenarios poses challenges, particularly for next-generation battery research and development (R&D), which is acutely sensitive to variations in temperature, composition, and physical formats [47]. Moreover, the complexity of manufacturing, marked by initial manufacturing variabilities (IMVs) and intricate processes from material preparation to assembly, leads to interwind degradation patterns, such as voltage fade and coupled electrochemical dynamics [48-51]. These degradation patterns, chiefly kinetics and thermodynamics, manifest as impedance increase, loss of lithium-ion inventory (LLI), and loss of active material (LAM), and are traditionally analyzed via destructive post-mortem methods, contradicting practical needs for timely quality feedback. Efforts to mitigate premature degradation prioritize automated production lines to reduce human error and stringent control of microstructural properties [52-54]. Unfortunately, the accurate and non-destructive characterizations of microstructural parameters are challenging due to the complexities and intertwined degradation patterns inherent in battery production processes.

Emerging technologies such as smart manufacturing and digital twins enable monitoring of battery production processes, yet integrating recent advances like novel sensors for internal detection faces challenges in production integration and long-term use [55-62]. Manufacturers find electric signals such as current, voltage, and capacity, integrated with machine learning, promising for predicting battery EOL and internal states under diverse conditions [18, 63-66]. However, these models struggle to predict intermediate degradation patterns, with lifetime trajectory prediction posing a greater challenge due to the need for extensive sensory data and prior degradation pattern knowledge [31, 67, 68]. Despite the potential of data-driven approaches, their viability hinges on the availability of long-term sensory data, where necessary data collecting time conflicts with the goals of reducing verification time and costs [69]. Though mitigating data requirements, empirical verification models extrapolate degradation trajectories with significant calibration for novel materials and temperature requirements. Therefore, degradation information on battery prototype is critical for sustainable battery manufacturing and remanufacturing to alert malfunctions, if any, before massive production. Moreover, such an early alert is promising in advising proper waste management for the recycling and remanufacturing of defective prototypes, favoring production sustainability in the R&D stage.

### 1.2.3 Reusing

Different from newly manufactured batteries with high SOH or slight degradations, retired batteries typically have lower SOH after extensive cycling use. However, retired EV batteries are still with a huge remaining capacity value under-exploited, typically over 80% of the rated capacity [70],[71]. If not handled properly, it leads to economic burdens [8] for manufacturers and users, as well as subsequent environmental societal issues [72], including resource wastage, supply chain risks, and carbon emissions [6, 73, 74].

Promising strategies to address concerns regarding the intermittent surging of battery retirement are reusing [71, 75, 76] and recycling[77]. In reuse, retired batteries are repurposed for applications such as grid-connected energy storage systems [78], residential power supplies [79], and low-speed vehicles. However, reuse requires preliminary pretreatments, including consistency screening [80], capacity sorting [81], and regrouping [11, 82] to meet application requirements [83]. Despite the initiation of many pilot programs [33, 84, 85], the absence of stringent standards for use scenarios and retirement pathways hamper the rational use of the residual capacity [86]. Recycling uses residual values of retired batteries by materials extraction or structural repair [87, 88], suitable for irreversibly degraded retired batteries. Compared with pyrometallurgy and hydrometallurgy [89, 90], direct recycling stands out for superior profitability, lower energy consumption, and reduced carbon emissions [18, 87, 88]. Leveraging lithium replenishment and post-processing of cathode materials, direct recycling achieves efficient material structure repair, and performance restoration [91]. However, the state of health (SOH) determines the required chemical reagent and anticipated lithium supplementation dosages in direct recycling strategy formulations [92-94]. Insufficient dosage can result in an incomplete repair, while excessive dosage can lead to the generation of residual alkali on the cathode surface, deteriorating the restoration performance [88, 95]. Unfortunately, SOH retrieval, especially for safety concerned reusing, requires invasive material characterization [16, 96], and lengthy capacity calibration tests, the non-invasive, rapid, and sustainable SOH residual assessment remains an outstanding challenge.

To secure consistent SOH residual value assessment, EV batteries are monitored and operational data are recorded by cloud platforms [32] and battery passports [33]. European Parliament adopted the Batteries Directive on June 14, 2023, to ensure that retired batteries could be reused, reconditioned, or recycled at the end of service life [97]. However, the required monitoring and recording procedure is merely accessible in the EV service phase, retired batteries are no longer connected to the in-vehicle monitoring unit. Thus, it is challenging to retrieve field-available SOH data. The lifetime data integrity remains a major challenge, calling for the SOH estimation only using field data, opposite to historical data or under controllable conditions [18, 40, 98]. One solution is to perform a capacity calibration test at the retired battery collection field, i.e., SOC conditioning, which is straightforward, however, it requires unaffordable test time and extra electricity costs. The hybrid pulse power characterization test is an alternative for SOH estimation by dynamic pulse injection, but the complete test sequence takes over 12 hours. Tao *et al.* utilized short pulses for SOH estimation of retired batteries assuming that an SOC conditioning to 5% SOC was performed, which was barely compatible with random retirement SOC conditions [40]. Recent advances in sensory-based measurements include X-ray imaging [99], electrochemical impedance [100, 101], optical fiber sensing [102, 103], acoustics sensing [104, 105], partial charging [106, 107], and pulse injection [40, 80]. Sensing-based techniques are in the laboratory stage due to invasion, while pulse injection features untangling the battery degradation without physical damage, simultaneously faster than partial charging and electrochemical impedance-based methods [80, 108-112]. However, pulse injection is a data-centric method that is only feasible under the ideal assumption that physically measured data encompasses retirement conditions of the model deployment phase, known as the common challenge of domain shift in the machine learning community. Despite advances in transfer learning and domain adaption methods [66, 113, 114], challenges persist since the target domain to align still requires prior information upon deployment, and existing learning methods can hardly be updated for varied retirement conditions [115]. An alternative is spanning data testing scale under unexplored retirement conditions to mitigate data scarcity and heterogeneity while it can lead to increased costs. Wang *et al.* proposed a temperature excavation method to interpolate reaction kinetic preferences at different intermediate states during a thermochemical process, allowing for the construction of training databases at minimal thermochemistry experimental scale and cost as a data augmentation for machine learning model training [116]. Besides, generative learning also demonstrates the possibilities of estimating SOH with augmented data from partially cycled profiles, saving required physically tested data [117, 118]. However, the data scarcity and heterogeneity in battery reusing context are even more complex due to a mixture of cathode material types, physical formats, capacity designs, and historical usages, restricting potential integration of condition-specific knowledge of degradations into machine learning models.

### 1.2.4 Recycling

Recent advances in battery recycling have been focused on the pyrometallurgical, hydrometallurgical, and direct recycling [119]. In contrast to the pyrometallurgical and hydrometallurgical methods, direct recycling stands apart as a distinct approach. This process does not inflict secondary damage on the material structure, enabling more efficient structural repair and performance restoration. Moreover, direct recycling exhibits higher profitability, as it entails lower energy consumption, reduced greenhouse gas emissions, and lighter environmental footprints [88, 120]. In actual production, however, battery recyclers frequently encounter LIBs comprising unknown components or battery modules that consist of a mixture of different cathode material types. Considering that direct recycling can be heavily cathode-specific, such a complexity renders the application of direct recycling infeasible for achieving value conversion of the retired batteries [91]. It is crucial to emphasize that even if the vital metals from mixed cathode material types can be extracted using conventional recycling strategies, the interplay between different cathode materials during the recycling process can adversely impact product quality [121]. Therefore, understanding cathode material type information on the recycling side markedly impacts the direct recycling route choice and ultimately improves product quality, profitability, and sustainability.

Human-assisted direct recycling has been proposed to identify retired battery cathode material type information in the pre-treatment link, which is still not financially viable when the recycling industry is scaling up [122]. To effectively retrieve the retired battery cathode type information, the scientific and industrial community has recently initiated a battery lifetime tracing system [123] and emerging concepts like battery passport [33] and battery data genome [29]. Although substantial batteries have been utilized before those initiatives, there is a growing consensus that battery information should be accessible throughout the life chain to facilitate second-life decision-making [11]. This is notably the case for the battery recycling sector, the last station of the battery's second life, as the recycling route can be heavily cathode-specific. However, battery lifetime tracing systems or battery passports are enabled by electronic gadgets like bar codes and near-field communications, which could introduce intensive investment and could be widely incompatible with different battery designers. Furthermore, electronic gadgets remain challenging to consistently manage throughout their lifespan, leading to worn-out devices and inaccessibility at the recycling stage since the modern manufacturing process of LIBs is still not production-to-recycling integrated [124]. Hence, more breakthroughs are urgently needed to achieve an efficient battery cathode type sorting only using easy-to-access field information [3, 28], opposite to the historical data recorded or the human-assisted manner, facilitating the adoption of direct recycling to improve the quality and profitability of recycled products.

In the past few years, machine learning has emerged as a viable tool to tackle open questions in all battery fields. In other battery-related topics, machine learning has recently allowed us to automatically discover complex battery mechanisms [101, 125, 126], predict remaining useful life (RUL) [30, 63, 66, 127, 128], evaluate the SOH [100, 101, 129], optimize the cycling profile [43, 130], approximate the failure distribution [41], even to guide the battery design [131, 132] and predict life-long performance immediately after manufacturing [64]. In the case of battery recycling, few works have investigated machine learning regarding cathode materials [133, 134], which blames the scarce battery data, especially for those cycled to the end-of-life stage. The vast majority of published studies showcase very limited sample sizes [135] and are even more limited in battery cathode diversity [136]. The scarcity is attributed to the intensive cost, the long testing time [137], and, most importantly, the data privacy due to commercial or interest concerns. Consequently, the privacy issue rigidifies the dilemma where the existing battery data, though substantial in volume and diversity from multiple parties such as battery manufacturers, practical applications, academic institutions, and third-party platforms, cannot be shared. Such a dilemma calls for studying the cathode material sorting to optimize battery recycling route choice in a collaborative while privacy-persevering fashion.

FL, as a distributed and privacy-preserving paradigm, has the potential to resolve both multi-party collaboration (equivalently, the battery data volume and diversity) and privacy issues through collaborative machine learning [138-140]. In each training iteration, the distributed data owners perform local training with their local computational power, encrypt the as-trained model parameters/results, and upload them to a central coordinator for aggregation. Facts that raw datasets never leave their respective data owners and that transferred parameters/results are properly encrypted to protect data privacy. FL has been extensively investigated in numerous applicative fields, including public health [141, 142], clinical diagnosis [143-145], e-commerce [146], Internet of Things [147], mobile computing [148], and smart grid [149-151]. This approach can revolutionize the data-driven research paradigm in wide energy sectors by enabling privacy-preserving collaboration, especially for those with limited data access. Regarding the battery recycling sector, FL assumes promising possibilities for leveraging the giant amount of battery data that already exists but cannot be shared due to privacy concerns. With such a collaborative while privacy-preserving paradigm, retired battery sorting can be implemented with high accuracy, efficiency, scalability, and generalization, optimizing the quality and profitability of recycled products. To our knowledge, FL studies focused on battery recycling have never been reported.

### 1.2.4 Diagnostics and Prognostics

The recent advances in battery remanufacturing, reusing and recycling have been examined in previous literature, however, the adaptability of the methods across different battery operation conditions remains a huge concern. Much previous research reported mechanism-driven and semi-empirical prediction methods. For the mechanism-informed methods, pseudo-two-dimensional model [152], single particle model [153], electrochemical impedance spectroscopy [100, 101], distribution of relaxation time [126, 154], equivalent circuit model [155], incremental capacity analysis [156] and differential voltage analysis [131] are advantageous in accurately predicting microscopic degradation, such as lithium plating [157], solid-electrolyte-interphase formation [158], loss of lithium inventory (LLI) [159], and LAM [160]. However, the diversified operation conditions, such as dynamic charging and discharging protocols [161], SOC [162], and ambient temperatures [163], can pose significant divergence in the primary degradation mechanisms, leading to poor performance in practical use. In contrast to the mechanism-driven method, semi-empirical methods are developed by assuming equivalent circuit model [164] and empirical battery degradation patterns by deliberately fitting the historical usage parameters into the future aging curve [165], which are more computational-friendly. Even if semi-empirical methods have already showcased high accuracy on dynamic driving conditions, it is still sensitive to measurement noise and can be hardly adapted to other operation conditions.

Recent research interests focus on statistical and machine-learning aspects for their balanced prediction accuracy and model adaptability, with little requirement for expert knowledge of degradation mechanisms [29, 129, 166]. Severson et al. first proposed $var(Q_{100}-Q_{10})$ as a novel feature. Combining other features, their work reported an error of 9.1% for the early lifetime prediction even though the charging policies are changing [63]. Typically, plentiful features were combined to increase the prediction accuracy further. For instance, features such as current [167], voltage [168], temperature [169], impedance [170], incremental capacity [171], differential voltage [172], degradation trajectory [173], the onset of knee point [174] and some of their combinations [98] were widely used in battery lifetime prediction tasks. After successfully realizing an accurate prediction, research was developed towards practical scenarios where the batteries might experience more complicated and constantly-changing operation conditions [28]. For instance, Chueh et al. optimized the lifetime expectation by investigating 224 fast-charging protocols [43]. Braatz et al. proposed a hierarchical Bayesian model using twenty-dimensional features to classify the batteries subject to different charging conditions before predicting the lifetime in each classified battery group [130]. Large amounts of features and long cycling data were used to improve the cross-condition prediction accuracy but they also increased the computation [28], which was outside industry expectations. Reis et al. studied the relationship between the data amount and prediction accuracy to tackle this issue. They found that until the 20$^{th}$ cycle, increasing the amount of cycling data did not bring a better prediction [174], suggesting that the prediction capability of the features varied upon the selected cycle [175]. To date, there lacks a universal rule to guide the cross-condition feature engineering and lifetime prediction, where big difficulty in adaptability comes from the large feature divergence between different usages and their time-varying responses against cycles. Unfortunately, current research progress little in identifying which features are adaptable and when such features are best to be adapted. Adaptive machine learning is powerful in relating the cross-condition degradation features in all methods. A bridge that connects different operation conditions could be established by adaptive machine learning, with the aim of developing of highly accurate, robust, and interpretable lifetime prediction algorithms for a wide range of battery applications, especially under changing operation conditions.

### 1.2.5 Task Allocations

Diagnostics and prognostics tasks, such as SOC estimation, SOH estimation, RUL prediction, degradation trajectory prediction, cathode material type identification, degradation patterns identification are reviewed, which are summarized in Figure 1. 2.

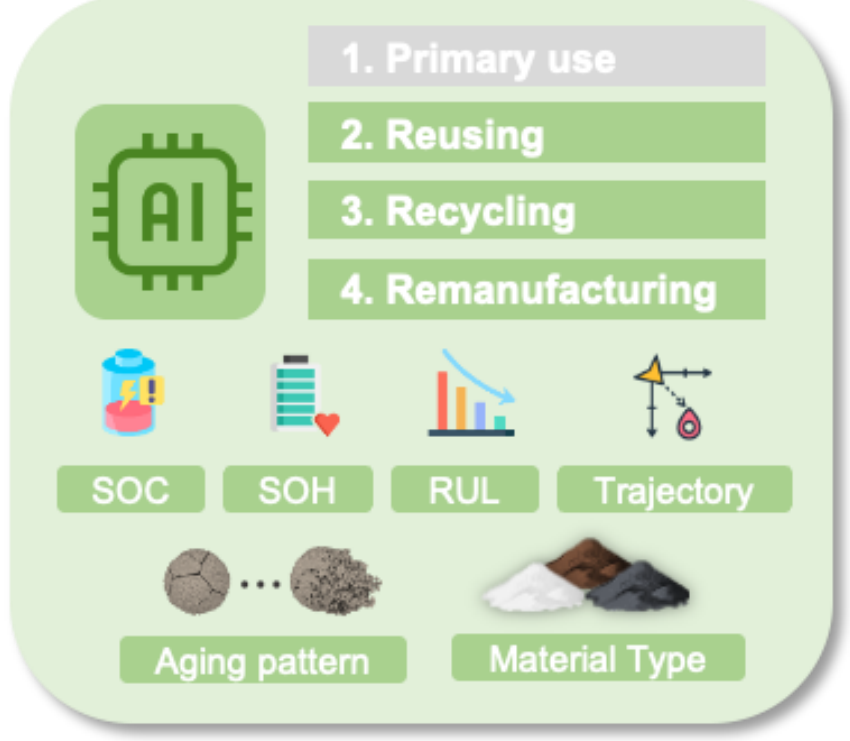

Figure 1. 2 Diagnostics and prognostics.

However, how to allocate these tasks among the different remanufacturing, reusing and recycling stages is crucial. In both EV applications and reuse, estimating SOC, SOH, and RUL is critical as batteries continue to be deployed, albeit under less stringent safety conditions, and provides essential insights into how much battery energy can still be utilized [176-178]. For recycling, SOC and SOH assessments offer insights into the extent of critical material recovery, with no requirements to estimate RUL since the operational life is irrelevant at this stage [18, 40]. For remanufacturing, SOC, SOH, and RUL serve as validation information with given known and standardized current protocols [179]. We note that the trajectory prediction is unique to remanufacturing since such information is predictable only if operation conditions are available in advance. Degradation pattern identification remains critical across all stages, ensuring safety in EV usage, reuse, and remanufacturing while optimizing recycling processes by rationally designing the recycling strategy and chemical reagent. Cathode material sorting is specific to second-life applications, where mixed batches of retired batteries, with missing battery tags and labels, require accurate identification of cathode material types for safe reuse, profitable recycling, and efficient remanufacturing. The diagnostics and prognostics tasks are summarized in Figure 1. 3.

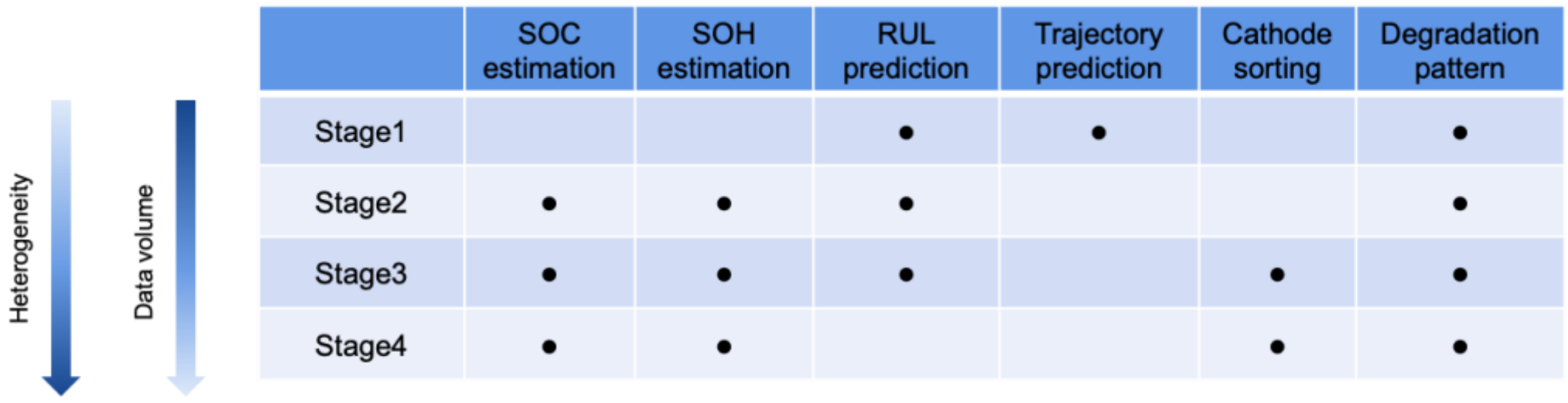

| | SOC estimation | SOH estimation | RUL prediction | Trajectory prediction | Cathode sorting | Degradation pattern |
|---|---|---|---|---|---|---|
| Stage1 | | | ● | ● | | ● |
| Stage2 | ● | ● | ● | | | ● |
| Stage3 | ● | ● | ● | | ● | ● |
| Stage4 | ● | ● | | | ● | ● |

Figure 1. 3 Diagnostics and prognostics task allocation.

## 1.3 Research Gaps

In this section, the data scarcity challenge, heterogeneity challenge, and the research problem statement are provided to highlight the research gap derived from the existing literature.

### 1.3.1 Data Scarcity

Data scarcity is a significant challenge in diagnostics and prognostics, particularly for retired batteries. This issue arises from several sources, including lack of continuous monitoring during operational life, especially in its later stages, when batteries are often disconnected with monitoring system. Furthermore, privacy concerns and commercial restrictions prevent the sharing of valuable field data from various stakeholders, such as manufacturers, operators, and third parties. Additionally, due to time, cost, and equipment limitations, comprehensive testing under diverse conditions is unfeasible, exacerbating the issue of data scarcity.

### 1.3.2 Data Heterogeneity

Data heterogeneity is another significant challenge in diagnostics and prognostics. This issue stems from the variability in battery types, including differences in chemistries, capacity designs, and manufacturing processes, resulting in diverse degradation patterns and performance characteristics. These differences complicate the integration of data-driven models, as models struggle to generalize across such varied datasets. Furthermore, the inconsistency in data collection methods and testing protocols across stakeholders, such as manufacturers, operators, and data providers, amplifies the heterogeneity issue.

### 1.3.3 Research Problem Statement

The primary challenge in this dissertation lies in developing robust and unified machine learning model for battery diagnostics and prognostics under data scarcity and heterogeneity for sustainable remanufacturing, reusing and recycling for adaptive lifetime management and sustainability. Research problem statements are summarized as below:

(1)Given very limited measurement data, how to develop quality control models for early-stage battery prognostics in the manufacturing and remanufacturing that can handle limited and heterogeneous measurement data to predict long-term performance accurately and rapidly?

(2)Given a few measurement data, how to design residual assessment models that can accurately evaluate the health status and residual value of retired batteries, given inaccessibility to the historical data and heterogeneity (material, format, capacity and historical usage) in real-world deployment scenarios?

(3)Given adequate but inaccessible measurement data, how to sort different types of cathode materials using the data from potential data contributors, such as the battery manufactures, practical applications, academic research, and third-party platform, while preserving the data privacy of the original data?

(4)Given developed models within one domain, how to effectively extend a model to generalize across different domains, diagnostics and prognostics tasks, and measurement techniques (such as long cycling and short pulse tests) in adaptive battery lifetime management with a unified framework?

## 1.4 Dissertation Overview and Contributions

This section presents the overview of this dissertation and the main contributions.

### 1.4.1 Dissertation Overview

Inspired by above research problem statements, this dissertation aims at solving the pressing challenges of data scarcity and data heterogeneity in battery remanufacturing, reusing and recycling using a unified and consistent machine learning framework for maximized battery lifecycle sustainability. The framework of this dissertation is presented in Figure 1. 4.

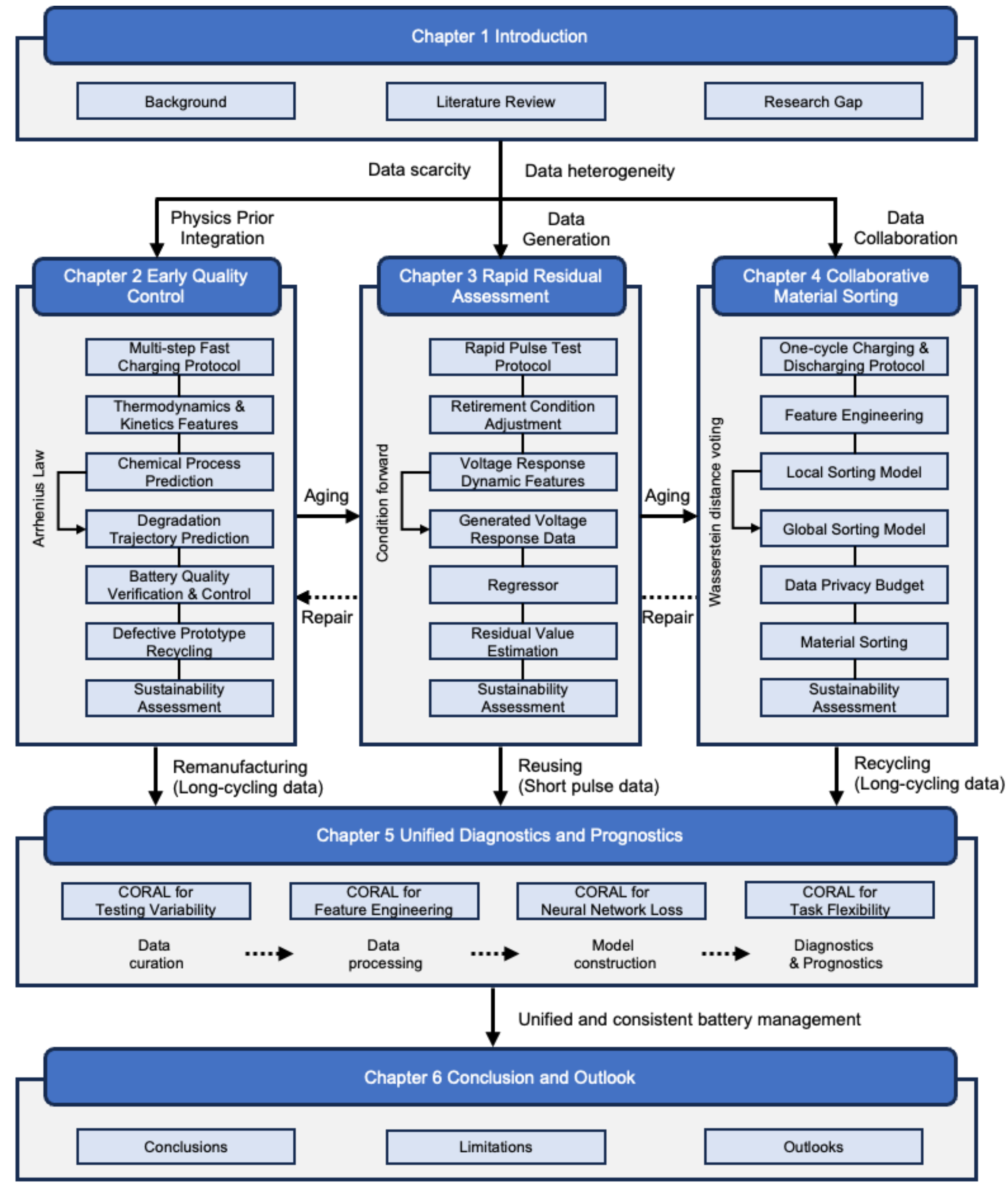

Figure 1. 4 The dissertation overview.

This dissertation starts from dealing with the data scarcity and data heterogeneity challenges by integrating physics prior knowledge, data generation and data generation collaboration using separate machine learning methods, respectively. After that, CORAL feature engineering and neural network loss integration are proposed to suit into different tests and learning paradigms as well as different lifetime stages with different data sources. Finally, long-cycling and short pulse test are validated for consistent battery diagnostics and prognostics, thus increasing the generalizability and robustness of proposed method in real-world applications. The effectiveness of the proposed CORAL method is validated through the general process of battery data curation, data processing, model construction, diagnostics and prognostics tasks.

### 1.4.2 Contribution

This dissertation makes significant contributions to the field of machine learning aided diagnostics and prognostics that address critical challenges in remanufacturing, reusing, and recycling for improved lifecycle sustainability. The key contributions are as follows:

(1) **CHAPTER 2 EARLY QUALITY CONTROL**: The dissertation proposes a physics-informed machine learning-based quality control model capable of predicting long-term battery performance from limited early-cycle observational data, otherwise traditional data required for long-term quality control are costly and time-consuming to obtain.

(2) **CHAPTER 3 RAPID RESIDUAL ASSESSMENT**: The dissertation introduces a novel residual assessment approach that accurately evaluates the health status and residual value of retired batteries by generating otherwise limited measurement data to extended data volume, despite inaccessibility of historical data and the diversity of battery materials, formats, and usage patterns.

(3) **CHAPTER 4 COLLABORATIVE MATERIAL SORTING:** The dissertation develops a federated learning (FL) strategy to sort different types of cathode materials for recycling using only one cycle of end-of-life cycling data, allowing collaboration among various battery stakeholders while preserving data privacy and ensuring high accuracy in material identification.

(4) **CHAPTER 5 UNIFIED DIAGNOSTICS AND PROGNOSTICS:** The dissertation offers a unified and consistent framework for adaptive battery diagnostics and prognostics tasks with flexibility, extending models developed in one specific domain to generalize across different domains, tasks (e.g., RUL prediction, SOC estimation and SOH estimation), and testing conditions (e.g., long cycling and short pulse tests) using CORAL, thus enhancing the robustness and applicability of predictive models in different application stages and lifetime stages.

These contributions collectively advance the sustainable management of LIBs by improving diagnostics and prognostics accuracy, enabling informed decision-making for battery remanufacturing, reusing and recycling by addressing the complexities of data scarcity and heterogeneity in real-world applications.

# CHAPTER 2 EARLY QUALITY CONTROL

## 2.1 Overview

In this chapter, we introduce a prototype battery verification method leveraging early cycle data (50 cycles, 4% of total lifetime) from both newly manufactured and accelerated aging batteries to supervise a machine learning model, as shown in Figure 2. 1. Our approach predicts battery lifetime trajectories, rather than EOL points in documented literature, by deliberately inferring multi-dimensional chemical processes before degradation trajectory prediction, using the principle that battery capacity is deterministic upon internal states. Without directly measuring the internal states, we use chemical process predictions to bypass the need for post-failure sensory information, typically obtained after lengthy and costly capacity calibration tests. Machine learning assisted verification of as-manufactured batteries solely using early data, and existing guiding samples from accelerated aging tests (e.g., high temperature).

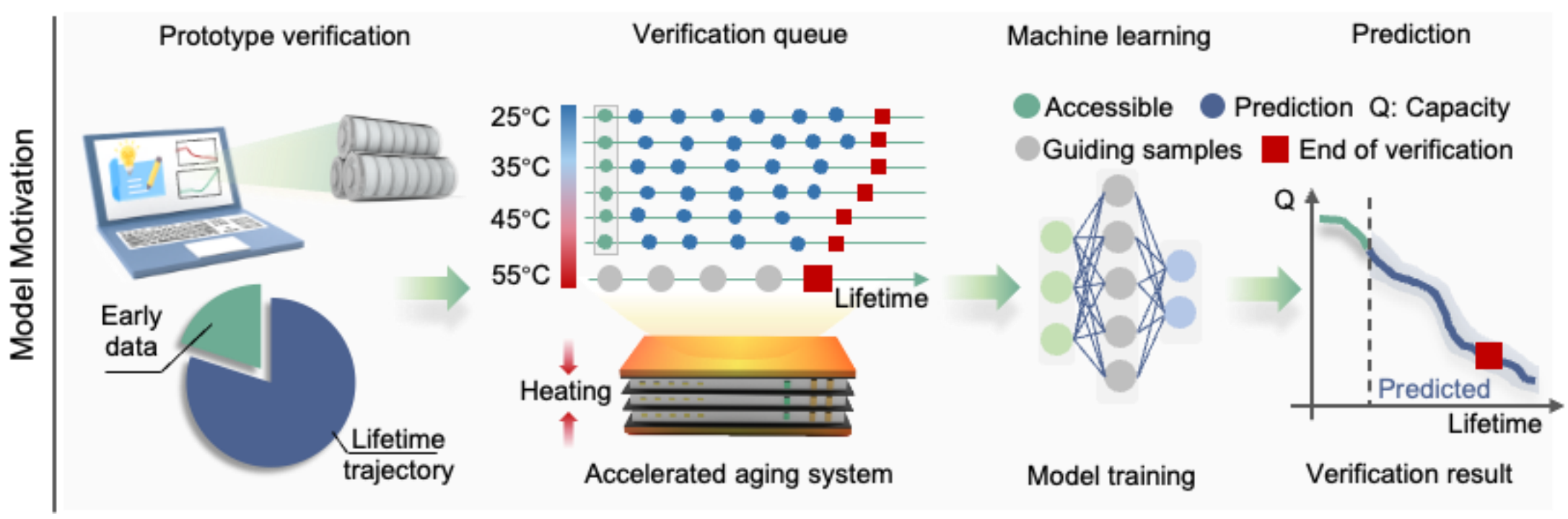

Figure 2. 1 The model motivation.

Figure 2. 2 illustrates initial manufacturing variabilities (IMVs) through a stepwise charge acceptance, aiding in calibrating our chemical process predictions. The early prediction model uses subtle yet significant initial manufacturing variabilities (IMVs) as conditions of battery aging, identified from a multi-step charging profile before cycling. Multi-dimensional chemical processes influenced by electrochemical mechanisms, including thermodynamic loss ($\Delta E$) and kinetic loss ($\eta$) can be evolved from IMVs by representing their lumped behaviors as stepwise charge acceptance. Physics-informed transferability metrics enable temperature adaptation, where $E_a$ is the activation energy, $k_B$ is the Boltzmann constant, $T_s$ is the source domain temperature and $T_t$ is the target domain temperature. The predicted state $s_{t+1}$ at time $t+1$ is continuously updated by previous state $s_t$ at time $t$, inspired by continuous electrochemical behaviors. Model interpretation from both data and electrochemistry aspects is performed to rationalize the spatially and temporally resolved decoupling of the thermodynamics and kinetics.

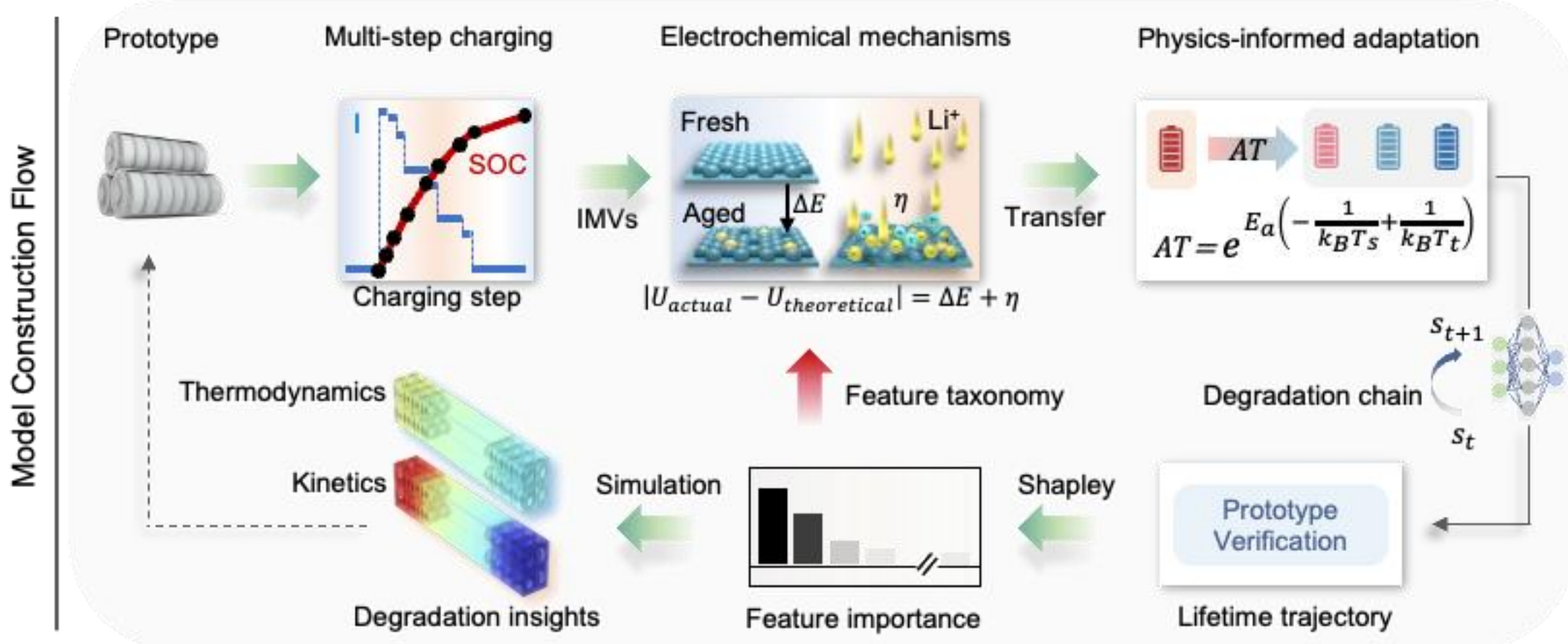

Figure 2. 2 The model components.

This model, informed by accelerated test samples, forecasts chemical processes across temperatures from early cycle data, offering a comprehensive view of battery lifetime without needing exhaustive tests. The proposed method prioritizes dynamic aging insights from multi-step charging to unravel degradation patterns, including thermodynamic and kinetic losses, and polarizations, significantly streamlining prediction complexity and cost.

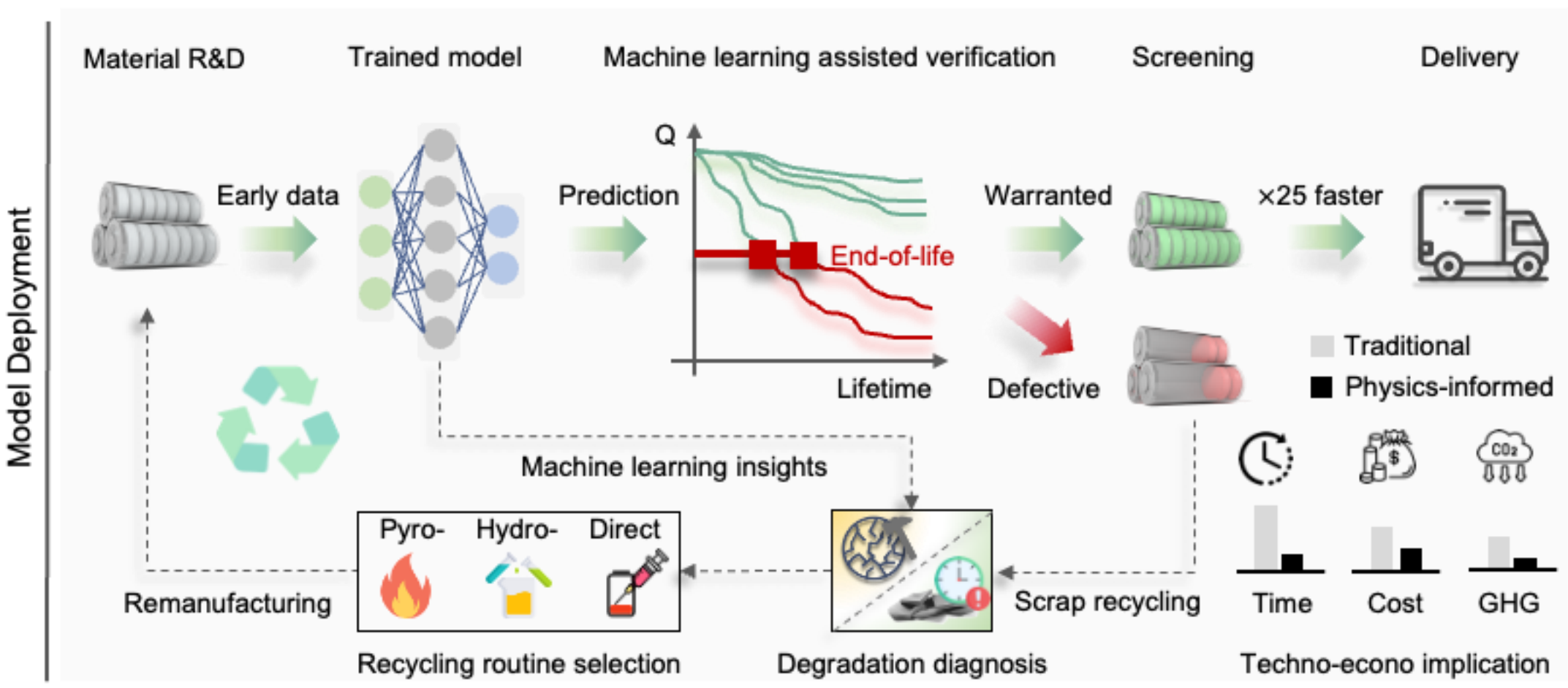

Figure 2. 3 The model deployment.

Figure 2. 3 details the application in battery R&D, enabling early quality identification and recycling of defective prototype batteries, thus minimizing economic and environmental losses. The trained physics-informed machine learning model can be deployed for accelerated product delivery of warranted batteries and optimized material recycling for defective (scrap) batteries. Routine (pyrometallurgy, hydrometallurgy, and direct recycling) selection, advised by machine learning insights, is promising in saving verification time, treatment cost, and greenhouse gas (GHG) emissions. Our findings highlight the model interpretability, the significance of non-destructive verification, and great sustainable promises in real-world applications of next-generation battery R&D verification, and scrap material recycling before massive production.

## 2.2 Thermodynamics and Kinetics

Batteries experience internal changes during aging, such as structural adjustments and LAM, leading to voltage loss. This manifests as charging voltages exceeding expected values, while discharge voltages fall below, due to electrochemical stability limits of battery materials. These constraints necessitate setting specific cut-off voltages for charging and discharging, where voltage loss accelerates reach of thresholds, consequently causing capacity decline. Therefore, tracking voltage loss offers a method to analyze capacity degradation. Despite interconnected mechanisms underlying battery degradation, it's possible to use a cathode material-independent formula to distinguish voltage loss. This formula calculates the difference between the actual electrode voltage and its theoretical counterpart, providing a method to analyze voltage loss (equivalently, capacity decline) amidst battery aging processes:

$$|U_{actual} - U_{theoretical}(*)| = \Delta E(SOC, SOH, T) + \eta(\boldsymbol{I}, SOC, SOH, T) \quad (2.1)$$

$U_{actual}$ is the actual working electrode voltage. $U_{theoretical}$ is the theoretical open-circuit voltage reflective of the essential characteristics of the battery material as-manufactured prototypes, denoted by the $*$ symbol. The $\Delta E$ is the thermodynamic voltage loss, attributed to the intrinsic material change due to aging, as a function of *SOC*, *SOH*, and environmental temperature *T*. $\eta$ is the current-induced polarization, which can be further subclassified into three parts, e.g., activation polarization ($\eta_{act}$), ohmic polarization ($\eta_{ohm}$), and concentration polarization ($\eta_{con}$) as follows:

$$\eta = \eta_{act} + \eta_{ohm} + \eta_{con} \quad (2.2)$$

This material agnostic formula quantifies contributions of thermodynamic and kinetic losses to the overall battery degradation, with their relative proportions changing as a function of *SOC*, *SOH*, environmental temperature *T*, and applied current *I*.

Particularly, applied current causes the battery working voltage to deviate from its OCV and cannot change the properties of materials, thus solely influencing $\eta$. In comparison, for the open-circuit state, voltage loss solely reflects thermodynamic loss contributions. As the applied current increases, kinetic loss becomes notably responsive. Therefore, by altering the applied current density, the relative proportions, equivalently the degradation pattern dominance, contributed by thermodynamic and kinetic loss can be modulated and quantified. Consequently, this seemingly simple and material-agnostic formula encapsulates nearly all factors pertinent to battery aging studies and enables *operando* decoupling of microscopic degradation mechanisms using macroscopic electric signals. Fundamentally, we use this formula as the theoretical support of our featurization taxonomy by comprehensively studying electric signals that can represent, at least partial information of, the voltage loss.

The core idea can be found in the Figure 2. 4. In Figure 2. 4a, the red line, when zero current is applied, reflect intrinsic material properties of as-manufactured prototypes) and the actual voltage curve (blue line). The difference between the lines can be divided into two major components, e.g., thermodynamic loss $\Delta E$ (reflective of the shift in intrinsic material properties when prototypes age), and kinetic loss $\eta$ (current-induced polarization). The theoretical OCV plus thermodynamic loss $\Delta E$ is the actual OCV, which further plus kinetic loss $\eta$ is the actual working voltage. Figure 2. 4b, the $\eta$ can be further categorized into three parts, e.g., activation polarization, ohmic polarization, and concentration polarization, with polarization effects (kinetic loss) becoming more pronounced as the applied current increases.

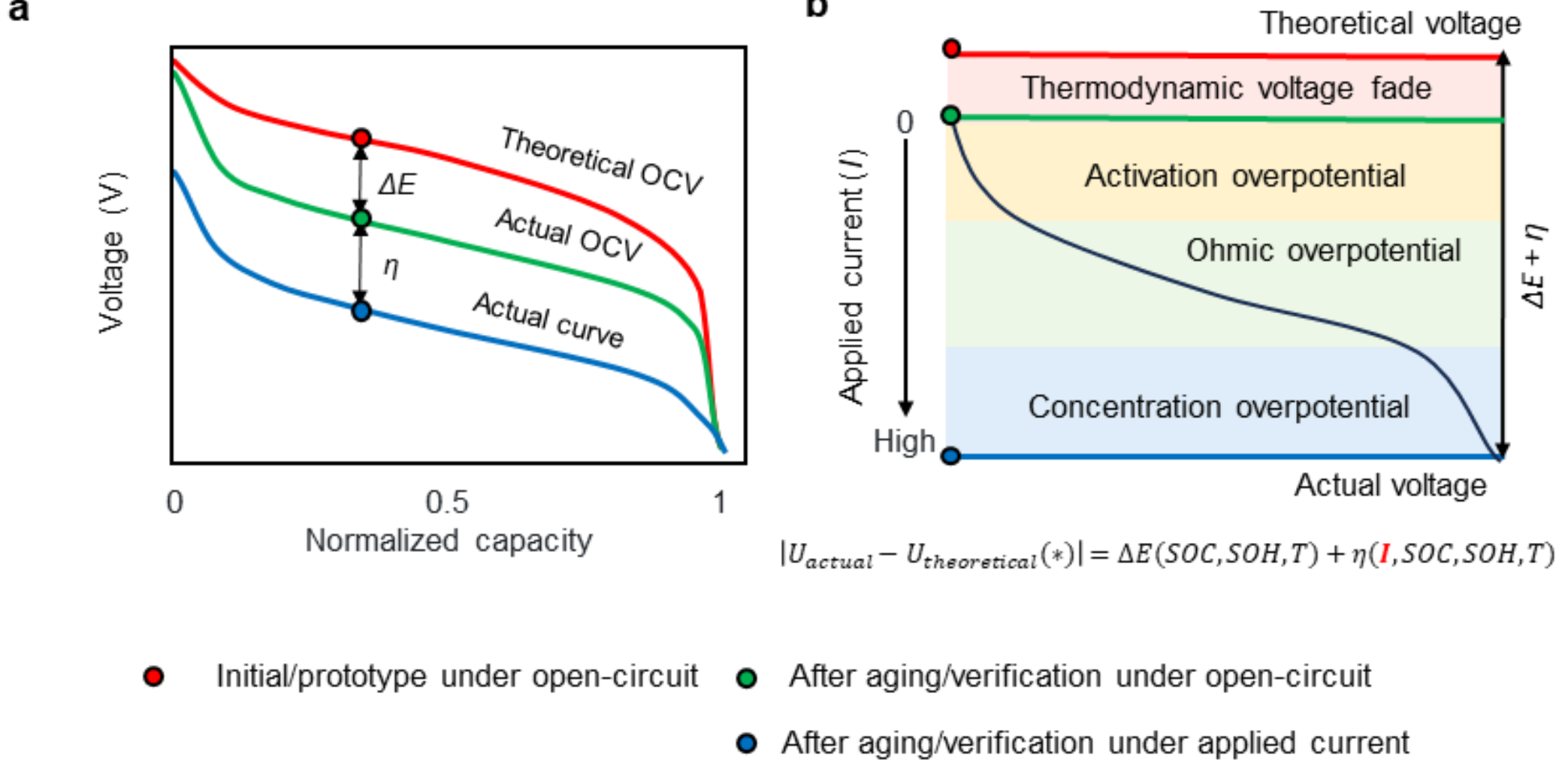

Figure 2. 4 The concept of thermodynamics and kinetics.

## 2.3 Data Collection and Processing

Prototype ternary nickel manganese cobalt LIBs ($LiNi_{0.8}Co_{0.1}Mn_{0.1}O_2$, with 1.1 Ah nominal capacity and 13wt% silicon oxide at graphite anode) were cycled under controlled temperature conditions (25, 35, 45, 55℃, see Figure 2. 5 from panels a to d) and multi-step charging (0.33C to 3C, where 1C is 1.1A) with 1C constant discharge beyond EOL thresholds (specifically, from 73% to 59% of nominal capacities).

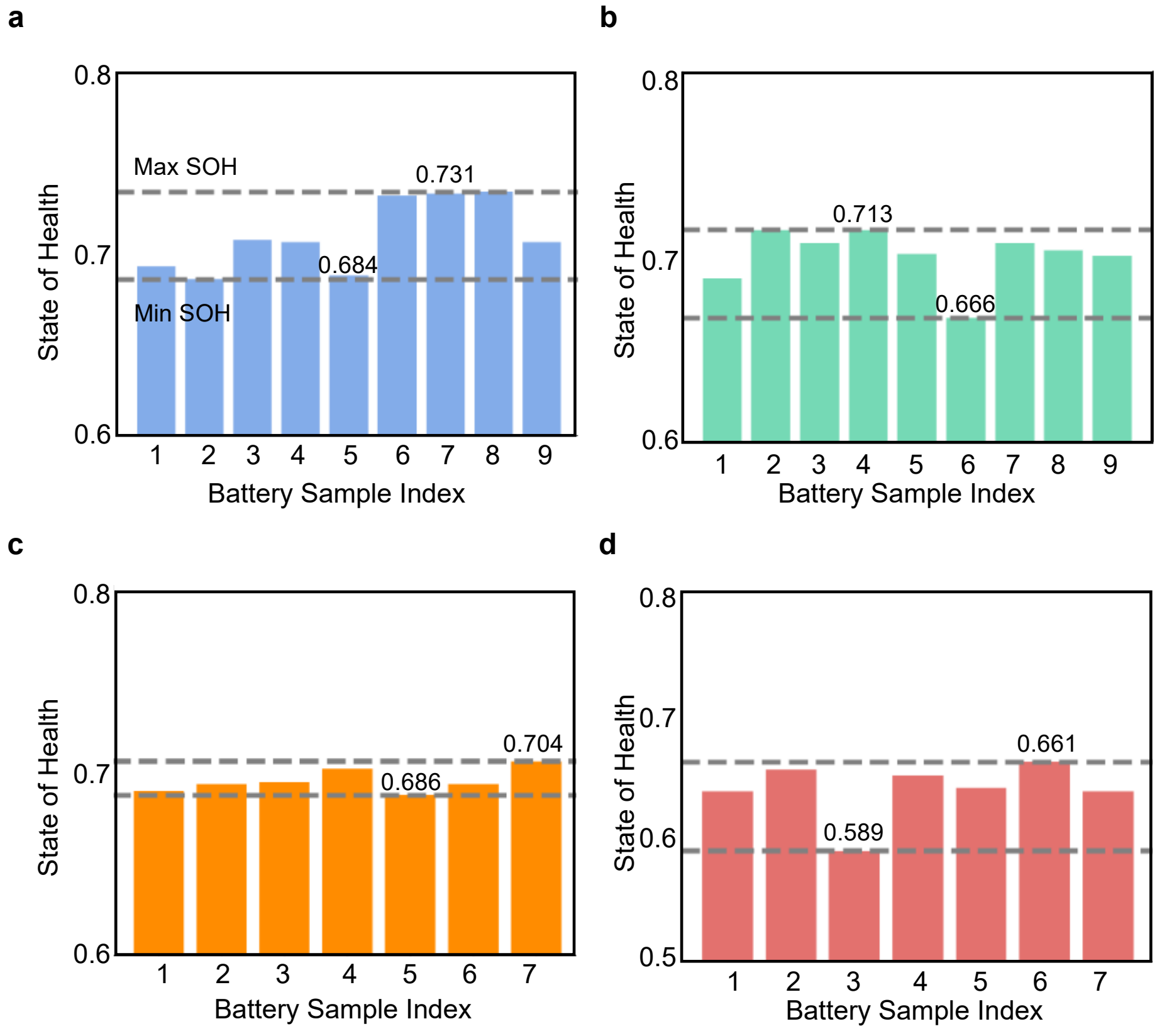

Figure 2. 5 State of health distribution.

Figure 2. 6 depicts battery capacity degrades over long-term cycles, with high temperatures reducing lifespan. Despite identical cycling and manufacturing conditions, variations in degradation trajectories are evident. The data is under 4 scenarios, i.e., 25, 35, 45 and 55℃.

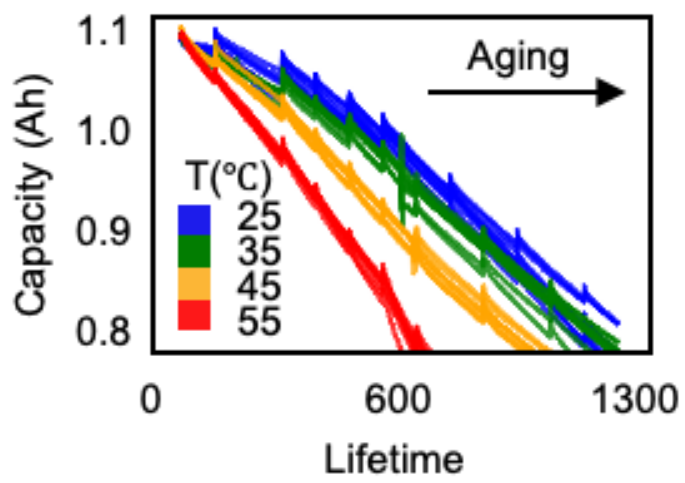

Figure 2. 6 Battery degradation trajectories.

We generate a unique battery prototype verification dataset spanning lifetimes of 480 to 1025 cycles (average lifetime of 775 with a standard deviation, i.e., STD, of 175 under EOL80 definition, see Figure 2. 7). The data is under (a) 25, (b) 35, (c) 45, (d) 55 ℃, and (e) all temperatures.

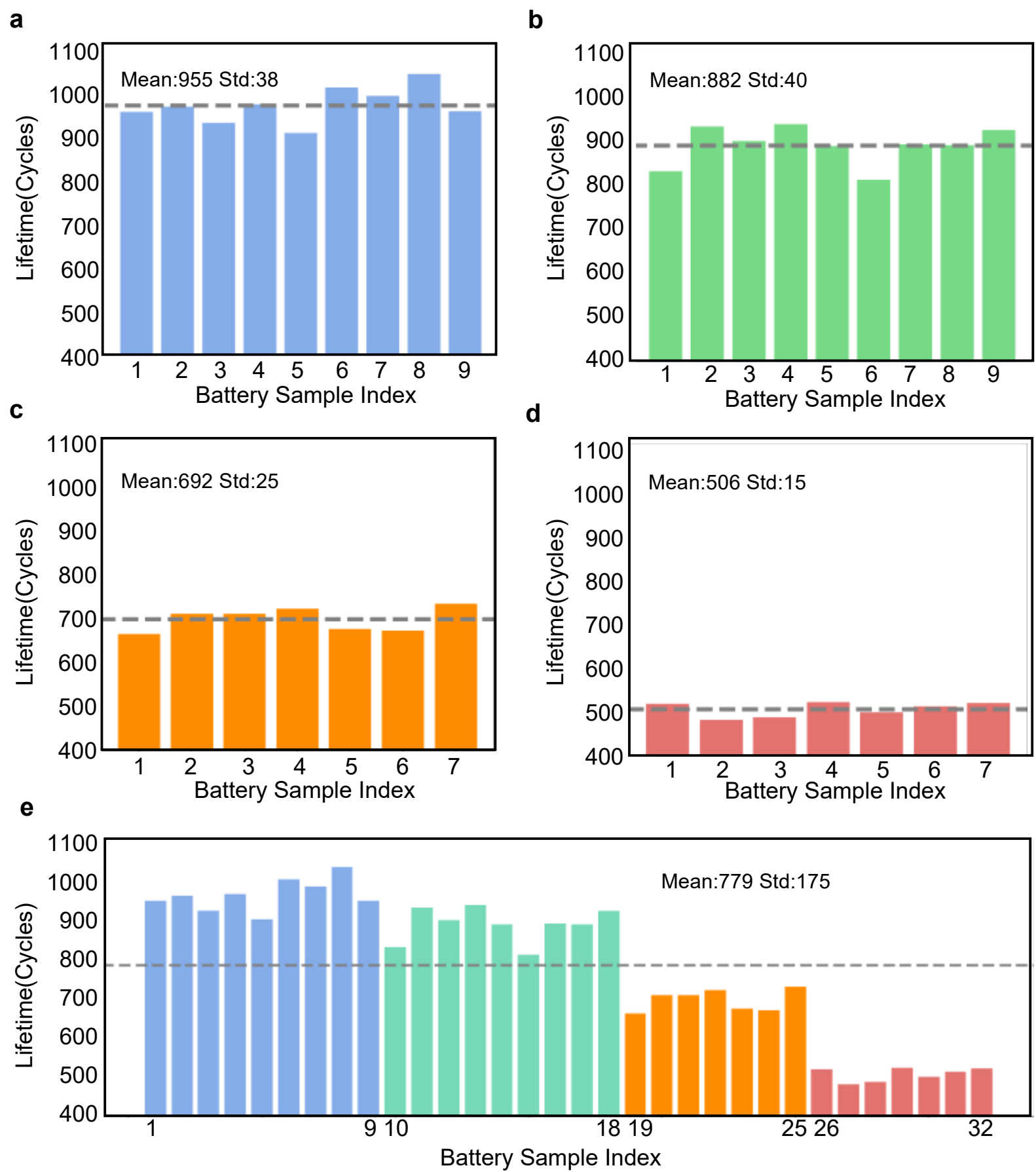

Figure 2. 7 Battery lifetime distribution at 80% of the nominal capacity.

The reference performance tests (RPTs) calibrated C rates, with SOH updates feasible in in-vehicle systems (see Figure 2. 8).

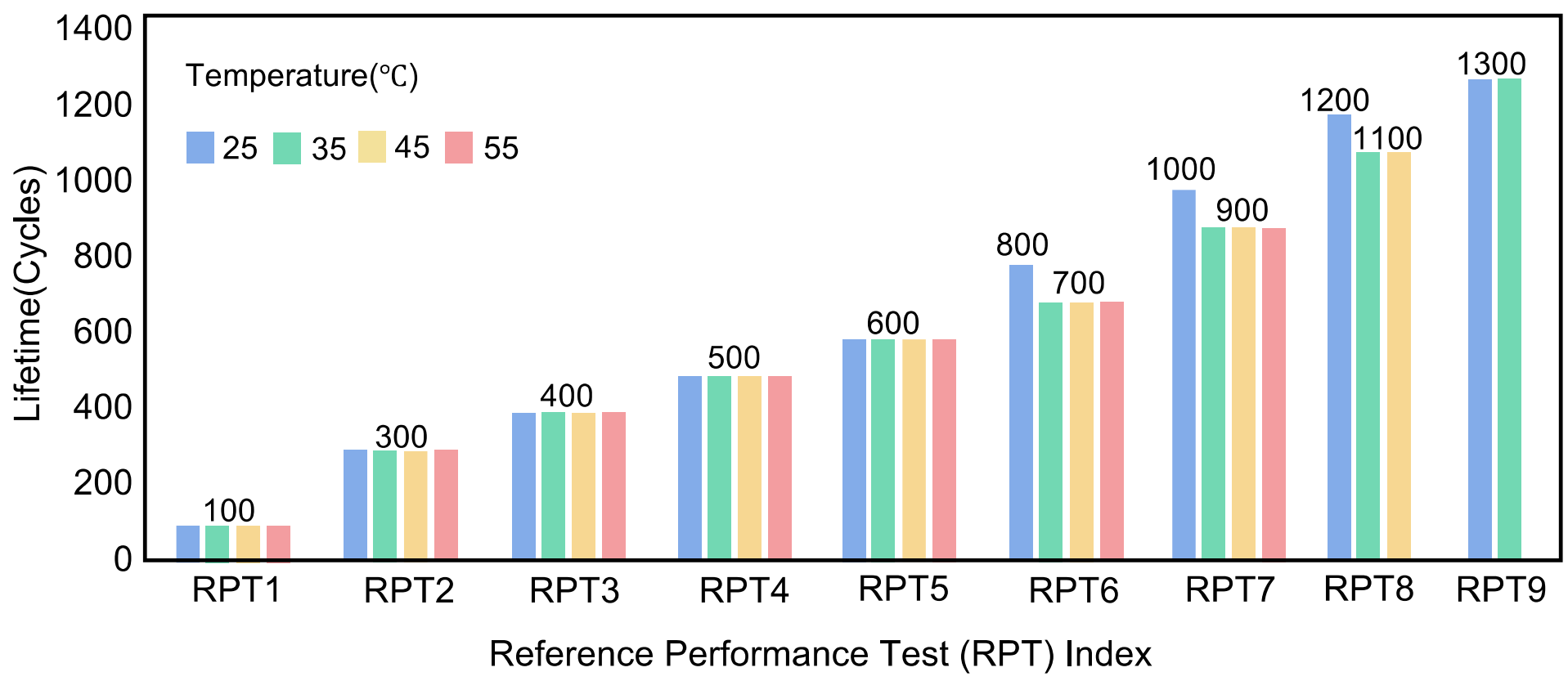

Figure 2. 8 Reference performance test for batteries.

### 2.3.1 Multi-Step Fast Charging Test Protocol

Figure 2. 9 demonstrates the key distinction from constant current verification, using a multi-step charge profile relevant to EV fast charging. This protocol charges 75% of the total SOC within 20 minutes, across a range of 0.33C to 3C with 9 consecutive steps. Details SOC allocations per step, with cut-off voltages (U1 to U9) representing charge acceptance at each SOC, are also presented.

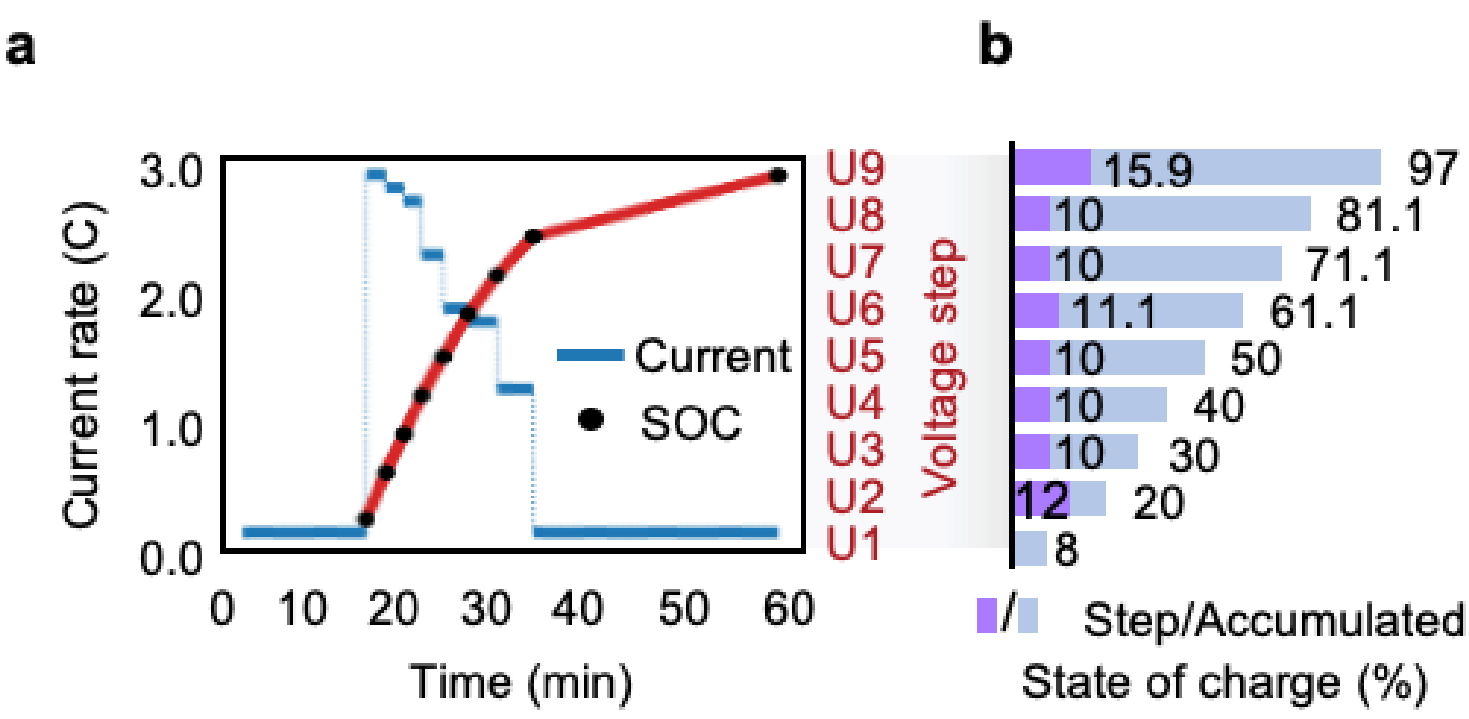

Figure 2. 9 The illustration of the multi-step fast charging.

The detailed charging profile can be found in the Table 2. 1. First, we set a fixed SOC target interval for each charging step, and apply constant current (CC) at different current rates for charging. When 1-3 cycles (Steps 3 to 14) of charging are completed, the voltages (average value) when hitting each SOC interval are used as the standard for the cut-off voltage of each stage in subsequent cycles. $C_{RPT}$ refers to the current rate as a function of updated nominal capacity after each RPT. For practical use, only Steps 4 to 12 (9-step) are required for data featurization.

Table 2. 1 The experimental design of multi-step fast charging test protocol.

| Charging details | Time duration (min) | SOC |
|---|---|---|
| Step1: Rest | 30.00 | |
| Step2: 0.33$C_{RPT}$ CC to 2.5V | - | |
| Step3: Rest | 30.00 | |
| Step4: 0.33$C_{RPT}$ CC to U1 | 14.54 | +8.0% |
| Step5: 3.00$C_{RPT}$ CC to U2 | 2.40 | +12.0% |
| Step6: 2.90$C_{RPT}$ CC to U3 | 2.07 | +10.0% |
| Step7: 2.80$C_{RPT}$ CC to U4 | 2.14 | +10.0% |
| Step8: 2.40$C_{RPT}$ CC to U5 | 2.50 | +10.0% |
| Step9: 2.00$C_{RPT}$ CC to U6 | 3.00 | +11.1% |
| Step10: 1.80$C_{RPT}$ CC to U7 | 3.33 | +10.0% |
| Step11: 1.40$C_{RPT}$ CC to U8 | 4.29 | +10.0% |
| Step12: 0.33$C_{RPT}$ CC to U9 | 28.93 | +15.9% |
| Step13: Rest | 120.00 | Summation: 97% |
| Step14: 1$C_{RPT}$ CC to (U10) | 56.40 | -94% |
| Step15: Rest | 60.00 | |
| Step16: Repeat | Steps 3 to 14 are repeated 3 times. Mean values of (U1-U9) are taken as cut-off voltages for subsequent cycling. | |

## 2.3.2 Feature Engineering of Charging Dynamics

We design the feature extraction from charging curves, taking two key aspects into account. Firstly, the controllability of charging over discharging in practical use ensures easier signal acquisition. Secondly, our multi-step charging protocol unveils varied internal state dynamics by cycling through different current rates and conditions. Features are derived from material-agnostic electrochemical principles, correlating electrical signals with underlying electrochemical states, traditionally requiring invasive methods to ascertain. Figure 2. 10a differentiates actual from theoretical battery voltage, partitioning voltage discrepancies into thermodynamic and kinetic losses—the former reflects intrinsic degradation such as LLI and LAM when idle, while the latter becomes pronounced with a high current load. We distinguish kinetics and thermodynamics by varying the current density, specifically, thermodynamics at lower currents, and kinetics at higher. The difference between actual electrode voltage $U_{actual}$ and the theoretical voltage $U_{theoretical}$ (when zero current is applied, reflective of intrinsic material properties), can be divided into two components, e.g., thermodynamic loss $\Delta E$ (reflective of the shift in intrinsic material properties), and kinetic loss $\eta$ (current-induced polarization).

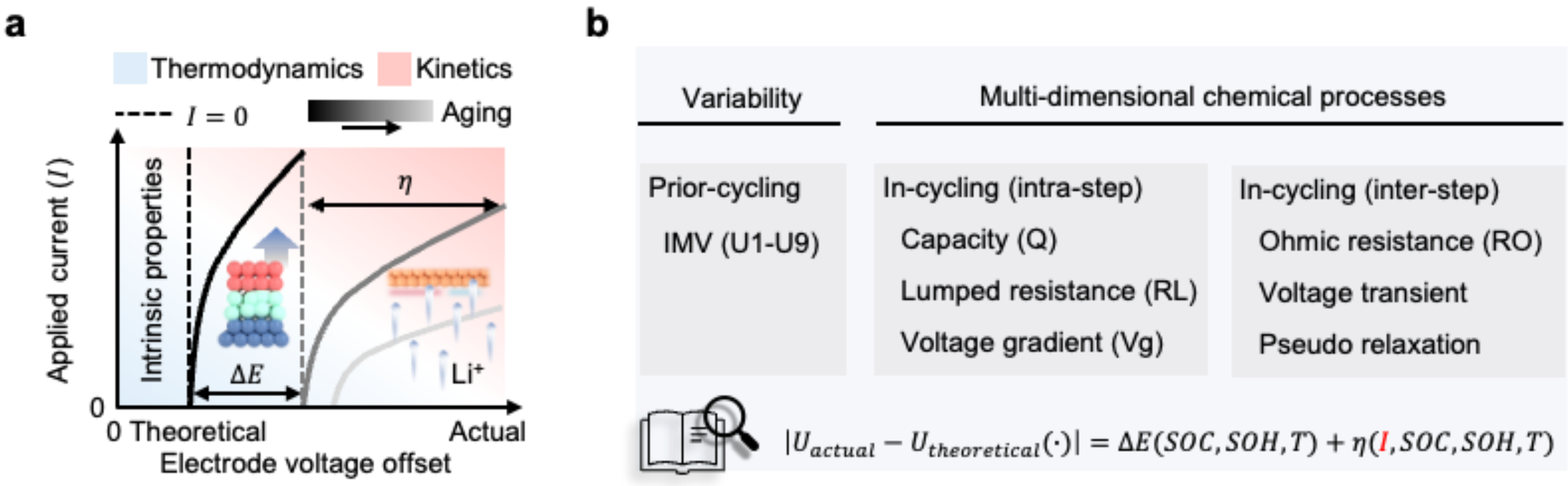

Figure 2. 10 The feature engineering from multi-step fast charging data.

Figure 2. 10b shows that the feature taxonomy captures variabilities and chemical processes before and during cycling, demonstrating disparities between actual and theoretical voltages. Extracted features are further categorized into prior- and in-cycling (intra-step and inter-step), where intra-step features are lumped representations of thermodynamic and kinetic loss, while inter-step features are purely linked to kinetic loss by current density switching. Specifically, intra-step features represent differences between $U_{actual}$ and $U_{theoretical}$, with current density deciding thermodynamic or kinetic dominance; inter-step features depict the $\eta$, including concentration, activation, and ohmic resistances behaviors. The feature taxonomy aims to decouple total capacity loss into its kinetic and thermodynamic components.

Figure 2. 11 illustrates extracted features of the chemical process over the first 800 cycles, with the ohmic resistance revealing an increase in ohmic resistance, i.e., a decline in kinetic capacity, across most switching stages. The color maps normalized feature values, and the size of bubbles maps the deviations across battery instances.

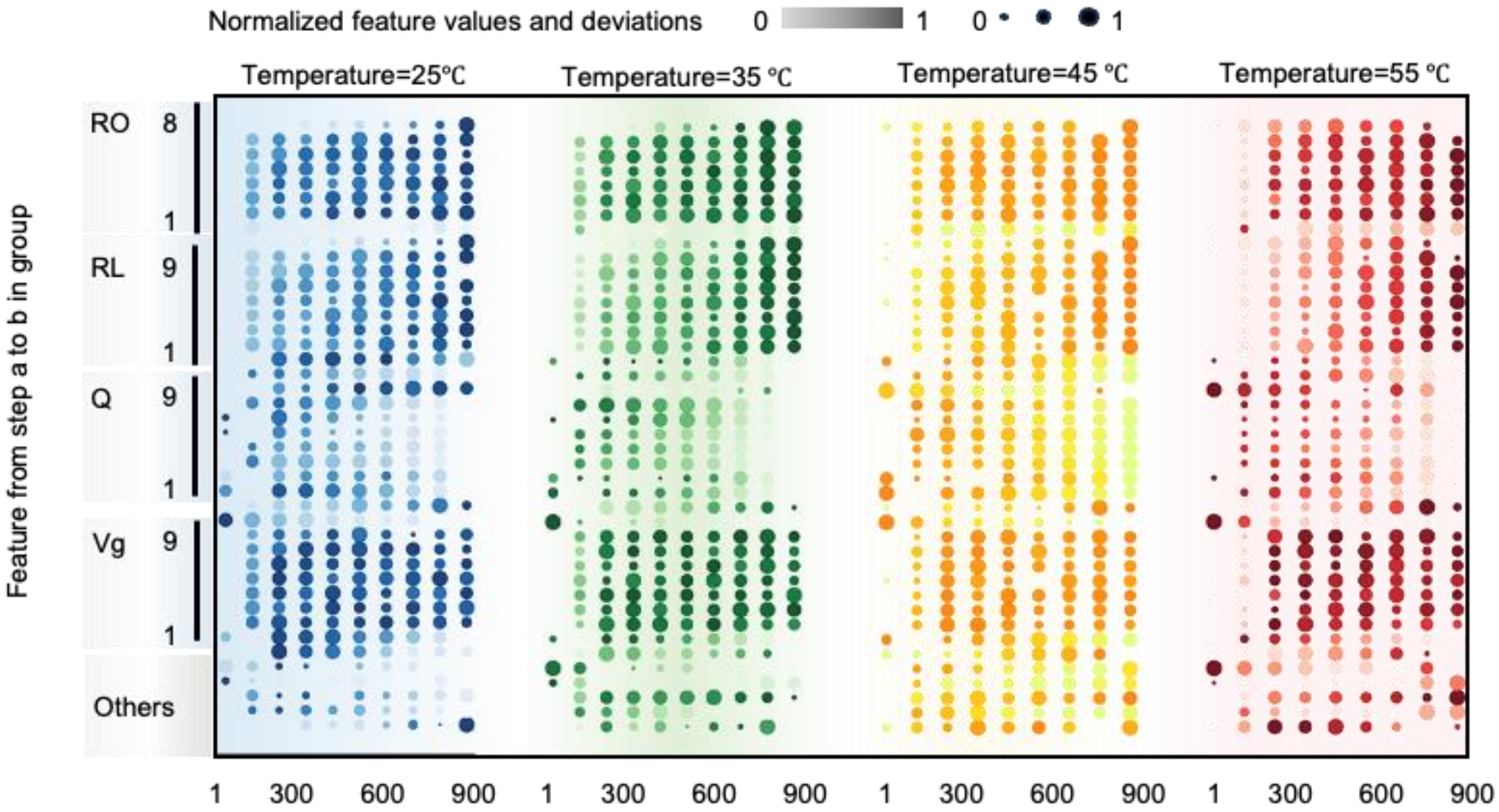

Figure 2. 11 Visualization of extracted features.

The feature taxonomy is designed to link statistical features to the physical meaning of chemical processes. Prior- and in-cycle features are extracted to characterize initial manufacturing variability (IMV) and chemical process evolution during long-term cycling. The in-cycle features are split into inter- and intra-step features thanks to the rich dynamic information provided by multi-step charging schemes. The feature identification number, taxonomy, name, description, and physical meaning are presented in the APPENDIX A.

## 2.4 Early Quality Control

As introduced in previous sections, we design the feature extraction from charging curves, taking two key aspects into account. Firstly, the controllability of charging over discharging in practical use ensures easier signal acquisition. Secondly, our multi-step charging protocol unveils varied internal state dynamics by cycling through different current rates and conditions. The interpretable features are fed into the machine learning model to predict the future degradation using early observational data, enabling the early quality control of the remanufactured batteries and accelerating the quality feedback turn overs.

### 2.4.1 Physics-Informed Transferability Metric

We propose a physics-informed transferability metric to quantitatively evaluate the effort in the knowledge transfer. The proposed transferability metric integrates prior physics knowledge inspired by the Arrhenius equation:

$$r = A\, e^{(-\frac{E_a}{k_B T})} \tag{2.3}$$

where, $A$ is a constant, $r$ is the aging rate of the battery, $E_a$ is the activation energy， $k_B$ is Boltzmann constant and $T$ is the Kelvin temperature. The Arrhenius equation provides us with important information that the aging rate of batteries is directly related to the temperature. We observe the domain-invariant representation of the aging rate ratio, consequently, the proposed Arrhenius equation-based transferability metric ($AT_{score}$) is defined as:

$$AT_{score}^{source \rightarrow target} = \frac{r_{target}}{r_{source}} = e^{(-\frac{E_a^s}{k_B T_s} + \frac{E_a^t}{k_B T_t})} \tag{2.4}$$

where, $E_a^s$ is the activation energy of the source domain, $E_a^t$ is the activation energy of the target domain, $T_s$ and $T_t$ are the Kelvin temperatures of the source domain and the target domain, respectively. The closer the $AT_{score}$ is to 1, the more similar the source domain and target domain are, so the better the knowledge transfer is expected. Since the dominating aging mechanism is unknown (characterized by $E_a$) as a posterior, we alternatively determine the aging rate by calculating the first derivative concerning the variations on the predicted chemical process curve:

$$r = \frac{\mathrm{d}\hat{\boldsymbol{F}}}{\mathrm{d}C} \tag{2.5}$$

where, $\hat{\boldsymbol{F}}$ is the predicted chemical process feature matrix. We linearize the calculation in adjacent cycles by sampling the point pairs on the predicted chemical process:

$$r = \frac{\sum_{i=0}^{n} F_{end+i} - F_{start+i}}{n * (end - start)} \tag{2.6}$$

where, $n$ is the number of point pairs, $start$ and $end$ are the cycle index where we start and end the sampling, respectively. $F_{start+i}$ and $F_{end+i}$ is the feature value for the $(start + i)$th and the $(end + i)$th cycle, respectively. This calculation mitigates the noise-induced errors, resulting in a more robust aging rate computation. For domains where the aging mechanism is already known (different domains share the same $E_a$), the $AT_{score}$ can be expressed in the following form:

$$AT_{score}^{source \rightarrow target} = e^{(-\frac{1}{T_s}+\frac{1}{T_t})\alpha} \tag{2.7}$$

where $\alpha = \frac{E_a}{k_B}$ is a constant value. This formula ensures that, in cases where the aging mechanism is known, we can calculate transferability between different domains using only the temperatures of the source and target domains. This allows the model for continuous temperature generalization without any target data.

### 2.4.2 Multi-Temperature Domain Adaptation

Multi-source domain adaptation is an effective solution to alleviate data scarcity in the target domain. Using the physics-informed transferability metric, we assign a weight vector $W_{1\times K} = \{W_i\}$ (where $K$ is the number of source domains, $W_i$ is ensemble weight for the $i$th source domain) to source domains to quantify the contributions when predicting the chemical process of target domain. The $W_i$ is defined as:

$$W_i = \left(|AT_{score}^{source\,i\rightarrow\,target} - 1| \cdot \sum_{j=1}^{K} \frac{1}{|AT_{score}^{source\,j\rightarrow\,target} - 1|}\right)^{-1} \tag{2.8}$$

where, $AT_{score}^{source\,i\rightarrow\,target}$ is the $AT_{score}$ from the $i$th source domain to the target domain. This mechanism ensures the source domain with better transferability has a higher weight, effectively quantifying the contribution of each source domain to the prediction of the target domain. From Equation (2.4) and Equation (2.8), we can obtain the aging rate of the target domain:

$$r_{target} = \sum_{i=1}^{K} W_i \cdot AT_{score}^{source\,i\rightarrow\,target} \cdot r_{source}^{i} \tag{2.9}$$

The multi-source transfer based on $AT_{score}$ includes the following three steps. Here we give an example for illustration. Detailed settings to reproduce the results in the manuscript are otherwise specified. First, we calculate aging rates $r$ for all target and source domains by using early-stage data, i.e., we set $start = 100$, $end = 200$, $n = 50$ in Equation (2.6). After calculating aging rates for all features or aging curves, we obtain a target domain aging rate vector $r_{1\times N}^{target}$ and a source domain aging rate matrix $r_{K\times N}^{source}$, where $K$ and $N$ are the number of source domains and the number of features, respectively. Second, we calculate the transferability metric $AT_{score}$ and weight vector $W_{1\times K} = \{W_i\}$ by Equation (2.4) and Equation (2.8). Third, we predict the late stage (cycles after 200) aging rate of the target domain ($r_{target}$) using Equation (2.9). Note that $AT_{score}^{source\,i\rightarrow\,target}$ and $W_i$ are obtained by both target and source domain early-stage data, which are used to measure the transferability from source domain to target domain based on their aging rate similarity. $r_{source}^{i}$ is obtained from all accessible data in the source domain, consistent with our definition of the early-stage estimate problem. Specifically, only early-stage data in target domain is available in practice, while source domains can provide more comprehensive aging information to assist the target prediction using complete data. For multi-source domain adaptation, the source domain temperature is set to 25°C, and 55°C, and the target domain temperature is set to 35°C, and 45°C, for practical verification purposes that intermediate temperatures should be studied. For uni-source domain adaptation, the source domain temperature is set to 55°C, and the target domain temperature is set to 25°C, 35°C, and 45°C, for practical verification purposes that use accelerated data (55 °C ) to rapidly verify battery performance under other temperatures.

### 2.4.3 Chemical Process Prediction Model Performance

Battery lifetime inconsistencies often stem from manufacturing process instabilities, or initial manufacturing variabilities (IMVs). We probe IMVs during an early cycling phase through a nine-step charging regimen, designating SOC levels at each phase, and measuring corresponding cut-off voltages to approximate IMVs. Essentially, the IMVs are the difference between the $U_{actual}$ and $U_{theoretical}$ at the initial cycling stage, reflective of the shift in intrinsic material properties of as-manufactured prototypes.

Considering the cut-off voltage is a scalar vector for each battery, we deliberately broadcast dummy cycling indexes to span the cut-off voltage vector $U_{m\times 9}$ to a cut-off voltage matrix $U_{(C\times m)\times 10}$ to predict continuous chemical process, where $m$ is the battery number and $C$ is the total length of cycling index of all batteries, equivalently the length of the entire lifetime. Given a feature matrix $\boldsymbol{F}_{(C\times m)\times N}$, where $N$ is the number of features, the model learns a composition of $L$ intermediate layers of a neural network:

$$\hat{\boldsymbol{F}} = f_{\Theta}(U) = \left(f_{\sigma}^{(L)}\left(f_{\theta^{(L)}}^{(L)}\right) \circ \cdots \circ f_{\sigma}^{(1)}\left(f_{\theta^{(1)}}^{(1)}\right)\right)(U) \tag{2.10}$$

where, $L = 3$ in this work. $\hat{\boldsymbol{F}}$ is the output feature matrix, i.e., $\hat{\boldsymbol{F}}_{(C\times m)\times N}$, $\Theta = \{\theta^{(1)}, \theta^{(2)}, \theta^{(3)}\}$ is the collection of the network parameters from each layer, $U$ is the broadcasted input voltage matrix $U_{(C\times m)\times 10}$, and $f_{\Theta}(U)$ is a neural network predictor. Here all layers are fully connected with $f_{\sigma}$, which is a leaky rectified linear unit (Leaky ReLU) activation function. The number of neurons in hidden layers $f_{\theta^{(1)}}^{(1)}$, $f_{\theta^{(2)}}^{(2)}$, and $f_{\theta^{(3)}}^{(3)}$ are 32, 64, and 32 respectively.

The chemical process prediction model is trained on selected temperatures (i.e., temperatures where already measured data are accessible), while the temperature-related chemical process variations are considered using the physics-informed transferability. In each selected temperature, we split the data into 75% and 25% for training and testing, respectively. We train the chemical process prediction model using Adam optimizer, with epochs of 30 and a learning rate of $10^{-4}$. The loss function of the chemical process prediction model is:

$$L_{loss_ChemicalProcess} = \frac{\sum_{i=1}^{C}(\boldsymbol{F}_i - \hat{\boldsymbol{F}}_i)^2}{C} + \lambda_1 * \sum_{i=1}^{C} |\boldsymbol{F}_i - \hat{\boldsymbol{F}}_i| \tag{2.11}$$

where $\boldsymbol{F}_i$ is the $i$th label of defined chemical processes, $\hat{\boldsymbol{F}}_i$ is the predicted chemical processes feature matrix for the $i$th cycle, $\lambda_1$ is the regularization parameter, which is set to $10^{-5}$.

### 2.4.4 Lifetime Trajectory Prediction Model

Battery chemical process degradation is continuous, which we call the "Chain of Degradation". We have predicted the $r_{target}$ aging rates of each feature in the target domain, which can be further used to predict the chemical process. Therefore, when using aging rates $r_{target}$ to calculate each target feature $\boldsymbol{F}_{(C\times m)\times 1}$ in the feature matrix $\boldsymbol{F}_{(C\times m)\times N}$, the $i$th cycle target feature $\boldsymbol{F}_{target}^{i}$ should be based on $\boldsymbol{F}_{target}^{i-1}$ and $r^{i-1}$:

$$\boldsymbol{F}_{target}^{i} = \boldsymbol{F}_{target}^{i-1} + \sum_{j=i}^{K} W_j \cdot A_{score}^{source\ j \rightarrow target} \cdot r_{source j}^{i-1} \tag{2.12}$$

where, the $\boldsymbol{F}_{target}^{i}$ is the feature value of target domain in the $i$th cycle, $r_{source\ j}^{i-1}$ is the aging rate of source domain $j$ at the $(i-1)$th cycle. We concatenate the $N$ feature vectors $\boldsymbol{F}_{(C\times m)\times 1}$ to get the feature matrix $\boldsymbol{F}_{(C\times m)\times N}$.

It is assumed that the chemical process of the battery deterministically affects the aging process, we therefore use the predicted chemical process to predict the battery degradation curve. Battery degradation trajectory model learns $L$ intermediate mappings:

$$\hat{D} = f_{\Theta}\left(\hat{\boldsymbol{F}}\right) = \left(f_{\sigma}^{(L)}\left(f_{\theta^{(L)}}^{(L)}\right) \circ \cdots \circ f_{\sigma}^{(1)}\left(f_{\theta^{(1)}}^{(1)}\right)\right)\left(\hat{\boldsymbol{F}}\right) \tag{2.13}$$

where, $L = 3$ in this work. $\hat{D}$ is predicted battery degradation trajectories, $\Theta = \{\theta^{(1)}, \theta^{(2)}, \theta^{(3)}\}$ is the collection of the neural network parameters from each layer, $\hat{\boldsymbol{F}}$ is the predicted battery chemical process feature matrix, and the $f_{\Theta}(\hat{\boldsymbol{F}})$ is a neural network predictor. Here all layers are fully connected with $f_{\sigma}$, which is a leaky ReLU activation function. The number of neurons in hidden layers $f_{\theta^{(1)}}^{(1)}$, $f_{\theta^{(2)}}^{(2)}$, and $f_{\theta^{(3)}}^{(3)}$ are 32, 64, and 32 respectively. Adam optimizer was used for training, with epochs of 100 and a learning rate of $10^{-3}$. The loss function is defined as:

$$L_{loss_DegradationTrajectory} = \frac{\sum_{i=1}^{C}(y_i - \hat{y}_i)^2}{C} + \lambda_2 * \sum_{i=1}^{C} |y_i - \hat{y}_i| \tag{2.14}$$

where $y_i$ is the $i$th label of defined battery capacity trajectory, $\hat{y}_i$ is the $i$th predicted battery capacity trajectory, $\lambda_2$ is the regularization parameter, which is set to $10^{-5}$.

### 2.4.5 Lifetime Trajectory Prediction Results

To demonstrate the robustness of the proposed method under unseen conditions, we introduce two models for early verification tailored to manufacturer needs under multi- and uni-source domain adaptation, utilizing data at varied temperatures. The full feature set is input into these models, with specifics in APPENDIX A and APPENDIX B.

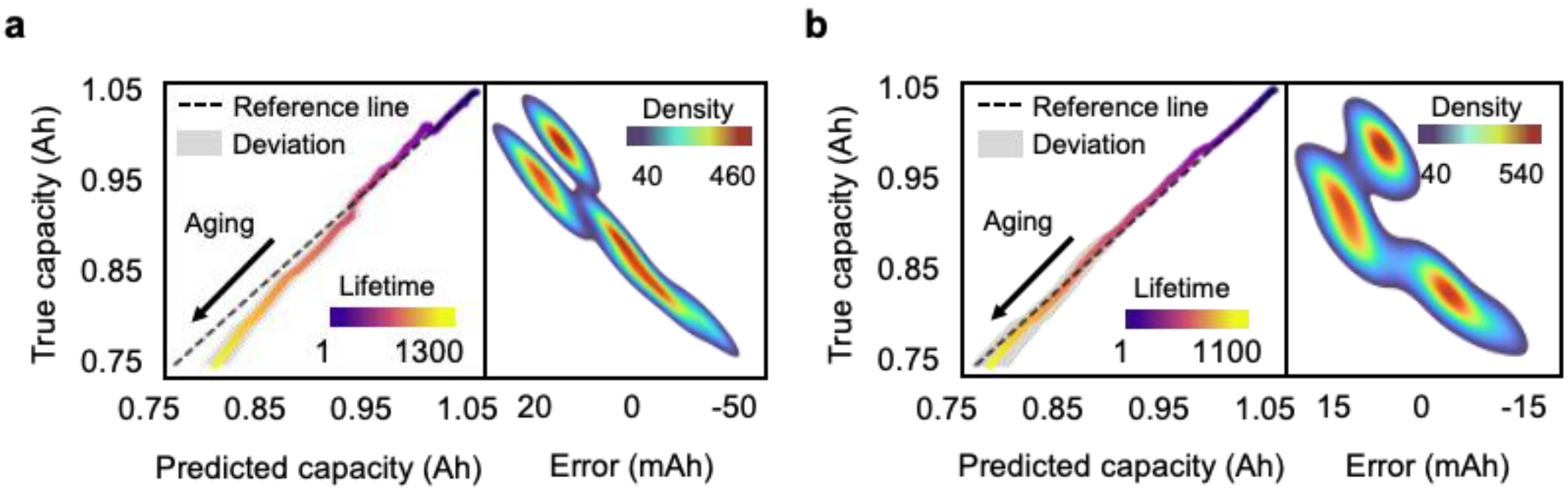

Figure 2. 12 Degradation trajectory prediction results.

All numerical results achieved in this plot are under the early verification setting that uses 20% of early data from batteries to be verified. Experimental settings are multi-source domain adaptation, i.e., data at 25 and 55℃ is accessible. Figure 2. 12a and Figure 2. 12b present parity plots and error distributions for the target domain at 35°C and 45°C, showing MAPEs (STD) of 1.4% (0.014) and 0.6% (0.006), respectively. Notably, overestimations occur as batteries near EOL, underscoring the verification challenge across the entire lifetime with early data and emphasizing the verification complexity. We compare the model against state-of-the-art methods across different lifetime phases (early, middle, and late, each representing 10% of the total lifetime) in the test set.

Figure 2. 13 contrasts the model performance with a long-short-term memory (LSTM) network (model 1), a model excluding IMVs (model 2), a model without physics-informed learning (model 3, lacking Arrhenius-based transfer), and a model using empirical formula-based model (model 4), detailed in APPENDIX C.

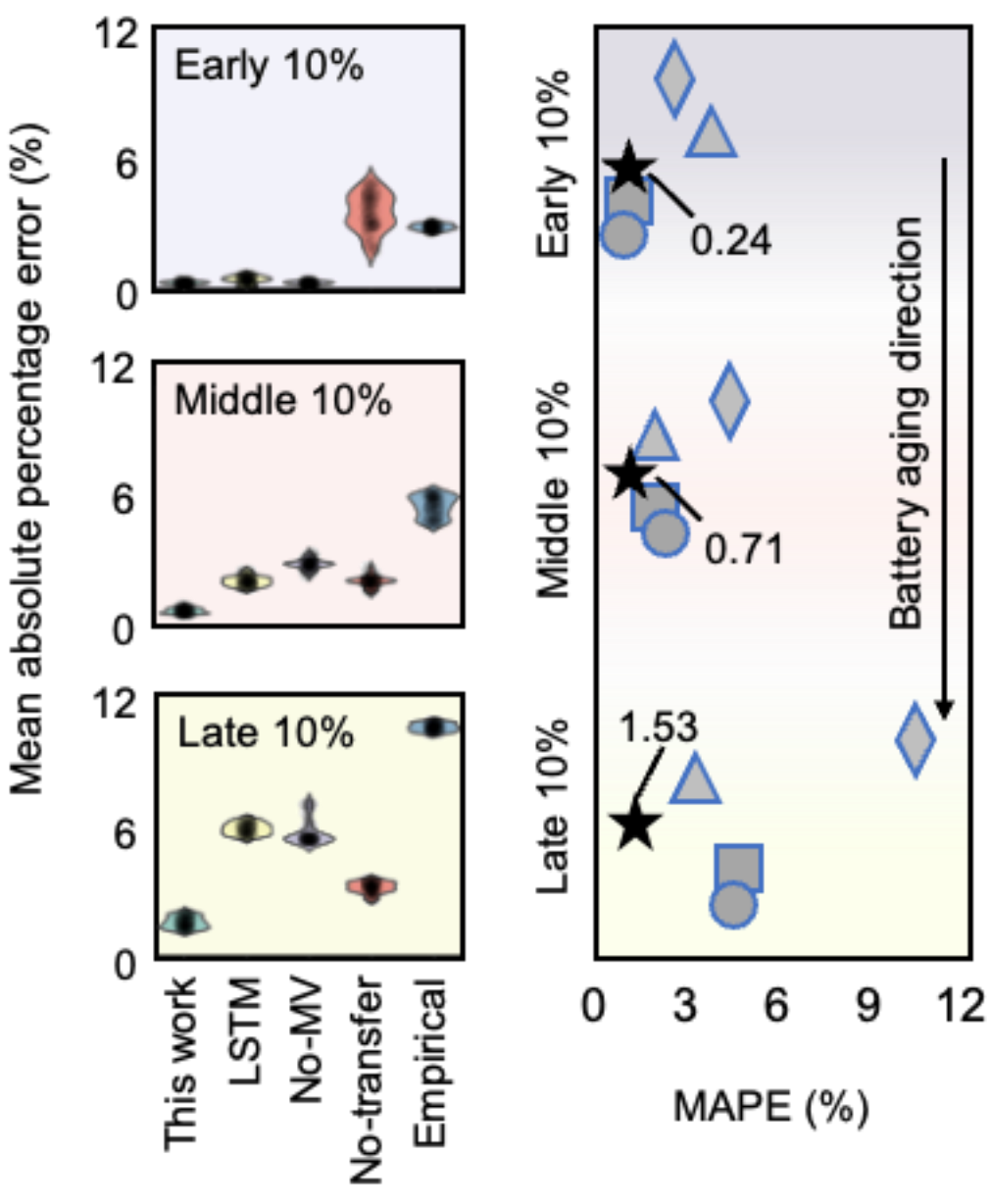

Figure 2. 13 Model performance comparisons.

In early cycles (first 10%), our method and model 2 both achieve a MAPE of 0.24%, while model 1 has a slightly higher error at 0.38%. Models 3 and 4, however, struggle with significant errors of 3.48% and 2.82%, respectively, highlighting difficulties with early temperature-induced lifetime deviations. Notably, model 2 worsens in the last 10% of cycles, with MAPE of 5.82%, underscoring the importance of temperature consideration, which model 2 lacks. Despite initial similarities, IMVs become crucial in later stages, with model 2 showing a late-cycle MAPE of 5.62%. Our model remains robust and precise across all stages, peaking at a MAPE of 1.53% in the last 10% of cycles, demonstrating the efficacy of incorporating IMVs and physics knowledge to address temperature-induced long-term variations. Moreover, the error distribution of the proposed method is significantly lower than the benchmark models, strengthening its practical relevance, especially in quality assurance contexts where prediction performances in extreme cases matter most.

### 2.4.6 Data Availability Analysis

Our analysis extends to sensitivities across varying EOL capacity values for customized verification scenarios. Figure 2. 14 indicates that benchmark models perform worse as the target capacity decreases, underscoring the difficulty of projecting future degradation with initial data alone. Yet, our model consistently surpasses others, with a maximum deviation of 33 cycles, despite predictions being supervised thousands of cycles in advance.

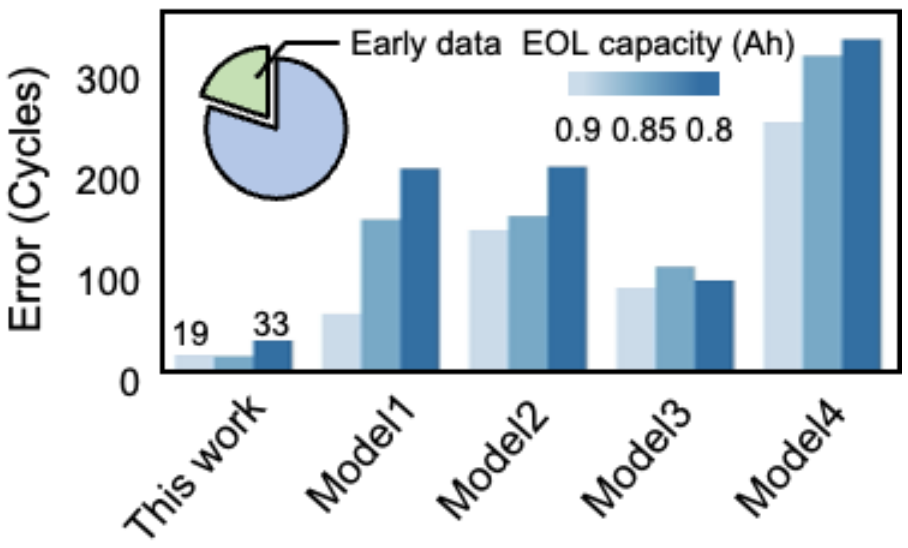

Figure 2. 14 Model sensitivity under different end-of-life (EOL) capacities.

To enhance verification speed and computational efficiency, we explore the reduced data availability, showing in Figure 2. 15a that errors remain below 2% MAPE at both 35 and 45°C with just 4.2% of lifetime data (50 cycles). Note that accessible data at 55℃ is assumed, which is used to predict the lifetime of 25, 35, and 45 ℃ verification scenarios. The challenge of data scarcity, particularly with a limited number of parallel samples due to constraints in cost or time, was also assessed. A specific test using only high-temperature (55°C) samples for accelerated verification demonstrates the impact of data paucity: a single sample results in high errors, but increasing to five samples significantly improves verification accuracy across various temperatures (Figure 2. 15b).

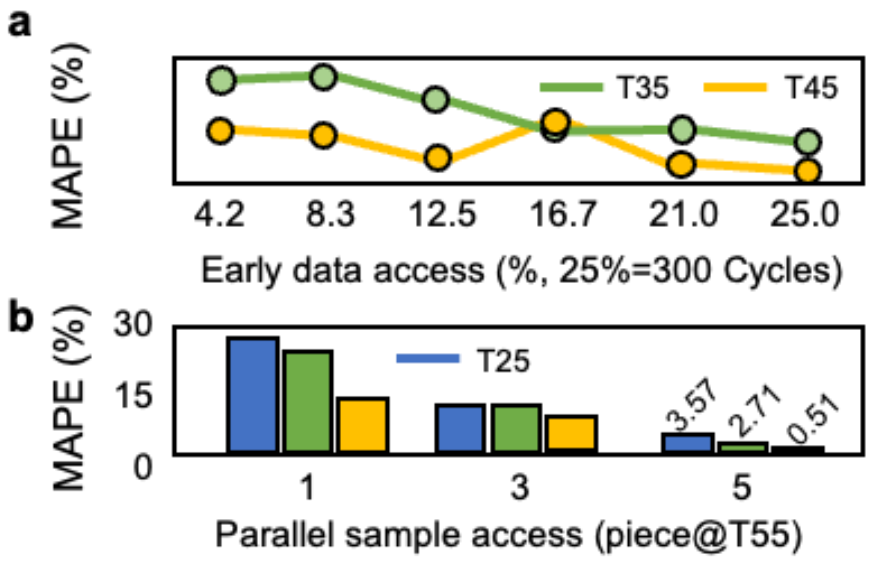

Figure 2. 15 Model prediction error.

Table 2. 2 Model early verification performance comparison.

| | The first 50 cycles are accessible | | | The first 25 cycles are accessible | | |
|---|---|---|---|---|---|---|
| | Verification temperature at 45℃ | | | | | |
| | MAPE (%) | STD | maxMAPE (%) | MAPE (%) | STD | maxMAPE (%) |
| Ours | 0.99 | 0.36 | 1.60 | 1.27 | 0.44 | 2.17 |
| Model1 | 67.75 | 7.91 | 81.37 | 87.95 | 9.86 | 104.68 |
| Model2 | 1.47 | 0.57 | 2.72 | 2.37 | 0.57 | 3.44 |
| Model3 | 7.13 | 0.64 | 8.10 | 7.00 | 0.57 | 7.94 |
| Model4 | 8.78 | 0.52 | 9.63 | 8.83 | 0.52 | 9.70 |
| | Verification temperature at 35℃ | | | | | |
| | MAPE (%) | STD | maxMAPE (%) | MAPE (%) | STD | maxMAPE (%) |
| Ours | 2.11 | 0.73 | 3.37 | 2.52 | 0.80 | 3.68 |
| Model1 | 87.99 | 14.68 | 107.38 | 89.78 | 11.76 | 107.01 |
| Model2 | 2.54 | 0.19 | 3.57 | 2.60 | 0.19 | 3.80 |
| Model3 | 11.56 | 0.59 | 11.97 | 11.04 | 0.79 | 11.44 |
| Model4 | 15.66 | 0.72 | 16.69 | 15.89 | 0.78 | 17.10 |
| | Verification temperature at 25℃ | | | | | |
| | MAPE (%) | STD | maxMAPE (%) | MAPE (%) | STD | maxMAPE (%) |
| Ours | 2.64 | 0.82 | 3.50 | 3.14 | 0.85 | 4.18 |
| Model1 | 73.66 | 21.18 | 108.56 | 78.94 | 25.2 | 138.87 |
| Model2 | 3.69 | 0.64 | 4.79 | 2.84 | 0.63 | 3.87 |
| Model3 | 9.51 | 0.91 | 13.41 | 11.78 | 0.82 | 12.76 |
| Model4 | 16.83 | 0.53 | 17.58 | 17.22 | 0.53 | 18.02 |

In an ultra-early verification setting, prioritizing time over accuracy and utilizing the first 50 cycles, our model achieves average MAPEs of 4.9% across 25, 35, and 45℃, outperforming benchmarks under similar data limitations (Table 2. 2). The MAPE (in %) and STD refer to the averaged MAPE and STD across batteries at 25, 35, and 45℃. The maxMAPE (in %) refers to the maximum MAPE at a selected temperature. The detailed benchmark model settings can be found in Appendix C. These findings affirm the viability of the proposed method in real-world verification contexts, offering adaptability in multi- and uni-source domain applications, and providing valuable insights for target domain evaluation with constrained data resources [63, 67].

## 2.5 Physics Interpretation

We use Shapley Additive Global Importance (SAGE) to quantify the feature importance in the battery degradation trajectory model. For the selected feature, we use a window length of 20 cycles to calculate the SAGE and slide the window in the entire battery lifetime. For the $i$th window $Win_i$, the feature importance is calculated as:

$$SAGE_{Win_i} = SAGE(\boldsymbol{X}_{Win_i}, Y_{Win_i}) \tag{2.15}$$

where $SAGE_{Win_i\,(1\times N)}$ is a vector containing SAGE values for $N$ features in window $Win_i$. $\boldsymbol{X}_{Win_i\,(20\times N)}$ and $Y_{Win_i(20\times 1)}$ are input matrix and output vector of the degradation trajectory prediction model in window $Win_i$, respectively. The correlation between two chemical processes in window $Win_i$ is defined as their 2nd-order Wasserstein distance. $SAGE$ is a function to calculate the feature importance using the mean squared error loss, which is calculated as:

$$SAGE = \frac{1}{d}\sum_{S\subseteq D\setminus\{i\}} \binom{d-1}{|S|}^{-1} E[Var(E\,[Y \mid X_S, X_i] \mid X_S)] \tag{2.16}$$

where, $Y$ is the output of the degradation trajectory prediction model, $X_S \equiv \{X_i \mid i \in S\}$ are subsets of features for different $S \subseteq D$, where $D$ is the set of all features and $D \equiv \{1, \ldots, d\}$. $\binom{d-1}{|S|}$ equals to combination numbers of features. $SAGE$ is a weighted average of conditional mutual information, which measures the reduction of uncertainty in output $Y$ given the inclusion of feature $X_i$ in all subsets $X_S$. The summation is for all possible feature subsets exclusive of feature $X_i$, thus it exhaustively calculates the importance of feature $X_i$ within each subset.

The average of $SAGE$ in all windows, i.e., across the entire lifetime, is defined as:

$$SAGE_{avg} = \frac{1}{w}\sum_{i=0}^{w} SAGE_{Win_i} \tag{2.17}$$

where, $w = \lceil C/20 \rceil$ is the round-up number of windows.

### 2.5.1 Statistical Feature Importance

We distinguish between kinetics and thermodynamics based on current stage densities, positing that machine learning-derived insights on thermodynamic loss enhance predictive accuracy at a single temperature while kinetic insights facilitate temperature adaptability. Group-wise analyses in Figure 2. 16a and Figure 2. 16b reveal a notable rise in the importance of capacity features in low-current areas versus their reduced significance at high currents, correlating with the observation that thermodynamic losses, not kinetic, predominantly affect degradation with a 79% share. Regarding other features, we note that the temperature impact on verification becomes negligible, indicating that physics-informed machine learning neutralizes temperature influence on predictive performance. Among these features, lumped resistance (RL), ohmic resistance (RO), and polarization speed (Vg) are prioritized for their contribution to verification accuracy. The contribution of Vg, influenced by SOC region sensitivities, is minimized due to its indirect relation to polarization resistance. RL and RO are more significant, incorporating concentration and ohmic polarization components.

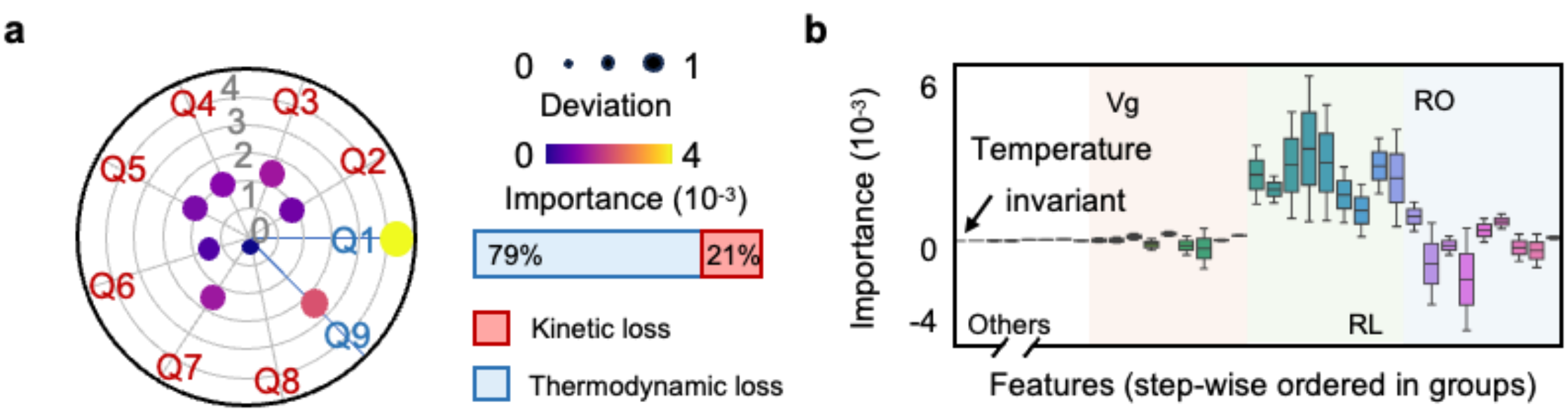

Figure 2. 16 Feature importance analysis.

However, the challenges of prototype verification are dualistic since a satisfactory explanation of dominating loss types does not guarantee a good verification, rather, it also depends on temperature adaptability associated with kinetic behaviors, see Table 2. 3.

Table 2. 3 Early verification model by feature selection.

| | MAPE (%) | | |
|---|---|---|---|
| Verification temperature | 25℃ | 35℃ | 45℃ |
| Thermodynamic loss(Q1+Q9) | 11.05 | 3.59 | 3.27 |
| Kinetic loss(Q2) | 5.32 | 1.10 | 1.12 |

We use Q1+Q9 and Q2 to present thermodynamic and kinetic loss, respectively. The experimental setting is the single-source domain adaptation (only guiding samples from 55℃ are available). 200 cycles of early data from the target domain are used. This is attributable to temperature primarily influencing kinetics, underlined by the Arrhenius law. Our model, predicated on this principle, reveals the expected diminished adaptability of thermodynamic responses to temperature changes. Thus, achieving a balanced verification requires weighing thermodynamic explanation capability against kinetic adaptability, supported by the experimental results shown in Table 2. 4.

Table 2. 4 Dualistic challenges of prediction capability and adaptability.

| | MAPE (%) | | | |
|---|---|---|---|---|
| Verification temperature | 25℃ | 35℃ | 45℃ | 55℃ |
| Exp 1 (Q1) | 0.43 | 0.29 | 0.47 | 0.67 |
| Exp 2 (Q2) | 0.44 | 0.39 | 0.48 | 0.68 |
| Exp 3 (Vg1) | 0.50 | 0.37 | 0.39 | 0.73 |
| Exp 4 (Vg2) | 1.02 | 0.62 | 0.83 | 0.93 |
| Exp 5 (Q1) | 2.70 | 1.35 | 1.88 | - |
| Exp 6 (Q2) | 2.65 | 1.11 | 0.91 | - |
| Exp 7 (Vg1) | 4.70 | 1.53 | 1.88 | - |
| Exp 8 (Vg2) | 2.84 | 1.60 | 1.55 | - |

We have demonstrated the dualistic challenges of early validation of battery prototypes due to a combination of prediction capability and transferability performance [65]. Therefore, we are motivated to find features that have both prediction capability and transferability performance. On one hand, the kinetic processes taking into account the influence of temperature shift are expressed by the high-current phase of our taxonomy framework. We need to find out features expressing dynamics that have better prediction capability. On the other hand, taking into account thermodynamic processes that are not affected by temperature and are expressed by the characteristics of the small current phase of our taxonomy framework, we need to find features expressing thermodynamics that have better transferability performance. Note that both prediction capability and transferability are not independent of each other, we interpret transferability (experiments designed with domain adaptation) by analyzing the relative reduction of the verification error with the identical feature input. For instance, when evaluating the transferability of feature, we first evaluate the single domain prediction error, noted $e_1$; we then evaluate the domain adaption prediction error, noted $e_2$. We take $e = |e_1 - e_2|$ as an evaluation metric. The smaller $e$, the better the transferability of the feature. Prediction capability is evaluated by $e_1$. The smaller $e_1$, the better the prediction capability of the feature.

### 2.5.2 Physics-Based Simulation

We utilize the finite element analysis (FEA) to elucidate complex physical processes that occur internally during degradation. The FEA methods are presented in APPENDIX D. Figure 2. 17a delineates the degradation patterns into three principal phases: initial SEI layer formation, the subsequent thickening, and lithium plating, aligning with the FEA [180, 181]. Contrary to bottom-up approaches that trace macro performance back to specific mechanisms, the proposed method employs a data-driven strategy to decouple loss types by correlating observable electrical signals with underlying thermodynamic and kinetic degradation processes, as illustrated in Figure 2. 17b.

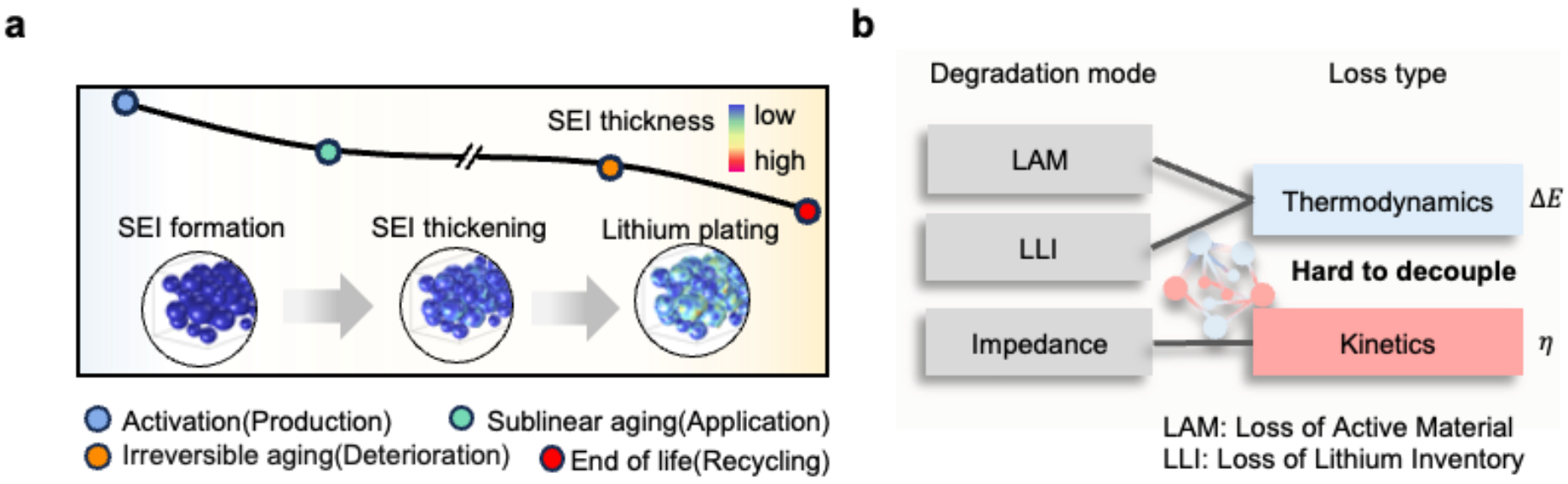

Figure 2. 17 Battery degradation mechanisms.

In Figure 2. 18a, we discern the interplay between concentration polarization and thermodynamic loss over a lifetime, observing an initial oscillation due to activation in early cycles that stabilizes into a clear dominance of thermodynamic loss and concentration polarization in the post-activation stage. Figure 2. 18b explores the correlation between these two degradation patterns across four temperatures, revealing consistent patterns that initial degradation shows a pronounced dip (marked by circle symbols), which becomes more profound and delayed at lower temperatures. This pattern aligns with activation processes including SEI layer formation and electrode structural change, impacting the primary degradation correlation. At lower temperatures, a temporary capacity restoration occurs, leading to a weaker correlation between concentration polarization and thermodynamic loss. At higher temperatures, quicker activation leads to a less dip, transitioning into a phase of predictable aging marked by SEI growth, increased LLI, and thus increased impedance. As batteries progress to the later stages of their lifetime, another critical point (marked by square symbols) signifies a shift towards irreversible degradation, characterized by significant LAM and the accelerated degradation processes. This phase sees a mixture of degradation mechanisms, leading to a notable decline in the correlation between concentration polarization and thermodynamic loss, with this shift manifesting earlier at higher temperatures.

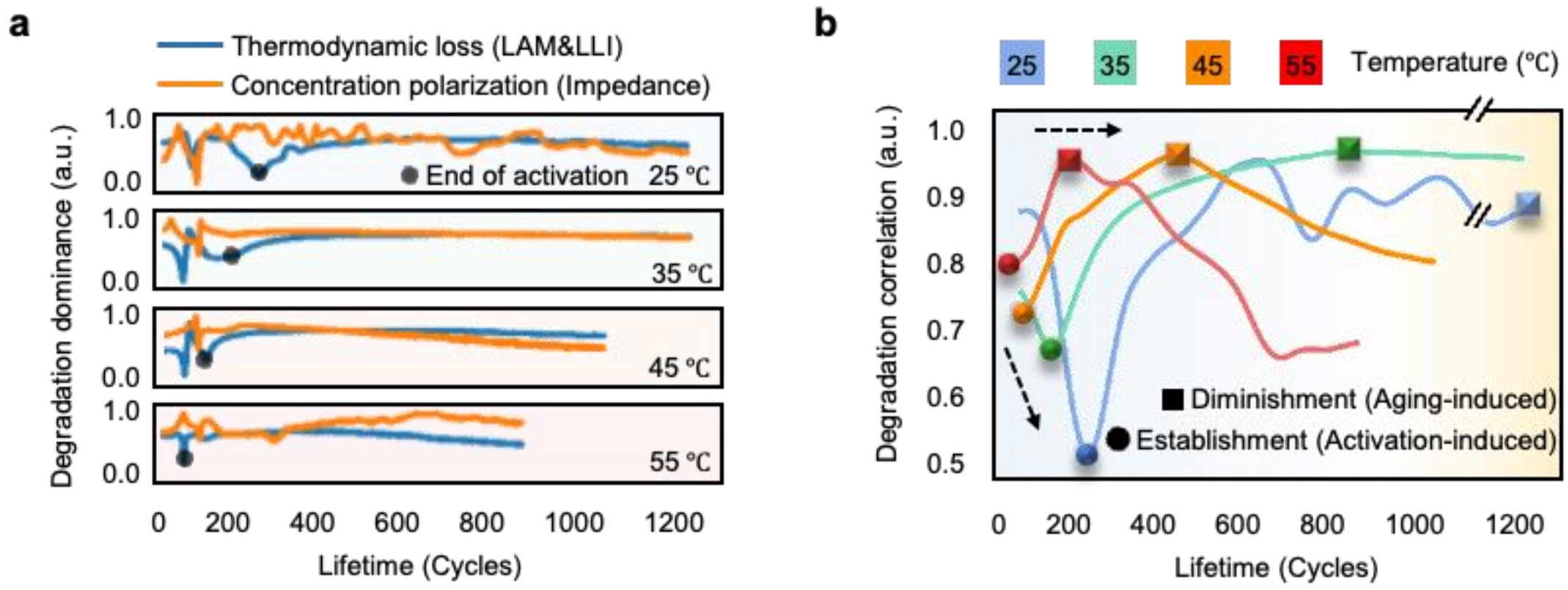

Figure 2. 18 Degradation dominance and correlation.

Analyzing quantified SAGE across early, middle, and late phases (APPENDIX E) reveals distinct aging behaviors at different lifecycle stages. Initially, capacity features fluctuate slightly, but as the battery ages and internal conditions worsen, changes in capacity and resistance features become evident. We note that such degradation patterns, elucidated through machine learning, correlate precisely with statistical predictions.

Further, incremental capacity analysis of discharge curves (Figure 2. 19a) confirms the existence of severe LLI and LAM, evidenced by reduced peak intensity in low SOC areas (indicating LAM at the anode) and peak shifts (signifying LLI)[182]. This analysis underscores the intensification of battery degradation in later lifetime stages, notably with severe LAM under the high-temperature accelerated aging test. Thermodynamic loss, averaging 85% of total degradation across temperatures, closely matches our 79% estimation for thermodynamic loss from machine learning interpretation (Figure 2. 19b). Note that this loss analysis is for the unit SOC at 25, 35, 45, and 55℃, respectively. Proportion comparison of thermodynamic (85%) and kinetic (15%) loss types, averaged over all temperatures. Normalized polarization types, i.e., concentration and other (ohmic and electrochemical) polarization, at 25, 35, 45, and 55℃, respectively. The proportion of concentration (82%) and other (18%) polarization averaged over all temperatures. The machine learning insight, i.e., the contribution of concentration polarization (74%) is indicated in Figure 2. 19c.

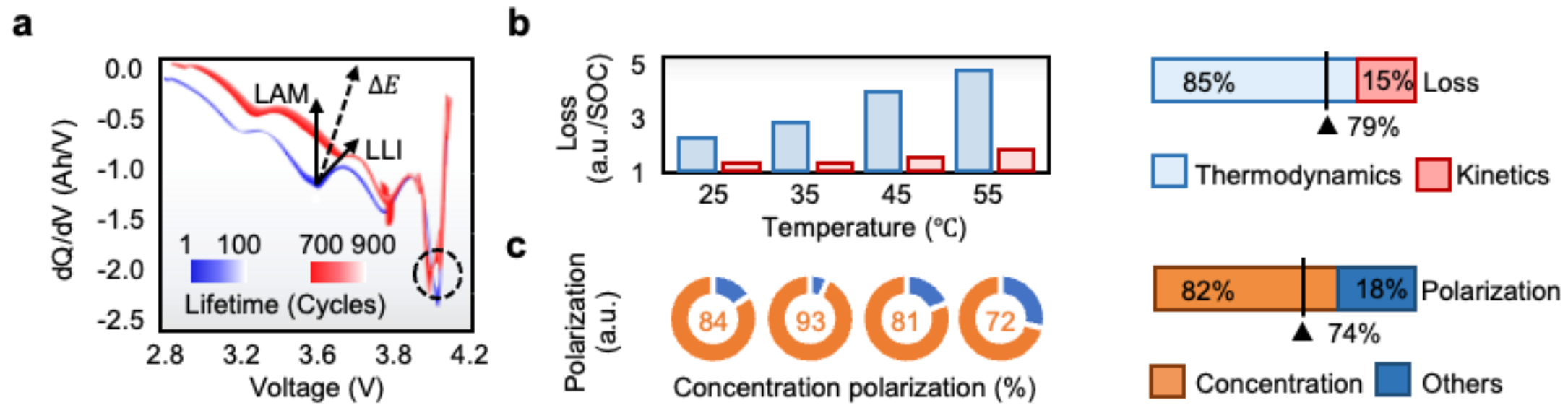

Figure 2. 19 Degradation mechanisms from machine learning.

## 2.6 Waste Management Application

The identification of critical points in Figure 2. 18b, marked by square symbols, has practical implications, especially highlighting the transition from sublinear to accelerated aging phases where degradation mechanisms notably shift. Such distinction aids in the development of nuanced recycling strategies for defective battery prototypes, suggesting pre-critical point lithium replenishment and post-critical point electrode repair as targeted recycling approaches [183]. This methodology offers a compelling use case for recycling materials from defective prototypes, employing non-destructive characterization to optimize recycling without compromising the integrity of functional prototypes. This approach streamlines recycling, ensuring efficient resource utilization while mitigating the environmental footprint of battery production, especially before putting prototype batteries, vulnerable to technical routine maturities, into massive production. Figure 2. 20 presents a forward-looking assessment of refined direct recycling and other conventional (direct, hydro-, and pyro-) recycling methods, concerning the economic viability and environmental impact. The detailed evaluation method is presented in APPENDIX F. This delineates the core idea of the physics-inspired characterization by enabling a non-destructive diagnosis of the defective prototypes in early stages, advising curtailed scrap material recycling, where the key difference is non-destructive knowledge of the internal structural information for curtailed recycling routine selection from (direct, hydro-, and pyro-) recycling methods.

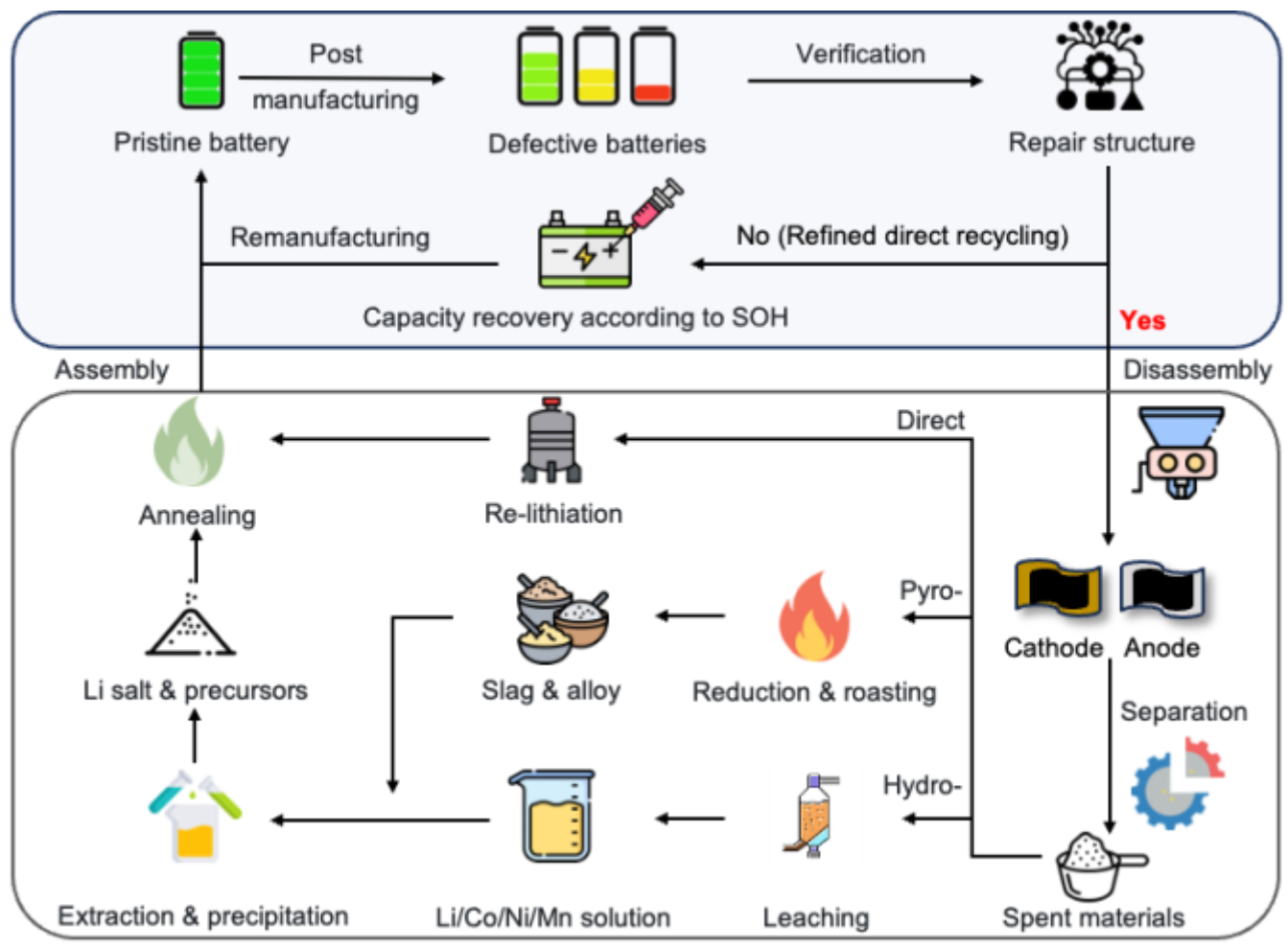

Figure 2. 20 High-level comparison of different recycling methods.

Figure 2. 21a shows that the refined direct recycling yields profit across all SOH levels, which can be rationalized by the saved time and cost of materials after the disassembly step. Other methods, especially hydrometallurgy recycling, only become profitable with higher SOH. While the advantage of refined direct recycling in Fig. 21b marginally decreases with increasing SOH compared with other recycling methods due to the increased price of recycling products using conventional methods, the superiority of refined direct recycling is stable thanks to the simplified process and affordable lithium supplement price. Note that the result in Figure 2. 21a and Figure 2. 21b are for both Nickel Manganese Cobalt (NMC811) and Lithium Iron Phosphate (LFP) batteries, considering battery SOH.

Here we amplify the implication of refined direct recycling by incorporating the Transport Impact Model (TIM), considering the interaction between battery prototype production, the elasticity of vehicle sales, the penetration rate of EVs, and technological advancements. The TIM was adopted to forecast the amount of rejected (defective) manufacturing-stage LIB prototypes in China [184-186]. The TIM gathers data on LIB installations and incorporates a range of factors to generate its forecasts. These factors include the growth of the Gross Domestic Product (GDP), the elasticity of vehicle sales, the penetration rate of EVs, and technological advancements. Utilizing this approach, TIM can predict the annual production and retirement of various kinds of batteries spanning from 2020 to 2060. For the years 2023 and 2030, TIM has estimated the scrap rates to be 7.67% and 4.34% of battery production, respectively [187]. These estimates are based on an exponential function fitted to historical and projected data, providing a methodological foundation for deriving future scrap rates used in this study. Through this analysis, TIM offers valuable insights into the lifecycle of LIBs in the Chinese market, enabling stakeholders to prepare for the future dynamics of battery recycling and disposal. Specifically, the total production, scrap rate, and scrap scale of new batteries (including NMC and LFP) from 2020 to 2060, were derived from the TIM [184-186]

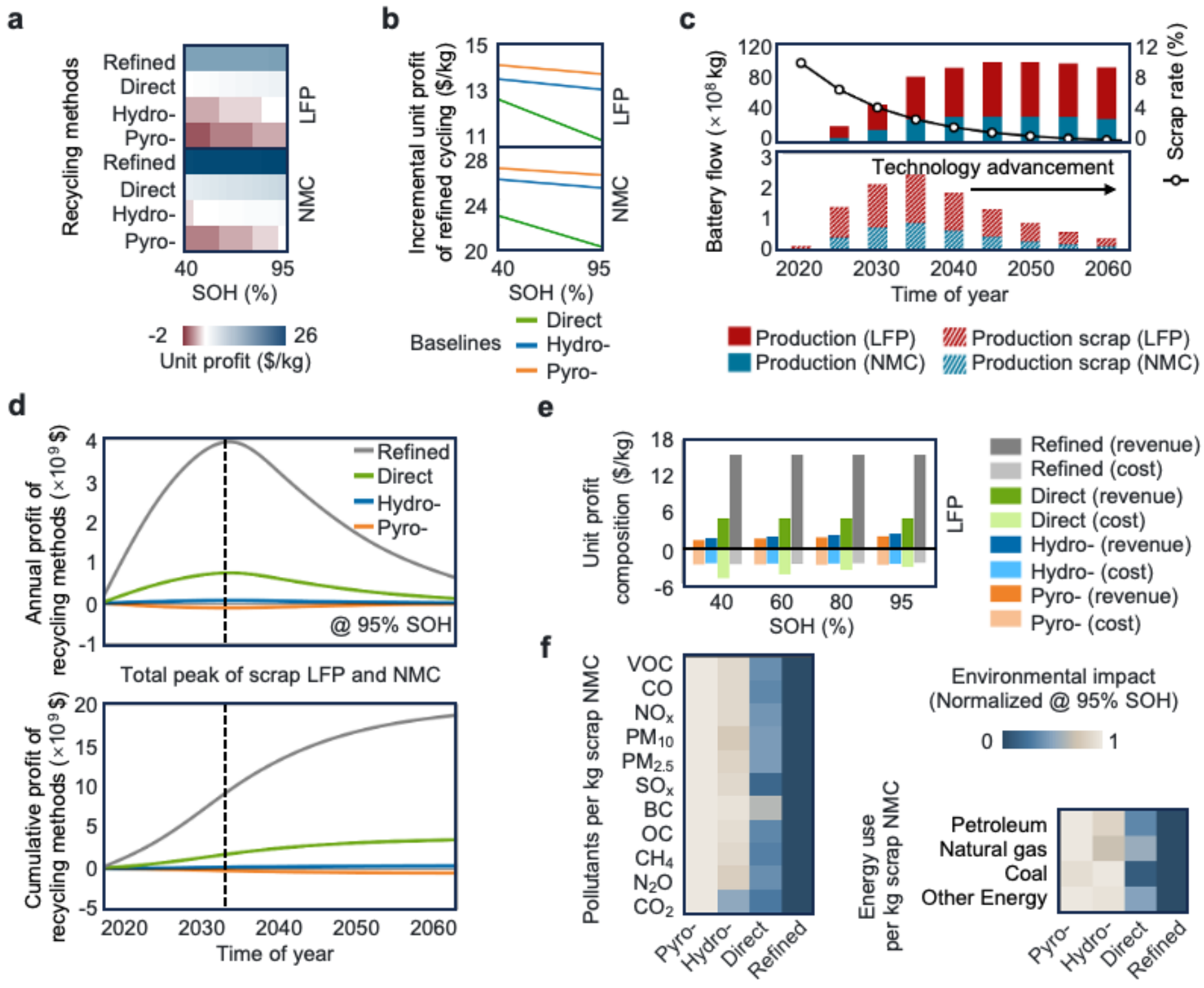

Figure 2. 21 Sustainability evaluation.

Figure 2. 21c forecasts battery production scale, scrap rate, and scrap scale over time from 2020 to 2060, showing a surge in scarp battery peaking circa 2035, reaching 230 million kg followed by a decline as advancements in production technology dampen the scrap rate to 0.38% by 2060. In this context, Figure 2. 21d shows the total projected profit over time by recycling defective prototype LFP and NMC battery at 95% SOH, aligning with the zenith of scrap amount depicted in Figure 2. 21c. The incremental profit anticipates a surge before 2035 and a slowed incremental speed, thereafter, demonstrating an up to 19.76 billion USD scrap recycling market by 2060. We emphasize such a market can be notably larger due to the material diversities in next-generation prototype batteries and the stubbornly higher scrap rate than our estimation in the R&D stage. Figure 2. 21e depicts the unit profit composition at different SOH for LFP prototype batteries, demonstrating consistently higher revenues (15.64$/kg) and lower costs (2.37-2.63$/kg) stemming from its procedural efficiency by inflicting no structure repair requirements, extra treatment reagents. In Figure 2. 21f, the unit environmental and energy impact of recycling slightly degraded NMC (95% SOH) prototype batteries, including Volatile Organic Compounds (VOC), Carbon Monoxide (CO), Nitrogen Oxides ($NO_x$), PM10, PM2.5, Sulfur Dioxide ($SO_x$), Black Carbon (BC), Organic Carbon (OC), Methane ($CH_4$), Nitrous oxide ($N_2O$), and Greenhouse Gases ($CO_2$) was evaluated. Concerning the environmental and energy impact, the refined direct recycling for NMC prototype batteries exhibits superior performance in all environmental pollutants and energy use, underscoring the sustainability potentials of the non-destructive characterization method by enabling scrap material recycling of defective prototype batteries before massive production.

## 2.7 Summary

In summary, our physics-informed machine learning approach enhances prototype verification, offering rapid assessment for both functional and defective prototypes, thus streamlining product delivery, sustainable defective prototype recycling and battery material research. Our model predicts entire battery lifetime trajectories across various temperatures with a modest 4.9% error using just 4% of total lifetime data (50 cycles), achieving verification 25 times faster than traditional methods. This efficiency not only cuts prototype verification costs but also reduces the carbon footprint, contributing to lifecycle sustainability. The approach leverages IMVs and machine-learned insights into degradation patterns (thermodynamic and kinetic) for an accurate, physics-based understanding of battery internal evolution. It offers a novel non-destructive solution for both proactive production scaling and recycling for production scraps, overcoming limitations of mechanism-centric and post-mortem verification. Overall, our findings illuminate the potential of leveraging physics-informed machine learning to predict complex system evolutions like battery degradation, enabling accessible insights into latent system behaviors.

We prospect that our findings are material-agnostic, thus being widely applicable for promoting the battery lifecycle sustainability, inclusive of prototype R&D in the manufacturing stage, moreover, primary applications (EVs), secondary applications (reuse), and recycling (for both in-production scrap materials and retired batteries). Future research should expand to include a wider range of material types and charging protocols to generalize findings, potentially informing the R&D and recycling of next-generation battery materials, particularly in pioneering post-lithium chemistries. Addressing the open-ended challenge of electrochemical-level separation of battery degradation patterns could further consolidate the statistical evidence presented. While this study primarily addresses the verification of the as-manufactured batteries, the potential applications of physics-informed machine learning can extend to complex predictive challenges, including autonomous material discovery, climate modeling, and physical system state estimation without explicitly known governing equations. Ultimately, this chapter demonstrates the promise of advanced online system management using a non-destructive characterization methodology, paving the way for enhanced sustainability in future battery technology R&D by studying complex system evolutions with physics-informed machine learning and non-destructive characterization methodologies.

# CHAPTER 3 RAPID RESIDUAL ASSESSMENT

## 3.1 Overview

In this chapter, we perform retired battery residual assessment, i.e., SOH estimation, using the pulse voltage response data generated by an attentional variational autoencoder at a saved physical measurement time and cost. The high-level process of general pretreatment steps, generative model training, and practical model deployment steps are illustrated in Figure 3. 1. The research idea is to use generative learning to exploit already measured data, for the pulse voltage responses exploration in continuous retirement conditions, i.e., SOC distribution in retired battery collection stage. Downstream SOH estimation is implemented without physically measured pulse response data, saving pretreatment electricity costs and carbon emissions otherwise required from conditioning SOC to the anticipated level. Compared with the ampere-hour integral method and the machine-learning pulse injection, generative learning underscores a faster generation speed and accurate estimation. With generated data, a simple regressor successfully estimates the SOH with low error under unseen or inaccessible SOC retirement conditions in the database, a challenging out-of-distribution (OOD) issue. The OOD issue is resolved by learning the inherited physical pattern between the SOC condition and voltage response induced by the rapid pulse. We demonstrate the generalizability of our method by verifying through 2700 physically measured samples, spanning 3 cathode material types, 3 physical formats, 4 capacity designs, and 4 historical usages (1 laboratory accelerated aging, 1 pure battery EV driving, and 2 hybrid EV driving). Generative learning suits interpolation and extrapolation by integrating prior knowledge of retirement conditions into latent space scaling of the model parameters, saving measurement time, electricity costs, and consequent carbon emissions while not compromising SOH estimation accuracy.

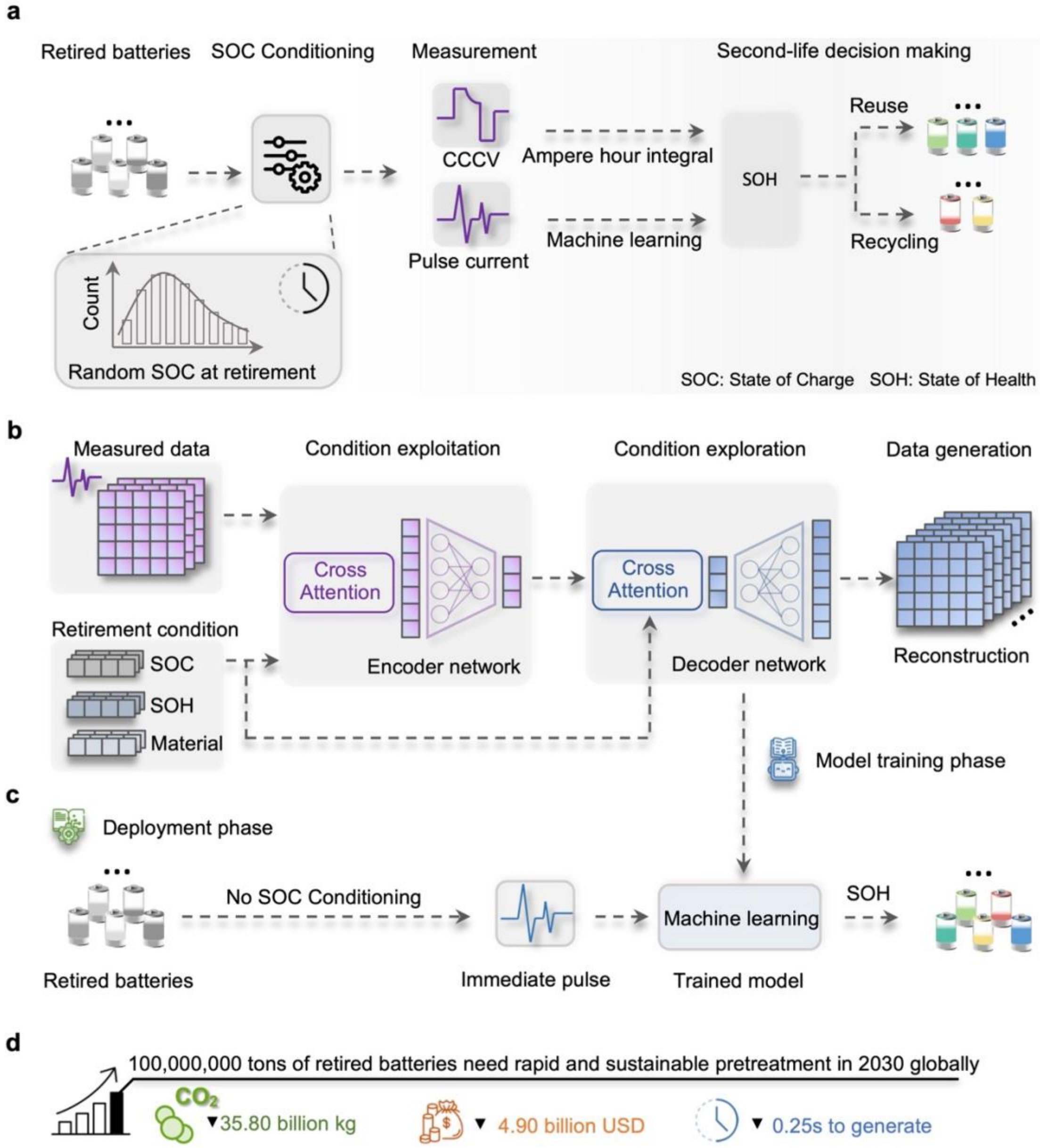

Figure 3. 1 Conceptualization of the residual assessment of retired batteries.

A conservative global case study of battery retirement in 2030 highlights the significance of rapid and sustainable battery residual estimation in pyrometallurgical, hydrometallurgical, and direct recycling circumstances. Through technical-economic analysis, generative learning-assisted SOH estimation could save $4.9 billion electricity

costs and 35.8 billion kg $CO_2$ emissions by 2030 worldwide. We discuss the model interpretability, recycling pretreatment implications, and broader aspects of future smart recycling pretreatment directions integrated with machine learning.

## 3.2 Random Retirement Conditions

The retired batteries are subject to random retirement conditions, which is uniform SOC distribution upon the collection. In Figure 3. 2, two generation scenarios, i.e., the interpolation and extrapolation are illustrated, where batteries have to be tested under cases of no data is available. This motivates me the adoption of generative methods to create data points in unobserved SOC conditions.

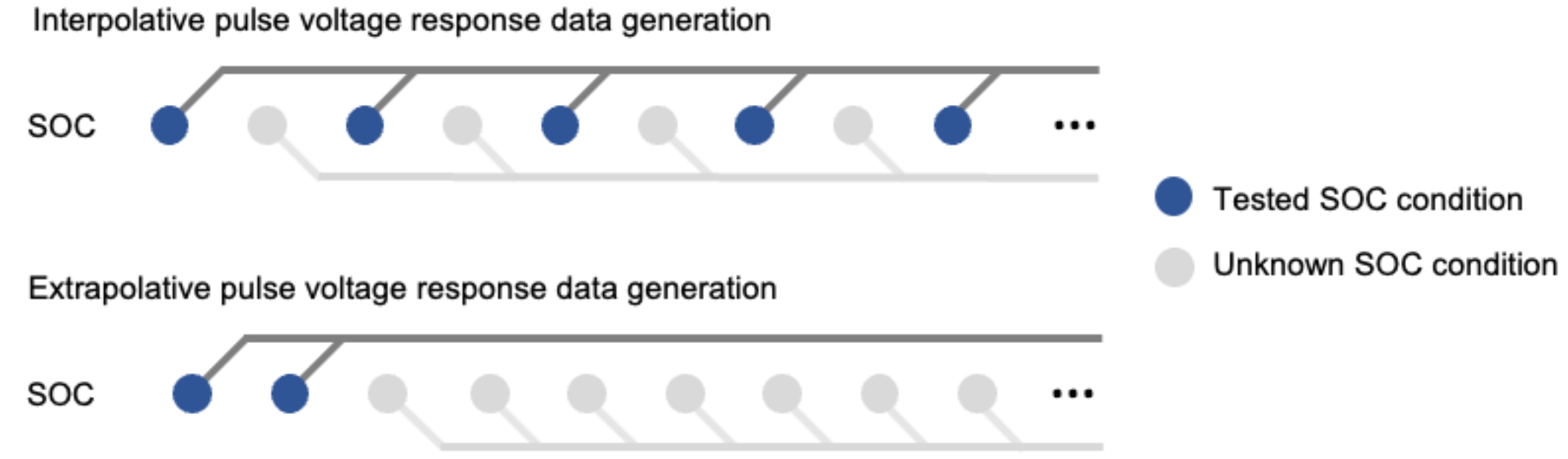

Figure 3. 2 The data interpolation and extrapolation.

## 3.3 Data Collection and Processing

Data scarcity and heterogeneity are major challenges in data-driven battery diagnosis and prognosis, especially for retired batteries. We collected 2700 samples from 270 physically retired batteries across wide SOC levels, i.e., from 5% to 50% with a 5% grain. In Figure 3. 3, the dataset covers 3 prevailing cathode material types, including 119 physically collected NMC (nickel manganese cobalt oxide), 56 LFP (lithium iron phosphate), and 95 LMO (lithium manganese oxide) retired batteries, spanning 3 physical formats (cylindric, pouch, prismatic) and 4 historical usage patterns (1 laboratory accelerated test, 3 EV usages, including 1 purely electrified power-train mode and 2 hybrid power-train modes). We stress that the inclusion of highly heterogeneous testing samples makes our dataset the largest rapid pulse-based dataset for SOH estimation of retired batteries up to date, facilitating the model generalizability beyond the physically measured datasets.

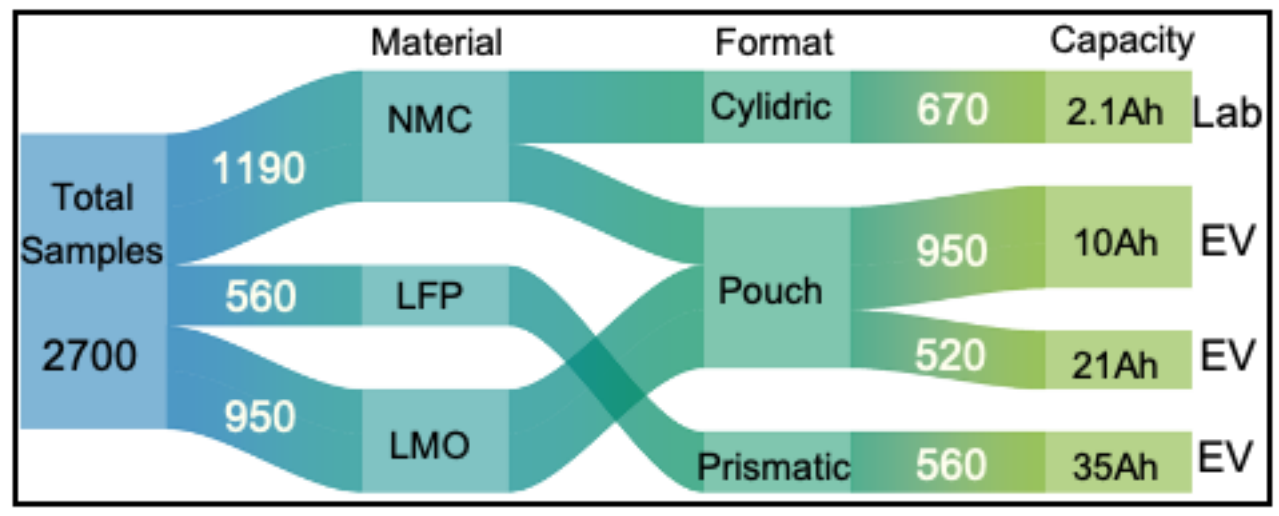

Dataset considering heterogeneities for retired batteries

Figure 3. 3 The Sankey plot for retired batteries distribution.

### 3.3.1 Multi-Dimensional Pulse Test Protocol

In contrast to regular constant current constant voltage (CCCV) testing, the pulse test inflicts no lengthy test time and extra damage to the retired batteries. Therefore, we use the CCCV results as the SOH benchmarking capacity values. Pulse test data are used as training data for the generative learning model. Even though the battery cathode material type varies, we perform a standardized and accessible feature engineering, ensuring compatibility in practical use. Here we elaborate on the battery capacity calibration and pulse injection experiment details. We use the constant voltage constant current (CCCV) method as the gold standard for determining the capacity of retired batteries. Even considering the different initial SOC distributions of retired batteries, we use a unified method of first discharging and then charging to determine the capacity of retired batteries. The experiment is divided into the following three steps: capacity calibration, SOC conditioning, and pulse injection.

Step 1: Capacity calibration. First, the retired batteries are discharged to lower cut-off voltage using a 1C constant current. Second, they are charged to the upper cut-off voltage using a 1C constant current, then charged using constant voltage until the current drops to 0.05 C. Third, they are then discharged to the lower cut-off voltage using a 1C constant current. Where C is the rated capacity of the retired battery, and the relevant information is included on the battery nameplate or serial number. We use the actual discharge capacity as the calibrated battery capacity and then let the battery stand for 20 minutes before the pulse injection. The cut-off voltage is in Table 3. 1.

Table 3. 1 The cut-off voltage setting.

| Material | Charging cut-off voltage(V) | Discharging cut-off voltage(V) |
| --- | --- | --- |
| LMO,10Ah | 4.2 | 2.0 |
| NMC,21Ah | 4.2 | 2.7 |
| LFP,35Ah | 3.65 | 2.5 |

where, LMO is lithium manganese oxide, NMC is nickel manganese cobalt oxide, and LFP is lithium iron phosphate. The protection voltage setting of charge and discharge is listed in Table 3. 2.

Table 3. 2 The protection voltage setting.

| Material | Charging protection voltage(V) | Discharging protection voltage(V) |
| --- | --- | --- |
| LMO,10Ah | 4.3 | 1.95 |
| NMC,21Ah | 4.3 | 2.65 |
| LFP,35Ah | 3.7 | 2.45 |

Step 2: SOC conditioning. After completing the rest of the calibrated battery, SOC conditioning is performed to inject pulses at the desired SOC levels. We consider different pulse parameters, specifically pulse width, pulse intensity, and pulse direction (a discharge pulse or a charge pulse). The pulse parameters are detailed in Table 3. 3.

Table 3. 3 Pulse injection parameter setting.

| Pulse width $t_1$ | Pulse rest time $t_2$ | Pulse amplitude ($\pm$C) |
|---|---|---|
| 30ms | 450ms | 0.5-1-1.5-2-2.5 |
| 50ms | 750ms | 0.5-1-1.5-2-2.5 |
| 70ms | 1.05s | 0.5-1-1.5-2-2.5 |
| 100ms | 1.5s | 0.5-1-1.5-2-2.5 |
| 300ms | 4.5s | 0.5-1-1.5-2-2.5 |
| 500ms | 7.5s | 0.5-1-1.5-2-2.5 |
| 700ms | 10.5s | 0.5-1-1.5-2-2.5 |
| 1s | 15s | 0.5-1-1.5-2-2.5 |
| 3s | 45s | 0.5-1-1.5-2-2.5 |
| 5s | 75s | 0.5-1-1.5-2-2.5 |

Step 3: Pulse injection. We take one specific SOC conditioning step as an intuitive illustration. Charge the retired battery with a constant current of 1C for 3 minutes to 5% SOC, and then cut off the charging current. The battery is then left to stand for 10 minutes to rest, expecting the battery to return to a steady state in preparation for subsequent pulse injection. Then, perform multiple consecutive pulse injections with different pulse widths. The pulse width and pulse resting time are as shown in Table 3.3, that is, for each pulse width and resting time (each row of the table), we consecutively perform pulse injection with pulse amplitude being 0.5-1-1.5-2-2.5(C) in order, including positive and negative pulses. Note that positive and negative pulses alternate to cancel the equivalent energy injection. For instance, at the 30ms pulse width, we inject 0.5C positive current pulse, then let the battery rest for 450ms, and then inject 0.5C negative current pulse, then again let the battery rest for 450ms. Other remaining pulses with different amplitudes follow the rest of the previous pulse injections. When the last pulse and rest are finished, all the required experiments at 30ms pulse width are finished. Repetitive experiments are performed until the remaining pulse widths are exhausted, indicating all pulse injections at 5% SOC are completed. Then charge the retired battery with a constant current of 1C for *another* 3 minutes to 10% SOC, followed by the same procedure as explained above.

The range of SOC conditioning is determined by the state of the health of the retired battery. Specifically, the upper bound of the SOC conditioning region is the state of the health value of the retired battery under test. For instance, when the retired battery has an SOH of 0.5, then the SOC conditioning region will be 0.05 to 0.45, with a grain of 0.05. Then repeat Step 2 and Step 3 until the SOC conditioning regions are exhausted. In addition, we set voltage protection during pulse injection to ensure the safety of the experiment. The specific voltage protection parameters are consistent with those in Table 3.2. Only EV-retired batteries (LMO 10 Ah, NMC, 21 Ah, and LFP, 35Ah) were subject to the above test procedure. All tests are performed with BAT-NEEFLCT-05300-V010, NEBULA, Co, Ltd, and the air conditioner temperature is set at 25 ℃. The detailed experimental setup can be found in APPENDIX G.

The batteries subject to accelerated aging are NMC (nickel manganese cobalt oxide), cylindric 18 650 batteries with 2.1 Ah nominal capacity. In each accelerated aging cycle, all batteries were charged using the recommended current, specifically, 2C constant current followed by a constant voltage at 4.2 V with C/200 or 30-minute cut-off condition. The same constant discharge current is applied with a cut-off voltage of 3.0V. Note that the term C stands for charge (discharge) rate when a 1 hour of charge (discharge) is performed. After each accelerated aging cycle, the batteries rest for 30 minutes to reach a steady state, preparing for the pulse injection. Besides the accelerated aging cycle, pulse injection is performed. The SOC conditioning region is 0.05 to 0.50, with a grain of 0.05, for all batteries. The SOC conditioning current is 1C and the conditioning time is 3 minutes for each 0.05 SOC change. Note that for all conditioned SOC levels, only a pulse of 5s width is performed and the pulse amplitude is $\pm$0.5C, $\pm$1C, and $\pm$2C. The air conditioner temperature is set at 25 ℃. Note that positive and negative pulses alternate to cancel the equivalent energy injection. We use identical rest time, i.e., 25s, for all injected pulses and all batteries.

### 3.3.2 Feature Engineering of Pulse Voltage Response

Figure 3. 4 demonstrates the feature extraction after pulse current injection. For the first five injected pulses, the turning points, i.e., the points with zero second-order derivative of the voltage responses are recorded as features, resulting in 21-dimensional features. Despite pulse time at a 5-second level, we also perform sensitivity measurement on pulse width with dual aims to verify pulse robustness and further shorten test time, ranging from 30ms to 5s.

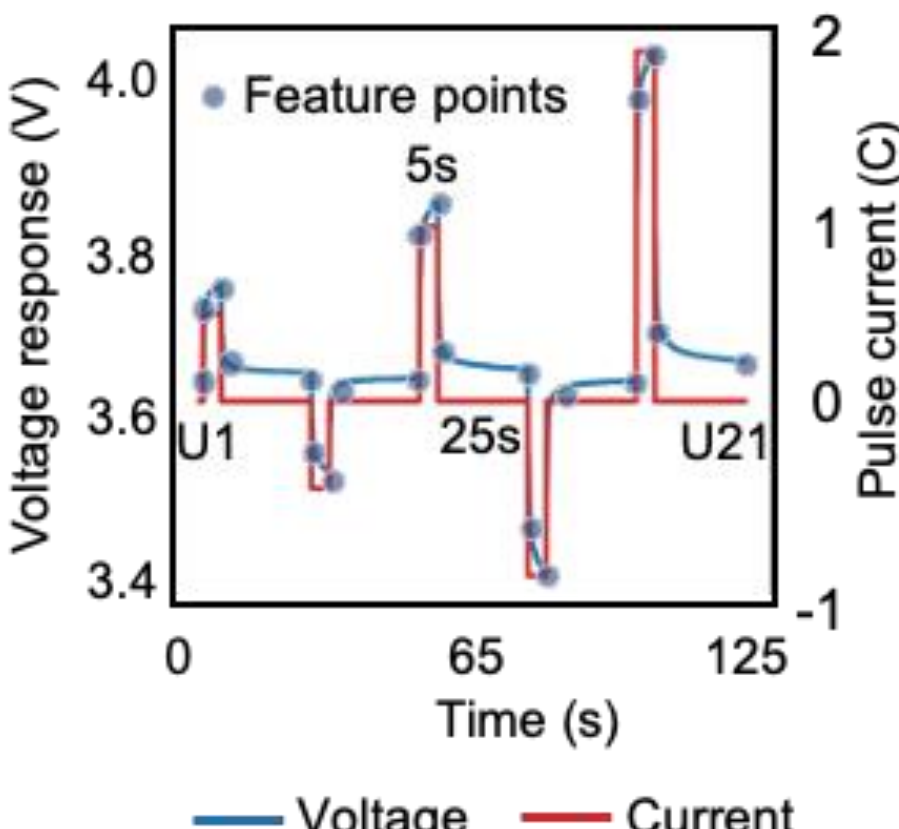

Figure 3. 4 Pulse current and the voltage response of the retired batteries.

The tested SOC region is under the assumption that retired batteries either exhibit low SOC or are subject to compulsory discharging for safe stationary warehouse storage requirements, thus the SOC upper limit is set at 50%. In Figure 3. 5, the first dimension of the extracted features U1 from accelerated aging batteries (NMC, 2.1Ah) and EV driving aging batteries (NMC, 21Ah) are illustrated. Despite differences in capacity designs and historical usages, the pulse voltage responses exhibit a consistent degradation pattern, so do other retired batteries for other feature dimensions.

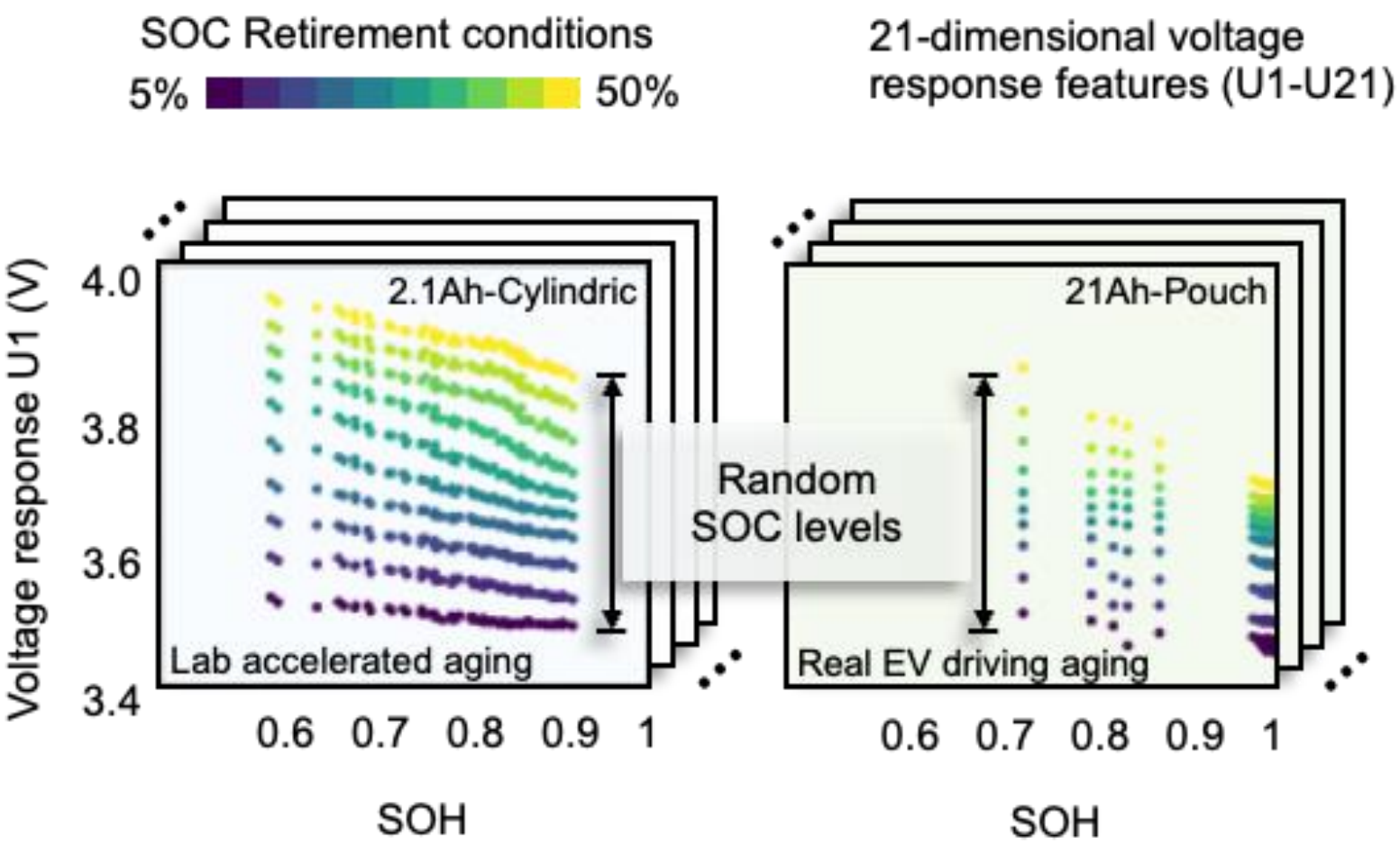

Figure 3. 5 Relationship between pulse voltage response and calibrated SOH.

## 3.4 The Generation Model

We first consider using already-measured pulse voltage response data to reconstruct themselves, rather than directly generating new data samples. The reconstruction means that the measured data are first compressed into a latent variable space while maintaining a learned structure from original statistical distributions in Figure 3. 6, and then decompressed to the original dimensions. In this chapter, the reconstruction is equivalent to supervising the encoder neural network model and decoder neural network model training phase. The core idea is to balance the exploitation of limited measured data and the exploration of extended data space using latent dependencies between retirement conditions and pulse voltage response data with cross-attention mechanisms.

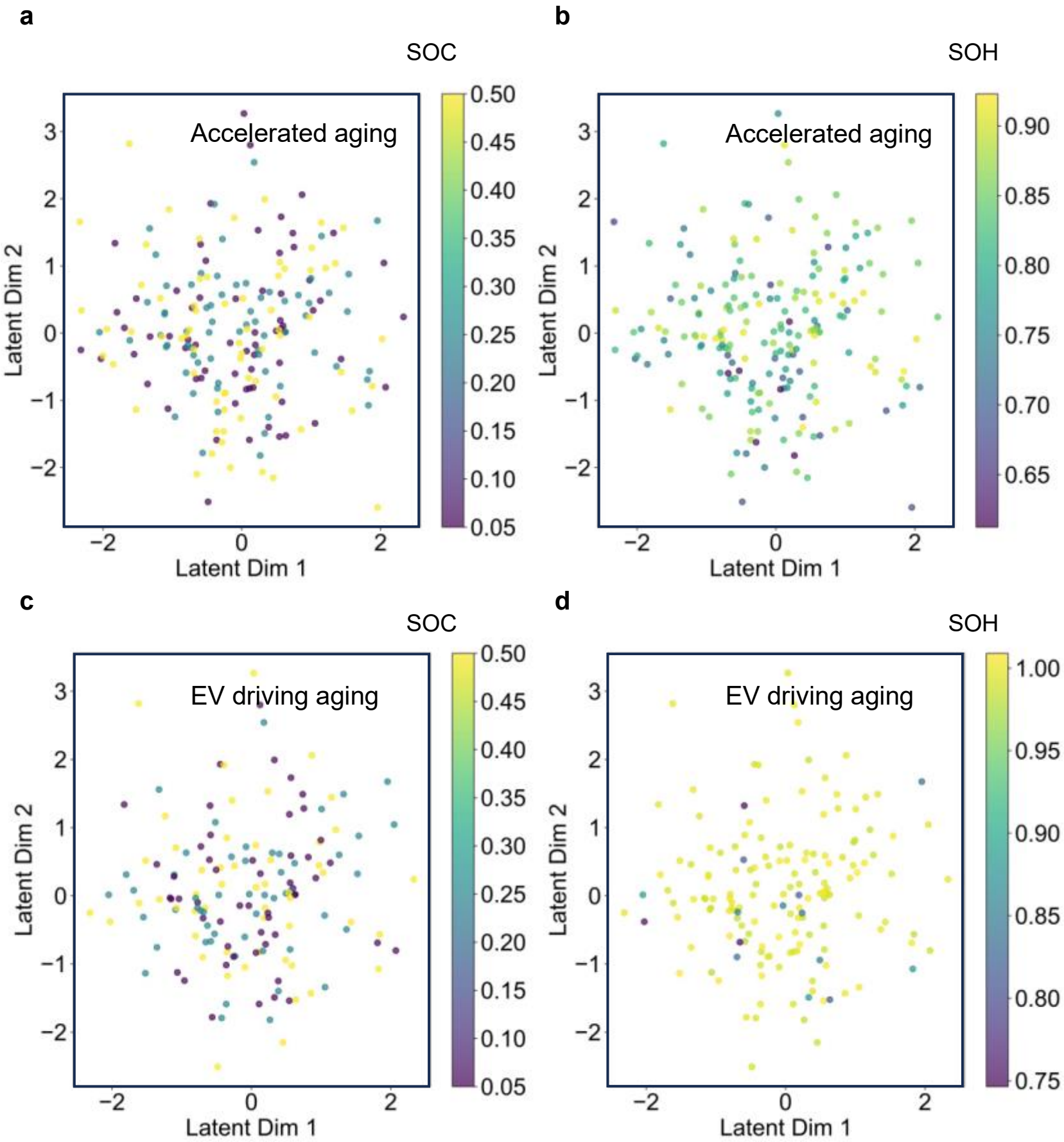

Figure 3. 6 The two-dimensional latent space of the encoder network.

Then measured data fused with conditional information are fed into the encoder neural network, obtaining latent variables containing retirement conditions. Compressed latent variables are decompressed by training a decoder neural network model and reconstructing the input data samples, also guided by the feed-forward retirement conditions. Another input of the decoder neural network is only random noise, subject to Gaussian assumption in Figure 3. 7, inflicting no physical experimental measurement otherwise involved with additional test time, energy consumption, and even safety hazards.

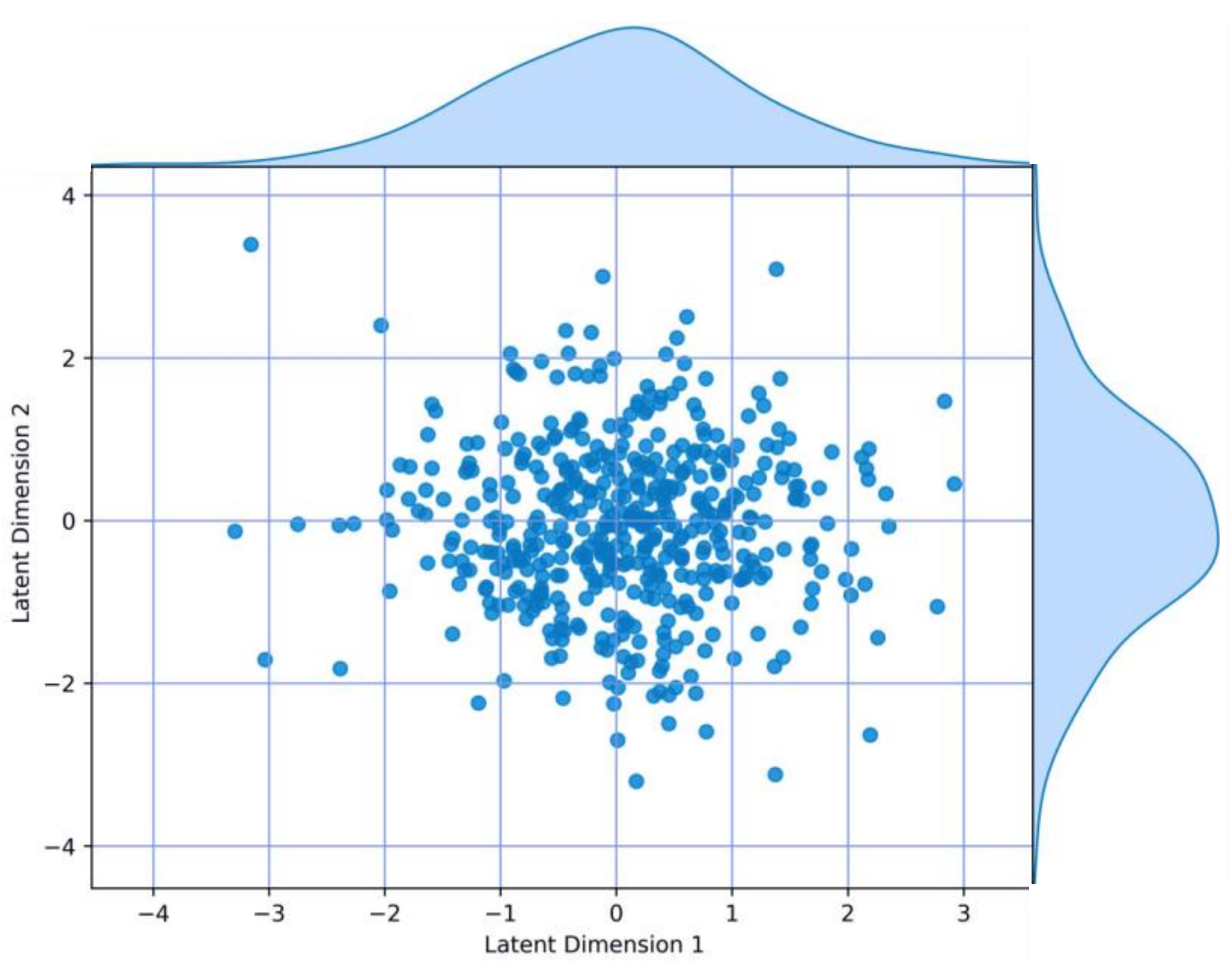

Figure 3. 7 The random Gaussian noise data to generate pulse voltage response data.

### 3.4.1 Cross-Attention Mechanism

Cross-attention in neural networks enables a model to focus on specific parts of one input, i.e., the query, based on the information in another input, i.e., the key and value. It is useful in scenarios where the relevance of certain features in one data stream depends on the additional information provided and has demonstrated successful applications in battery health diagnosis and prognosis [188-191]. In battery recycling pretreatment, retired batteries are under random retirement conditions, i.e., SOC distributions. From expert knowledge, the pulse voltage response exhibits considerable shift with SOC. Therefore, the cross-attention enables the exploration of conditional dependencies between pulse voltage responses and the SOC retirement conditions. The general formulation of the cross-attention mechanism is:

$$\text{Attention}\ (Q, K, V) = \text{softmax}\left(\frac{QK^T}{\sqrt{d_k}}\right)V \tag{3.1}$$

where, $Q, K, V$ represent the query, key, and value sequence, respectively. $d_k$ is the scaling factor, typically the dimension value of the key $K$. $T$ is the transpose operator. The softmax function normalizes the input vector so that the sum of the probabilities is 1, making the calculated attention a valid probability distribution. In the cross-attention, the softmax is used to calculate the weights representing the importance of different elements in the input sequence. Given a vector $\mathbf{v} = [v_1, v_2, \cdots, v_P]$ of real numbers, the softmax function for the $i$-th element of this vector is given by:

$$\text{softmax}\,(\mathbf{v})_i = \frac{e^{v_i}}{\sum_{p=1}^{P} e^{v_p}} \tag{3.2}$$

where, the denominator is the sum of the exponentials of all elements $v_i$ in the vector $\mathbf{v}$, $i = 1, \dots, P$. $P$ is the number of elements in the vector $\mathbf{v}$.

### 3.4.2 Encoder Neural Network with Cross-Attention

The encoder network in the variational autoencoder is designed to compress input data into a latent space. It starts by taking the 21-dimensional battery voltage response feature matrix $\mathbf{x} \in \mathbb{R}^{N\times 21}$ as main input and retirement condition matrix $\mathbf{cond} = [SOC, SOH] \in \mathbb{R}^{N\times 2}$ as input, where $N$ is the sample size. The condition input is first transformed into an embedding $\mathbf{C}$, belonging to a larger latent space with 64-dimension. The conditional embedding $\mathbf{C}$ is formulated as:

$$\mathbf{C} = \text{ReLU}\,(\mathbf{cond} \cdot \mathbf{W}_\mathbf{c}^\mathbf{T} + \mathbf{b}_\mathbf{c}) \tag{3.3}$$

where, $\mathbf{W}_\mathbf{c} \in \mathbb{R}^{64\times 2}, \mathbf{b}_\mathbf{c} \in \mathbb{R}^{\mathrm{N}\times 64}$ are the condition embedding neural network weighting matrix and bias matrix, respectively.

The main input matrix $\mathbf{x}$, representing pulse voltage response features, is also transformed into this 64-dimensional latent space $\mathbf{H}$:

$$\mathbf{H} = \text{ReLU}\left(\mathbf{x} \cdot \mathbf{W}_\mathbf{h}^\mathbf{T} + \mathbf{b}_\mathbf{h}\right) \tag{3.4}$$

where, $\mathbf{W}_\mathbf{h} \in \mathbb{R}^{64\times 21}, \mathbf{b}_\mathbf{h} \in \mathbb{R}^{\mathrm{N}\times 64}$ are the main input embedding neural network weighting matrix and bias matrix, respectively.

Both $\mathbf{H}$ and $\mathbf{C}$ are then integrated via a cross-attention, allowing the network to focus on the voltage response matrix $\mathbf{x}$ conditioned by the additional retirement condition information $\boldsymbol{cond}$:

$$\mathbf{AttenEncoder} = \text{Attention}\ (\mathbf{H}, \mathbf{C}, \mathbf{C}) \tag{3.5}$$

where, $\mathbf{H} \in \mathbb{R}^{N\times 64}$ and $\mathbf{C} \in \mathbb{R}^{N\times 64}$ are embeddings of pulse voltage response data and retirement condition data, respectively. $\mathbf{AttenEncoder} \in \mathbb{R}^{N\times 64}$ is the cross-attended output matrix from the voltage response embedding $\mathbf{H}$ and the retirement condition embedding $\mathbf{C}$.

$\mathbf{z_{mean}}$ and $\mathbf{z_{log_var}}$ constitute a Gaussian distribution for each $L = 2$ dimensional latent space:

$$\mathbf{z_{mean}} = \mathbf{AttenEncoder} \cdot \mathbf{W}^{\mathbf{T}}_{\mathbf{z_{mean}}} + \mathbf{b}_{\mathbf{z_{mean}}} \tag{3.6}$$

$$\mathbf{z_{log_var}} = \mathbf{AttenEncoder} \cdot \mathbf{W}^{\mathbf{T}}_{\mathbf{z_{log_var}}} + \mathbf{b}_{\mathbf{z_{log_var}}} \tag{3.7}$$

where, the $\mathbf{z_{mean}} \in \mathbb{R}^{N\times L}$, $\mathbf{z_{log_var}} \in \mathbb{R}^{N\times L}$ are the mean and logarithm of the variance of the Gaussian distributions, respectively. $\mathbf{W}_{\mathbf{z_{mean}}} \in \mathbb{R}^{L\times 64}$, $\mathbf{b}_{\mathbf{z_{mean}}} \in \mathbb{R}^{N\times L}$ are the trainable weighting matrix and bias matrix, respectively. $\mathbf{W}_{\mathbf{z_{log_var}}} \in \mathbb{R}^{L\times 64}$, $\mathbf{b}_{\mathbf{z_{log_var}}} \in \mathbb{R}^{N\times L}$ are the neural network weighting matrix and bias matrix for the latent space variance embedding, respectively.

### 3.4.3 Latent Space Scaling and Sampling

Certain retirement conditions, e.g., extreme SOH and SOC can be under-represented in the battery recycling pretreatment due to data scarcity or measurement budget. Specifically, the retired batteries exhibit concentrated SOH and SOC, leading to poor estimation performance when confronted with OOD batteries. Collected retired batteries are discharged lower than a certain voltage threshold due to the safety concerns of the warehouse storage requirement, resulting in an upper-limit SOC typically lower than 50%. Even if the explicit battery retirement conditions are still unknown, we can use this approximated prior knowledge to generate enough synthetic data to cover the actual retirement conditions. Given two data generation settings, namely, interpolation and extrapolation, we use different latent space scaling strategies. In the interpolation setting, the scaling matrix $\mathbf{T} \in \mathbb{R}^{N\times N}$ is an identity matrix $\mathbf{I} \in \mathbb{R}^{N\times N}$ assuming the encoder network and decoder network can learn inherited data structures without taking advantage of any prior knowledge. In the extrapolation setting, however, the assumption cannot be guaranteed due to the OOD issue, a general challenge of machine learning models. Here we use the means of training and testing SOC distributions to define the scaling matrix, then the latent space is scaled as:

$$\hat{\mathbf{z}}_{\mathbf{mean}} = \mathbf{T}_{\mathbf{mean}} \cdot \mathbf{z}_{\mathbf{mean}} \tag{3.8}$$

$$\hat{\mathbf{z}}_{\mathbf{log_var}} = \mathbf{T}_{\mathbf{log_var}} \cdot \mathbf{z}_{\mathbf{log_var}} \tag{3.9}$$

where, $\mathbf{T}_{\mathbf{mean}} \in \mathbb{R}^{N\times N}$ and $\mathbf{T}_{\mathbf{log_var}} \in \mathbb{R}^{N\times N}$ are the scaling matrices defined by the broadcasted mean, and variance ratio between the testing and training SOC distributions, respectively. We emphasize that the SOH distributions are irrelevant to such a scaling. This is because these identical SOH values could be seen as representing physically distinct batteries, i.e., they do not affect the scaling process. Thus, feeding the model with the same SOH values during training and reconstruction does not present an OOD problem. On the other hand, for the SOC dimension, our goal is to generate data under unseen SOC conditions, where physical tests cannot be exhausted.

### 3.4.4 Decoder Neural Network with Cross-Attention

The sampling step in the VAE is a bridge between the deterministic output of the encoder neural network and the stochastic nature of the scaled latent space. It allows the model to capture the hidden structure of the input data, specifically the pulse voltage response $\boldsymbol{x}$ and $\boldsymbol{cond}$ to explore similar data points. The sampling procedure can be formulated as:

$$\mathbf{z} = \mathbf{z}_{\mathbf{mean}} + e^{\frac{1}{2}\cdot \mathbf{z}_{\mathbf{log_var}}} \cdot \boldsymbol{\epsilon} \tag{3.10}$$

where, $\boldsymbol{\epsilon} \in \mathbb{R}^{\mathrm{N}\times\mathrm{L}}$, is a Gaussian noise vector sampled from $\boldsymbol{\epsilon} \sim \boldsymbol{\mathcal{N}}(\mathbf{0}, \mathbf{I})$. The exponential term $e^{\frac{1}{2}\cdot \mathbf{z}_{\mathbf{log_var}}}$ turns the log variance vector to a positive variance vector. $\mathbf{z} \in \mathbb{R}^{\mathrm{N}\times\mathrm{L}}$ is the sampled latent variable.

Decoder Network transforms the sampled latent variable $\boldsymbol{z}$ back into the original dataspace, reconstructing the input data or generating new data attended on the original or unseen retirement conditions. The first step in the decoder is a dense layer that transforms $\boldsymbol{z}$ into an intermediate representation:

$$\mathbf{H}' = \mathrm{ReLU}\left(\mathbf{z} \cdot \mathbf{W}_{\mathbf{d}}^{\mathbf{T}} + \mathbf{b}_{\mathbf{d}}\right) \tag{3.11}$$

where, $\mathbf{W}_{\mathbf{d}} \in \mathbb{R}^{64\times\mathrm{L}}$, $\mathbf{b}_{\mathbf{d}} \in \mathbb{R}^{\mathrm{N}\times 64}$ are the neural network weighting matrix and bias matrix for the latent variable decoding embedding, respectively. $\mathbf{H}' \in \mathbb{R}^{\mathrm{N}\times 64}$ is the embedded latent variable.

$\mathbf{H}'$ is then integrated via a cross-attention, allowing the network to focus on relevant aspects of the voltage response matrix $\mathbf{x}$ conditioned by the additional retirement condition embedding $\mathbf{C}'$:

$$\mathbf{AttenDecodeder} = \mathrm{Attention}\left(\mathbf{H}', \mathbf{C}', \mathbf{C}'\right) \tag{3.12}$$

where, $\mathbf{H}' \in \mathbb{R}^{\mathrm{N}\times 64}$ and $\mathbf{C}' \in \mathbb{R}^{\mathrm{N}\times 64}$ are embedded latent variable and retirement condition embedding, respectively. $\mathbf{AttenDecodeder} \in \mathbb{R}^{\mathrm{N}\times 64}$ is the cross-attention output matrix from the embedded latent variable $\mathbf{H}'$ and the retirement conditions embedding $\mathbf{C}'$. When training, we let $\mathbf{C}' = \mathbf{C}$, and the decoder reconstructs the input pulse response data. When generating new data, $\mathbf{C}'$ is the SOH and SOC conditions of the new data to be generated. The reconstructed (generated) data $\hat{\mathbf{x}} \in \mathbb{R}^{\mathrm{N}\times 21}$ is calculated as:

$$\hat{\mathbf{x}} = \sigma(\mathbf{AttenDecodeder} \cdot \mathbf{W_o^T} + \mathbf{b_o}) \tag{3.13}$$

where, $\mathbf{W_o} \in \mathbb{R}^{21\times 64}$, $\mathbf{b_o} \in \mathbb{R}^{\mathrm{N}\times 21}$ are the neural network weighting matrix and bias matrix for the output transformation, respectively. $\sigma$ is sigmoid activation function.

### 3.4.5 The Generative Model

The loss function consists of two parts, i.e., the reconstruction loss and the Kullback-Leibler (KL) divergence loss. The reconstruction loss, i.e., the mean square error (MSE) loss $Loss_{MSE}$ between original and reconstructed (generated) data is:

$$\mathrm{Loss_{MSE}} = \frac{1}{N}\sum_{i=1}^{N}(\mathbf{x_i} - \hat{\mathbf{x}}_\mathbf{i})^2 \tag{3.14}$$

where, $\mathbf{x_i} \in \mathbb{R}^{1\times 21}$ and $\hat{\mathbf{x}}_\mathbf{i} \in \mathbb{R}^{1\times 21}$ are the original and reconstructed (generated) pulse voltage response data in each sample, respectively. $N$ is the sample size.

KL divergence loss $Loss_{KL}$, i.e., the KL divergence between original and generated data is:

$$\mathrm{Loss_{KL}} = -\frac{1}{2}\sum_{i=1}^{N}\left(1 + \mathbf{z}_{\mathbf{log_var}\ \mathbf{i}} - \mathbf{z}^2_{\mathbf{mean}\ \mathbf{i}} - \mathrm{e}^{\mathbf{z}_{\mathbf{log_var}\ \mathbf{i}}}\right) \tag{3.15}$$

The total loss is the linear combination of $Loss_{MSE}$ and $Loss_{KL}$:

$$\mathrm{Loss} = \omega_{\mathrm{xent}} \cdot \mathrm{Loss_{MSE}} + \omega_{\mathrm{KL}} \cdot \mathrm{Loss_{KL}} \tag{3.16}$$

where, $\omega_{\mathrm{xent}}$ and $\omega_{\mathrm{KL}}$ are set to 0.5 to achieve a balance between the generation accuracy and the diversity, respectively [192]. $N$ is the sample size.

### 3.4.6 Generative Model Performances

Figure 3. 8a shows the data reconstruction results of the first feature dimension U1 at selected SOC values (35% and 50%), with a MAPE lower than 1%, indicating that the model learned the dependencies between pulse voltage response data and SOC conditions.

Such dependencies are of generality in a wide SOH range, from 0.6 to 0.95. Regarding other feature dimensions, the data reconstruction results are satisfactory, with a MAPE lower than 1%, see Figure 3. 8b. The reconstruction results for other types of retired batteries (NMC, 21Ah, LMO, 10Ah and LFP, 35Ah) are presented in APPENDIX H, demonstrating the model generality and flexibility. It is noted that lower SOC regions exhibit comparatively higher reconstruction MAPE, though still lower than 1%, which can be rationalized by the sensitivity of the polarization response at low SOC regions. It is therefore recommended to perform pulse injections at a middle SOC region for reliable representation of battery degradation. However, as we highlight the random retirement conditions with different SOCs, pulse injection at fixed SOC needs extensive conditioning time and energy consumption, exacerbating a dilemma between pulse injection regions and unaffordable conditioning time and cost.

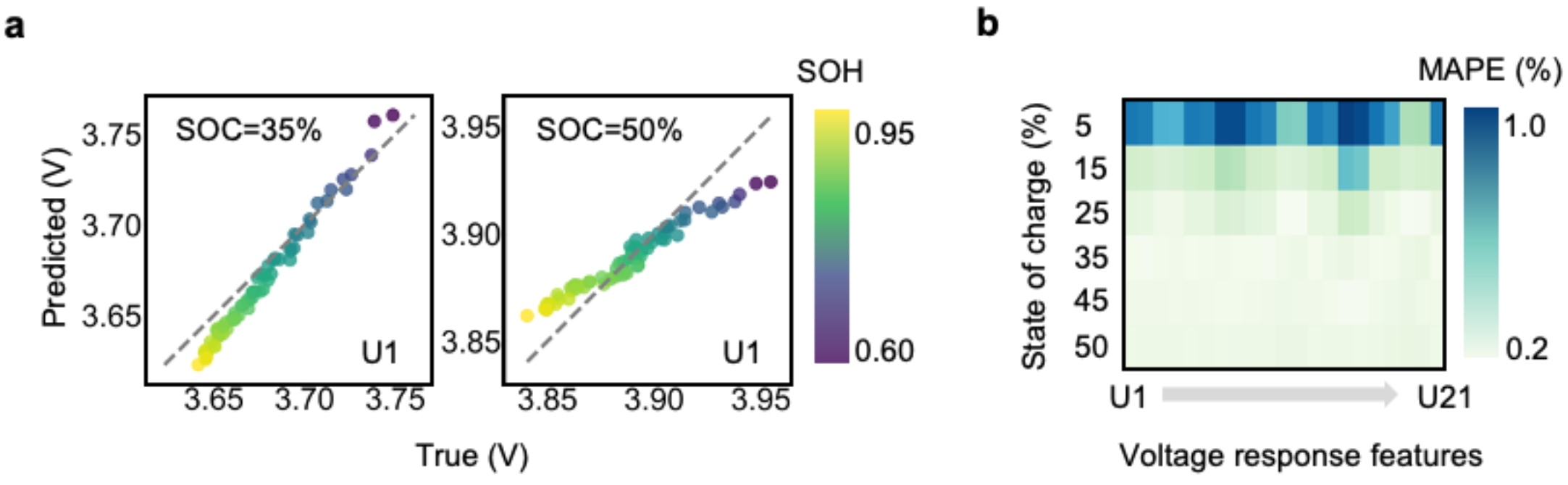

Figure 3. 8 The generative model performance.

With the proposed generative learning methodology, we stress that the already-measured data for training purposes is at the cost of extra physical tests, however, the generation is free of extra cost. In Figure 3. 9, we validate eight data generation cases, including interpolation and extrapolation. In the interpolation cases (from Case0 to Case3), the lower and upper bounds of the SOC for training data are fixed at 5% and 50%, respectively. In comparison, the extrapolation cases (from Case4 to Case7) mean that the already-measured data are exclusive with unseen OOD data to generate, a challenging and open issue in machine learning communities. Here we use prior knowledge of retired batteries to scale the latent distribution in the encoder neural network, generalizing already-measured data to OOD conditions. We note that Case4 and Case5 are toy examples since the training costs of these cases are higher than the cost savings of generating the same volume of data under unseen SOC conditions. However, we still use the toy cases to demonstrate the robustness of latent space scaling in more generalized battery retirement settings. In practical use, therefore, the users could customize the data generation strategy based on the trade-off between model training cost and data generation accuracy.

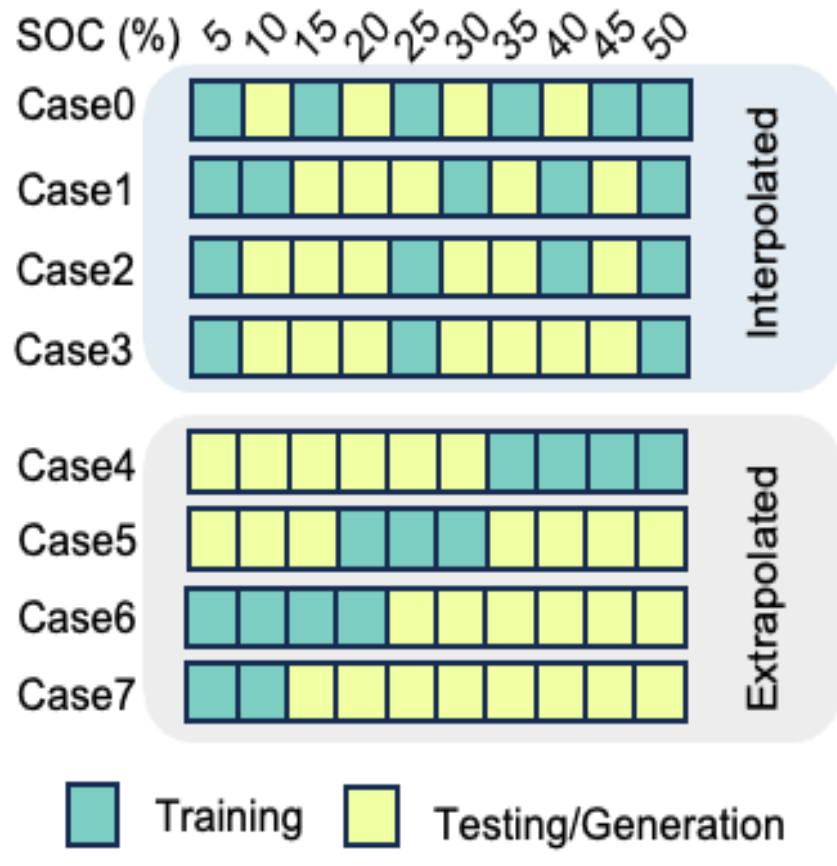

Figure 3. 9 SOC simulation of the retired batteries.

In Figure 3. 10, the distance-based data generation performance under unseen retirement conditions is illustrated. It is noted that the as-trained generative learning model is used to generate more data samples, especially under, but not limited to unseen retirement conditions. For instance, Case3 uses the physically tested pulse voltage response data at 5%, 25%, and 50% SOCs to generate data at 10%, 15%, 20%, 30%, 35%, 40%, and 45% SOCs. All interpolation cases exhibit a low MAPE error below 2%, even if the model is never trained with such data. From intuition, the extrapolative Case6 and Case7 are more promising in saving pretreatment costs since the used training data are time and cost-efficient to retrieve. We demonstrate that even in OOD cases, the generative learning strategy successfully guides the already-measured data to generalize for OOD data. Without latent space scaling, the data generation exhibits a clear increasing error when the physically measured training data is far away from the condition to generate.

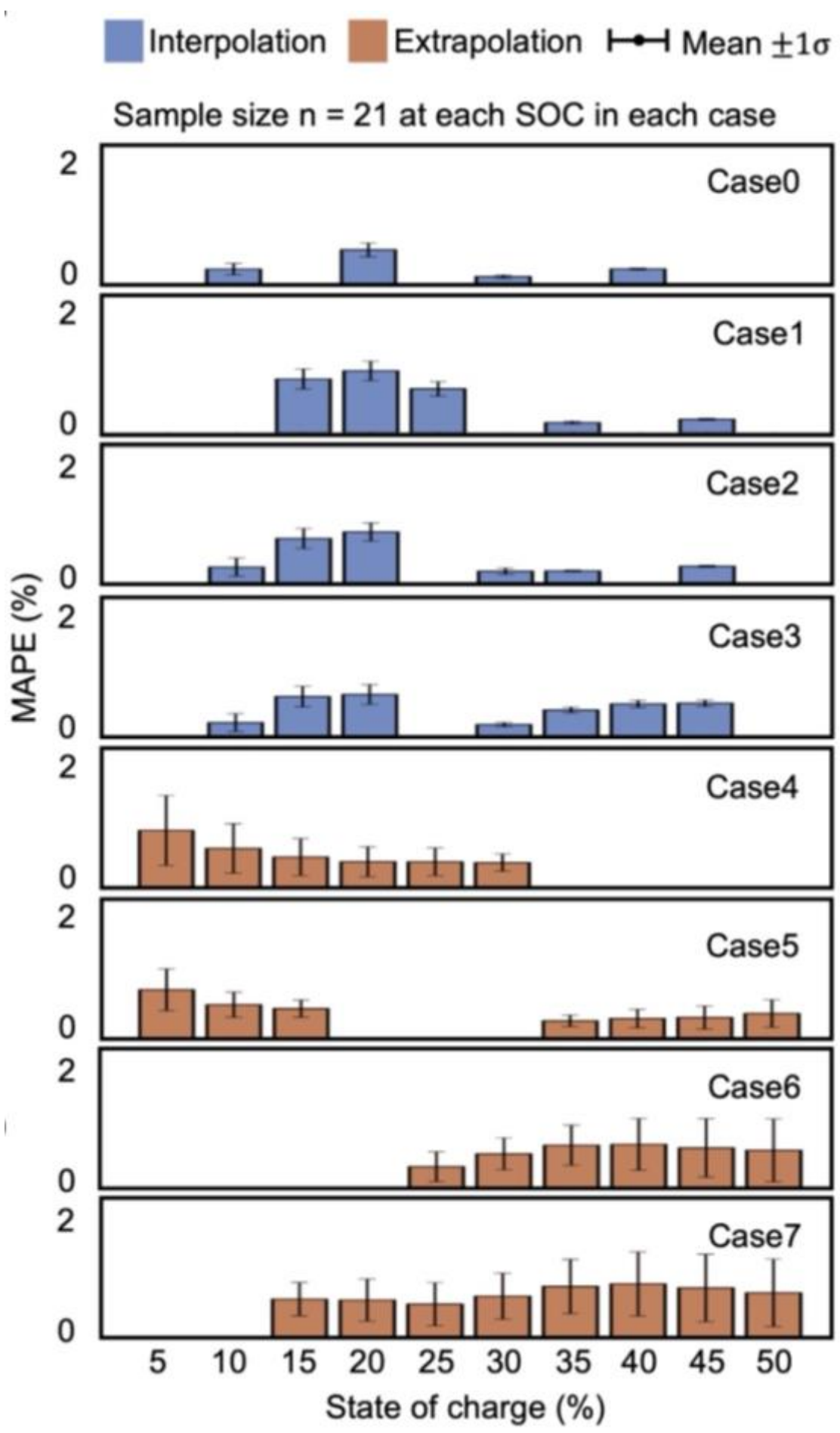

Figure 3. 10 The data generation performance under random retirement scenarios.

This phenomenon has an attractive implication that one can exploit the measured data to explore the unseen data to retrieve the data of higher cost using customized generation strategies. The model achieves low Kullback-Leibler divergences across verification settings, suggesting that the model can automatically learn the distribution of already measured data to create diversified data instances, see APPENDIX I.

## 3.5 Residual Value Evaluation Using Generated Pulse Voltage

In this section, we evaluate the fidelity of the generated pulse voltage response data by feeding it into simple regressor to examine the model performance under random battery retirement conditions.

### 3.5.1 Random Forest Regressor

We adopt a random forest algorithm to perform SOH estimation due to the suitability of tabular data after feature engineering of the pulse voltage response curves, which can be formulated as:

$$\bar{\mathbf{y}} = \bar{h}(\mathbf{X}) = \frac{1}{K}\sum_{m=1}^{M} h(\mathbf{X}; \boldsymbol{\vartheta}_{\mathrm{m}}, \boldsymbol{\theta}_{\mathrm{m}}) \tag{3.17}$$

where $\bar{\mathbf{y}}$ is the predicted SOH value vector. $M$ is the tree number in the random forest. $\boldsymbol{\vartheta}_{\mathrm{m}}$ and $\boldsymbol{\theta}_{\mathrm{m}}$ are the hyperparameters, i.e., the minimum leaf size and the maximum depth of the kth tree in the random forest, respectively. The hyperparameters are set as equal across different cases, i.e., $\mathrm{M} = 20$, $\boldsymbol{\vartheta}_{\mathrm{m}} = 1$, and $\boldsymbol{\theta}_{\mathrm{m}} = 64$, for a fair comparison. Implementations are in the Sklearn Package (version 1.3.1) in the Python 3.11.5 environment, with a random state at 0.

### 3.5.2 Residual Value Evaluation Performance

We use the generated pulse voltage response features to estimate the SOH of retired batteries. In Figure 3. 11, the SOH estimation performance under different battery retirement cases is illustrated, where shaded scatters and the void regions stand for the testing and training conditions of the generation model, respectively. Taking Case 3 as an example, physical pulse measurements are performed at 5%, 25%, and 50%, which are used to train the generation model. The as-trained model generates unseen data at 10%, 15%, 20%, 30%, 35%, 40%, and 45% SOCs, utilized as input data of a regressor to realize SOH estimation in these unseen SOC levels. Regardless of data scarcity, we demonstrate that the generated data successfully facilitates accurate SOH estimation with an average MAPE and STD of 4.9% and 2.9% in both interpretative and extrapolative cases, respectively. However, the interpolative data generation leads to more accurate and stable estimations, compared with extrapolative generation. The average MAPE and STD in interpolative cases are 4.4% and 2.5%, respectively; the average MAPE and STD in extrapolative cases are 6.0% and 2.7%, respectively. This phenomenon can be rationalized by the challenges in OOD issues, however, which is calibrated thanks to the integration of prior knowledge into the latent space scaling.

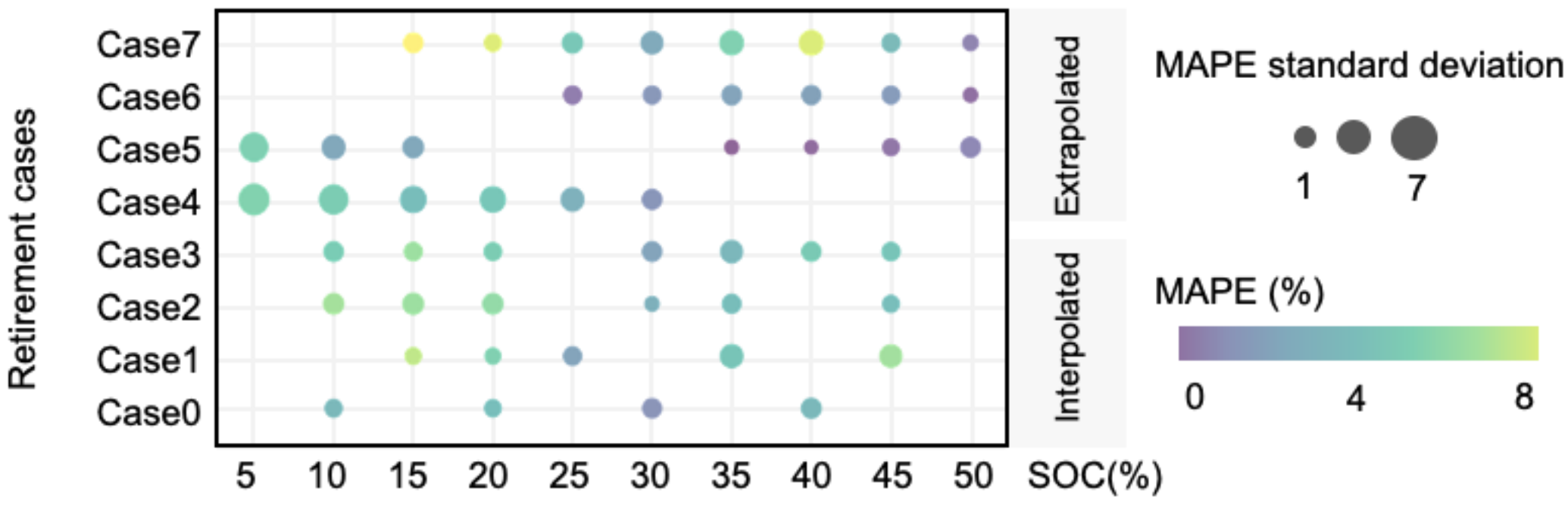

Figure 3. 11 SOH estimation performance under different cases.

SOH estimation results under selected SOC values are presented in Figure 3. 12 with parity plots, indicating a successful data generation across wide SOC conditions. We observe an interesting overestimation and underestimation of SOH under higher and lower SOC regions, which can be interpreted by the zero mean and unit variance Gaussian distribution in its latent space. Specifically, taking the underestimation as an example, generated pulse voltage responses conditioned on low SOC regions exhibit higher values, otherwise characterized by a more degraded battery with lower SOH, and *vice versa* for the overestimation[193]. The observation has an important implication that the data generation model can be tuned by manipulating the latent space for unseen but conditional data generation.

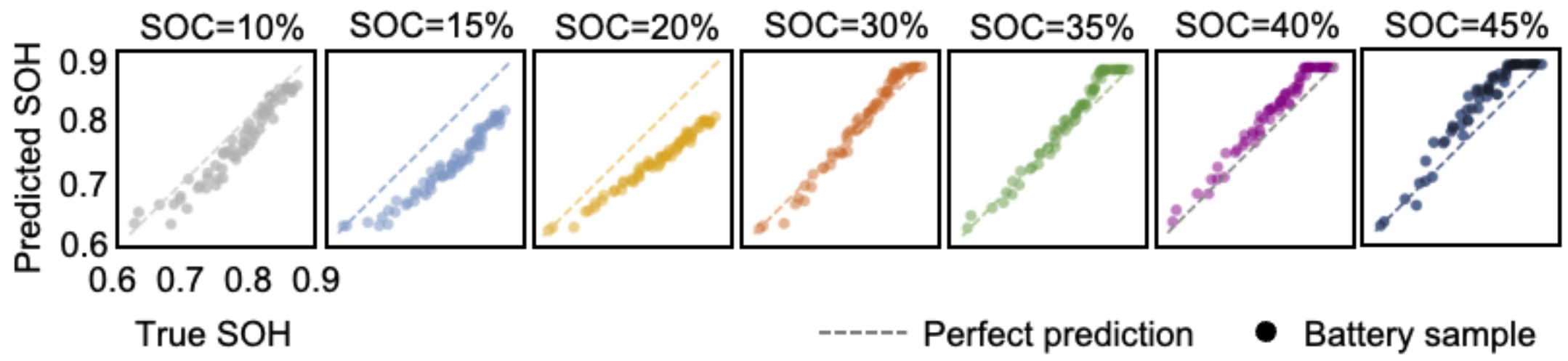

Figure 3. 12 The parity plot of true and estimated SOH in Case3.

Here we compare the SOH estimation results with and without data generation, highlighting its necessity under highly random retirement conditions. In Figure 3. 13, an interpretative case shows limited data access in some discrete SOC values, indicated by gray-shaded regions. Using these data, the regressor produces inaccurate estimations biased to the centroid of the available data distribution, with an average MAPE of 17.3% and a STD of 7.3. In comparison, the generative learning-assisted SOH estimations show a rigorous increase. The average MAPE is 5.4% and a STD of 2.7, indicated by stars in the plot.

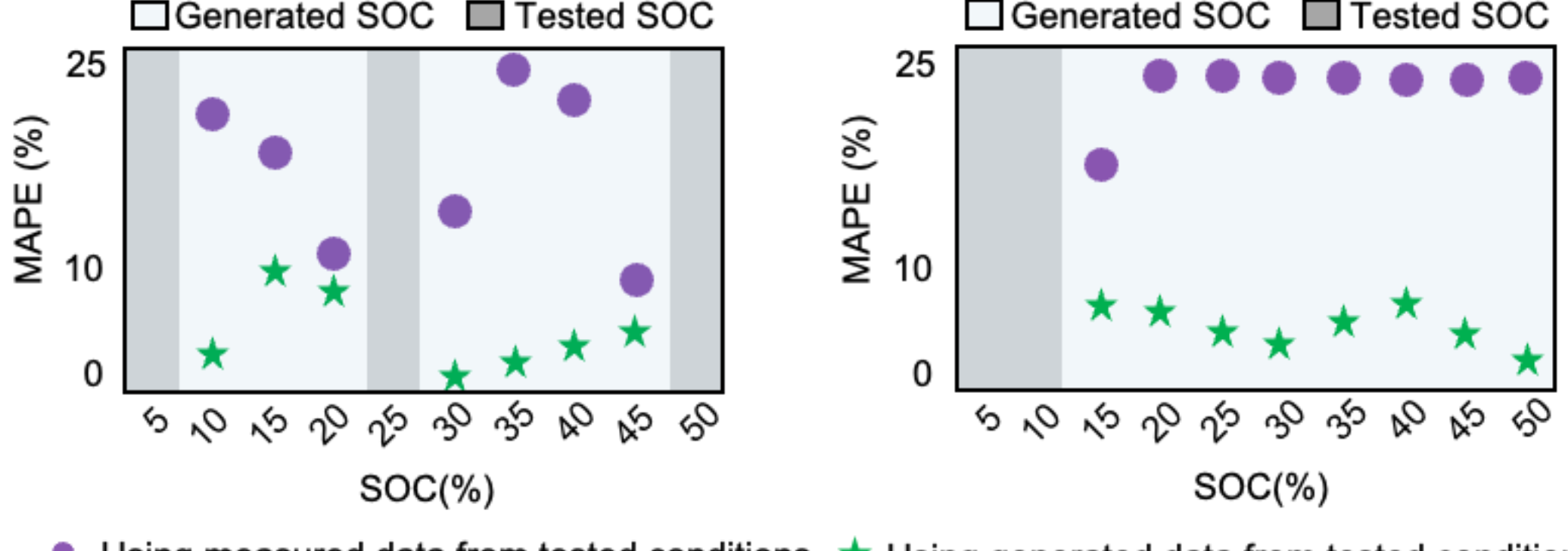

Figure 3. 13 SOH estimation performance comparison.

We consider an extreme, yet more preferred situation where the accessible data can be at low SOC regions. This situation has a clear physical meaning that data curation in these regions can minimize the SOC conditioning time. However, the data curation strategy brings more difficulties in accurate estimations, despite that the curation time and costs are saved. In Fig. 3.13 (right subplot), we take accessible data at 5% and 10% SOC for an illustration. It shows an asymptotically increased error when using the accessible data to make SOH estimations. The asymptotic effect can be interpreted as the shift from the accessible data distribution increases with SOC values. Such a challenge results in an average MAPE of 23.8% and a STD of 7.3 when using data in low SOC regions. In comparison, the estimation performance is increased to a MAPE of 6.0% and a STD of 2.9 using generated data. The estimation results for other retired batteries, with different cathode material types, physical formats, and historical usage patterns, are consistently improved, see Figure 3. 14 (a) 21Ah NMC and (b) 10Ah LMO.

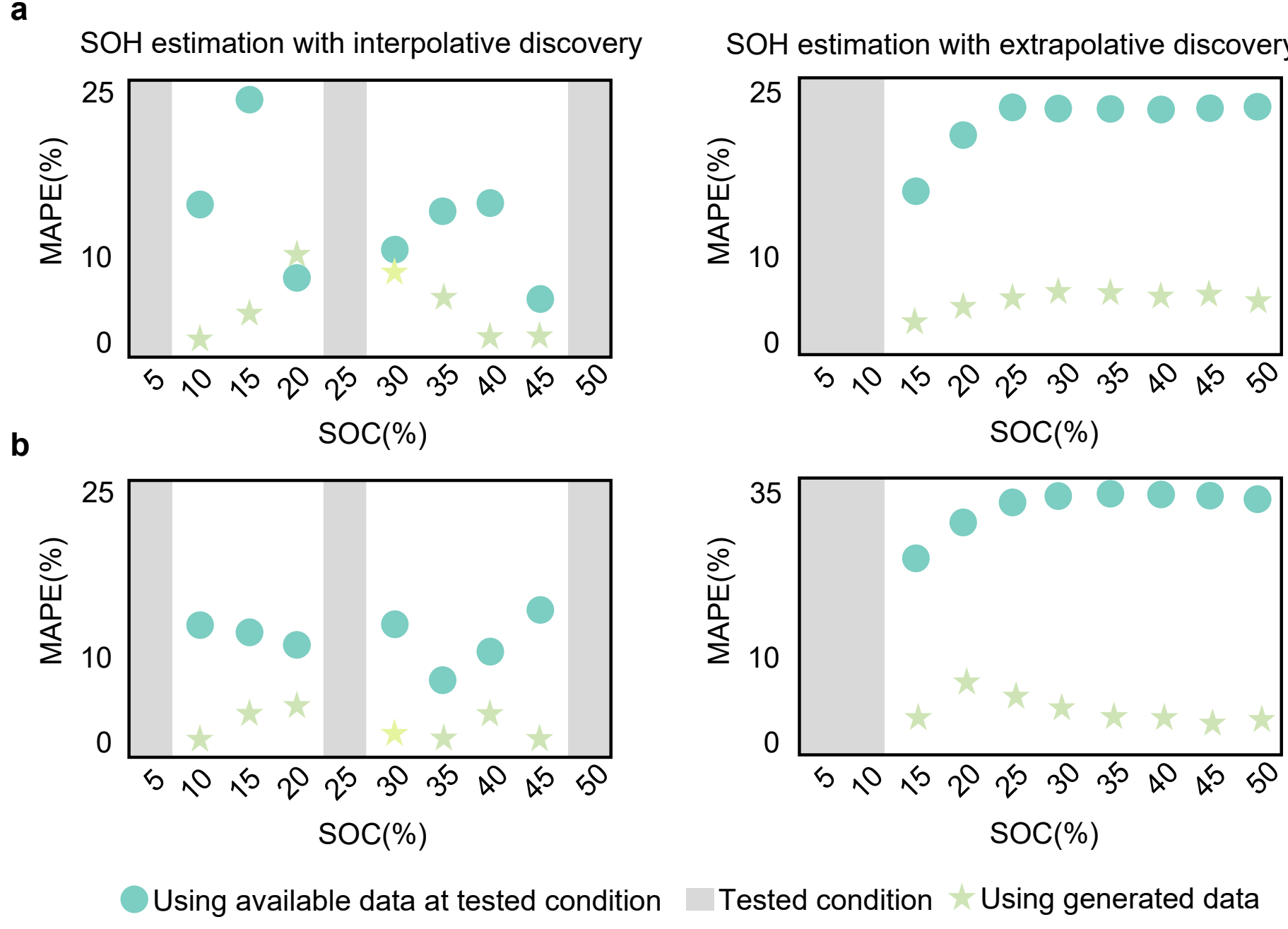

Figure 3. 14 SOH estimation on other materials.

Therefore, the proposed generative learning empowers rapid SOH pretreatment of retired batteries, by generating data in unseen SOC conditions at saved time requirements otherwise required by SOC conditioning[40], thus advancing efficient and sustainable battery reuse and recycling in a data-driven manner.

## 3.6 Sustainability evaluation of Data Generation

In this section, we assess the sustainability of the data generation by calculating the energy consumption that is needed to secure the data as well as the time that is needed to trained the generative model.

### 3.6.1 Cost and Carbon Reduction from Pulse Test

We evaluate the electricity and $CO_2$ emission savings using different data generation cases. Figure 3. 15 shows the electricity and $CO_2$ emission savings in an ascending case order and the color bar maps the according values. Except for Case4 and Case5, two toy examples, the extrapolation strategy can save up to 60,000 USD in electricity costs and 460,000 kilograms of $CO_2$ emissions assuming a 1000-ton battery retirement scale.

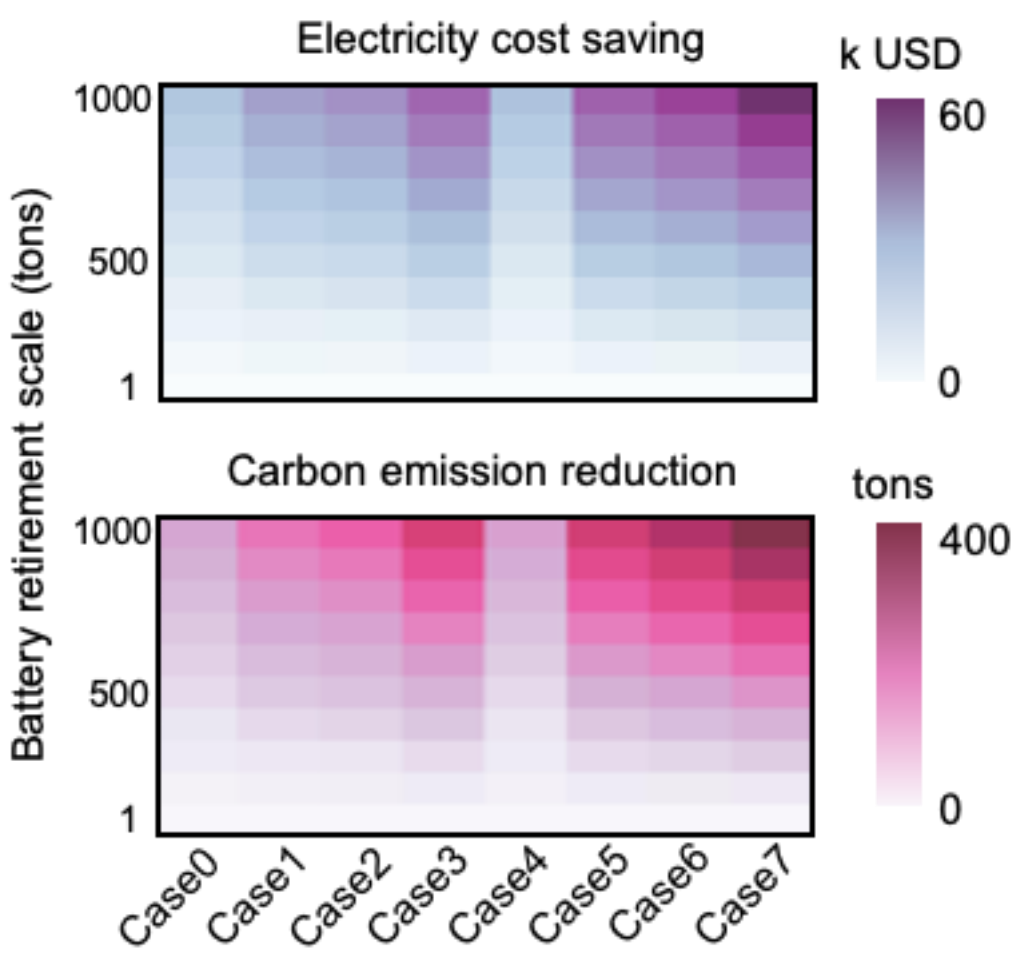

Figure 3. 15 The electricity cost savings and $CO_2$ emission reductions.

This priority is achieved by generating expensive data, which needs SOC conditioning costs, from cost-efficient physically measured data. Inside the interpolation generation, we found that the model does not require stringent intermediate points to interpolate on, leading to a usage priority in Case3, using 5%, 25%, and 50% SOC data. Despite different case difficulties, the data generation model is easy to train, converging in less than 50 iterations within a milliseconds level, see APPENDIX J. Once the generative model is properly trained, the users can generate unlimited data across wide retirement conditions, without any measurement costs, facilitating rapid, accurate and sustainable retired battery SOH estimation for reuse and recycling decision-making.

### 3.6.2 Carbon and Cost Savings from Pulse Test Data Generation

In Figure 3. 16, the pyrometallurgy exhibits the highest $CO_2$ emissions, followed by the hydrometallurgy and direct recycling. $CO_2$ emissions using CCCV in pyrometallurgy, hydrometallurgy, and direct recycling are 1104-, 729-, and 579-kilograms eq. per ton of retired NMC batteries, respectively, with 80% and 50% of the retirement SOH and SOC conditions assumed. We found that with our generative learning-assisted SOH estimation, $CO_2$ emissions were reduced to 753-, 379-, and 228-kilograms eq. per ton of retired NMC batteries, respectively. Besides $CO_2$ emissions, the electricity cost savings using the identical pretreatment are in presented in APPENDIX K for NMC and LFP retired batteries, respectively.

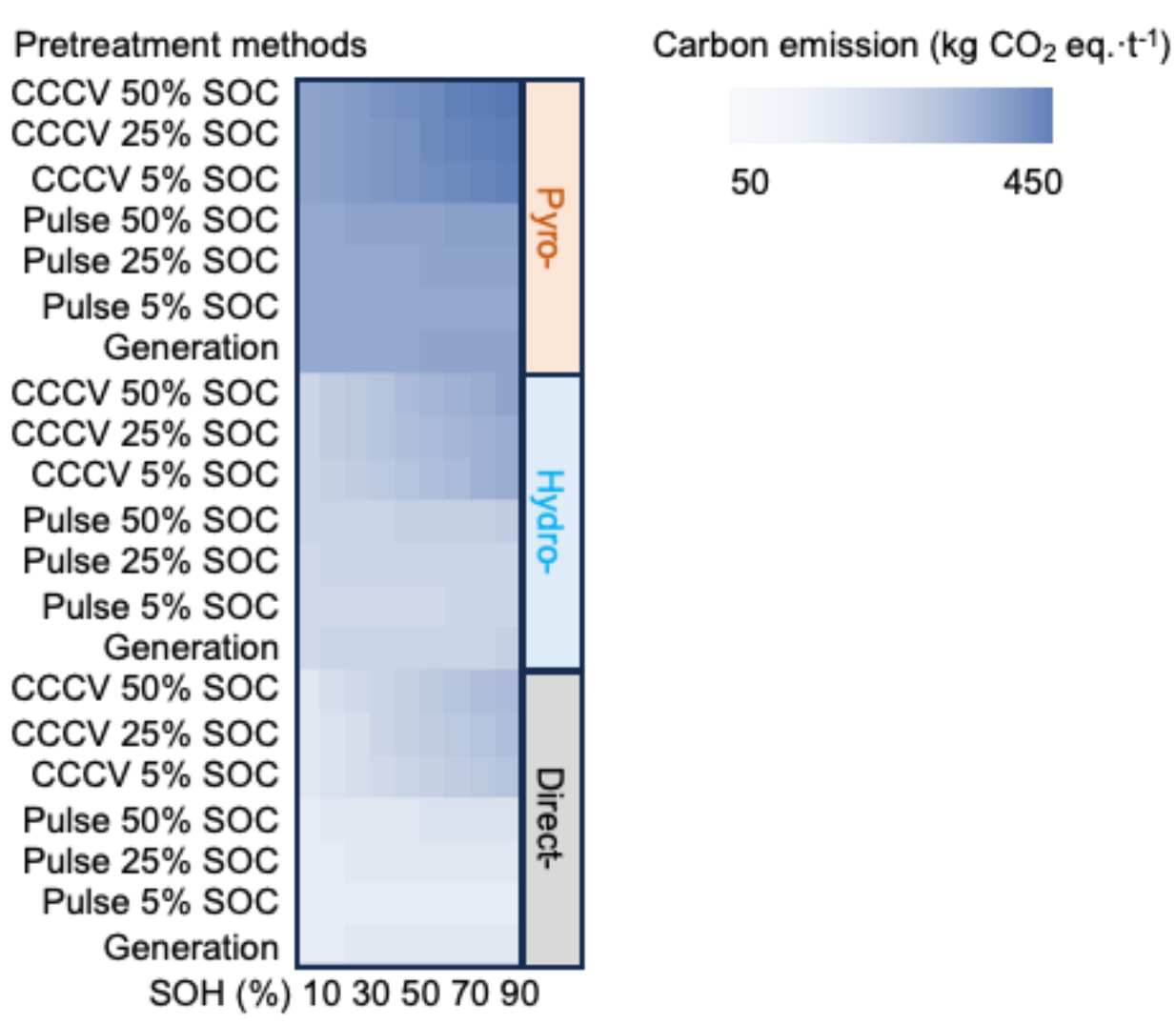

Figure 3. 16 CO2 emission reduction assessment.

In Figure 3. 17, we clarify the impact of an efficient pretreatment by providing a comparative analysis of electricity costs against the battery retirement scale at 80% SOH, excluding the pulse injection electricity costs in the post-training phase. At the 1.07 tons retirement scale, generative learning costs are lower than that of pulse tests for retired batteries with 25% SOC, making it the second-best pretreatment method in terms of cost.

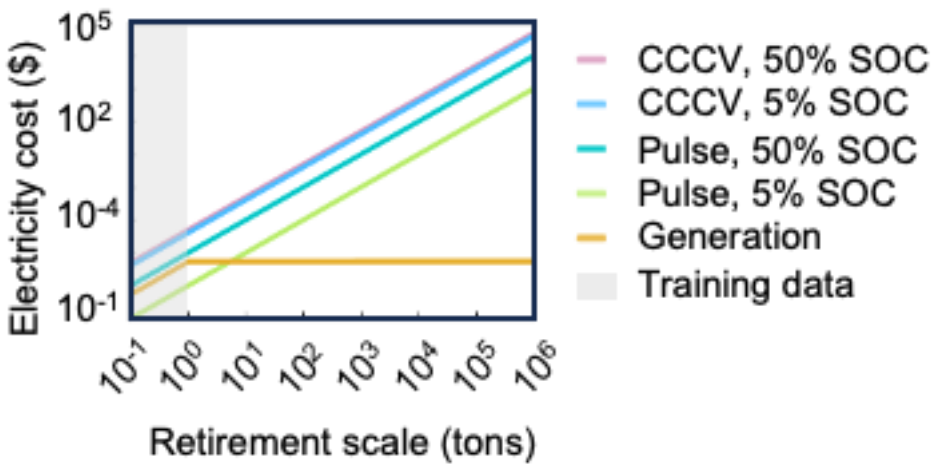

Figure 3. 17 Comparative analysis of electricity cost against battery retirement scale.

At the 5.33 tons retirement scale, the costs of the generative learning are lower than that of pulse tests for retired batteries with a 5% SOC, establishing it as optimal. Moreover, the advantage of generative learning does not rely on a retirement scale of retired batteries when it goes beyond 5.33 tons, suitable for massive retirement battery processing. It is conservatively estimated that data collected from 1 ton of retired batteries (25,500 samples of 18650 batteries) are adequate to supervise the data generation model. However, our model only uses 10.6% of such a training data scale (2700 samples), indicating that the model training costs for data generation can still be further reduced to that is lower than $5.7 \times 10^{-3}$ USD.

### 3.6.3 Cost Breakdown of Residual Assessment

Direct recycling is selected for evaluation due to its cost-sensitive nature. We analyze the impact on overall cost proportions with electricity cost reductions. In the upper (lower) part of Figure 3. 18, the ring plot and stacked bar plot of pretreatment costs are illustrated to show the changes in electricity consumption for 1 ton of retired NMC (LFP) batteries with the relative and absolute value, excluding material flows, respectively. When the retirement condition is at 80% SOH and 50% SOC, the electricity consumption proportion using deliberate data generation decreases from 13.3% to 6.6% compared to the traditional CCCV capacity calibration test. The absolute electricity cost, inclusive of the pretreatment and recycling, decreases from 80 to 36 USD per ton of retired batteries.

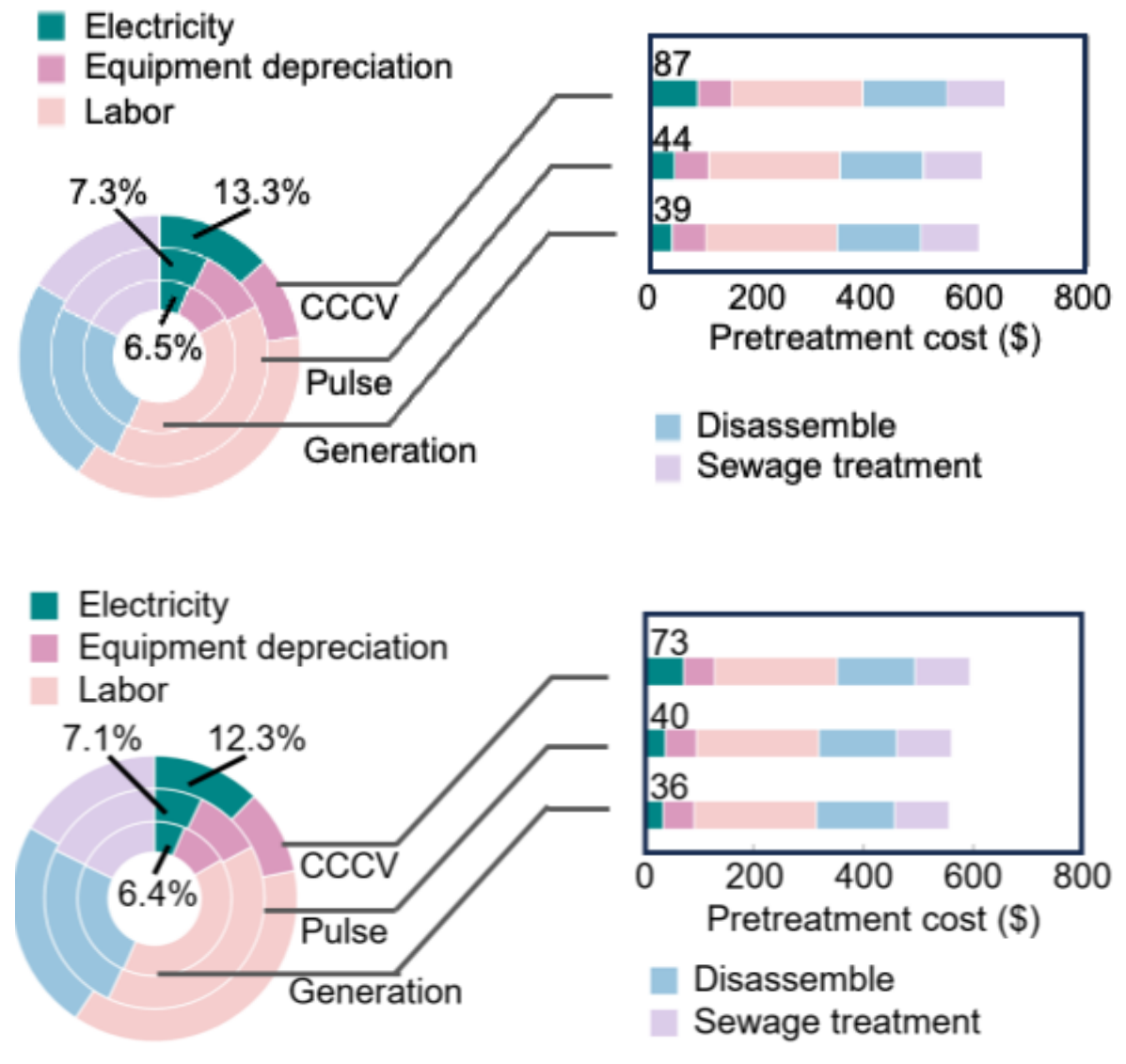

Figure 3. 18 Pretreatment cost breakdown analysis.

### 3.6.4 Sustainability Potential in Future Retirement Scenarios

Considering a battery retirement scenario in 2030 globally, as illustrated in Figure 3. 19, pretreatment electricity saving exponentially increases with retirement scales up to a $10^9$ USD level. Note that the calculation is with pulse electricity in post-training phase included, towards the 2030 battery retirement scenario, assuming 80% SOH and 50% SOC at retirement. The generative learning saves up to 4.9 billion USD in electricity costs and reduces 35.8 billion kilograms of $CO_2$ emissions compared to traditional CCCV-based capacity calibration tests. Regarding LFP batteries, a similar scale effect can be observed in electricity cost saving, and $CO_2$ emission reduction under the 2030 scenario as well. The cost saving and $CO_2$ emission reduction do not solely stem from priorities of direct recycling, but more importantly, from saving exhaustive data curation at random retirement conditions using the proposed generative learning-assisted SOH estimation.

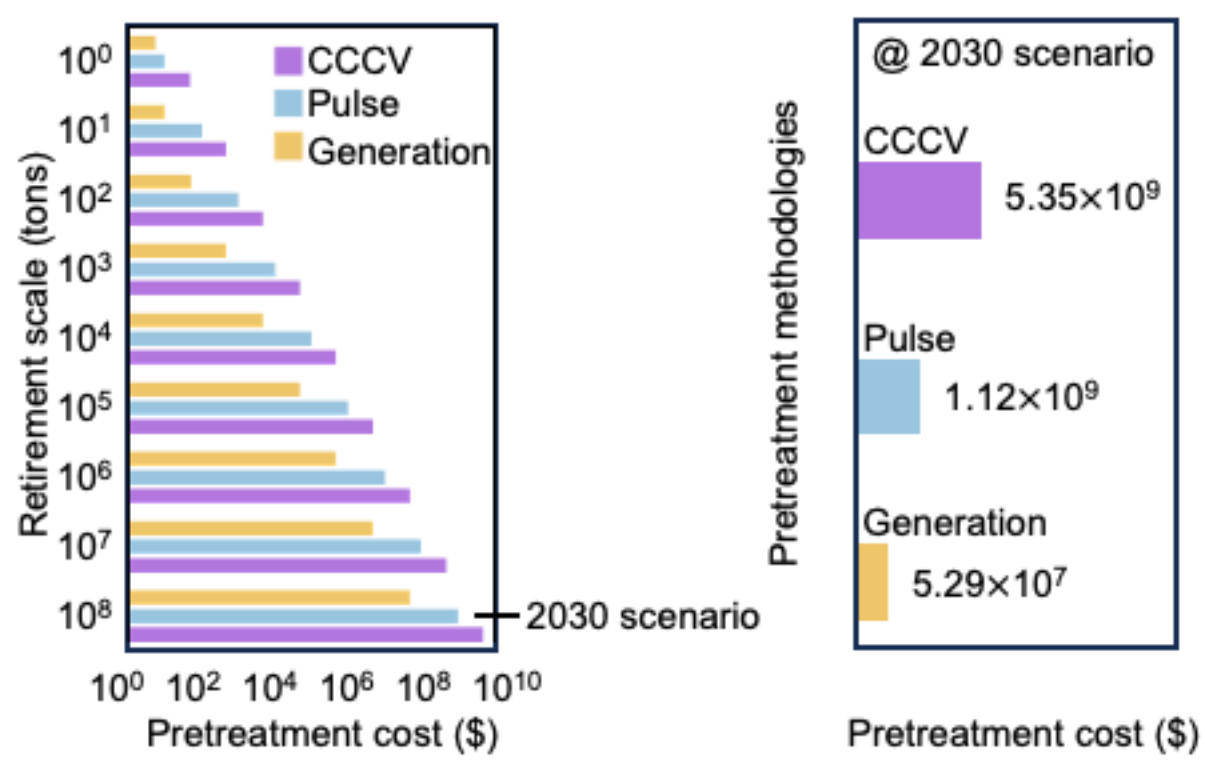

Figure 3. 19 Pretreatment costs against the battery retirement scale.

## 3.7 Summary

We have elaborated on the success of using generative learning in generating high-fidelity pulse voltage response data of retired batteries while saving conditioning costs and time. The success is attributed to the generation model to learn the dependency between retirement conditions and pulse voltage responses, inclusive of heterogeneities in cathode material types, physical formats, capacity designs, and historical usages. Diverging from the traditional SOC conditioning using capacity calibration tests, even the state-of-the-art pulse tests combined with machine learning at a specific SOC level [40], the proposed data generation model underscores a balance between exploiting already measured data and exploring potential data space with random SOH and SOC retirement conditions. The proposed method inflicts no extra physical measurements, reducing secondary energy use and environmental burden in practical use. With a portion of the generated data, downstream SOH estimation tasks still perform consistently compared to the sufficient data situations, a useful implication for reducing computational cost and improving estimation accuracy in real-world cases, see APPENDIX L.

In conclusion, the proposed generative learning demonstrates consolidated promises in estimating the SOH of retired batteries rapidly, accurately, and sustainably. We present a generative learning model supervised by a few physically measured pulse voltage response data that can effectively generate new data across wide SOC and cathode material types, with an MAPE of interpretative and extrapolative cases below 2%. With generated data, a regressor obtains an MAPE below 6% for SOH estimation, including unseen SOC conditions. The results are verified through 3 prevailing cathode material types (NMC, LFP, and LMO), 3 physical formats (cylindric, pouch, and prismatic), 4 capacity designs (2.1Ah, 10Ah, 21Ah, and 35Ah), and 4 historical usages (1 laboratory accelerated aging and 3 different EV-driving aging patterns). We showcase the economic and environmental viability of the data generation in an upcoming battery retirement scenario of 2030 globally, by saving 4.9 billion USD electricity costs and reducing 35.8 billion kilograms $CO_2$ emissions. Generally, the proposed data generation method enables sustainable battery reuse and recycling decision-making [17], especially for direct recycling, by devising appropriate use of lithium supplements and other chemical reagents, critically important to recycling costs and product qualities. Broadly, the generative learning method inspires the promises of exploiting already measured data to explore the unexhaustive data space, as opposed to separate physical measurements, alleviating the data scarcity and heterogeneity issues in critical estimation and predictive applications where data are time-consuming, expensive, and polluting to retrieve.

# CHAPTER 4 COLLABORATIVE MATERIAL SORTING

## 4.1 Overview

In this chapter, we perform a cathode material sorting of the retired batteries, leveraging the existing battery data from multiple collaborators, such as battery manufacturers, practical application operators, academic research institutions, and third-party platforms, in a collaborative while privacy-preserving machine learning fashion as illustrated in Figure 4. 1.

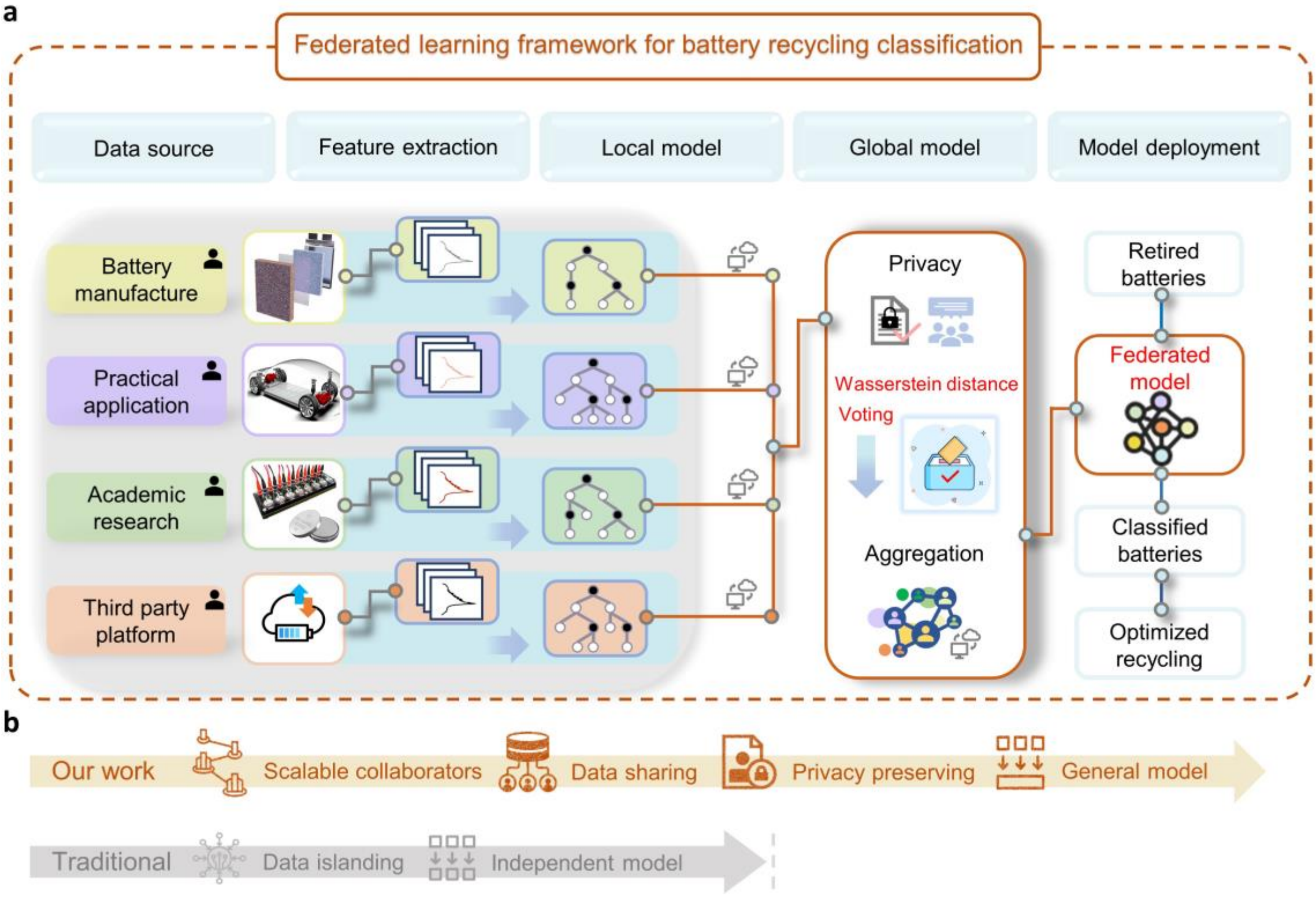

Figure 4. 1 Multi entity battery data and privacy.

Multiple data sources, such as battery manufacturers, practical application operators, academic research institutions, and third-party platforms, can be data contributors. The battery data are neither exchanged between contributors nor uploaded to the battery recycler. Instead, the data contributors train local models and share model parameters with the battery recycler to build a global model. The proposed Wasserstein-distance voting technique fuses the local models into the global model, which is robust to data imbalance and noise. Battery recyclers can use the jointly-built model for battery sorting, combined with the easy-to-access field testing data. Our FL framework encourages collaborators to sharing the data while preserving data privacy different from the traditional data islanding paradigm.

Our FL model was trained using only one cycle of field-testing data via a standardized feature extraction process, without any prior knowledge of the historical operation conditions. We compare the predictive power of our FL model with that of independently learned local models based on local data under both homogeneous and heterogeneous battery recycling circumstances. The heterogeneity issue is resolved by our proposed Wasserstein-distance voting strategy. An economic evaluation of retired battery recycling using our proposed FL framework is conducted, highlighting the relevance and necessity of accurate sorting of retired batteries. We comprehensively discuss the model interpretability, battery recycling implications, and broader prospects of the future battery recycling practice integrated with FL.

## 4.2 Data Collection and Processing

The unique battery kinetics in different battery types are often high-dimensional and hard to characterize due to divergent operating cases, manufacturing variability, and historical usages [194]. We collected and standardized 130 retired batteries with 5 cathode material types from 7 manufacturers to construct an OOD, equivalently heterogeneous dataset, as presented in Table 4. 1.

Table 4. 1 The classified battery groups and detailed information.

| Manufacturer | Cathode | Class | Cells | Composition | Q (Ah) |
|---|---|---|---|---|---|
| CALCE [195-199] | LCO | 1 | 7 | $LiCoO_2$ | 1.35 |
| HNEI [200] | NMC /LCO | 2 | 13 | $LiCoO_2$ and $LiNi_4Co_4Mn_2O_2$ | 2.80 |
| MICH-Expa [201] | NMC | 3 | 8 | NMC111: CB: PVDF (94:3:3) | 5.00 |
| MICH-Form [64] | NMC | 4 | 39 | NMC111: C65: PVDF (94:3:3) | 2.36 |
| OX [202] | LCO | 5 | 8 | $LiCoO_2$ | 0.74 |
| SNL [203] | LFP | 6 | 21 | $LiFePO_4$ | 1.10 |
| SNL [203] | NCA | 7 | 14 | $LiNi_xCo_yAl_{1-x-y}O_2$ | 3.20 |
| SNL [203] | NMC | 8 | 15 | $LiNi_xMn_yCo_{1-x-y}O_2$ | 3.00 |
| UL-PUR [204] | NCA | 9 | 5 | $LiNi_{0.8}Co_{0.15}Al_{0.05}O_2$ | 3.40 |
| Total | | | 130 | | |

Given different historical usages, the capacities of the collected batteries are below 90% of the nominal capacity. The battery cathode materials are lithium cobalt oxide (LCO), nickel manganese cobalt (NMC), lithium ferrophosphate (LFP), nickel-cobalt-aluminum oxide (NCA), and NMC-LCO blended types, which are further grouped into 9 classes based on the manufacturers. We intentionally include batteries with divergent historical usages, from laboratory testing to EV driving profiles, to train a generalized model for the battery recycler independent of historical usages and battery types.

### 4.2.1 Open-Source Dataset Standardization

For standardization, all data required from the recycler are the currently-probed (field-testing) cycle with one charging and discharging test, which is easy to implement in practical cases. The as-probed data are first denoised by filling in missing values, replacing outliers, and performing median filtering. Human-induced and cathode-heterogeneity-induced noises are deliberately retained, though, to make the model robust to imperfect inputs. The data are then linearly interpolated for curve filling, as presented in Figure 4. 2.

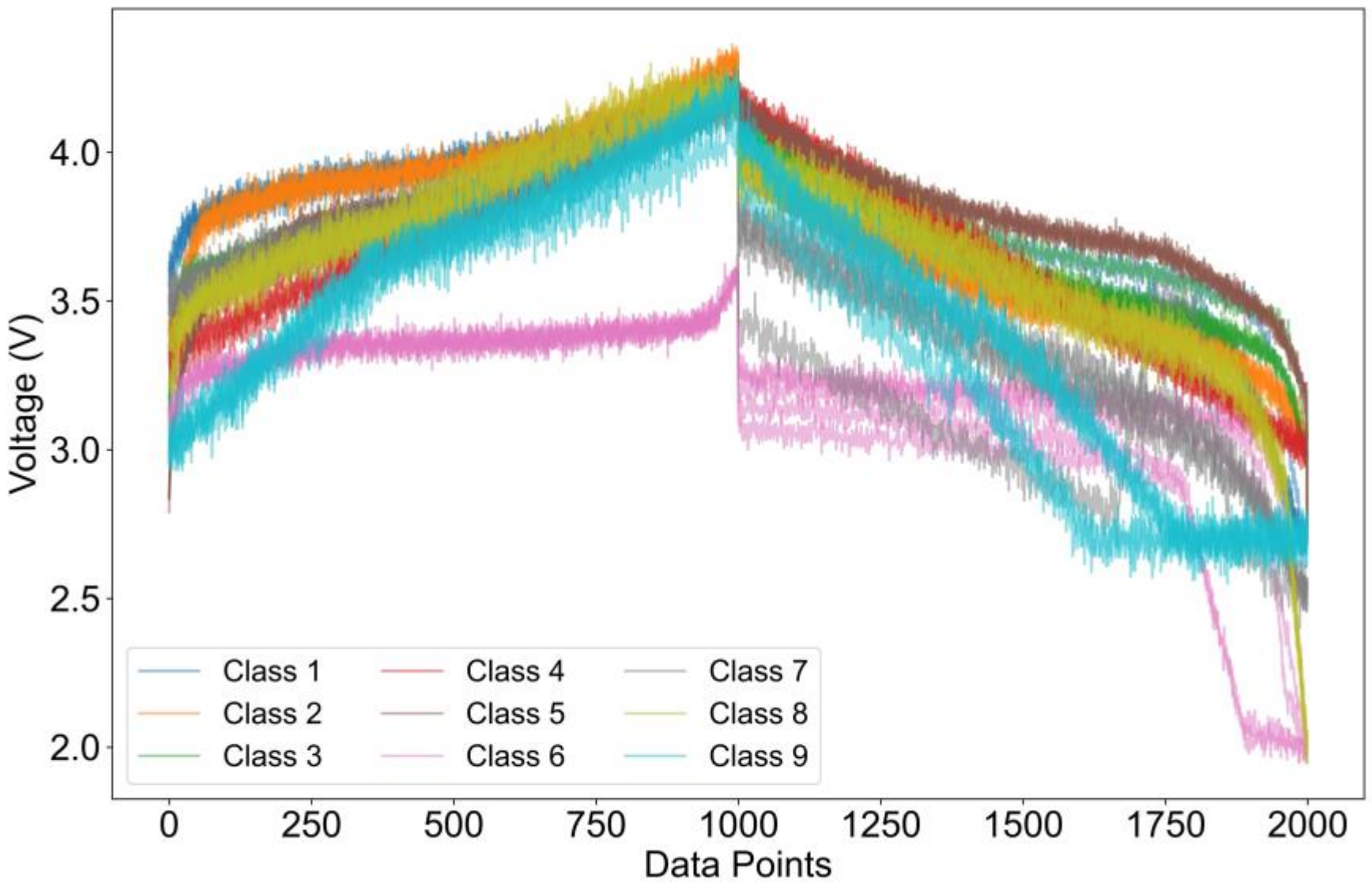

Figure 4. 2 Charging and discharging curve.

### 4.2.2 Feature Engineering

Features extracted from the standardization pipeline are well interpretable, a concern of significant commercial interest. To the best of our knowledge, it is the first time that heterogeneous battery data from multiple sources and historical usages are utilized to assist in the strategy design of battery recycling. Figure 4. 3 demonstrate the feature engineering process. We focus on the charging and discharging curve of the retired batteries in the last cycle, i.e., one charging and one discharging cycle (See APPENDIX M). In the charging cycle, 15 features are extracted from the voltage-capacity and dQ/dV curves, where V and Q refer to the voltage and capacity values, respectively. The same set of features are extracted for the discharging cycle. As a result, 30 features are extracted in total, as indicated from F1 to F30. Refer to APPENDIX N for a detailed explanation of the features.

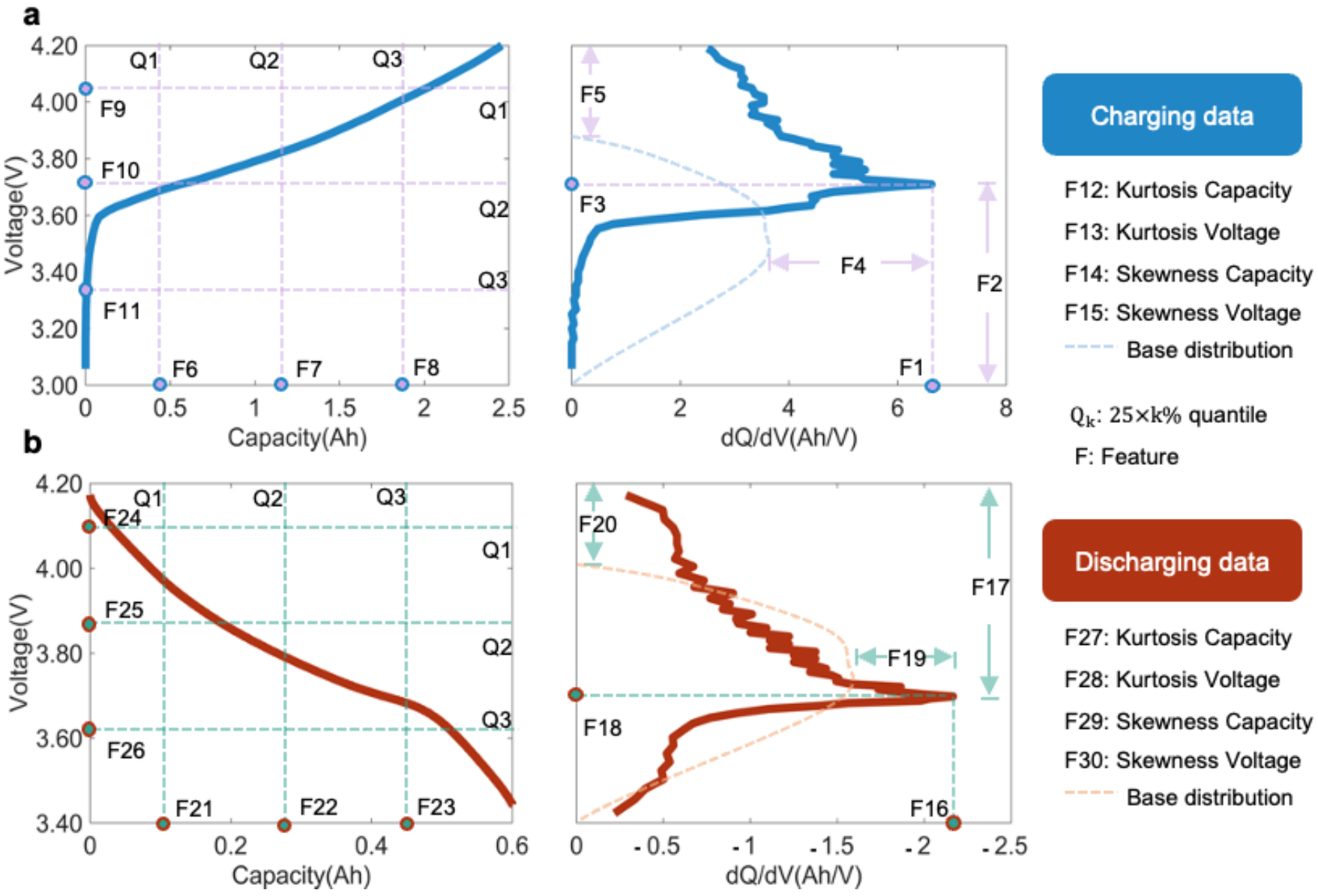

Figure 4. 3 Feature engineering.

Figure 4. 4 showcases the absolute and relative feature values of the selected batteries from each class. Most relative feature values in different classes overlap in the -1 to 0 region (with the light green color) and are indistinguishable, illustrating the difficulty in classifying battery type using one cycle of battery data.

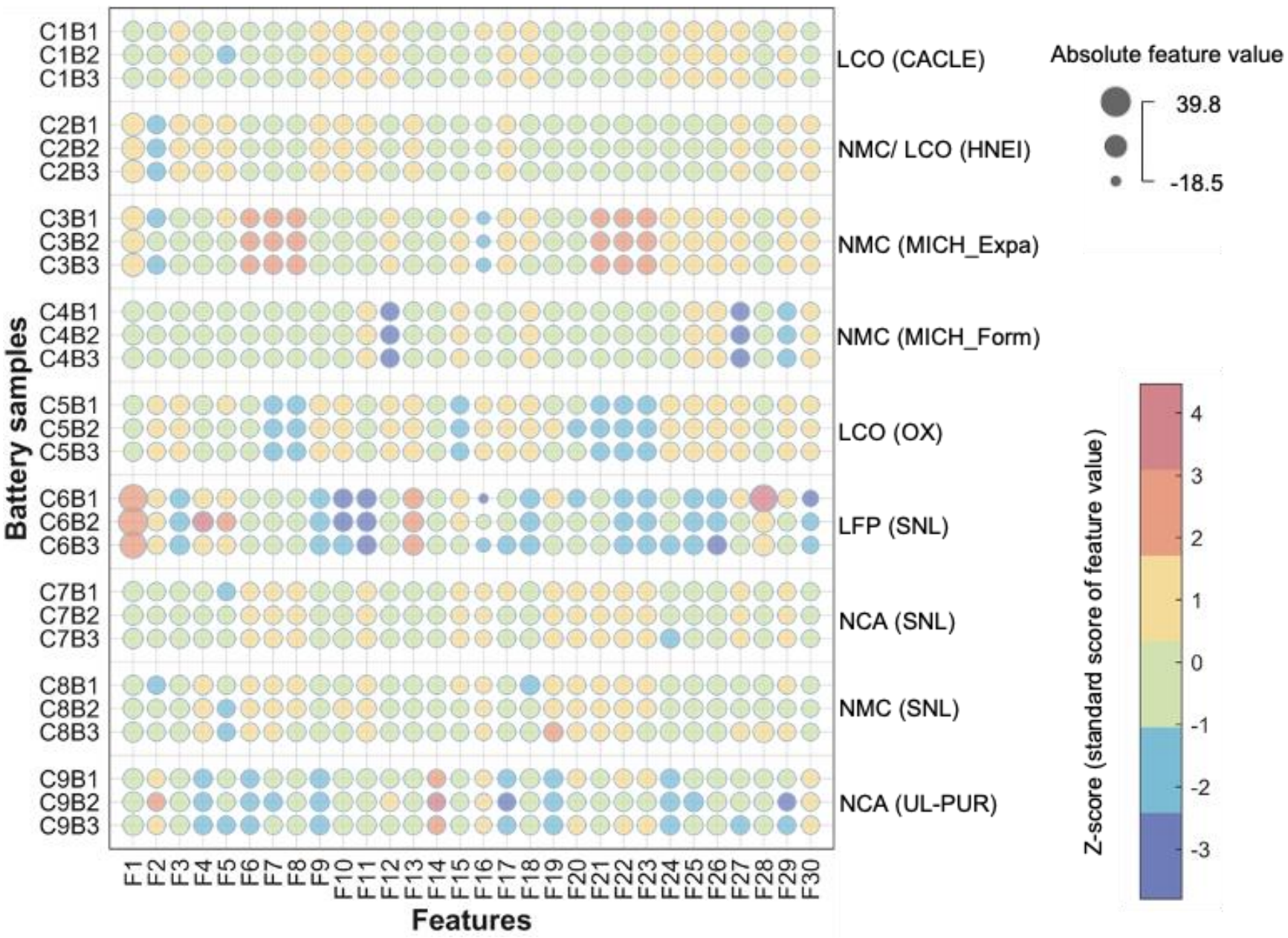

Figure 4. 4 Feature engineering results.

The difficulty is expected because the divergent historical operation conditions can influence the charging-discharging kinetics of the batteries so that the extracted features can be largely correlated despite the different battery type, see Figure 4. 5.

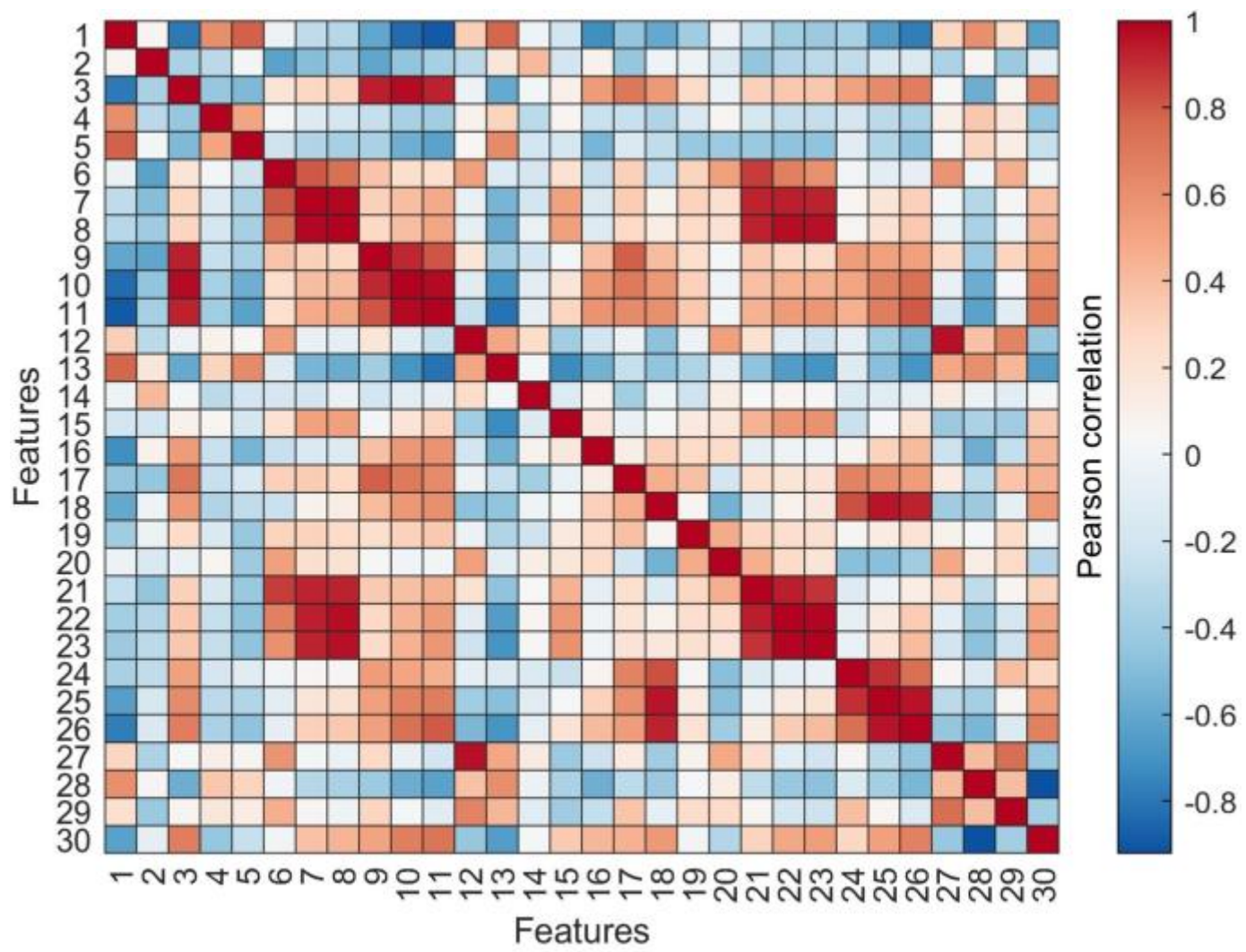

Figure 4. 5 Feature correlation analysis.

As a result, rather than directly interpreting the extracted features using expert knowledge, we employ an alternative data-driven approach that automatically leverages the latent patterns across various battery types.

## 4.3 Collaborative Material Sorting Model

In this section, we elaborate on the data augmentation, client generation, client model, voting strategy design and the feature importance analysis methodologies.

### 4.3.1 Data Augmentation with Privacy Budget

We perform data augmentation with two considerations: (1) more diversified data for training a generalized model and (2) protecting data privacy by preventing adversarial reconstruction (reverse engineering to eavesdrop on the private data). Given a feature matrix $\mathcal{F}_{N\times M}$ and a class label vector $C_{N\times 1} = \{c^i\}$, $i = \{1,2,\cdots,9\}$, where $N$ and $M$ are the numbers of measured batteries (known as "observations") and associated features, respectively. The data augmentation includes the following three steps: First, we index the feature matrix $\mathcal{F}$ for a subset $F^i$ using each unique class label $c^i$. Second, we augment $F^i$ into $F^i_{\text{Aug}}$ by resampling with replacements using Bootstrapping. The resampling size (or the number of bootstrapped observations) of $F^i_{\text{Aug}}$ is set to $S = 200$. For the class label of $F^i_{\text{Aug}}$, we have $\{\tilde{c}^i_{\text{Aug}}\} = \{c^i\}$, which means the augmented class labels $\{\tilde{c}^i_{\text{Aug}}\}$ is identical to the original class labels. Third, we add random Gaussian noise to each observation of $F^i_{\text{Aug}}$ to evaluate the robustness and the privacy budget (PB) of the trained model, then we have a noisy feature matrix subset $\tilde{F}^i_{\text{Aug}}$. A hyperparameter NSR, i.e., the noise-to-signal ratio in percentage, controls the noise intensity, defined as the ratio of noise power to signal power. Intuitively, the model performance $A$ can be deteriorated by increasing NSR, denoted by a function $A(\text{NSR})$. The definition of the PB is given:

$$\text{PB} = 100\% \times \max\left(\text{NSR} \middle| A(\text{NSR}) \geq \underline{A}\right) \tag{4.1}$$

where $\underline{A}$ denotes the lower bound of acceptable accuracy of the model, depending on specific application requirements.

Finally, we stack the augmented data of each class to get the augmented feature matrix $\tilde{\mathcal{F}}_{9S\times M} = \{\tilde{F}^i_{\text{Aug}}\}$ and the class label vector $\tilde{C}_{9S\times 1} = \{\tilde{c}^i_{\text{Aug}}\}$, where $i = \{1,2,\cdots,9\}$. We use 80% (the primary split) of the augmented data to generate the client samples as the training set. We use 40% (the secondary split) of the remaining augmented data as the testing set. Both primary and secondary splits are in a stratified manner to ensure samples required in each client are sampled. The detailed data split setting here is for illustration and can be modified to further investigate the minimum data sample requirement for collaborators.

### 4.3.2 Client Simulation and Client Model

The FL framework involves multiple collaborators, known as clients, to train a global model jointly. One client serves as a data contributor for the battery type sorting task in our setting. In this work, we simulate 10 clients, each possessing different classes and different observations of battery data. To be specific, each $\mathrm{Client}_k$ is defined over a triplet, i.e., $\mathrm{Client}_k \triangleq (\mathrm{lb}_k, \mathrm{ub}_k, \mathrm{NC}_k)$, $k = \{1,2,\cdots,10\}$, where $\mathrm{lb}_k$ and $\mathrm{ub}_k$ are the minimum and maximum number of observations in $\mathrm{Client}_k$. The value of $\mathrm{lb}_k$ and $\mathrm{ub}_k$ are set to 100 and 200 for all clients, respectively. $\mathrm{NC}_k$ stands for the minimum number of classes in each $\mathrm{Client}_k$, quantifying the level of client-wise heterogeneity (namely, the heterogeneity index). Then, random observations are subsequentially drawn from the augmented data $\{\tilde{\mathcal{F}}, \tilde{C}\}$ for the client based on the as-defined triplet.

The random forest is a decision-tree-based machine-learning algorithm, with each tree defined over a collection of random variables. Formally, for an $m$-dimensional feature vector $\mathbf{X} \triangleq [x_1,\cdots,x_m]^T = \tilde{\mathcal{F}}$, and a response vector $\mathbf{Y} \triangleq \tilde{C}$, the goal is to learn a prediction function $g(\mathbf{X})$ for predicting $\mathbf{Y}$. The prediction function $g(\mathbf{X})$ is determined by minimizing the expectation of the loss function $L$:

$$E_{XY}\left[L\big(\mathbf{Y}, g(\mathbf{X})\big)\right] \tag{4.2}$$

where, the subscripts denote expectations on the joint distribution of $\mathbf{X}$ and $\mathbf{Y}$.

The $j$-th decision tree, or the $j$-th base learner, is denoted as $h_j(\mathbf{X};\ \Theta_j)$, where $\Theta_j$ parameterizes a random collection of a set of random variables of $\mathbf{X}$. In the sorting, a class assignment rule to every leaf node $t$ considering a zero-one loss function gives:

$$h_j\big(\mathbf{X};\ \Theta_j\big) = \underset{y\in\tilde{C}}{\mathrm{argmax}}\, p(\, y \mid t\,) \tag{4.3}$$

where, we pick the class with the maximum posterior probability.

The random forest constructs $g$ by learning a series of base learners, $h_1(\mathbf{X};\ \Theta_1),\cdots,h_J(\mathbf{X};\ \Theta_J)$. These base learners are combined to give an ensembled prediction function $g$, determined by the most frequently predicted classes:

$$g(\mathbf{X}) = \underset{y\in\tilde{C}}{\operatorname{argmax}} \sum_{j=1}^{J} I\left(y = h_j(\mathbf{X};\ \Theta_j)\right) \tag{4.4}$$

where, $I$ is the indicator function. $I\left(y = h_j(\mathbf{X};\ \Theta_j)\right) = 1$ if $y = h_j(\mathbf{X};\ \Theta_j)$ and 0 otherwise. The number of trees in each random forest is fixed at ten, i.e., $J = 10$ for a balanced classification accuracy and computation cost. We deliberately let collaborators (clients) learn the most suitable random forest structure, i.e., the model parameters, by themselves, rather than fixing the parameters since each collaborator could have very different battery numbers and cathode material types. By only presetting the number of trees in the random forest, the collaborators could have enough flexibility to train the best model that suits their own data distribution. The bottom-level random forest algorithm (client model) is implemented using readily available MATLAB packages, more specifically, the TreeBagger function in the Statistics and Machine Learning Toolbox. The MATLAB version is R2022a, and the code runs on a personal computer with Intel (R) Core (TM) i5-10400 CPU @ 2.90GHz RAM 8 GB.

### 4.3.3 Voting Strategy Design

In the proposed FL framework, local random forests are first trained on each client with its own local data in a parallel fashion. Then the local client models are aggregated into a global model by means of a proper voting strategy from the local sorting results. In our work, battery class distribution across clients can be heterogeneous, which brings difficulty in aggregating the biased client models. To this end, we propose a Wasserstein distance voting (WDV) method to aggregate the client models rather than the traditional majority voting (MV). Our model aggregation method is robust to heterogeneous class distributions across clients. The core idea of the WDV is to reduce the weightings of clients whose observations are similar to ones in the global model. The Wasserstein distance measure is defined as:

$$W_q(\cdot,\cdot) = \left( \inf_{\gamma \in \mathrm{MP}} \int_{\Omega_1 \times \Omega_2} |x_1 - x_2|^q d\gamma(x_1, x_2) \right)^{\frac{1}{q}} \tag{4.5}$$

where $\gamma$ is a transport operator, referring to the transport of arbitrary attributes pairs, i.e., $(x_1, x_2)$, from the global feature space $\Omega_1$ to the client feature space $\Omega_2$. MP stands for a measurably preserved transport.

The WDV term $\omega$ is defined as:

$$\omega_k = \alpha^{-\lambda(1-\mathcal{M}_k(W_q))} \tag{4.6}$$

where, $\alpha > 0$ and $\lambda > 0$ are voting hyperparameters. $\mathcal{M}_k$ is the average operator on the pairwise Wasserstein distance between feature spaces of $\mathrm{Client}_k$ and the global feature space, i.e., the recycler.

The aggregated global model $G(x)$ can be obtained as:

$$G(x) = \underset{y \in \tilde{C}}{\mathrm{argmax}} \sum_{k=1}^{K} \omega_k I\big(y = g_k(x)\big) \tag{4.7}$$

where, $K$ is the number of clients. $I$ is the indicator function. $I\big(y = h_j(\mathbf{X};\, \Theta_j)\big) = 1$ if $y = h_j(\mathbf{X};\, \Theta_j)$ and 0 otherwise. Finally, to maximally protect data privacy, the local votes are properly encrypted before being uploaded to the recycler. Most encryption methods, e.g., secure Hash Algorithms, shall suffice without compromising the sorting accuracy. The high-level FL framework, including the Wasserstein-distance voting and the transfer of parameters, is implemented from scratch.

### 4.3.4 Feature Importance

We use permutation importance to measure the feature importance in the client model, i.e., the random forest algorithm. The core idea of the permutation importance is to use out-of-bag data to examine the effect of feature permutation using the trained random forest model. In the first step, a prediction is made on several observations of the out-of-bag data. In the second step, the feature $\theta_m, m = 1,\cdots,m$ is randomly permutated for observations of out-of-bag data, then the modified out-of-bag data is passed down each tree in the random forest. In the first and second steps, two predictions are made by the trained random forest model, i.e., $\hat{C}$ and $\hat{C}^*$. The permutation feature importance of $\theta_m$ is defined as:

$$G\mathrm{Imp}_n^m = \frac{1}{J_n}\sum_{j\in\mathcal{J}_n} I\left(y_n \neq \hat{C}_{n,j}^*\right) - \frac{1}{J_n}\sum_{j\in\mathcal{J}_n} I\left(y_n \neq \hat{C}_{n,j}\right) \tag{4.8}$$

where, $\mathcal{J}_n$ is the cardinality of the $n$ out-of-bag observations, $J_n$ is the number of trees in the random forest considering $n$ out-of-bag observations. The feature importance of $\mathrm{Imp}_n^m$ is averaged over all observations as a global importance in the client model. Similarly, feature importance in the FL framework is averaged over all clients.

### 4.3.5 Evaluation Metrices

We use a one-vs-all prediction strategy to predict a multi-class classification problem, such that the original problem is converted into several binary classification problems. The accuracy of each binary classification sub-problems is defined as follows:

$$\text{accuracy} = \frac{\text{Number of correct predictions}}{\text{Total number of predictions}} = \frac{\mathrm{TP}+\mathrm{TN}}{\mathrm{TP}+\mathrm{FN}+\mathrm{FP}+\mathrm{TN}} \tag{4.9}$$

where, $\mathrm{TP}, \mathrm{FP}, \mathrm{FN}, \mathrm{TN}$ refer to the number of true positive, false positive, false negative, and true negative predictions.

The prediction accuracy for the multi-class classification problem gives:

$$\text{Accuracy} = \frac{1}{\text{NC}}\sum_{i=1}^{\text{NC}} \text{accuracy}_i \tag{4.10}$$

where, $i$ is the class label, NC is the number of battery classes.

The accuracy could not provide an adequate model evaluation when classification samples are imbalanced, i.e., heterogeneous data distribution. Thus, the F1-score is used, whose definition is as follows:

$$\text{F1} = \frac{2 \times \text{Precision} \times \text{Recall}}{\text{Precision} + \text{Recall}} \tag{4.11}$$

where, Precision=TP/(TP+FP) and Recall=TP/(TP+FN).

## 4.4 Collaborative Material Sorting Model

In this section, we elaborate on the model performance under different data access and evaluate the physical meanings of salient features.

### 4.4.1 Homogenous Data Access

We first consider a setting where the battery data are homogeneously distributed across the collaborators (namely, the clients). The homogeneity means that each client offers to share the battery data across all 9 classes, even though the specific number of batteries is not restricted. The detailed configuration can be found in Table 4. 2.

We train our FL model without requiring information on the historical use of the retired batteries. In this case, the recycler and the clients only need to test the retired batteries at the current (field-testing) cycle, specifically, with a complete charging-discharging cycle for a standard feature engineering process initiated by the recycler. Local models are trained based on features extracted from their private battery data. The FL framework aggregates the local model parameters, rather than the private battery data.

Table 4. 2 Homogeneous data access.

| Client | Class (Battery number) | Total |
|---|---|---|
| 1 | 1 (10), 2 (15), 3 (7), 4 (18), 5 (17), 6 (10), 7 (10), 8 (12), 9 (10) | 9 (109) |
| 2 | 1 (18), 2 (17), 3 (15), 4 (19), 5 (16), 6 (16), 7 (15), 8 (19), 9 (11) | 9 (146) |
| 3 | 1 (11), 2 (12), 3 (11), 4 (20), 5 (14), 6 (12), 7 (9), 8 (9), 9 (8) | 9 (106) |
| 4 | 1 (16), 2 (18), 3 (14), 4 (17), 5 (15), 6 (12), 7 (14), 8 (12), 9 (10) | 9 (128) |
| 5 | 1 (13), 2 (14), 3 (21), 4 (14), 5 (20), 6 (19), 7 (25), 8 (23), 9 (21) | 9 (170) |
| 6 | 1 (27), 2 (16), 3 (16), 4 (19), 5 (16), 6 (18), 7 (19), 8 (19), 9 (19) | 9 (169) |
| 7 | 1 (15), 2 (20), 3 (19), 4 (21), 5 (18), 6 (13), 7 (23), 8 (17), 9 (12) | 9 (158) |
| 8 | 1 (21), 2 (19), 3 (25), 4 (20), 5 (17), 6 (17), 7 (18), 8 (17), 9 (12) | 9 (166) |
| 9 | 1 (24), 2 (12), 3 (25), 4 (15), 5 (19), 6 (25), 7 (15), 8 (26), 9 (18) | 9 (179) |
| 10 | 1 (20), 2 (11), 3 (12), 4 (17), 5 (11), 6 (6), 7 (15), 8 (11), 9 (11) | 9 (114) |
| | Total battery number | 1445 |

Figure 4. 6 compares two FL methods, i.e., the MV and our proposed WDV, with the independent learning (IL) paradigm. It should be noted that the accuracy for the IL is averaged over all clients in a non-federated manner. Compared with the IL, the MV does not sacrifice sorting performance, with an average accuracy of 95%, while being capable of protecting data privacy and mitigating computational burden. However, 3 classes are missorted using the MV. For instance, 3 batteries in NMC (SNL, class 8, 15 in total) are missorted into NCA (SNL, class 7), resulting in a sorting accuracy of 80%. The sorting accuracy for NCA (UL-PUR, class 9) is 81%, with 2 batteries missorted into NMC (MICH_Form, class 4) and 1 battery missorted into NMC/LCO blended type (HNEI, class 2), respectively. In contrast, the WDV outperforms the MV since it only missorted one battery, resulting in a sorting accuracy of 99%. We also evaluate the prediction probability of each class for the MV and WDV, respectively. It turns out that the WDV makes a more confident sorting than the MV since the prediction probabilities of the WDV are generally right-skewed to a higher probability value. Therefore, our proposed WDV produces higher sorting accuracies across all classes, and the sorting is of richer probability confidence margins.

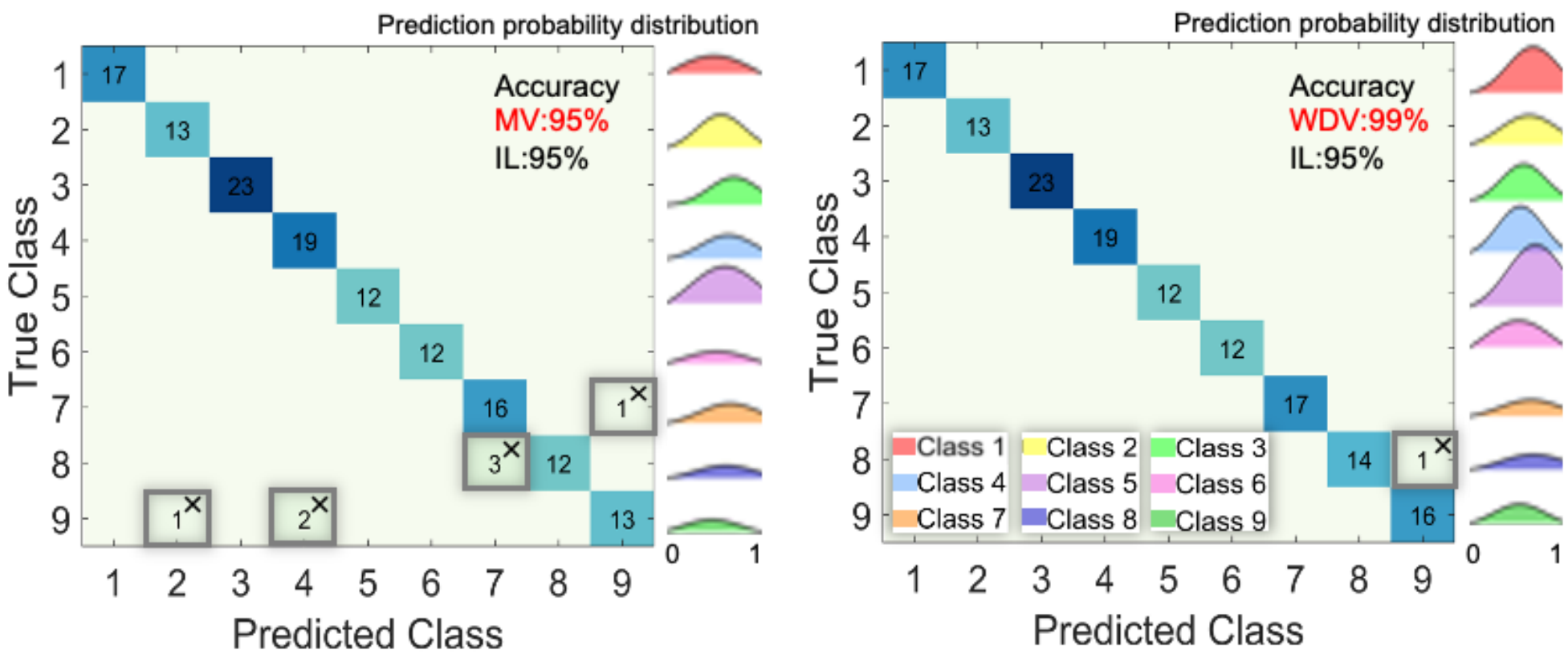

Figure 4. 6 Material sorting results with homogenous data access.

We also evaluate the PB, considering that client data might be vulnerable to reverse engineering by eavesdropping on private data [205]. In this regard, we add random Gaussian noise to the client data with different intensities. The intensity of the randomness is controlled by a noise-to-signal ratio (NSR), ranging from 1% to 10%.

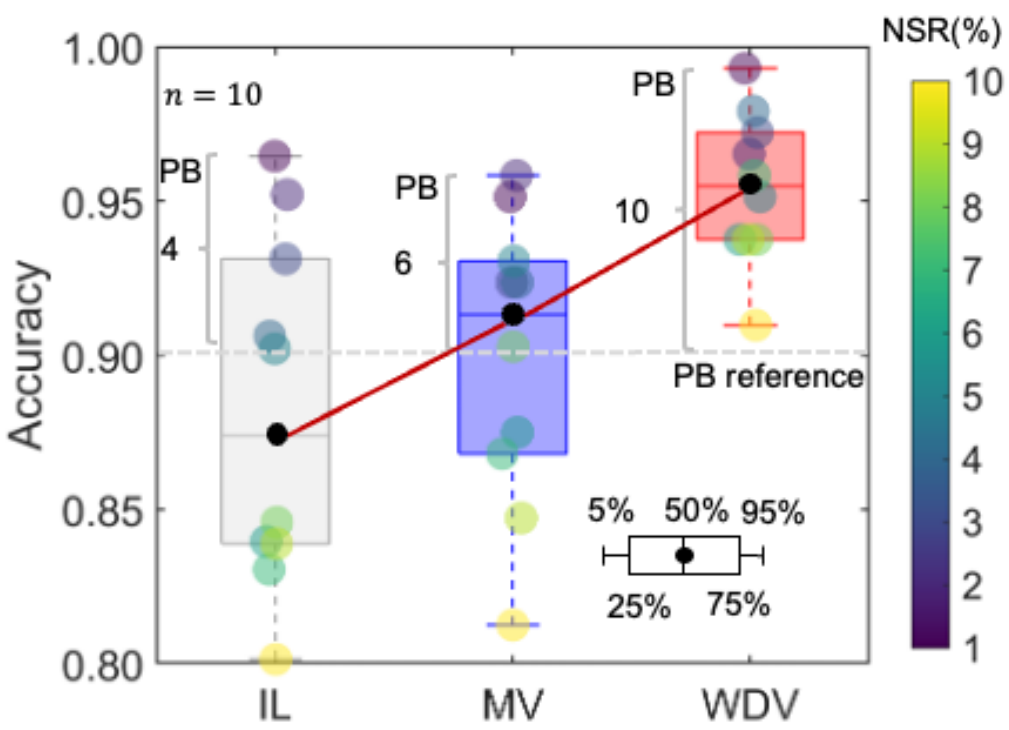

Figure 4. 7 Material sorting results with data privacy budgets.

Figure 4. 7 shows the accuracy and PB comparison when using IL, MV, and WDV, respectively. The sorting accuracy of the MV decreases from 95% to 82%, similar to that of the IL when the noise intensity increases from 1% to 10%. In this noise range, the median sorting accuracy of the MV and IL is 92% and 86%, respectively. In contrast, the sorting accuracy of the WDV is still above 90% in the presence of 10% noise, which is a stringent noise level in practical cases. The WDV has a median sorting accuracy of over 95% in the same noise range. Taking an average sorting accuracy of 90% as an acceptable

reference level, the PB values of IL, MV, and WDV are 4, 6, and 10, respectively. Therefore, applying FL produces a more privacy-secure sorting than IL, hence being capable of preventing data eavesdropping. Furthermore, our proposed WDV is more accurate and performs much better (with nearly doubled PB values) in the privacy-accuracy trade-off than the MV. In addition, the robustness to stringent noise when using the WDV also implies a good tolerance of the battery measurement requirement, reducing the expensive battery testing disbursement.

Noticing that a high sorting accuracy does not necessarily imply an acceptable sorting for a specific class, we also consider within-class sorting performances. Figure 4. 8 shows the F1-score and PB of the IL, MV, and WDV in each predicted class. The result shows that the IL has smaller F1-scores than the FL manner in all the classes, making poor sorting. Regarding FL, WDV outperforms the MV in each predicted class by producing higher average F1 scores. The deviation range of the F1 scores for WDV is smaller than that of the MV, indicating that the WDV is more robust (See APPENDIX O). Therefore, our proposed WDV not only has a better overall sorting accuracy among all nine classes but also within each class, compared with the MV.

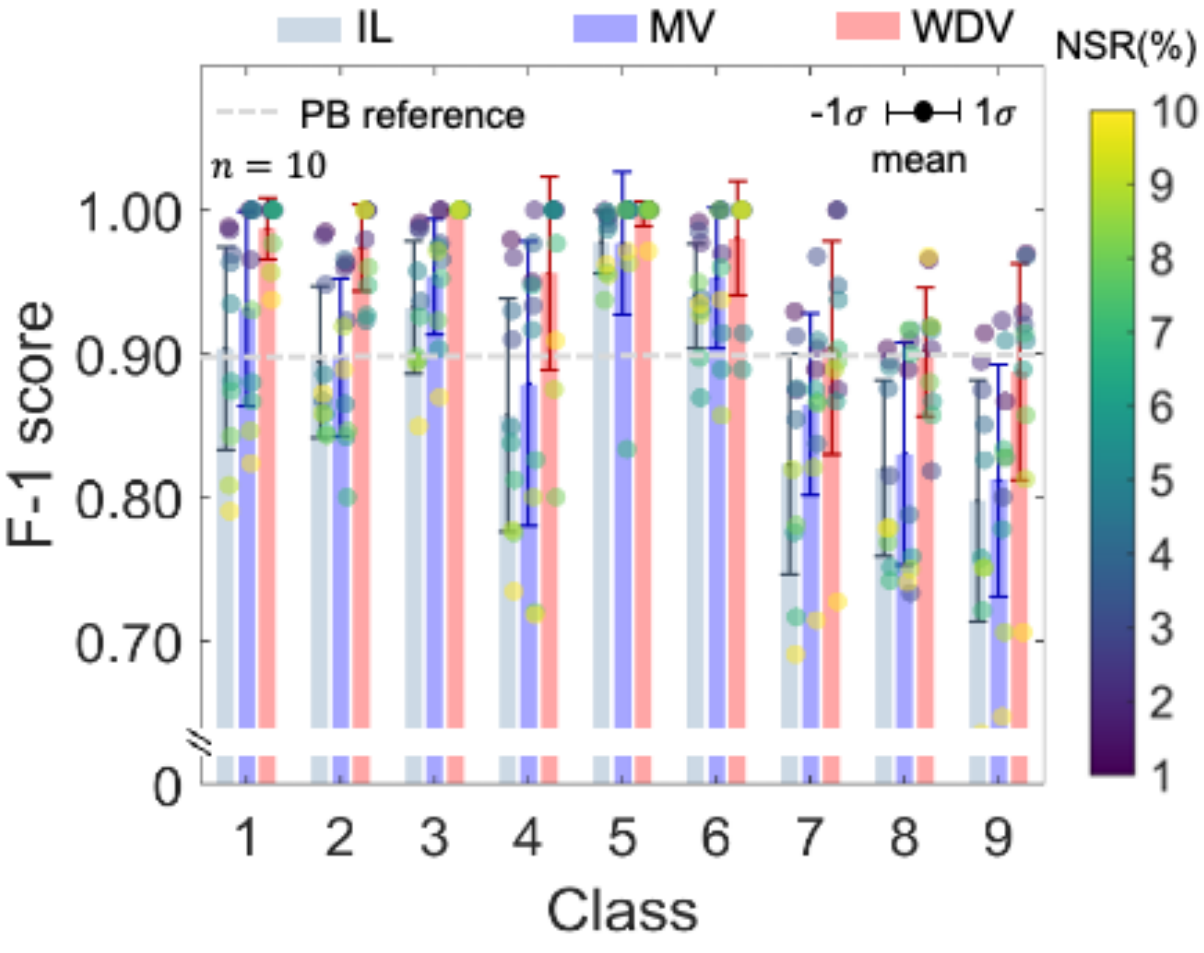

Figure 4. 8 F1-score of different classification methods.

Table 4. 3 The privacy budgets when referenced at a 0.9 F1-score level.

| Method | Non-federated | Federated | |
|---|---|---|---|
| Class | IL | MV | WDV |
| LCO (CALCE, Class1) | 5 | 6 | 10 |
| NMC\ LCO (HNEI, Class2) | 5 | 5 | 10 |
| NMC (MICH-Expa, Class3) | 8 | 9 | 10 |
| NMC (MICH-Form, Class4) | 2 | 6 | 9 |
| LCO (Ox, Class5) | 10 | 10 | 10 |
| LFP (SNL, Class6) | 8 | 8 | 9 |
| NCA (SNL, Class7) | 2 | 2 | 5 |
| NMC (SNL, Class8) | 2 | 4 | 6 |
| NCA (UL-PUR, Class9) | 0 | 2 | 6 |
| Average | 4.67 | 5.78 | 8.33 |
| Increased PB (%) using WDV | + 78% | + 44% | - |

Regarding the PB value when using the non-federated IL, referenced at a 0.9 F1-score level, is significantly lower than the federated way across all classes, demonstrated in Table 4. 3. This indicates a more severe risk of data leakage for IL compared with FL. When further applying our proposed WDV, the private budget can increase by 78% and 44% compared with the non-federated IL and the federated MV, respectively. The results demonstrate that the WDV successfully leverages the battery-chemistry-related insights hidden in clients while effectively preserving client data privacy.

### 4.4.2 Heterogeneous Data Access

We also consider an extreme, while a more actual situation where the data can be exclusively scattered among clients, i.e., the data distribution is heterogeneous. In this situation, the heterogeneity issue poses more challenges to battery type sorting since the clients are prone to train biased models and deteriorate global accuracy, which is still an open question in FL. In this section, we explore a more challenging situation rather than having homogeneous data access among each client. In contrast to the homogeneous data distribution among clients, the heterogeneous data distribution is more compatible with the real-world situation. We consider this situation by running the Client simulation process when setting the $NC_k$ to 2. It means the minimum number of battery classes, namely the heterogeneity index in each client, is two to nine. The possible combinations of the clients under different random situations are also considered using the Monte Carlo method. Specifically, the Client simulation process is run with 50 Monte Carlo trials. The random seed is retrieved using a sequence from 50 to 100 in MATLAB 2022a with the "v5uniform" method. Since we simulate the heterogeneity index from two to nine, there are in total of 400 trials in our heterogeneous data distribution setting. For each trial, the number of classes and batteries in one specific class are randomly shuffled. The model performance evaluations are conducted for each trial. We demonstrate that our FL framework can still classify retired batteries based on the standard feature engineering process at the current (field-testing) cycle without any knowledge of the previous operation conditions.

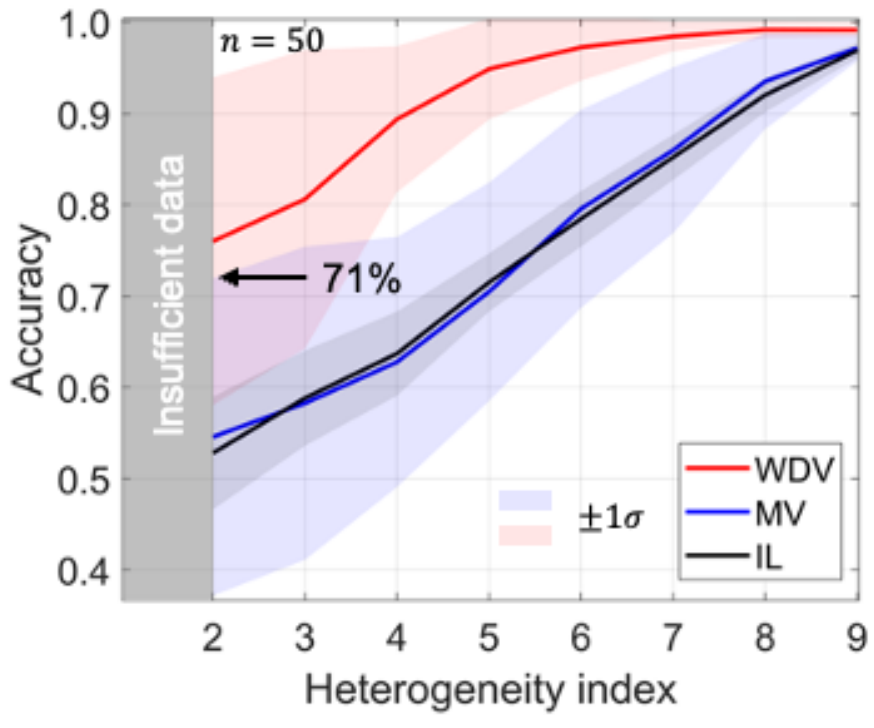

Figure 4. 9 Sorting accuracy as a function of heterogeneity index.

We consider the heterogeneity index, defined as the minimum number of battery classes for each client in each Monte Carlo simulation run. A higher heterogeneity index indicates a less heterogeneous battery data distribution. The heterogeneity index is no smaller than two such that one client can train a local model for a sorting task. Figure 4. 9 shows average sorting accuracy when the heterogeneity index varies. The average accuracies are plotted with solid lines, with the $\pm 1$ STD range indicated in the shaded region. As the heterogeneity index decreases from 9 to 2, the performance of the MV and the IL rapidly deteriorates at a sublinear rate. The average sorting accuracy of the MV is 0.55, slightly better than the IL, equivalent to a random guess when the heterogeneity level is two. This observation shows that the MV can help little to aggregate the local models under heterogeneous data access. In contrast, the WDV outperforms its MV counterpart in all heterogeneity levels, successfully mitigating the heterogeneous data distribution issue. Moreover, the WDV shows an interesting asymptotic effect when the heterogeneity index increases. This indicates that the WDV can potentially support optimal allocation of client battery data to reduce collaboration cost in practical situations.

Under heterogeneous data distribution setting in Table 4. 4, we further compare the class-wise and client-wise sorting performance of the MV and the WDV to the non-federated IL with two considerations: (1) the significance of our FL framework and (2) why our proposed WDV outperforms the MV.

Table 4. 4 Heterogeneous data access.

| Client | Class (Battery number) | Total |
|---|---|---|
| 1 | 2 (41), 3 (41), 4 (34), 5 (36), 6 (45) | 5 (197) |
| 2 | 1 (20), 2 (11), 4 (12), 5 (15), 6 (16), 7 (12), 8 (10), 9 (17) | 8 (113) |
| 3 | 1 (15), 3 (22), 4 (14), 5 (16), 7 (21), 9 (23) | 6 (111) |
| 4 | 2 (27), 3 (23), 5 (33), 6 (25), 7 (22), 9 (29) | 6 (159) |
| 5 | 2 (82), 8 (79) | 2 (161) |
| 6 | 2 (38), 5 (33), 6 (39), 8 (36), 9 (26) | 5 (172) |
| 7 | 2 (58), 4 (43), 8 (49), 9 (45) | 4 (195) |
| 8 | 1 (20), 2 (25), 3 (17), 5 (20), 6 (20), 7 (16) | 6 (118) |
| 9 | 1 (40), 3 (51), 9 (49) | 3 (140) |
| 10 | 1 (26), 2 (23), 3 (27), 4 (25), 5 (20), 7 (21), 9 (27) | 7 (169) |
| | Total battery number | 1535 |

First, we evaluate the client-wise sorting accuracy, shown in the lower side of Figure 4c. Client 5 achieves an average sorting accuracy of 25%, ranking last among all clients.

Meanwhile, client 2 achieves an average sorting accuracy of 86%, ranking first among all clients. However, the average sorting accuracy is only 55%, close to a random guess. Therefore, the client performance using the non-federated IL depends heavily on data access as shown in Figure 4. 10. In fact, without our FL framework, the battery recycler is equivalent to a single client, and the battery recycler can only make sorting on the battery types stored in its local database. This non-federated paradigm could not handle various types of retired batteries if the recycler did not build a database covering all the battery types it would handle. With our FL framework, the recycler can collaborate with several clients, even if under heterogeneous data situations.

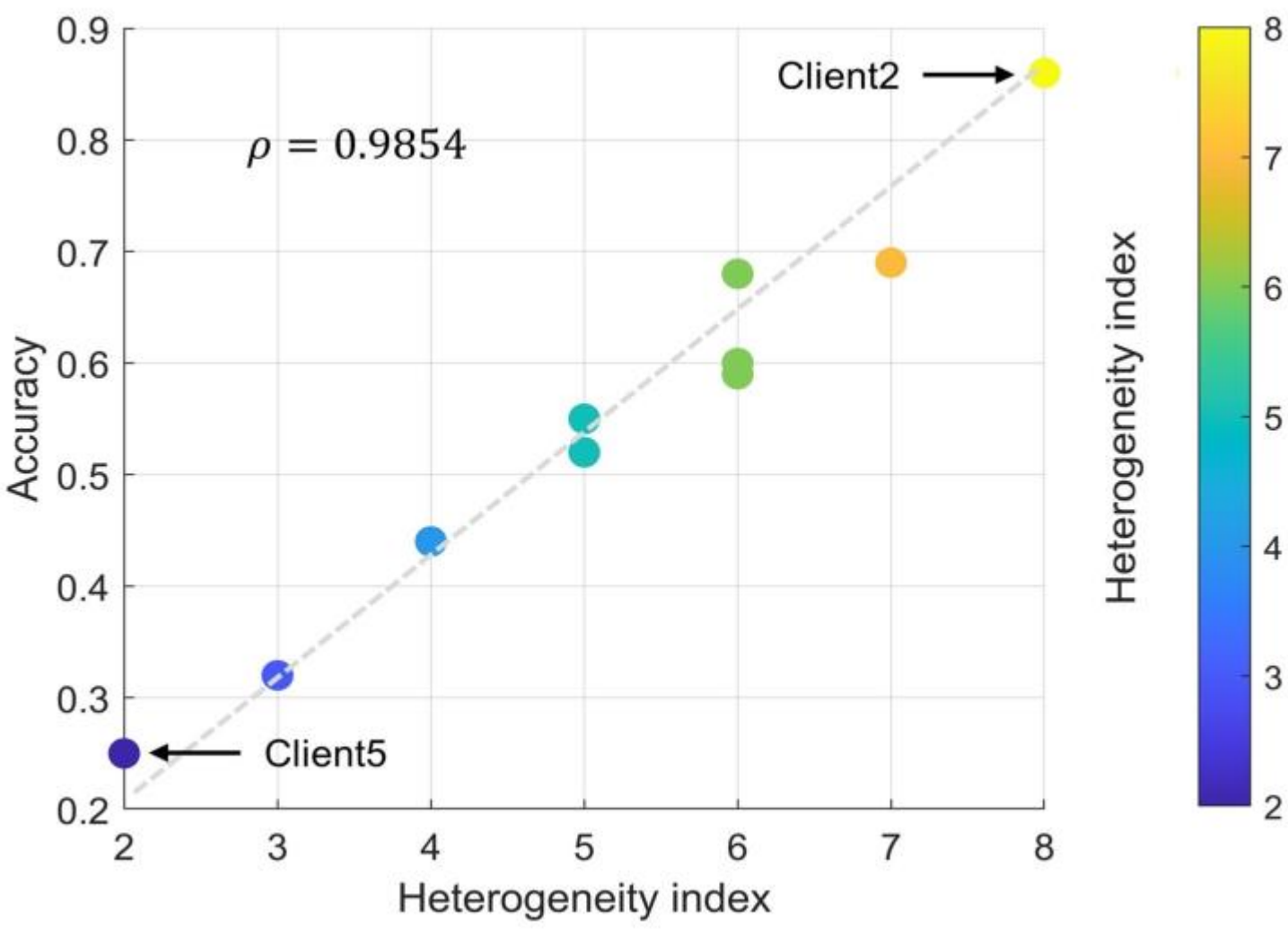

Figure 4. 10 The correlation between heterogeneity and classification accuracy.

We turn to analyze how to collaborate with clients under heterogeneous data access settings. The upper part of Figure 4. 11 shows the class-wise accuracy of the MV and WDV. It is noticed that the average sorting accuracy after using the MV is better than the non-federated way, which is 79%, as indicated in the lower side. It demonstrates the success of applying the FL framework to address the heterogeneous data distribution issue in this case. However, the MV totally missorted LFP (SNL, class 6) and NMC (SNL, class

8) with zero accuracy. The failure of the MV in specific classes can be rationalized by its core idea of giving more weight to the clients who contribute more battery samples while not guaranteeing diversity in battery types. For instance, the contribution of client 7 will be strengthened by the MV due to a large number of batteries (specifically, 195 augmented batteries, ranking second among clients), despite only contributing four classes of batteries. As a result, the MV will lead the aggregated model to be biased towards large client such as client 7 in Table 4.4.

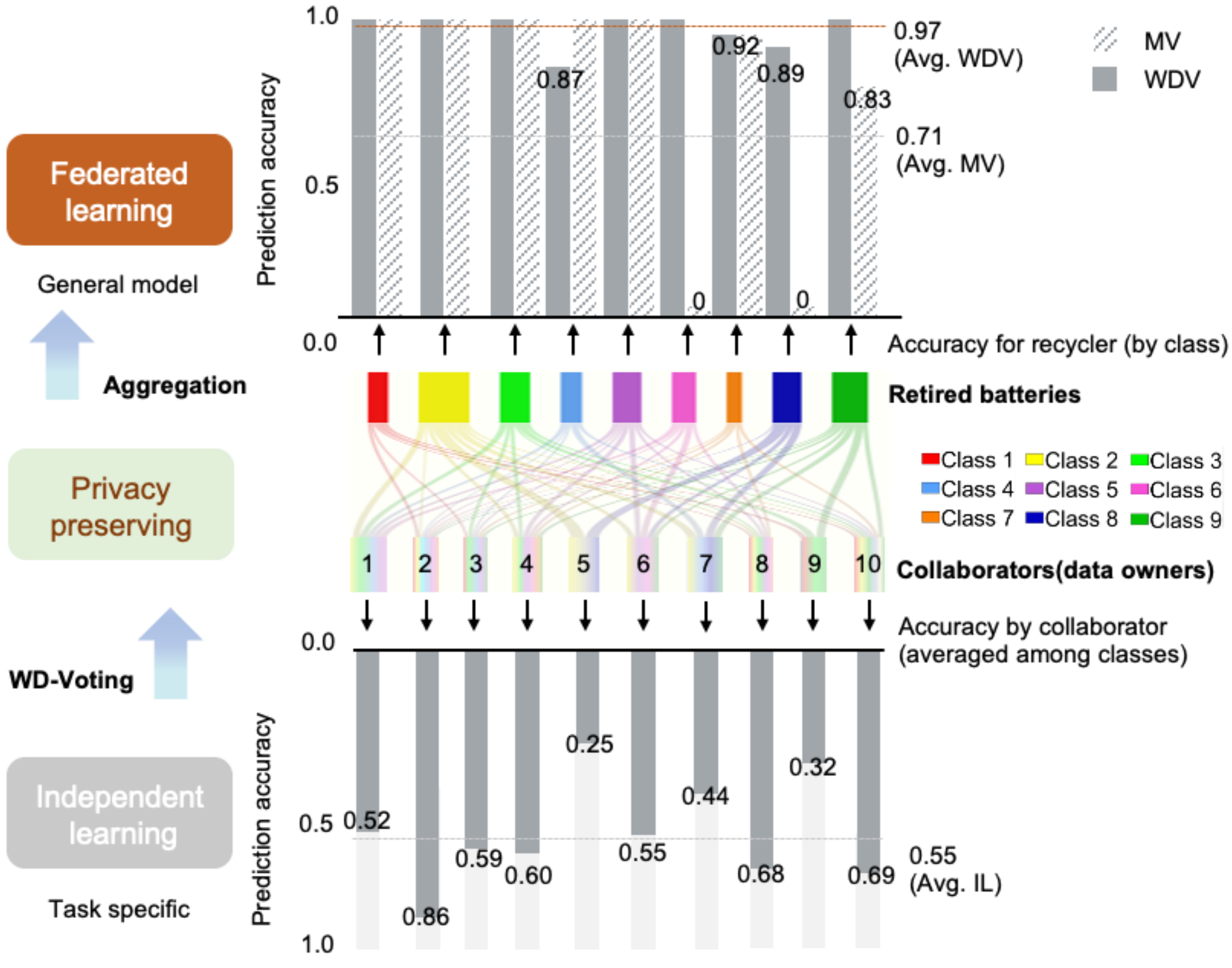

Figure 4. 11 Class-wise (upper part) and client-wise (lower part) sorting accuracy.

The biased phenomenon is evidenced by the as-described zero sorting accuracy for LFP (SNL, class 6) since the large client, such as client 7, never contributed any batteries in class 6. Similarly, client 1, the largest client with 197 augmented batteries, failed to contribute helpful information to the recycler regarding classifying NMC (SNL, class 8), which is consistent with zero accuracy in class 8. In contrast to the MV, our proposed

WDV focuses on the battery similarities between the recycler and each client by measuring the pairwise distance. We aim to assign fewer weightings to the clients with biased data distributions (equivalently, a higher heterogeneity), whose batteries are of higher similarities with the recycler, such that the recycler can have generalized information from each client. The results show that our proposed WDV successfully leverages helpful information from heterogeneous data distribution among clients. The WDV achieves 100% and 89% sorting accuracy for the otherwise missorted batteries in LFP (SNL, class 6) and NMC (SNL, class 8), respectively. The overall sorting accuracy using the WDV is up to 97%, with only 5 batteries missorted out of 144 samples.

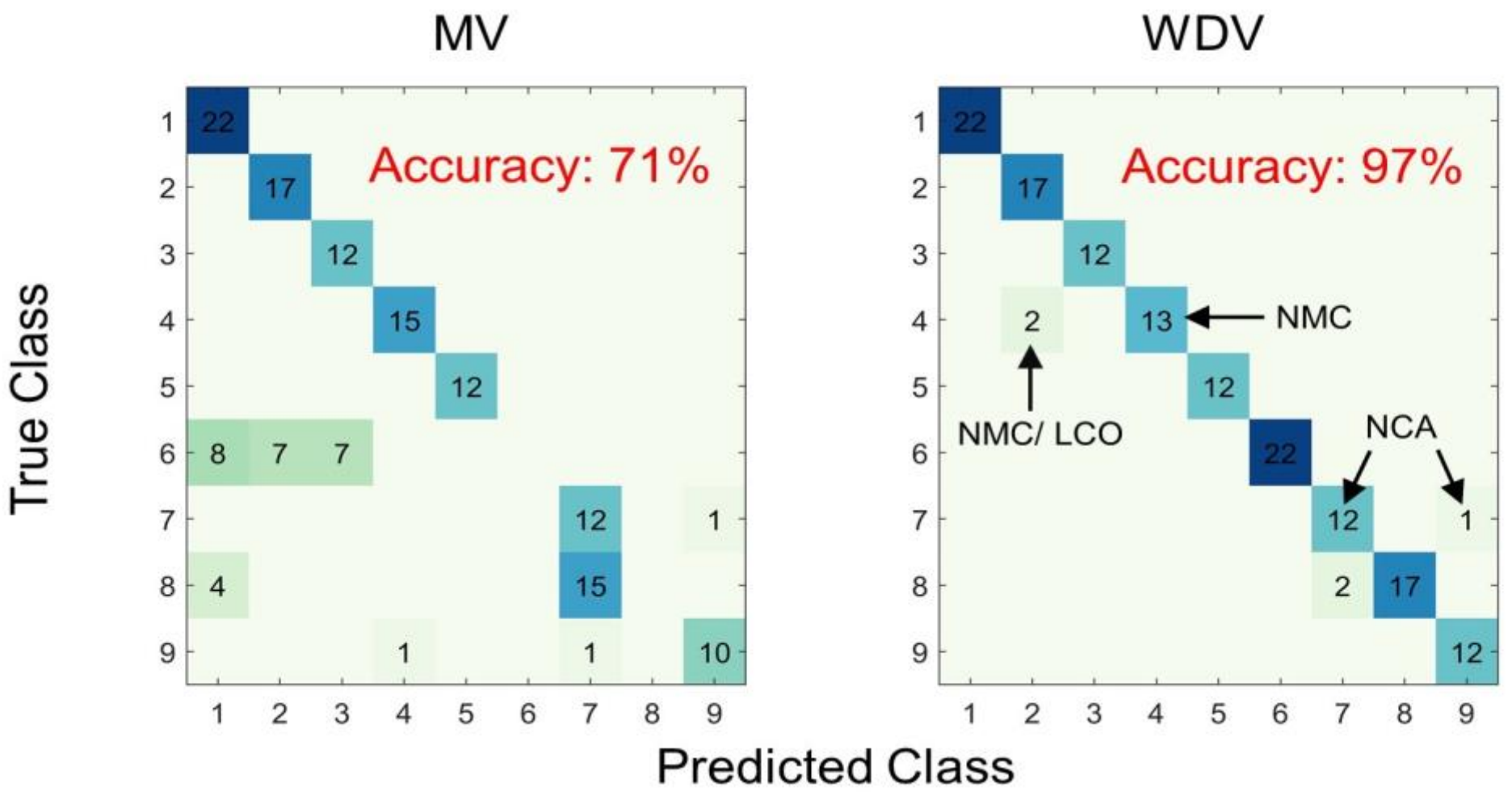

Figure 4. 12 The confusion matrix for the MV and WDV.

In Figure 4. 12, we also notice that the missorted batteries are of similar cathode materials. Specifically, 2 batteries with NMC cathode material were missorted into the NMC/LCO blended type; while 1 battery with the NCA cathode material was correct in material type but missorted into another manufacturer. On the contrary, the missorted results produced by the MV can spread to either many irrelevant classes or manufacturers. Therefore, we conclude that WDV can aggregate helpful client insights by distinguishing inherited differences in cathode material types. Inspired by this, the WDV also suggests that the clients are encouraged to contribute more battery data in diversity rather than more data in some specific classes. The recycler can optimize the benefit distribution based on helpful client information provided. Thus, our FL framework enables the recycler to know the battery cathode material type, even if without their own data access to various battery data, while preserving data privacy of potential clients.

### 4.4.3 Partial Missing Data

The framework is implemented under the ideal assumption that all collaborators are fully cooperative, such that the uploaded local results are assumed to be reliable. In Figure 4. 13, despite such an ideal assumption, the random forest model, incorporated with the WDV, is naturally robust against random parameter transfer losses even if parameters from a few collaborators end up missing. Note that the receiving end treats the lost client model sorting result as a random guess (a 0.5 probability).

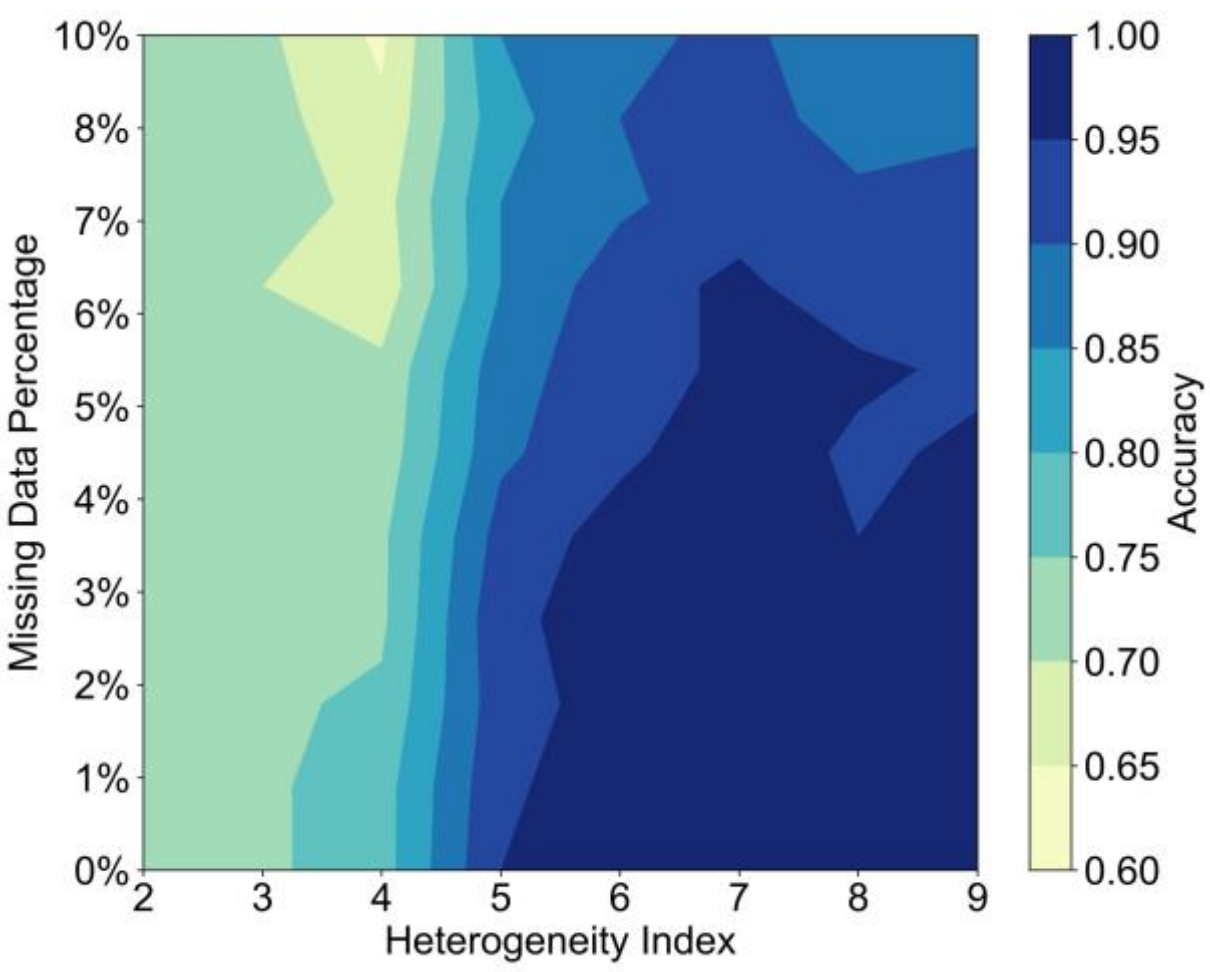

Figure 4. 13 Model performance under unrecoverable parameter loss.

The sorting accuracy only slightly degrades given the same heterogeneity setting. In the case of a blockchain-like environment with numerous collaborators of unknown trustworthiness and reliabilities, our research could be extended in search of a proper incentivization mechanism such that all recycling collaborators would fully contribute to their respective local model instead of attempting to become total free-riders. We prospect quantifying the helpful information that the recycling collaborators provided to design a benefit distribution strategy and a free-rider detection scheme to make the federated battery recycling ecosystem economically feasible.

### 4.4.4 Feature Importance Analysis

We then interpret our FL model by evaluating the most salient features correlated with battery cathode chemistry. Fig. 4.14 shows the importance of the features in descending order. The error bar indicates the importance deviation. Features F1 and F16 rank the top two features regarding out-of-bag importance. Interestingly, these two features have a clear physical interpretation of the battery dynamics, which we will further discuss in later sections. Here, we rationalize these two features by plotting the grouped battery samples in the feature space spanned by features F1 and F16.

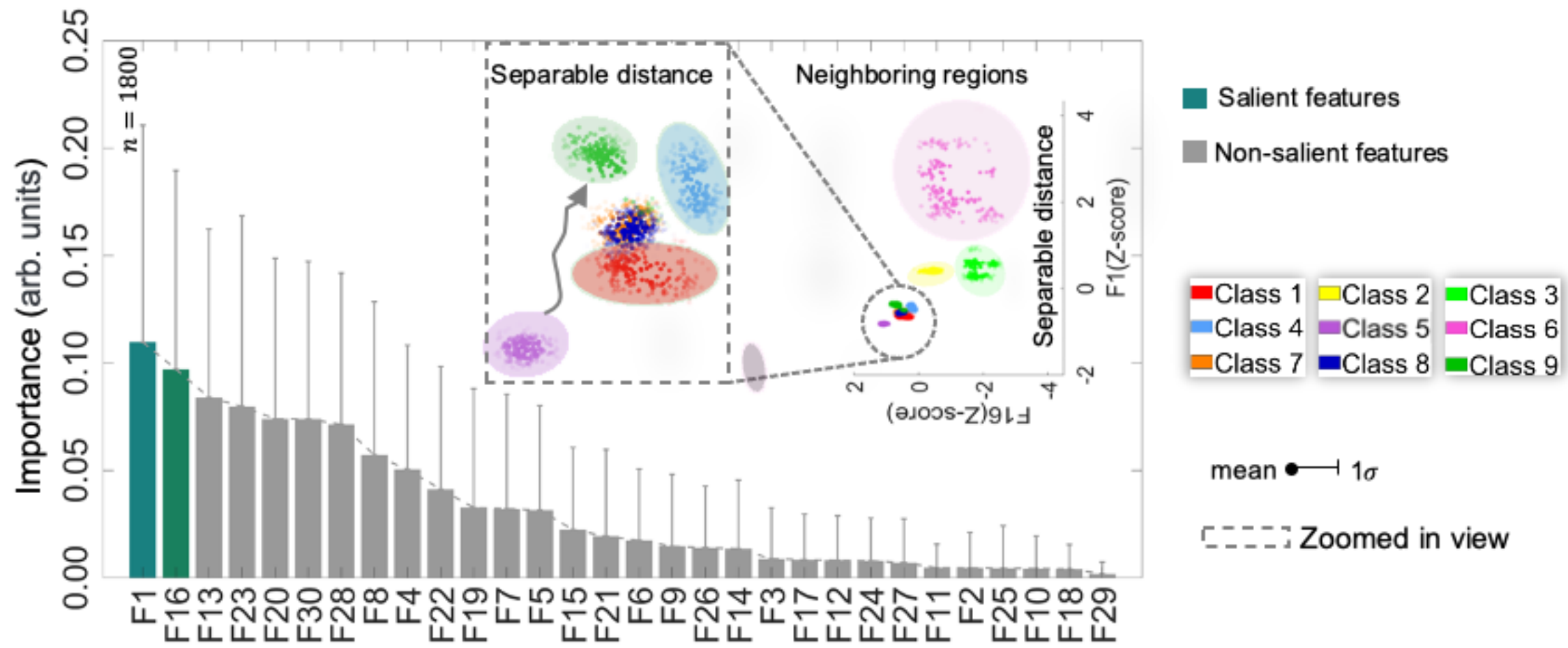

Figure 4. 14 Feature importance rationalization.

The subplot of Figure 4. 14 shows that NMC/LCO blended type (HNEI, class 2), NMC (MICH_Expa, class 3), and LFP (SNL, class 6) (sharing the color with Figure 3a) are clearly separable in the spanned feature space. For the remaining classes, the batteries are still separable (see the zoomed-in view), though in relatively more minor grains. On the contrary, the non-salient features have a relatively weaker sorting ability due to the non-separable feature space (See APPENDIX P). As a result, our FL framework successfully discovered useful mechanism insights to guarantee sorting accuracies. Such an insight could be further extended to simplify the model for light computation, hence less investment. Once the client models classify the batteries, the recycler can aggregate the client results to make a final decision on the battery cathode material types underpinned by the salient features.

Here we aim to rationalize the physical interpretation of these two salient features, i.e., F1 and F16. Features F1 and F16 are extracted from the dQ/dV curve, commonly used to analyze phase reactions in electrochemistry, though agnostic to underlying mechanisms. Regarding battery thermodynamics, the number of dQ/dV peaks and the corresponding voltage values can be used to analyze reaction on electrodes and to judge the composition of the cathode material. Regarding battery kinetics, the shape of the dQ/dV curve can help analyze the transport capacity of electrons and ions inside the battery, from which the chemical properties of battery materials can be deduced. Here, the Gibbs phase law can further help the rationalization: $F=C-P+n$, where F represents the degree of freedom, C represents the number of independent components, P represents the number of phase states, and n represents external factors. When studying electrode materials, constant temperature and pressure are assumed; thus, $n=0$. The number of independent components on the cathode is $C=2$. Since the discharging process of LFP is a phase change process, there are two phases, i.e., $P=2$. Since LCO, NCM, and NCA are solid solutions, only one phase exists during the discharging process i.e., $P=1$. Therefore, the degree of freedom of LFP ($F=0$) is lower than that of LCO, NCM, and NCA ($F=1$). As a result, the voltage of LFP does exhibit significant change during the reaction process, consequently, there is a noticeable peak on the dQ/dV curve. On the comparison, during the charging and discharging process of LCO, NCM, and NCA ($F=1$), the slope of the voltage change in the plateau is more significant than that of LFP, which can be reflected on the dQ/dV curve accordingly. Although LCO, NMC, and NCA ($F=1$) have similar structures, their components, and Li-ion mobility during the charging-discharging process differ, resulting in different dQ/dV peak values, which can be interpreted from F1 and F16. While other features are possible to decipher battery kinetics, they demonstrate less importance since more complicated expert knowledge is required for further processing. However, we highlight the power of our machine learning model by automatically utilizing the information provided by F1 and F16, which have a clear physical interpretation and underpin a general and high-accuracy model. Such good accuracies are independent of historical usages and use only one cycle of end-of-life charging and discharging data. Consequently, the battery recycling collaborators realize good sorting accuracies with our proposed salient features aided by machine learning.

## 4.5 Impact of Mixed Material on Recycling Cost and Profit

To help understand the relevance and necessity of battery sorting in actual recycling practice, also to verify the significance of our proposed WDV strategy, an economic evaluation is performed. Three recycling methods (pyrometallurgy, hydrometallurgy, direct recycling), two battery cathode types (LFP-graphite and NMC-graphite), two recycling modes (individual, hybrid), three sorting accuracy levels (97%, 71%, 55%) induced by the federated and non-FL methods (WDV, MV, IL) are included in the evaluation. The detailed calculation procedure and results can be found in APPENDIX Q. Figure 4. 15 shows a schematic diagram of three recycling methods, including pyrometallurgy, hydrometallurgy, and ML-direct recycling. The final product of pyrometallurgy is metal alloy. While final products of hydrometallurgy are lithium salt and precursor, which should be further processed to assemble batteries. Compared to the other two non-machine learning-aided methods, ML-direct recycling has the shortest process flow since the product is standard battery materials, which brings about the largest possible convenience and the least possible environmental footprints. It should be stressed that such convenience is enabled by accurate sorting, a vital link in pretreatment for actual battery recycling practice, thanks to our FL framework.

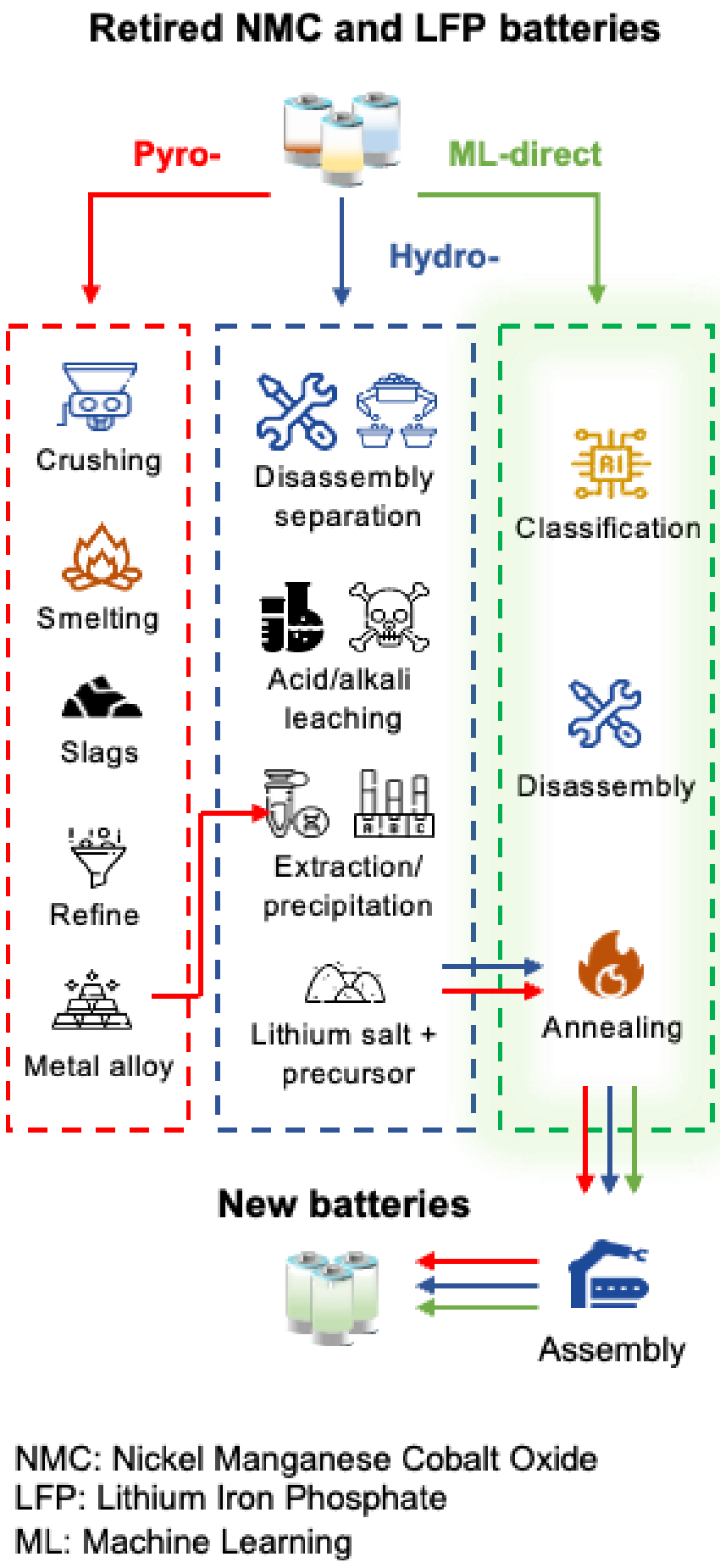

Figure 4. 15 Comparison between recycling with and without material sorting.

### 4.5.1 Without Mixed Material

The cost analysis of LFP and NMC batteries using different recycling methods is shown in Figure 4. 16a, including raw material, reagent, average labor, electricity & water, equipment depreciation, and sewage treatment. It can be observed that the raw material accounts for the largest proportion of the cost. As a result, the cost of NMC is always higher than LFP in any method owing to the large price difference between NMC and LFP. Besides, for the same type of batteries, the cost of ML-direct recycling is the largest while the pyrometallurgy is the least, owing to the large expense of reagents. Considering the reagents can be heavily cathode material specific, the profitability of ML-direct recycling largely depends on the sorting accuracy of the mixed retired batteries.

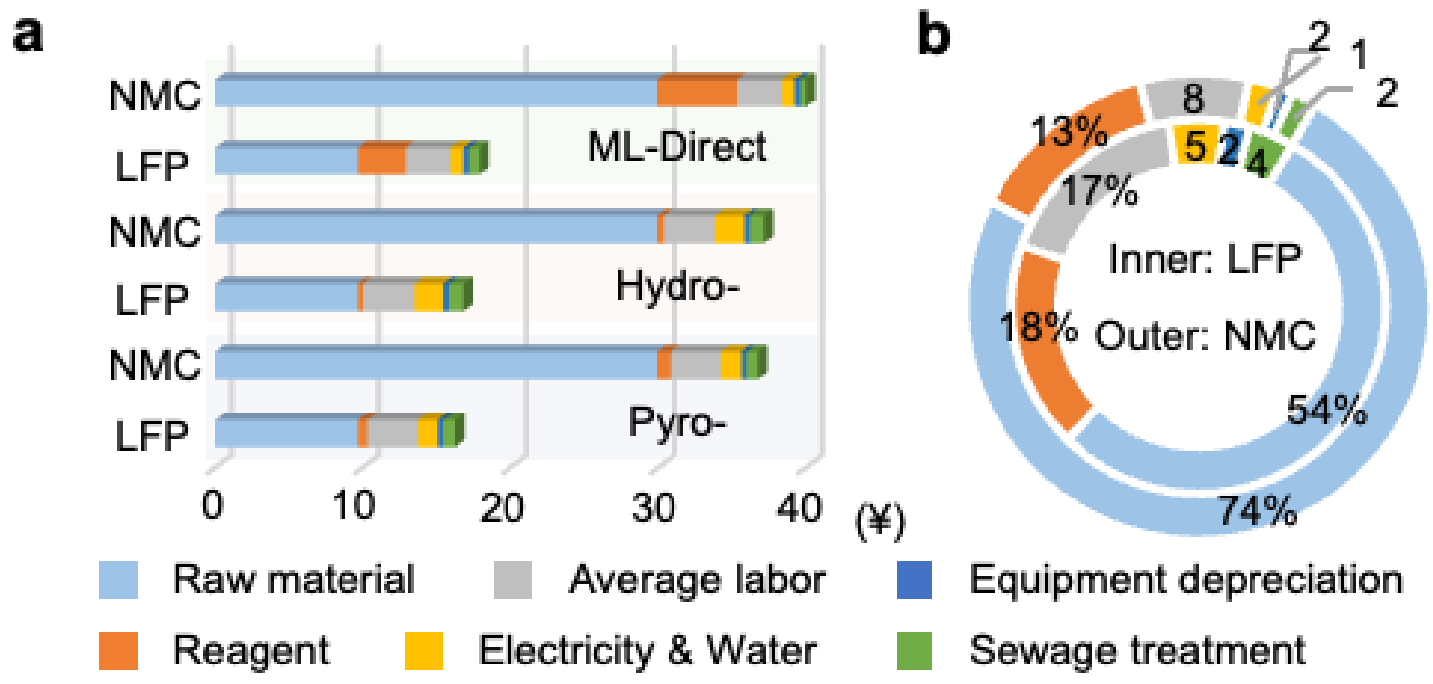

Figure 4. 16 Cost breakdown without mixed material.

Further analysis of the detailed proportion of cost structure in ML-direct recycling is in Figure 4. 16b. The outer and inner annuluses are NMC and LFP batteries, respectively. Except for raw material and reagent, the sum of the other costs is the same in price (5620 ¥/t) but more than twice the difference in percentage (NMC for 28%, LFP for 13%). The cost of raw materials for NMC (29900 ¥/t, accounting for 74%) is three times that of LFP (9687.5 ¥/t, accounting for 54%), which again indicates the profit of ML-direct recycling is sensitive to the sorting accuracies.

Figure 4. 17 lists the cost, revenue, and profit of LFP and NMC batteries using different recycling methods. For the largest profit option, NMC battery using ML-direct recycling (29944.25 ¥/t) is 2.25 times the second largest profit option (LFP batteries using ML-direct recycling, 13279.51 ¥/t). It can be summarized that ML-direct recycling has the largest revenue and profit. Moreover, it is also noticed that the profit of recycling NMC is always larger than LFP, highlighting the significance of efficiently sorting high-value recycling candidates from a bulk of mixed retired batteries.

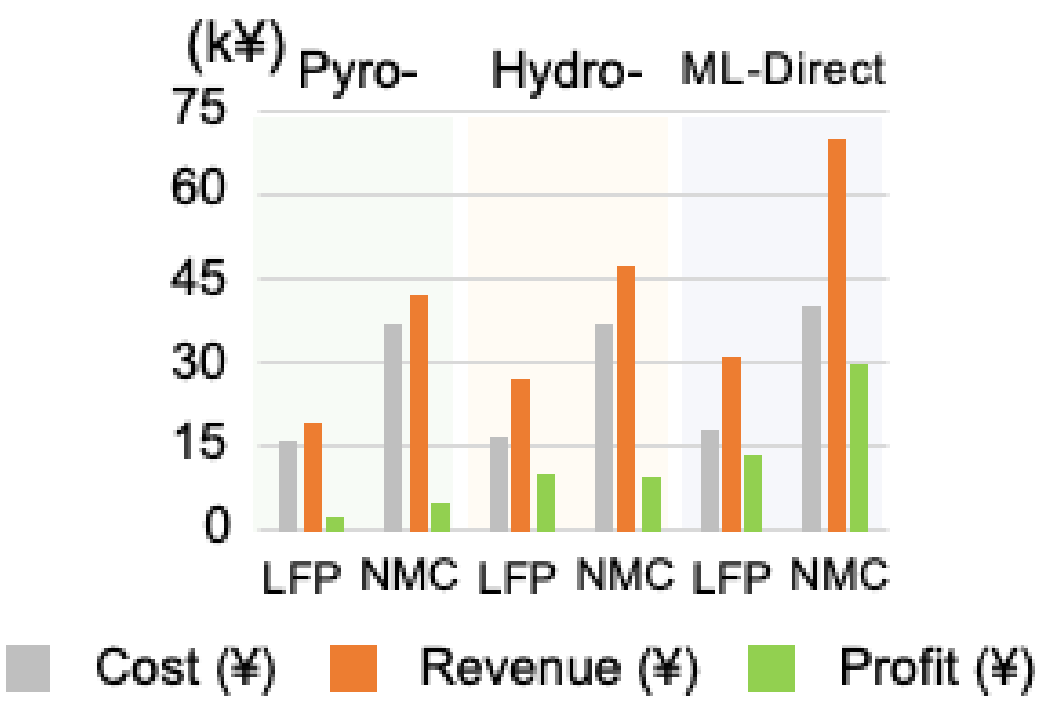

Figure 4. 17 Cost and revenue analysis without mixed material.

### 4.5.2 With Mixed Material

In a practical scenario, collected retired batteries could be expensive and even impossible to sort by human-aided pretreatment, especially when the recycling is scaling up. On the contrary, ML-direct recycling has the unique advantage of efficiently sorting the retired batteries by leveraging existing data sources from multiple battery recycling collaborators. An economic analysis using different machine learning paradigms (IL, and FL using MV and WDV) is carried out in Figure 4. 18. Due to the high sorting accuracy of WDV, the two types of batteries (LFP and NMC) can be completely sorted and the final product can be utilized to assemble new batteries directly. On the contrary, the MV and IL produce significant errors in distinguishing cathode materials, thus leading to low-value products (impure materials) that are unable to be directly utilized, requiring refining. As a result, the profit decreases asymptotically for MV and IL methods when sorting accuracy is lower than WDV, specifically 97%. NMC battery recycling using WDV-based ML-direct recycling has a high profit of 24389.33, 21611.88, and 18834.42 ¥/t for the LFP/NMC ratio of 1:2, 1:1 and 2:1, respectively, which are higher than those of pyrometallurgy (4372.32, 3994.46 and 3616.61 ¥/t) and hydrometallurgy (9957.45, 10039.27 and 10121.09 ¥/t).

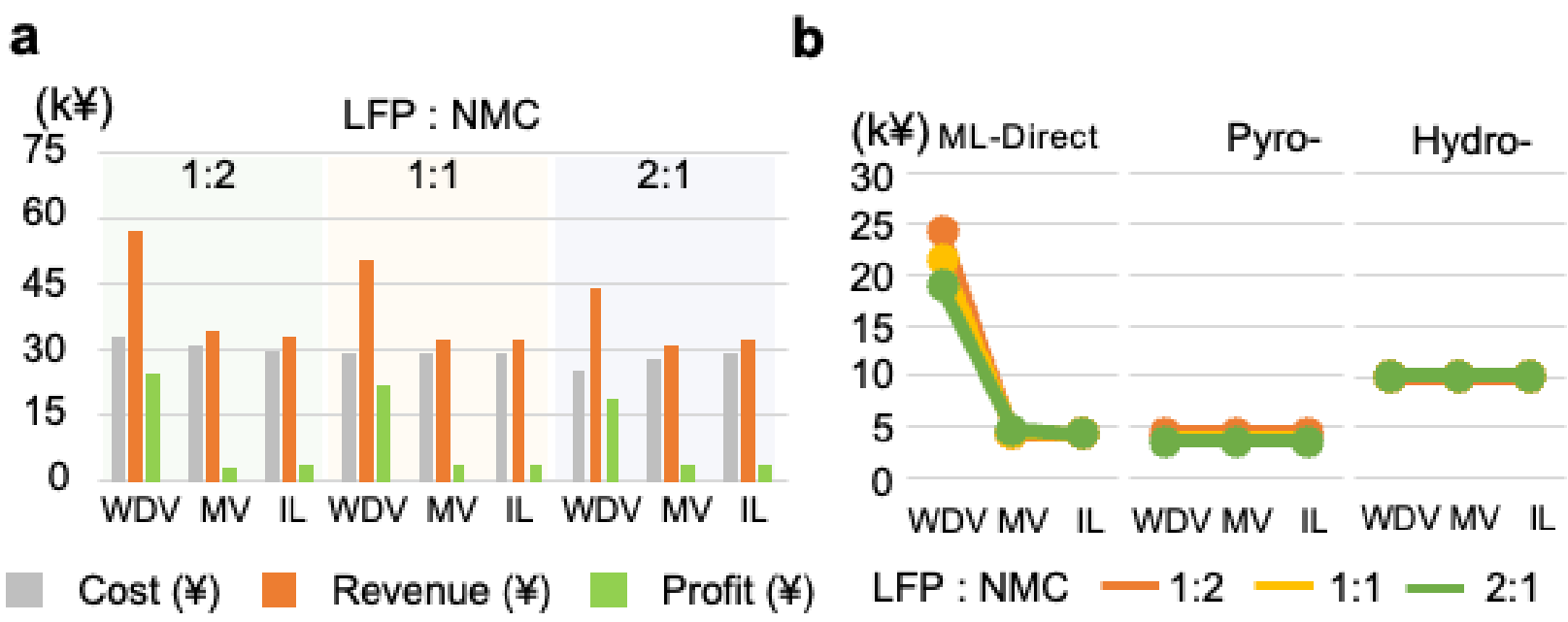

Figure 4. 18 Cost and revenue analysis with mixed material.

Given the profits of pyrometallurgy and hydrometallurgy are not sensitive to sorting accuracy, these methods do not require stringent cathode material information. Such a high profit from ML-direct recycling not merely stems from the inherited advantage of direct recycling but is enabled by our effective and accurate retired battery sorting. ML-direct recycling performs noticeable advantages in environmental protection [88, 206], operation simplicity, privacy, data sharing, and profit. Our ML-direct recycling method has huge socioeconomic values and can quickly accelerate the development of the battery recycling industry, especially when next-generation batteries are even more complex in cathode material diversities.

## 4.6 Summary

We have successfully demonstrated our FL framework, especially our proposed WDV strategy, serving as a key to profitable battery recycling from a practical perspective. Such success is achieved by leveraging existing data sources to train a general data-driven battery sorting model, rather than an expensive human-aided sorting. Our model features a collaborative while privacy-preserving fashion, enabling the direct recycling methodology, which is heavily cathode-specific and sensitive to the recycling candidates. We discuss the merit of our work from a multi-level perspective, including the fundamental mechanism of battery sorting, the implication of profitable recycling, and the advantage of the federated battery recycling paradigm.

When the recycling collaborators successfully sort the retired batteries from the recycler, a voting procedure is performed. Noting that the data volume and data diversity of each recycling collaborator may differ, the voting results could be biased to the specific cathode depending on the data distribution. This still hinders the profitability of battery recycling given low sorting accuracies, highlighting the significance of our WDV strategy. The result shows using our WDV-based FL framework, the battery recycling industry has a high possibility of transforming from the current human-aided battery sorting to an automatic, collaborative, and privacy-preserving fashion, with high sorting accuracy. Such effective retired battery sorting serves as a key to the battery recycling practice using direct recycling, equivalently, our ML-direct recycling. Without our method, the economic benefits of direct recycling could be greatly reduced to a level lower than traditional pyrometallurgy and hydrometallurgy even with small errors in sorting accuracy. In next-generation battery recycling, there could be various battery types involving different chemical compositions, including Si anode, Na ion, lithium-sulfur, and zinc-air batteries, etc. The data collection for model training will be even more challenging due to data privacy and data heterogeneity (battery diversity), calling for FL to address such issues. In addition to battery type information, direct recycling also requires sorting for SOH since it directly determines the amount of reagents added to the direct recycling process, accounting for the majority of the recycling cost. Excessive or insufficient reagents will lead to a declined product quality, which in turn leads to a decline in revenue and sustainability. Different from the sorting problem, SOH estimation is more challenging since it requires historical data to formulate a regression problem. Moreover, the SOH can be heavily dependent on historical usage, while such information is difficult to retrieve at the end-of-life stage due to poor lifecycle management. Using only field-available information to determine SOH remains a critical challenge. It should be noted that the profit calculation assumes an 80% SOH for the retired batteries. Future work should consider SOH information to increase the profitability of ML-direct recycling, which is a great commercial concern.

As mentioned above, the core idea of adopting FL into battery recycling is leveraging the existing data information in a collaborative while privacy-preserving manner, which is intuitively consistent with the distributed nature of battery data. We note that the cost of sorting through machine learning is not considered, which attributes to a lack of relevant industry data and conversion standards. Under a FL setting, the recycler only needs to process battery information, thus the machine learning cost is not sensitive to the recycling scale. We, therefore, assume that the once-for-all machine learning cost will be covered when the battery recycling scale enlarges.

Even though more in-depth investigations on the accuracy-privacy-cost balance should be conducted, we emphasize that the proposed FL framework tackles the common concerns in collaborative learning, including privacy, efficiency, and fairness, which can be addressed consistently and elegantly. We begin by noticing that the data privacy of the collaborating clients is fully protected, as neither the raw battery data nor the extracted features are leaked out of their respective data sources; even the as-trained local models are kept confidential to the data sources themselves. The only information being transferred, the local cathode sorting result, can be appropriately encrypted before transferal to eliminate potential eavesdroppers, ensuring PB. Also, with the full support of parallelized local training and only one round of result transfer, the proposed framework is highly efficient in computation and communication, which remains a huge challenge in commercialization in other fields. Specifically, the selection of random forests as the bottom-level machine learning algorithm, instead of more advanced neural network architectures, is made with full consideration of the feature engineering settings and cost-effectiveness requirements. Feature engineering, which prepares the data for FL with expert-knowledge-based information extraction, transforms the raw gigabyte-scale sequential data into kilobyte-scale tabular data. Decision trees such as random forests are more adept at learning from such low-dimensional data with heterogeneous features, whether in terms of accuracy, efficiency, or interpretability. Thus, our proposed framework is light and scalable without requiring intensive investment in the battery recycling sector, which is of significant interest to industrial practice. The framework further achieves fairness by assigning a local training task of the same scale, though in different cathode types and sample sizes, to all clients under the recycling collaboration. Even when compared with other FL frameworks, the proposed framework is still better in terms of these metrics, as those alternative frameworks would most likely require various rounds of model updates with considerable parameter transferals, which would compromise efficiency and expose collaborators to an immense level of privacy leakage.

Currently, the framework is implemented under the ideal assumption that all collaborators are fully cooperative, such that the uploaded local results are assumed to be reliable. Despite such an ideal assumption, the random forest model, incorporated with the WDV, is naturally robust against random parameter transfer losses even if parameters from a few collaborators end up missing. The sorting accuracy only slightly degrades given the same heterogeneity setting. In the case of a blockchain-like environment with collaborators of unknown trustworthiness and reliabilities, our research could be further extended in search of a proper incentivization mechanism such that all recycling collaborators would fully contribute to their respective local model instead of attempting to become total free-riders. We prospect quantifying the helpful information that the recycling collaborators provided to design a benefit distribution strategy and a free-rider detection scheme to make the federated battery recycling ecosystem economically feasible.

To conclude, FL is a promising route for retired battery sorting and enables emerging battery recycling technologies, especially direct recycling, in their development, practical application, and optimization. We create a retired battery sorting model using only one cycle of end-of-life charging and discharging data as opposed to any historical data while preserving the data PBs of multiple battery recycling collaborators. In the homogeneous setting, we obtain a 1% cathode material sorting error; in the heterogeneous setting, we obtain a 3% cathode material sorting error, thanks to our Wasserstein-distance voting strategy. Such a level of accuracy is achieved by (1) automatically exploring the unique patterns in the salient features without assuming any prior knowledge of historical operation conditions and (2) using our proposed Wasserstein-distance voting strategy to correct heterogeneous data distribution among recycling collaborators. An economic evaluation showcases the relevance and necessity of accurate retired battery sorting to the profitable battery recycling industry using direct recycling. In general, our approach can complement the existing first-principle-based recycling route research paradigms on actual battery recycling practice, where retired batteries are necessary while challenging to sort. Broadly speaking, our work enlightens the possibilities of leveraging existing data from multiple data owners, rather than time-consuming and expensive data generations, to develop and optimize complex decision-making procedures such as the battery recycling route design in a collaborative while privacy-preserving fashion.

# CHAPTER 5 UNIFIED DIAGNOSTICS AND PROGNOSTICS

## 5.1 Overview

The adaptive diagnostics and prognostics are essential to ensure the proposed method's transferability in previous sections. This chapter starts from presenting an adaptive machine-learning prediction framework across varied operation conditions, i.e., the fast-charging ($D_S$) and extremely fast-charging ($D_T$) conditions. As presented in Figure 5. 1, the framework uses CORAL to correct feature divergence from different fast-charging conditions. The CORAL-aided multi-layer perceptron (MLP) realizes a 93.1% cross-operation-condition prediction accuracy using a full feature full cycle model, outperforming an 85.7% prediction accuracy produced by the state-of-the-art model fine-tuning method. To decouple feature-wise and cycle-wise contribution to such prediction, five classes of features (10 in total) in the first 320 cycles are investigated. The wasserstein distance is adopted to evaluate the adaptability of the CORAL-transformed features.

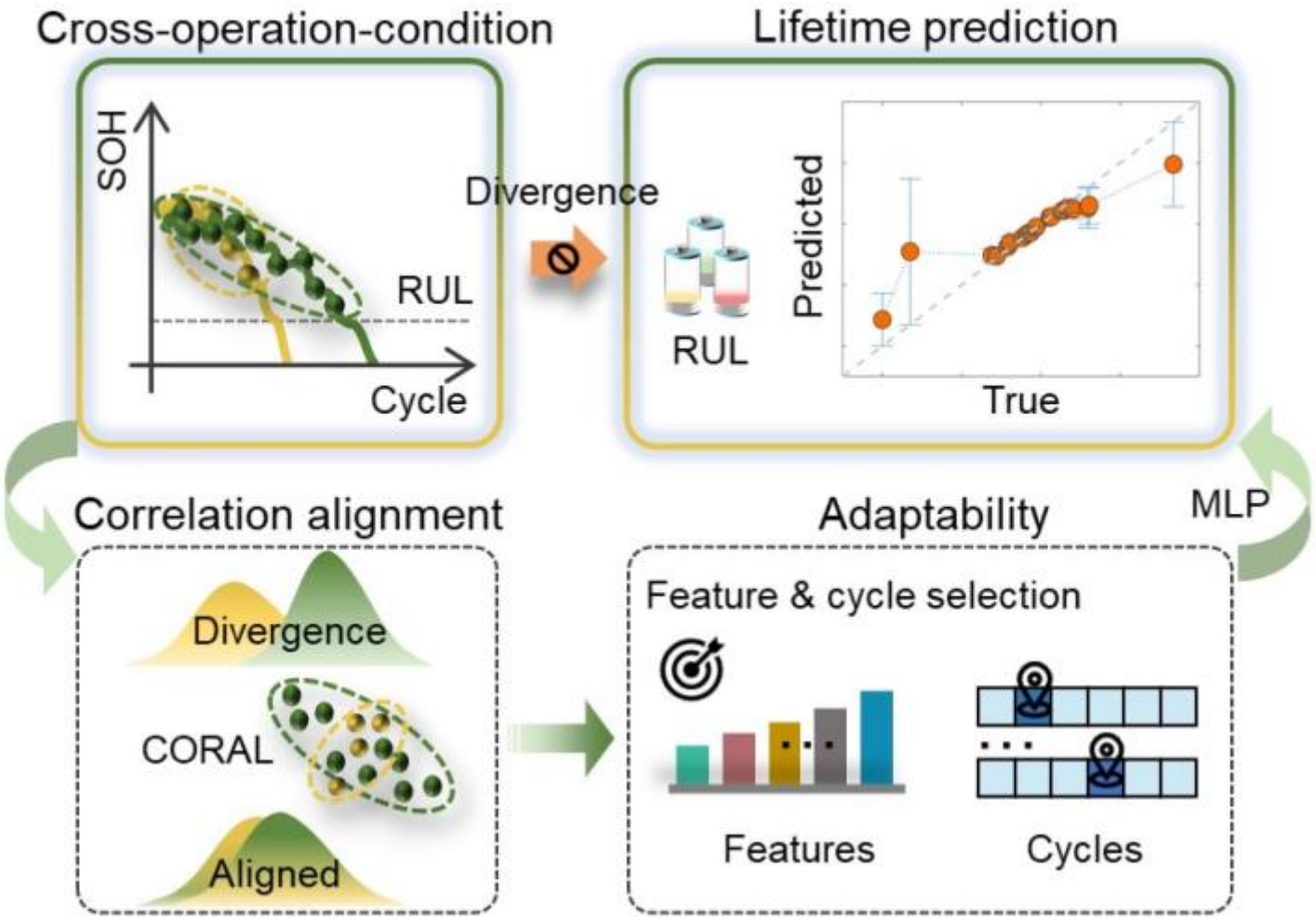

Figure 5. 1 General workflow of adaptive lifetime management.

The power of our adaptive machine-learning method is illustrated in feature-wise and cycle-wise models by effectively removing the redundant feature and cycle data while producing state-of-the-art prediction accuracy. The CORAL-aided MLP using one feature of only one cycle can realize a 93.3% prediction accuracy, saving up to 99.4% of the computation compared with the full feature full cycle model. Our findings suggest that by evaluating the feature-wise and cycle-wise adaptability, one can expedite the feature selection process without random training of the models. We also present a prediction capability-adaptability matrix to guide the feature selection for cross-condition lifetime prediction. Finally, the generalization of the proposed method is validated across two independent fast-charging datasets with different charging policies.

## 5.2 Data Collection and Processing

The fast-charging datasets reported by Braatz et al. (Dataset1) [63] and Chueh et al. (Dataset2) [43] are used. The batteries used in these two datasets are commercial LiFePO4/graphite cells with a nominal capacity of 1.1Ah.

Table 5. 1 Overview of the datasets by Braatz et al. and Chueh et al.

| Parameters | Dataset1 [63] | Dataset2 [43] |
|---|---|---|
| Testing chamber | APR18650M1A forced convection to 30°C | |
| Cell chemistry | lithium-ion phosphate (LFP)/graphite | |
| Nominal capacity | 1.1 Ah | |
| Nominal voltage | 3.3 V | |
| Discharging profiles | 4C | |
| Lifetime definition | 80% of the nominal capacity | |
| Cutoff potentials | 3.6 V (upper) and 2.0 V (lower) | |
| Charging profiles | C1(Q1)-C2 | C1-C2-C3-C4 |

The datasets share identical testing temperatures (forced convection to 30°C), cutting-off potential (3.6V for charging; 2.0V for discharging), and discharging rate (4C). Variables recorded in the two datasets are cycling policy, discharging capacity, internal resistance, charging time, and cell temperature. The charging profiles of Dataset1 and Dataset2 follow a multi-stage format of C1(Q1)-C2 and C1-C2-C3-C4, respectively. The detailed experimental setups for the two datasets can be found in Table 5. 1. We train and test our model on Dataset1 and validate the model performance on Dataset2, which has exclusive and more complex fast-charging operation conditions.

We reconstruct Dataset1 for different lifetime distributions to investigate the cross-operation-condition lifetime prediction. Figure 5. 2 shows a clear lifetime drop when the second-stage charging rate exceeds 5C. The batteries are in fast-charging and extremely fast-charging groups indexed by the second stage of charging, i.e., C2 in Dataset1.

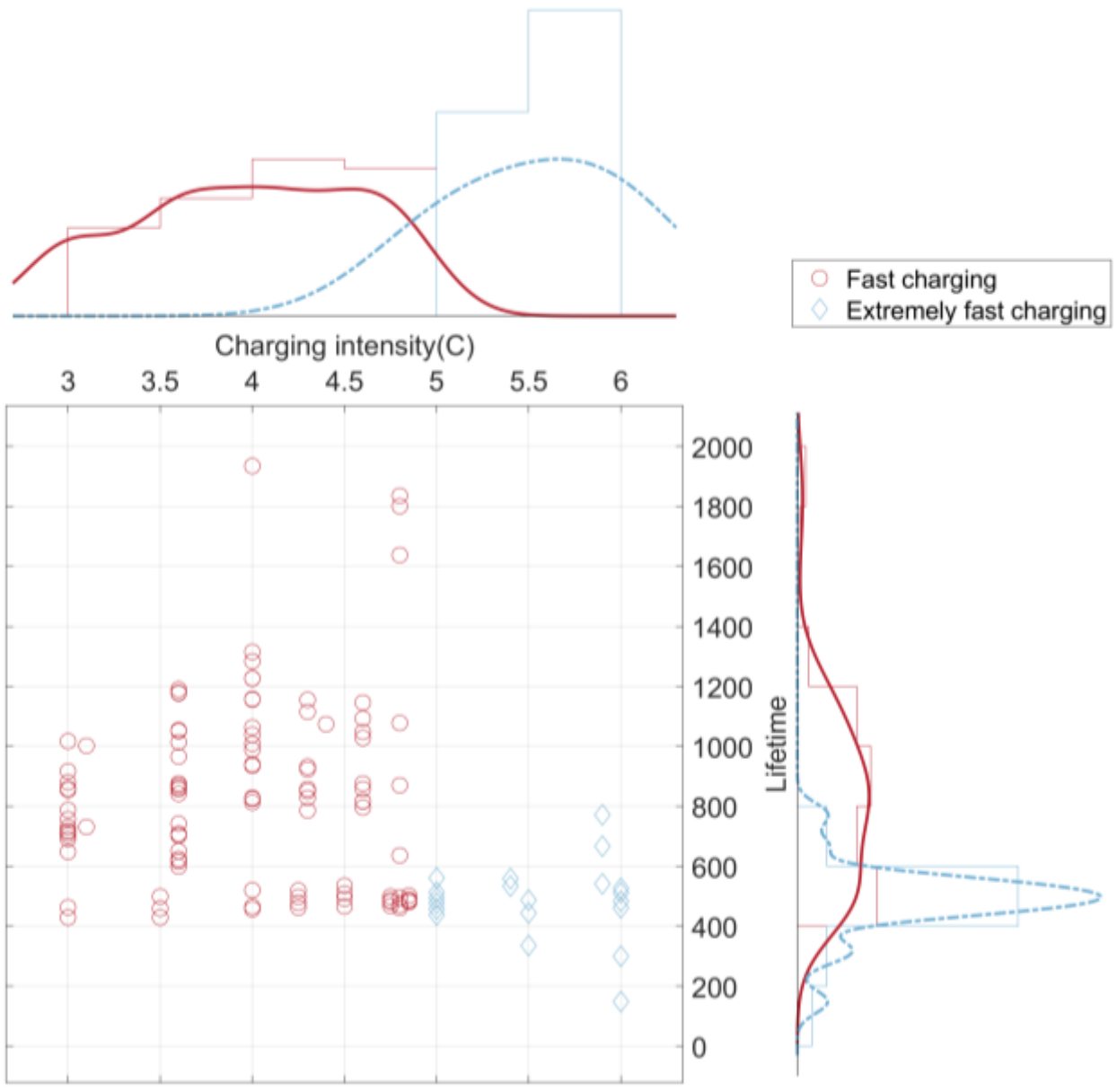

Figure 5. 2 The lifetime distribution by second-stage charging rate.

On the contrary, in Figure 5. 3, the lifetime exhibits no clear trend with first-stage charging rate. The batteries are in fast-charging and extremely fast-charging groups indexed by the first stage of charging, i.e., C1 in Dataset1. Dataset1 is reconstructed by indexing the second stage charging rate. As a result, the two operation conditions are fast (104 samples) and extremely fast-charging conditions (20 samples), respectively.

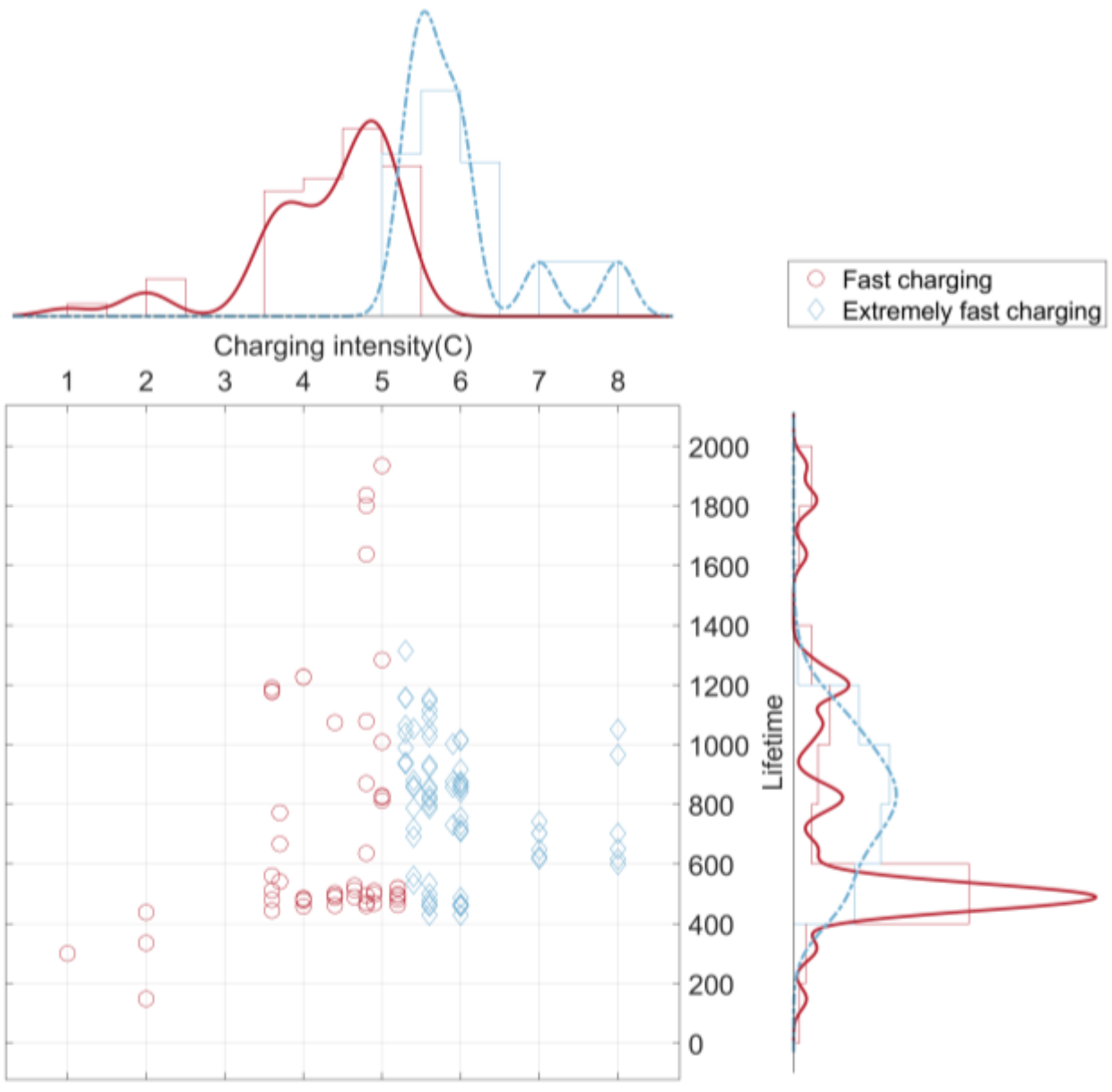

Figure 5. 3 The lifetime distribution by first-stage charging rate.

Specifically, the maximum lifetime divergence between the fast and extremely fast-charging operation conditions is 1163, which is 150% and 786% larger than the average and minimum lifetime, respectively. See Table 5. 2 for the detailed charging policies of the reconstructed Dataset1.

Table 5. 2 The summary of the charging policies.

| Policy | Operation conditions (C, second-stage charging rate) | |
|---|---|---|
| | Fast | Extremely fast |
| Charging rate | 3.0, 3.1, 3.5, 3.6, 4.0, 4.25, 4.3, 4.4, 4.5, 4.6, 4.75, 4.8, 4.85 | 5, 5.4, 5.5, 5.9, 6 |
| Number of cells | 104 | 20 |

To study such a divergence, manual feature engineering is performed. In contrast to refined feature engineering for a specific feature, we aim that the selected features can cover as many different aging phenomena as possible, such as changes in time response, capacity, resistance, geometrical pattern, and thermal aspects. Even though there could be causal interferences with these aspects, which are not independent of the physical aging process, we deliberately hold these aspects for comparative analysis. The comparative analysis aims to identify which features are more directly related to the physical aging process, thus having better adaptability under cross-operation-condition settings. We, select 10 features with simple manipulations such as maximization, minimization, and deviation from five classes, i.e., the charging time, capacity deviation, internal resistance, geometrical, and temperature.

The extracted feature classes are presented in Figure 5. 4, where two typical cells with 461 and 1009 lifetimes are selected, and the presented features are at the 100th cycle for illustration. lb and $\text{ub}_i$ denote that the features are extracted between $lb^{\text{th}}$ and $\text{ub}_i{}^{\text{th}}$ cycle, respectively. The explanation of these features is presented in Table 5. 3 The explanation to the extracted features.

| Features | Full name (from $lb$ to $ub_i$) | Formula |
|---|---|---|
| act | Averaged charging time | $\text{act}_i = \frac{\int_{lb}^{ub_i} t}{ub_i - lb}$ |

Continued Table 5. 4. Note that the $lb$ and $ub_i$ denote the lower and upper bounds in cycles of feature extraction, respectively. Except for dact, $lb$ = 10, $ub_i$ = 20, 40, 60, …, 320. For dact, $ub_i$ = 20, 40, 60, …, 320, and $lb_i$ = $ub_i$-20.

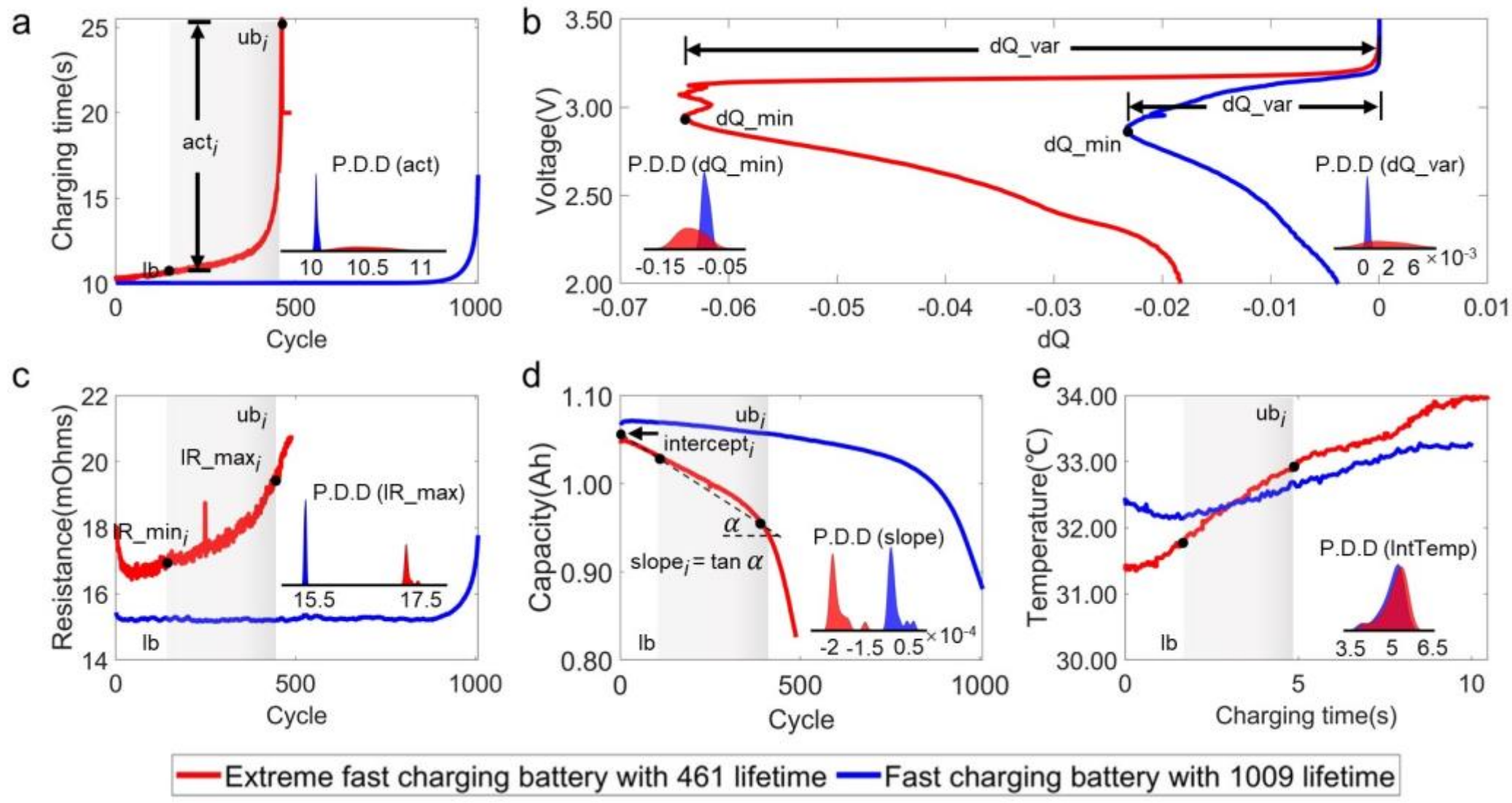

Figure 5. 4 Feature extraction and feature divergence between typical cells.

For instance, we notice the temperature rising speed is higher for the extremely fast-charging condition than the fast-charging condition. Other feature value evolutions of these two conditions are presented in Figure 5. 5. Red circles note outliers in each observed cycle. The feature evolution trend is illustrated by a red dashed line connecting the mean value in each cycle, where the apparent feature divergences bring about challenges in cross-operation-condition lifetime prediction.

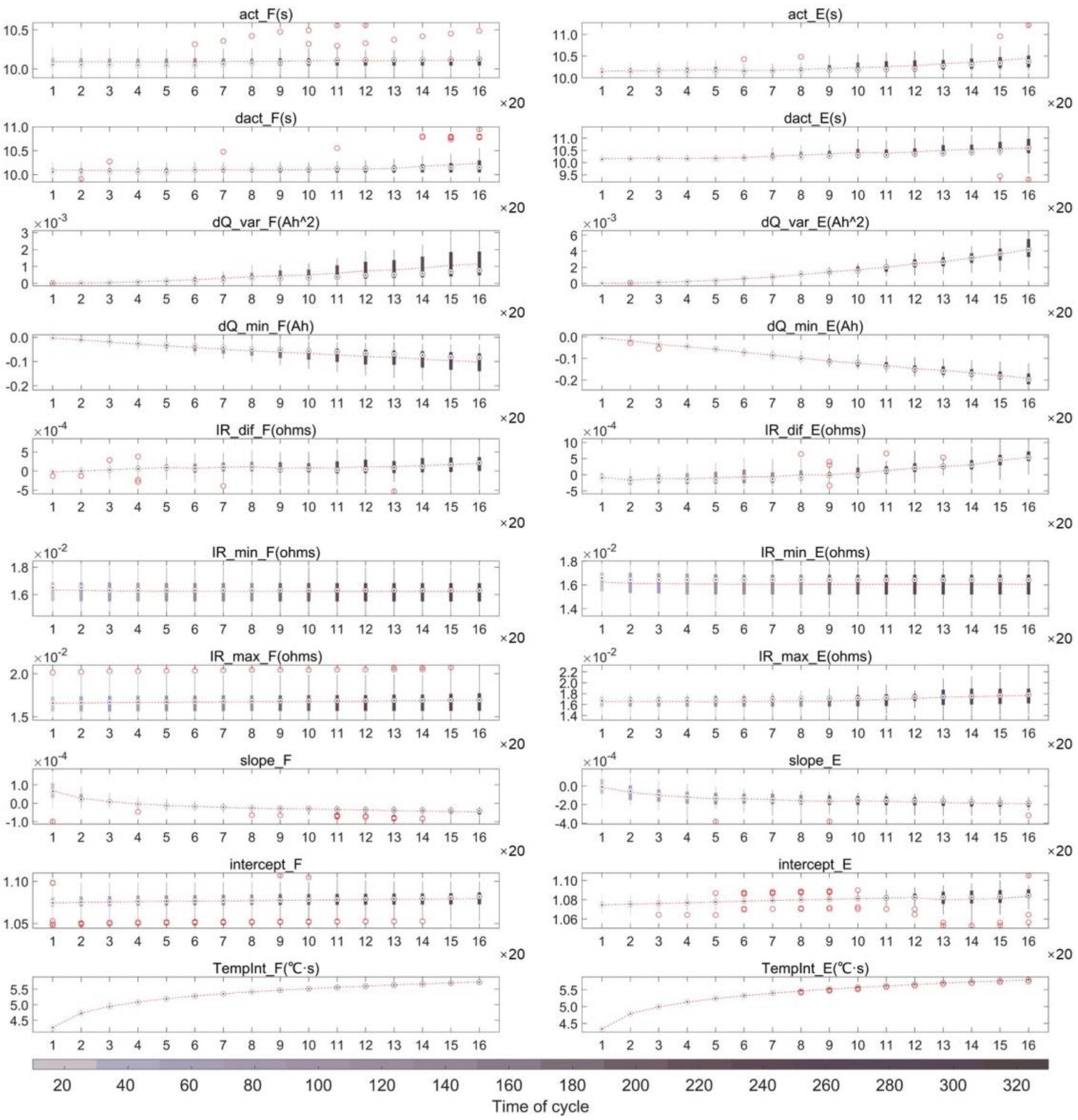

Figure 5. 5 Cycle evolution in fast and extremely fast-charging conditions.

Table 5. 3 The explanation to the extracted features.

| Features | Full name (from $lb$ to $ub_i$) | Formula |
| --- | --- | --- |

| act | Averaged charging time | $\text{act}_i = \frac{\int_{lb}^{ub_i} t}{ub_i - lb}$ |
|---|---|---|

Continued Table 5. 4 The explanation to the extracted features.

| Features | Full name (from $lb$ to $ub_i$) | Formula |
|---|---|---|
| dact | Deviation of averaged charging time | $\text{dact}_i = \frac{\int_{ub_i}^{ub_{i+1}} t}{ub_{i+1} - lb_i}$ |
| dQ_var | Variance of deviation capacity | |
| dQ_min | Minimum deviation capacity | |
| IR_dif | Deviation of internal resistance | $\text{IR_dif}_i = \text{IR_max}_i - \text{IR_min}_i$ |
| IR_min | Minimum internal resistance | |
| IR_max | Maximum internal resistance | |
| slope | Slope of the capacity-lifetime curve | |
| intercept | Intercept of the capacity-lifetime curve | |
| IntTemp | Time integral of temperature (in $\log_{10}$) | $\text{IntTemp}_i = \int_{lb}^{ub_i} \text{T}$ |

## 5.3 Correlation Alignment (CORAL)

The core idea of using CORAL is to correct the divergence of the extracted features. See Table 5. 3 The explanation to the extracted features.

| Features | Full name (from $lb$ to $ub_i$) | Formula |
|---|---|---|
| act | Averaged charging time | $\text{act}_i = \frac{\int_{lb}^{ub_i} t}{ub_i - lb}$ |

Continued Table 5. 4 for an explanation of the feature extraction process. As we highlight a cross-operation-condition lifetime prediction, which means that we deliberately train the model on the source domain and test the trained model on the target domain, we aim to use the CORAL-transformed source domain features to predict target domain labels. In this way, the large number of accessible source domain samples and target domain information can be leveraged simultaneously. Here we define the source domain, i.e., the fast-charging condition $D_S$ by a feature matrix $X_S$ in the feature space $\chi_S$ and a label vector $Y_S$ in the label space $\Upsilon_S$, i.e., $D_S = \{X_S, Y_S\}$, where $X_S = \{x_1, \cdots, x_n\}^{(m)} \in \chi_S$ and $Y_S = \{y_1, \cdots, y_m\} \in \Upsilon_S$. n and m are the feature dimension and sample size, respectively. When evaluating the full feature full cycle and best feature full cycle model, the feature vector dimension $n$ is 160 and 16, respectively. Similarly, the target domain, i.e., the extremely fast-charging condition, is defined as $D_T = \{X_T, Y_T\}$. Let $C$ be the feature covariance matrix. Note that all features are normalized to [-1,1], and the covariance matrix is calculated as follows:

$$C_S = Cov(X_S) + \lambda I_p \tag{5.1}$$

$$C_T = Cov(X_T) + \lambda I_p \tag{5.2}$$

where $I_p$ is an identity matrix. $\lambda > 0$ is the regularization term. Smaller $\lambda$ corresponds to more alignment, while larger $\lambda$ corresponds to less alignment. $Cov(\cdot)$ is the covariance operator. Given a feature matrix $X_{m \times n}$, $Cov(X_{m \times n})$ is an $n \times n$ square matrix.

To minimize the feature divergence between $D_S$ and $D_T$, a linear transformation $A$ is applied to the $D_S$ features. The Frobenius norm is used to quantify the feature divergence:

$$\min_A \|C_{\hat{S}} - C_T\|_F^2 = \min_A \|A^T C_S A - C_T\|_F^2 \tag{5.3}$$

where $C_{\hat{S}}$ is the covariance of the CORAL-transformed source domain features. It has been proved that if $rank(C_S) \geq rank(C_T)$. $\|\cdot\|_F$ stands for the Frobenius norm:

$$\|\cdot\|_F = \sqrt{\sum_{i=1}^{m} \sum_{j=1}^{n} |a_{ij}|^2} \tag{5.4}$$

where, $a_{ij}$ denotes the arbitrary element at the $i^{th}$ row and $j^{th}$ column of the matrix.

In the cross-operation-condition setting, the core issue is to analyze the features' generality, i.e., the adaptability when the operation condition changes. Such adaptability is measured by the Wasserstein distance between CORAL-transformed source domain features and target domain features. Adaptability is quantitatively defined as:

$$Adaptability = 1 - W_q \tag{5.5}$$

where, $W_q$ is the Wasserstein distance:

$$W_q(\cdot,\cdot) = \left( \underset{\gamma \in MP}{inf} \int_{\Omega \times \Omega} |X_{\hat{S}} - X_T|^q d\gamma(X_{\hat{S}}, X_T) \right)^{\frac{1}{q}} \tag{5.6}$$

where $\gamma$ refers to the transportation of arbitrary CORAL-transformed feature points in the fast-charging condition $X_{\hat{S}}$ to the extremely fast-charging condition feature $X_T$. $\Omega$ is the original feature space. *MP* stands for such transportation and is measurability preserved. Here, $q = 1$ is chosen to guarantee the robustness of such a measure since it is not sensitive to measurement noises.

## 5.4 CORAL For Interpretable Feature Engineering

Traditional machine-learning lifetime prediction methods are subject to lifetime and feature divergence. To be specific, models trained by the fast-charging conditions ($D_S$) could achieve high prediction accuracy on the domain with similar distribution to $D_S$ while such models exhibit poor adaptability for extremely fast-charging ($D_T$), illustrating the difficulty of the cross-condition lifetime prediction tasks [30]. Such difficulty is attributed to the heterogeneous data distribution of the battery lifetime and the features. One solution to correct the divergent features is to leverage the $D_T$ information to fine-tune the pre-trained MLP model under$D_S$, such that condition-specific information can be learned.

### 5.4.1 Performance of CORAL-Aided Neural Network

In this section, fine-tuning (FT) aided MLP is regarded as the baseline model for its state-of-the-art cross-condition prediction ability. However, FT requires sufficient data under $D_T$ and it is time-consuming in model retraining, which can hardly be satisfied in practical use. In this regard, we adopt CORAL to correct the feature divergence between $D_S$ and $D_T$. Figure 5. 6 shows that the battery lifetime is successfully aligned by proper CORAL regularization parameters, i.e., $\lambda = 100$.

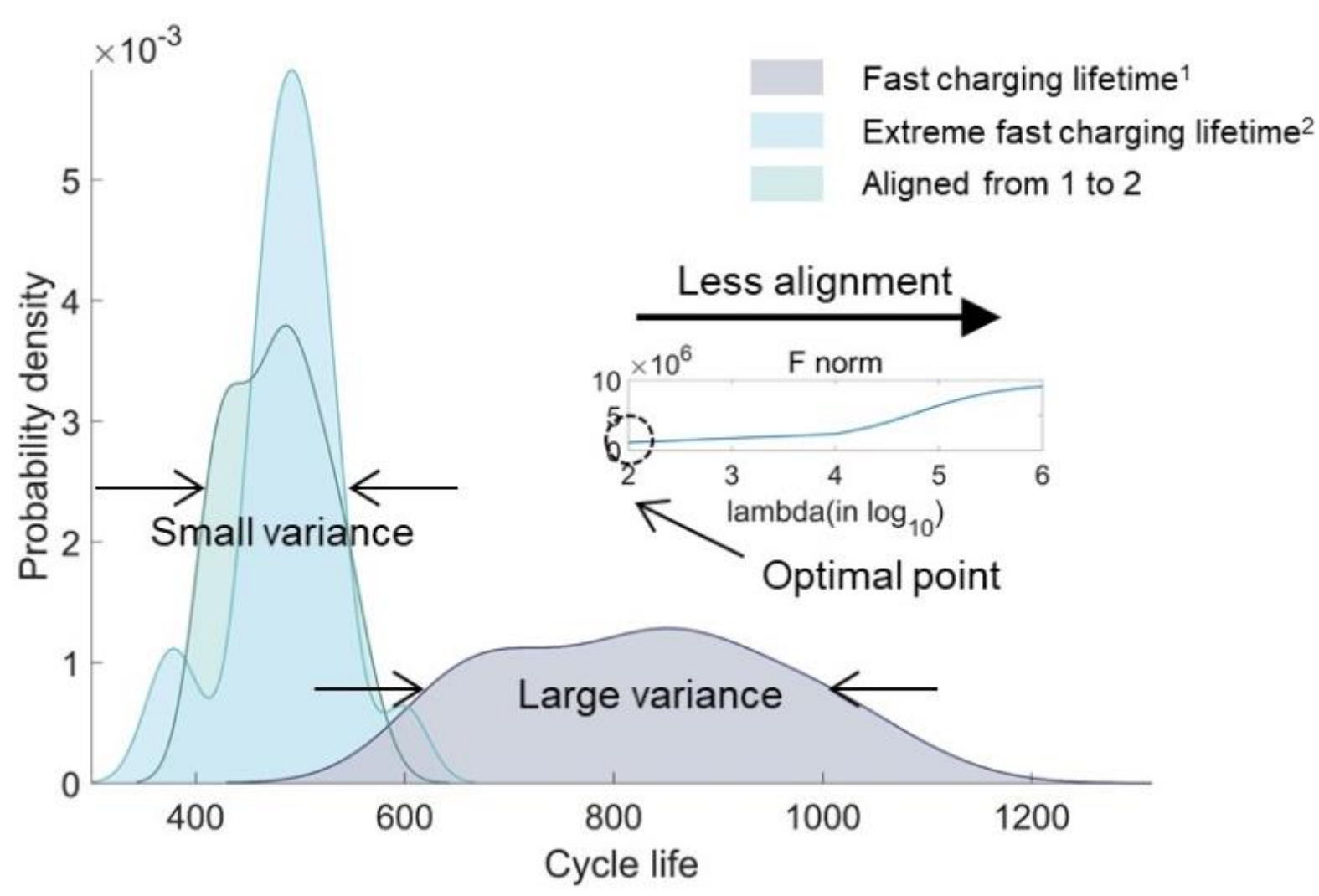

Figure 5. 6 The battery lifetime alignment using CORAL.

In addition, Figure 5. 7 shows that feature divergence is corrected with small residual values, indicating a good alignment performance. Our core idea is to feed the CORAL-transformed $D_S$ features to the MLP, namely the CORAL-aided MLP, to directly predict the lifetime in $D_T$. The Frobenius norm of normalized feature covariance between fast-charging and extremely fast-charging conditions are calculated, respectively. The optimal regularization point is $\lambda = 0.01$.

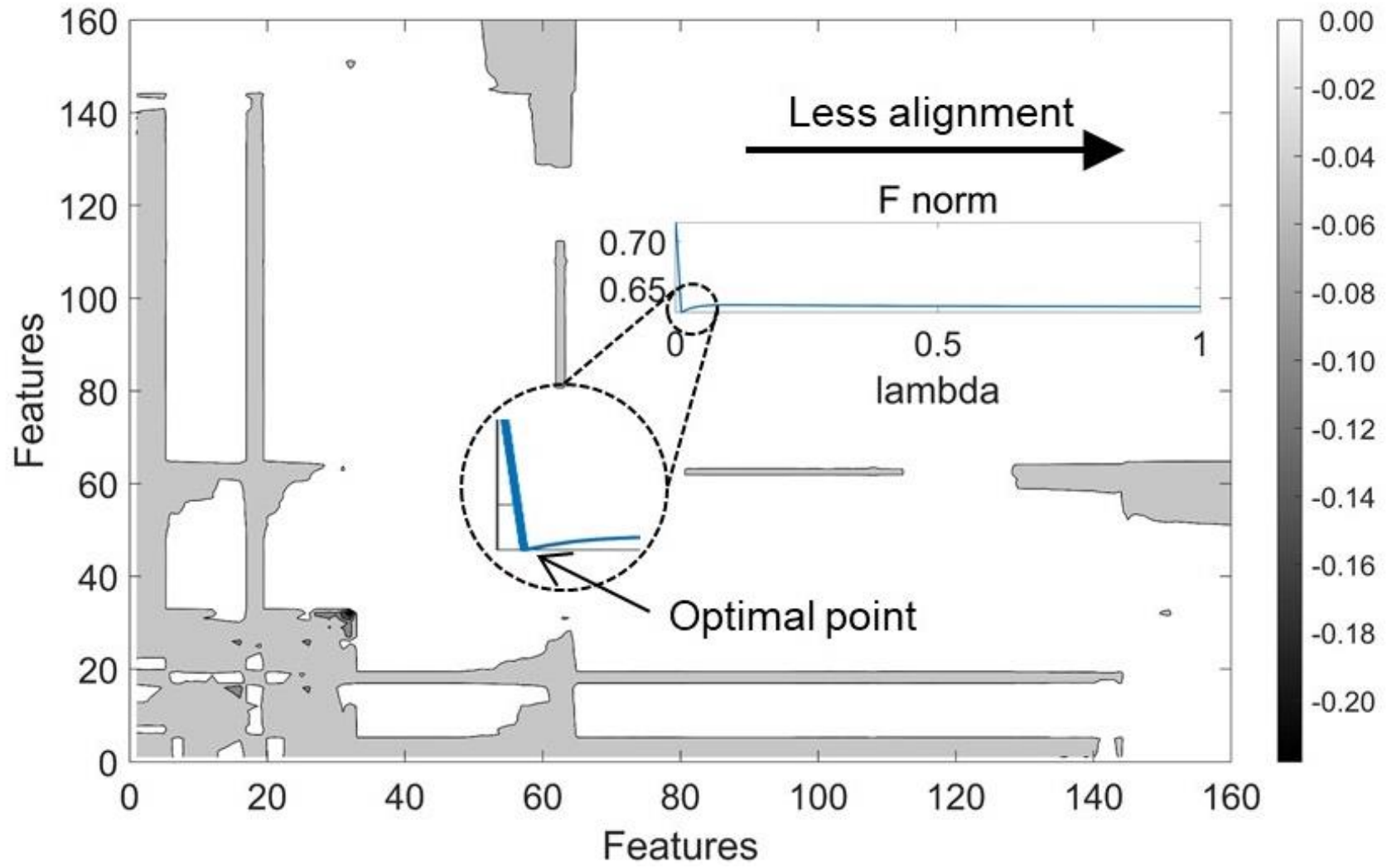

Figure 5. 7 The residual error matrix calculated by $C_{\hat{S}} - C_T$.

To generate prediction uncertainties, 200 Monte Carlo experiments are carried out, ensuring the testing result is validated over the whole sample space [63, 98]. The Monte Carlo experiment is performed when splitting training set (including Cross-dataset validation), random drawing secondary testing samples, and model fine-tuning such that the models are unbiased to specific long-lived or short-lived cells. In this way, different primary training-secondary testing (fine-tuning testing) pairs can be evaluated to guarantee statistical confidence. The random seeds for the Monte Carlo experiments are from 1 to 100 in MATLAB R2022a with the generation keywords as mcg16807. As a result, such 100 Monte Carlo experiments generate 200 testing situations for secondary testing and fine-tuning testing, respectively.

The predicted lifetimes produced by CORAL-aided MLP and FT are compared by each Monte Carlo random experiment in Figure 5. 8a. The CORAL-aided MLP generally better tracks the ground truth lifetime, especially for extremely long and short-lifetime batteries. While the FT model fluctuates around the mean lifetime expectation, indicating an inadequate condition-specific degradation pattern is learned. Such prediction performances as probability confidence are compared in Figure 5. 8b. The CORAL-aided MLP exhibits a 2.28% error rate (in median), accounting for a 50% probability confidence level. Concerning the FT, one has only a 4% confidence probability of reaching an equivalent prediction result. If one wants an 80% probability confidence to attempt a cross-operation-condition prediction, the corresponding error rate of such prediction is less than 9.17% using CORAL-aided MLP. In the subplot of Figure 5. 8b, the parity plot shows that the CORAL-aided MLP gives more close predictions around the perfect line than FT. Moreover, the prediction uncertainties using CORAL-aided MLP are smaller than those of FT, indicating that CORAL-aided MLP is robust to prediction uncertainties.

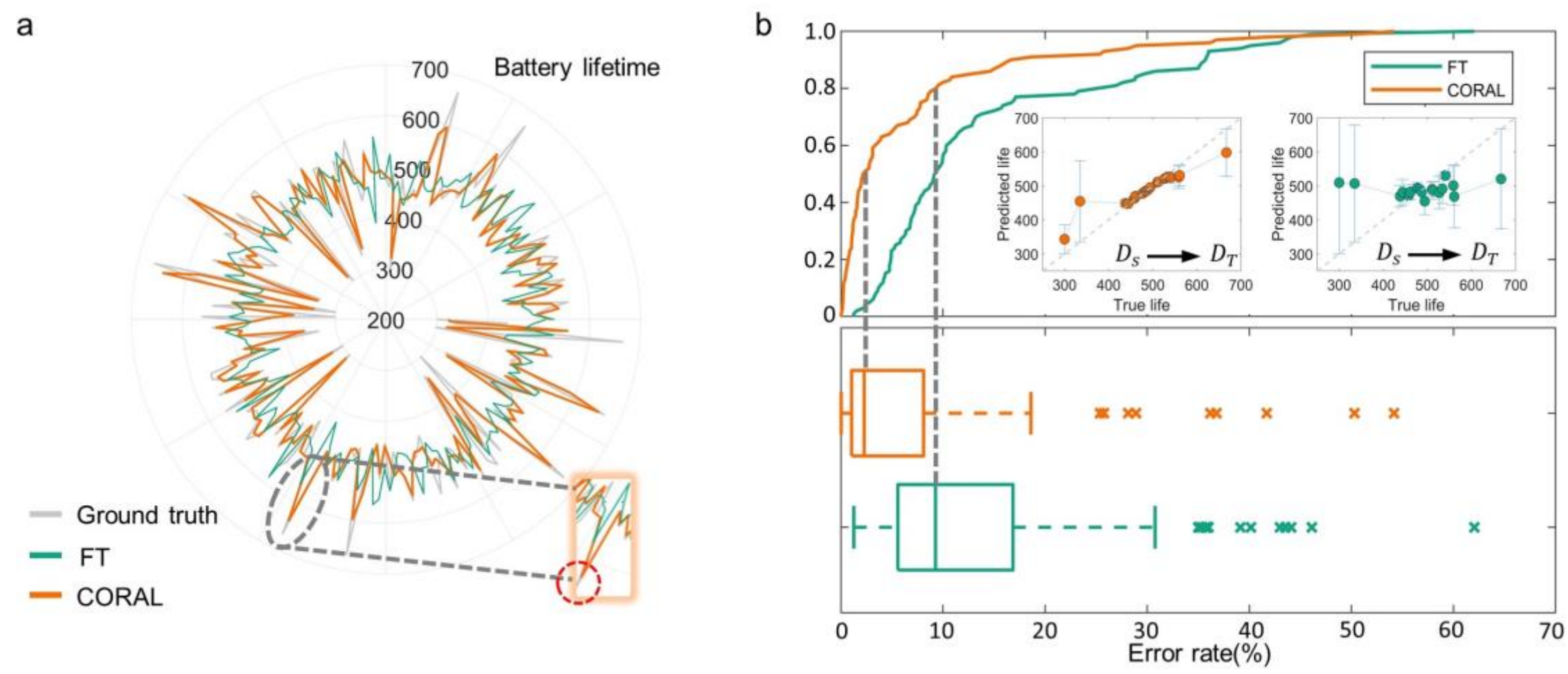

Figure 5. 8 Performance comparison between CORAL-aided MLP and FT.

Figure 5. 9 shows the mean absolute error of CORAL-aided MLP and FT is 31.8 and 59.7 cycles (equivalently, the accuracy is 93.1% and 85.7%), respectively. It is also observed that the FT fails to perform well on the long-lifetime batteries, which could be attributed to the fact that the model is fine-tuned with the short-lifetime batteries in $D_T$ and the FT model could consequently overfit short-lifetime ones. In contrast, CORAL-aided MLP realizes more accurate predictions regardless of the lifetime value. As a result, the error distribution is more concentrated, leading to a smaller STD error using CORAL-

aided MLP.

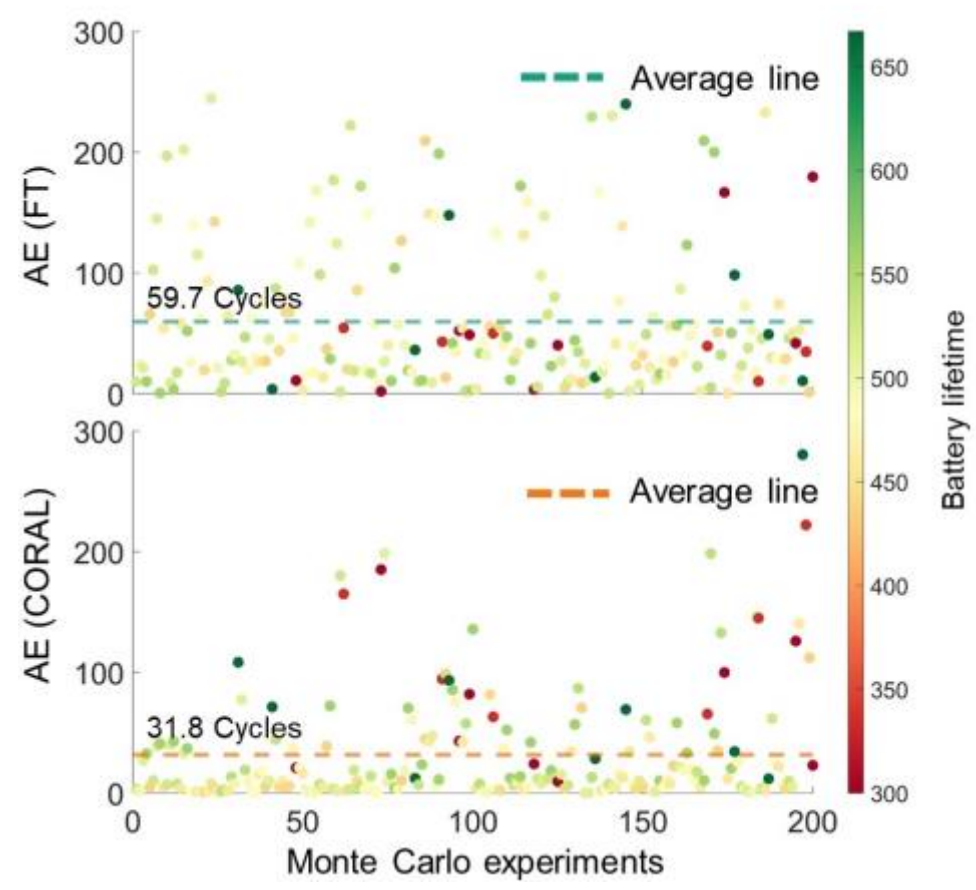

Figure 5. 9 Absolute error under each Monte Carlo experiments.

In Figure 5. 10, the standard deviation error for FT and CORAL-aided MLP are 22.73 and 12.96 cycles, respectively.

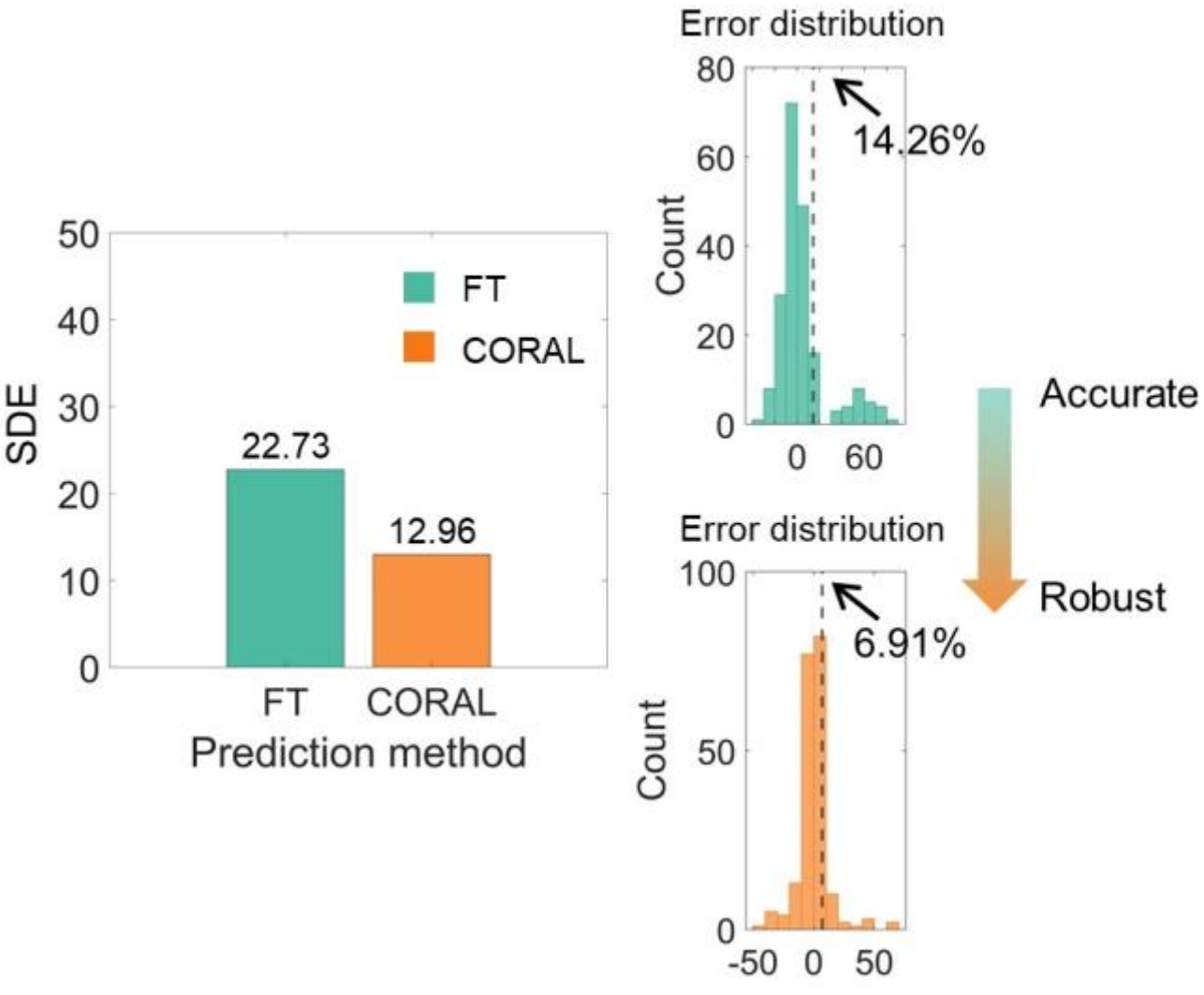

Figure 5. 10 Stability analysis of CORAL and FT-aided MLP.

### 5.4.2 Performance of Feature-wise Prediction

Having demonstrated the efficacy of the CORAL-aided MLP using a full feature full cycle model, it remains unclear to which extent the features contribute to cross-condition prediction accuracy. In this section, feature-wise prediction performance is evaluated. In Figure 5. 11, the mean absolute error of the prediction using only one feature with full cycle (FC) data and with one best cycle (BC) data are presented, respectively.

In Figure 5. 12, we determine such BC data by calculating the Pearson correlation (ρ) between the feature and lifetime values. As shown in Figure 5. 11, capacity deviation feature class (dQ_min and dQ_var) outperforms the other four feature classes. dQ_min is better than dQ_var, with a lower median absolute error of 4.4 and 7.9 cycles in BC and FC, respectively. It can be explained that dQ_min and dQ_var reflect the capacity degradation information of LIBs, which is directly related to the capacity-determined battery lifetime.

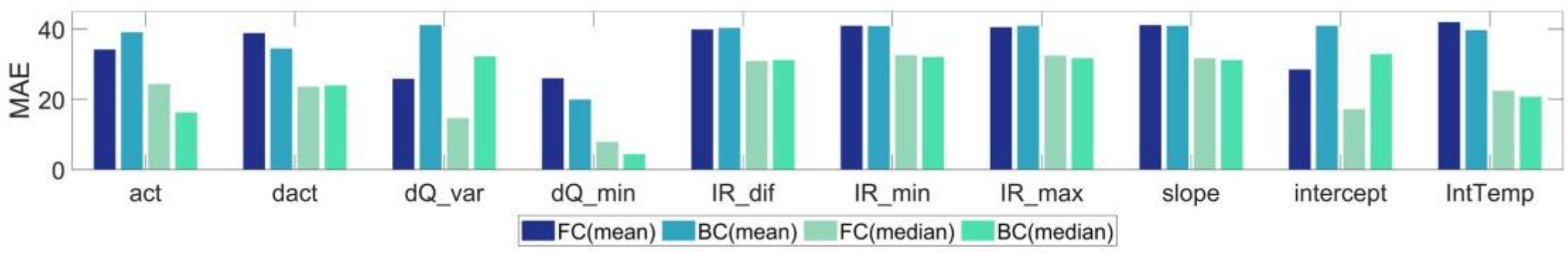

Figure 5. 11 Feature-wise prediction error.

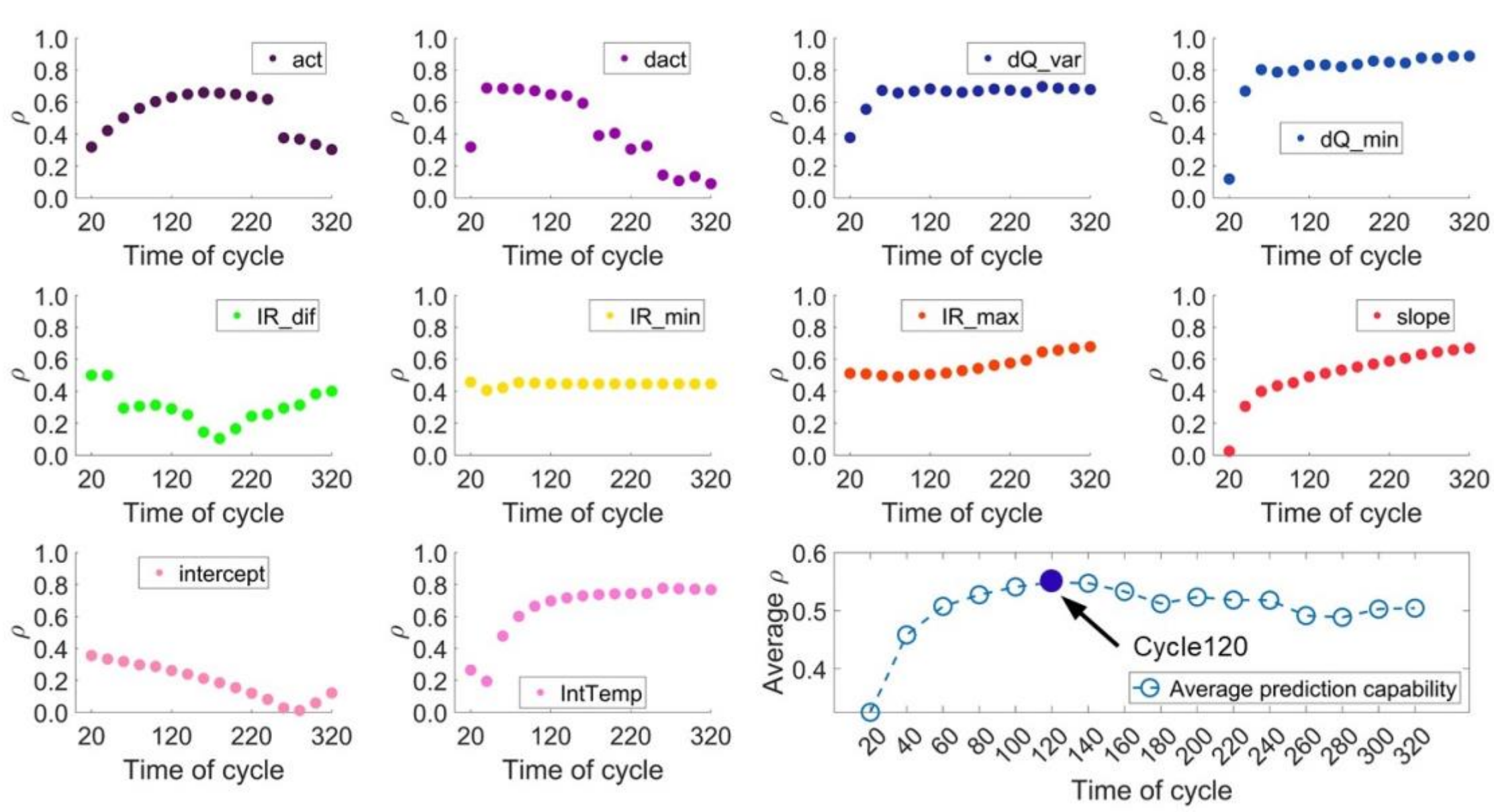

Figure 5. 12 Prediction capability of each feature against the cycle.

### 5.4.3 Performance of Cycle-wise Prediction

The batteries are time-varying dynamic systems, leading to different time responses. As a result, the prediction capabilities of the features also vary with the cycle, influencing the prediction accuracy. The time variation in dQ_min prediction is already observed in Figure 5. 12 that using more data does not produce a more accurate prediction result. For instance, the absolute error is 4.4 cycles in BC data, while the counterpart is 7.9 cycles in FC. Similar results are for the act feature and IntTemp feature. A prediction capability ($\rho$) matrix is presented in Figure 5. 13, illustrating the prediction capabilities with time variations.

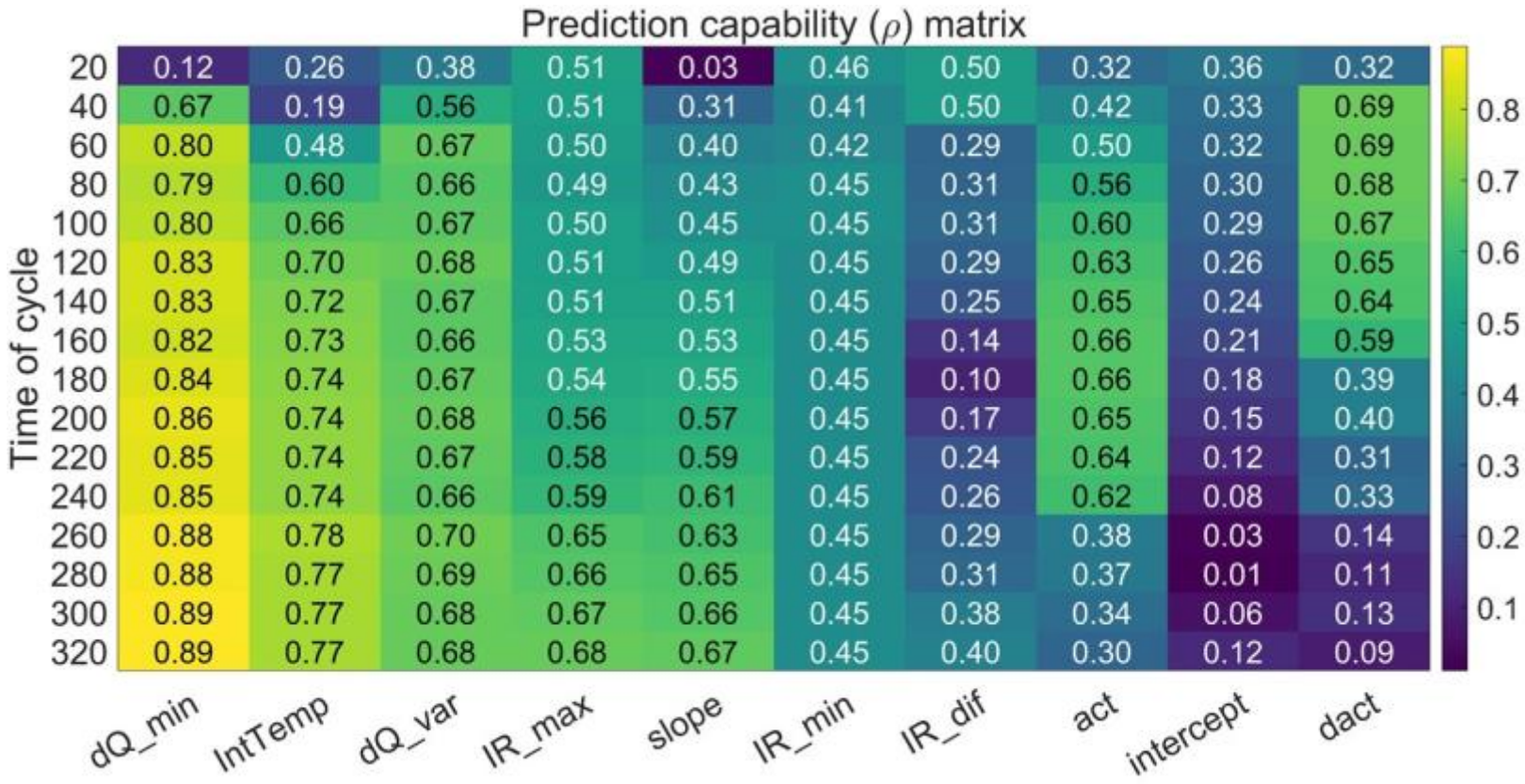

Figure 5. 13 Feature prediction capability matrix.

The prediction capability generally accounts for prediction error in BC. For instance, the prediction capability of dQ_min is 0.79 on average (in FC), while it is 0.89 in BC. Such a difference can be rationalized by the critical time for the feature to capture a stable degradation pattern, with a similar idea in a previous study that the features should be extracted from the middle part to make the feature more stable [207]. In this sense, the prediction capability matrix is a quantitative reference for cycle selection. Regarding the capacity deviation and temperature class, the time variation in the prediction capability exhibits an increasing trend with cycles. However, it is worth noting that such an increase slows down upon cycles, exhibiting a saturation phenomenon. For dQ_min, dQ_var, and IntTemp, the prediction capability accounts for up to 90% of that at their BC at cycle 120, adhering to that of Severson et al. [63] and a number of their following works using cycling data before certain cycles. Those detailed results are presented in APPENDIX R. In Figure 5. 14, the overall prediction capability ($\rho$) for each feature is determined by its cycle-wise distribution, with outlier points indicated. Detailed cycle-wise comparison of error rate distribution using each feature when the time of cycle changes is presented in APPENDIX S. It can be observed that dQ_min, dQ_var, and IntTemp exhibit the lowest error rate with a concentrated distribution. Such concentrated error rate distribution despite cycle changes also indicates these predictive features are promising in early lifetime prediction since they need less time to capture the degradation patterns.

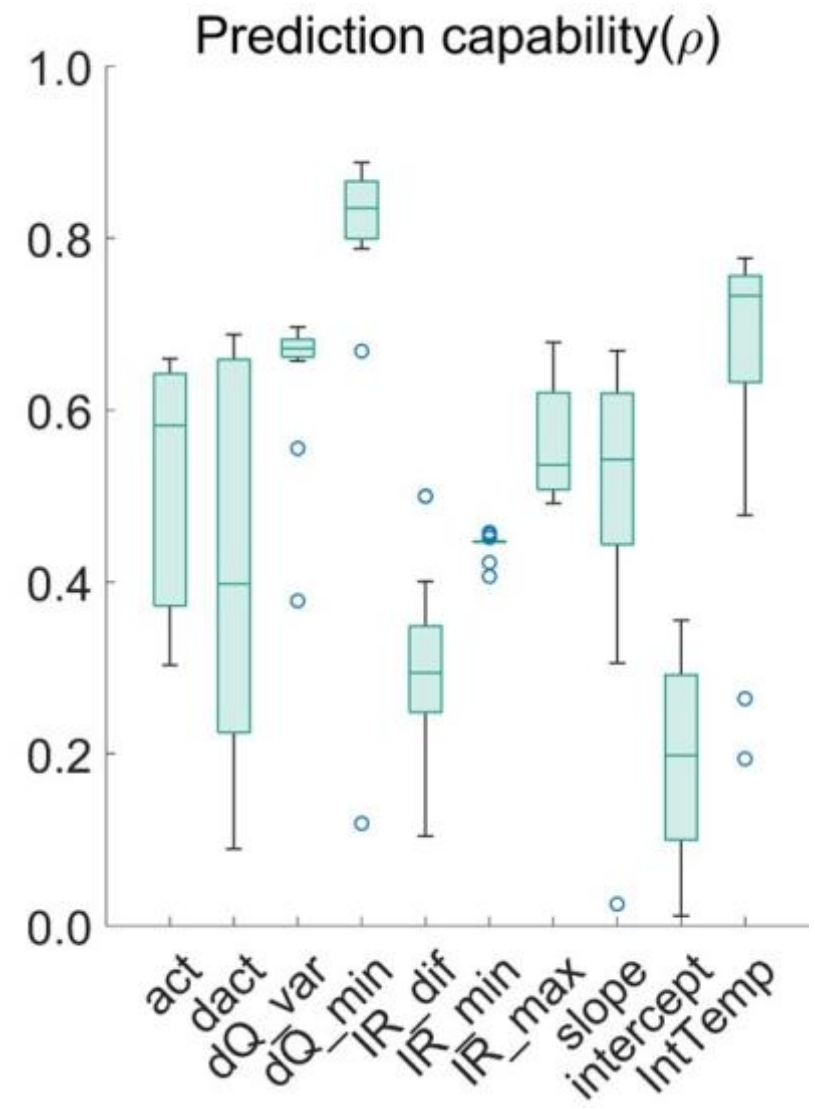

Figure 5. 14 Statistical summary of the prediction capability ($\rho$).

The predictions are attempted using five feature classes (a 10-dimensional feature vector) in each cycle to optimize the prediction performance by the cycle selection. The according result is presented in Figure 5. 15. It turns out that the prediction error decreases until the 120th cycle, with an increasing trend following. Again, it suggests that increasing cycles do not necessarily produce a more accurate prediction, adhering to the saturation phenomenon mentioned earlier in prediction capability. Such a saturation phenomenon can be rationalized by the average prediction capability of the feature vector in Figure 5.

12.

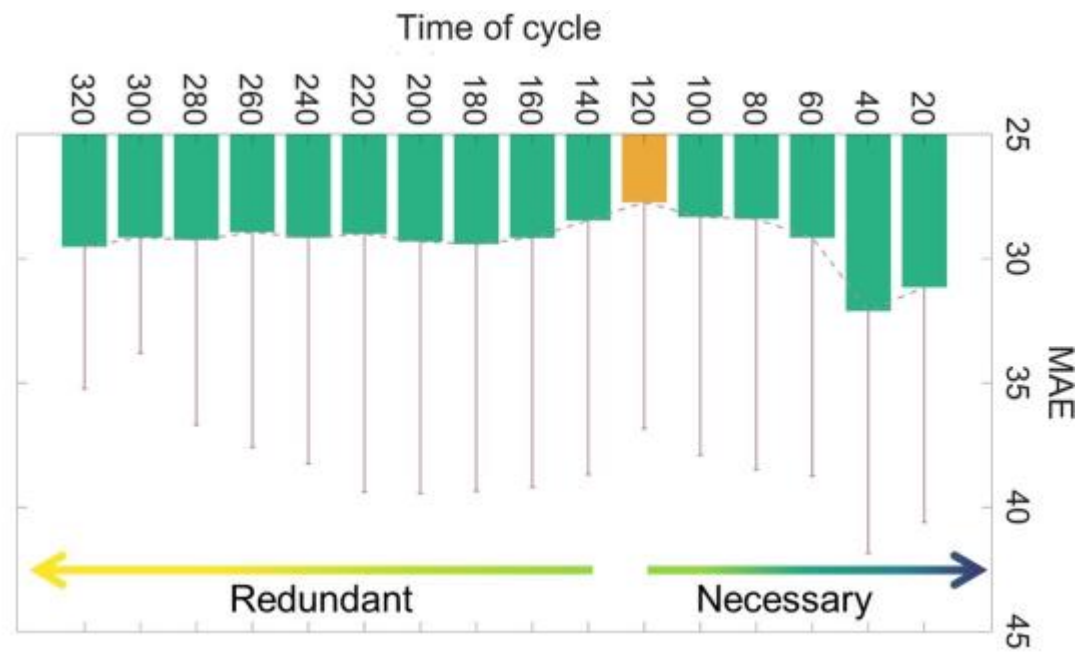

Figure 5. 15 Cycle-wise prediction error using all features.

Specifically, the prediction capability hits its maximum value at the 120th cycle and then slightly decreases. Considering that the sampling interval resolution might pose uncertainty on the cycle selection due to potential information loss, Shannon information entropy is used to validate the information fidelity of interval selection. We evaluate sampling interval resolutions to find a maximized interval to reduce computational effort while keeping information fidelity. Figure 5. 16 shows that the interval of 20 keeps 93% of the information entropy compared with the interval of 15, while the interval of 25 only holds 74% of the information entropy. Thus, the minimized prediction error until the 120th cycle is convincing due to a high information fidelity using a sampling interval of 20. The calculation method of Shannon entropy can be found in APPENDIX T.

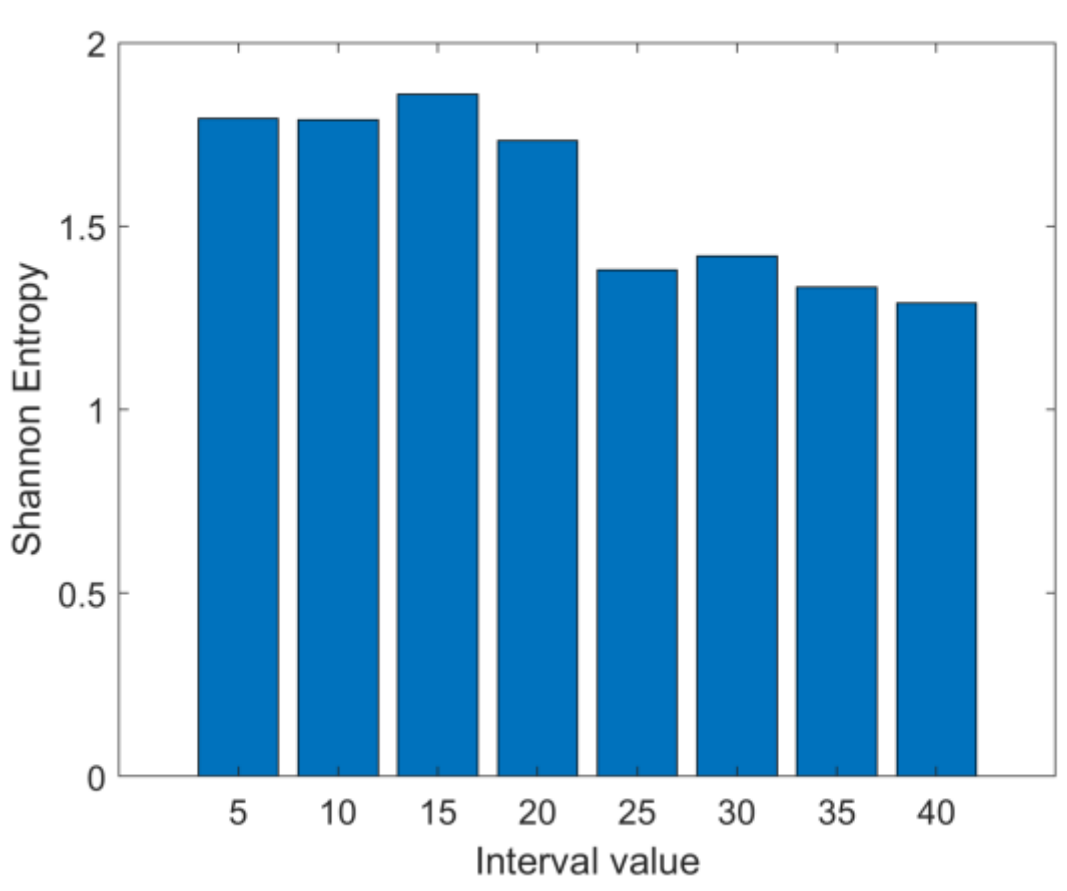

Figure 5. 16 The Shannon entropy against sampling interval values.

### 5.4.4 Sensitivity of CORAL Regularization

Noting that CORAL is a parameterized method, we also examine the sensitivity of such predictions when the model regularization parameter changes. In Figure 5. 17, the result shows that CORAL-aided MLP exhibits strong stability when regularization parameter significantly varies. Note that, the discrete regularization terms for feature alignment are 0.001, 0.01, 0.02, 0.05, 0.07, 0.1, 0.2, 0.5, 0.7, and 1. The discrete regularization terms for lifetime alignment are 0.01, 1, 2, 5, 7, 10, 20, 50, 70, and 100 in $10^4$. Thus, in total, 100 regularization parameter experiments are conducted. This stability indicates that such a method can easily perform parameter tuning in real applications.

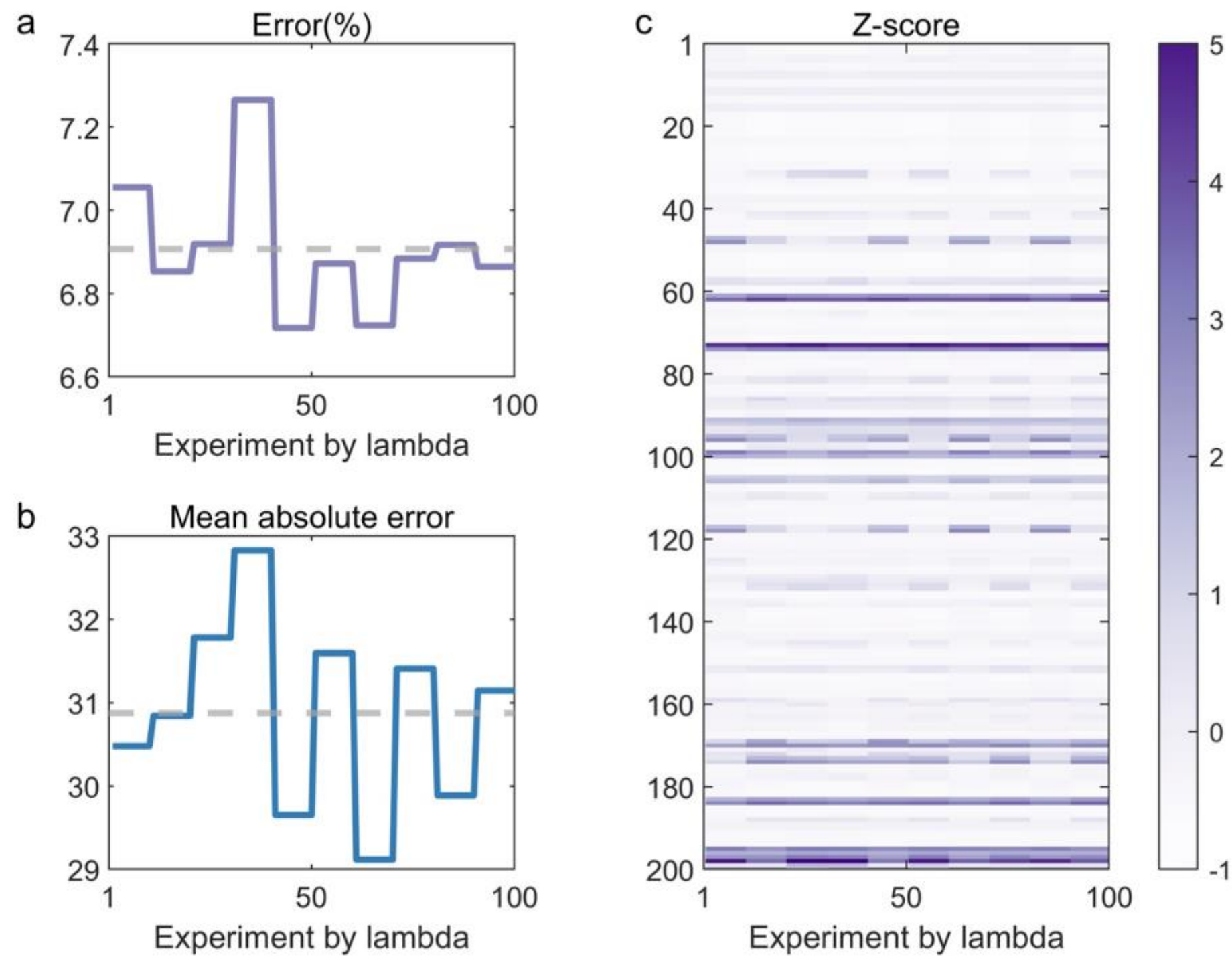

Figure 5. 17 Model parameter stability analysis.

### 5.4.5 Adaptability Evaluation

The feature-wise and cycle-wise prediction analyses are performed in the previous sections. For feature-wise contribution, the capacity deviation and temperature classes are rationalized by their direct relation to primary degradation mechanisms. For cycle-wise contribution, the 120th cycle is identified as a sufficient point to perform whole lifetime prediction, considering there exists a steady state for battery kinetics. In this section, such rationalizations are quantified as the adaptability to facilitate both feature and cycle selection for cross-operation-condition lifetime predictions. The adaptability is the Wasserstein distance of the features between the two fast-charging conditions. This is intuitive because batteries with similar kinetics or equivalent degradation mechanisms should have a slight feature divergence. A high adaptability value indicates a slight feature divergence across operation conditions. From a machine-learning perspective, high adaptability yields a similar feature space, which is advantageous for cross-operation-condition prediction tasks.

The feature-wise adaptability is presented in Figure 5. 18a with the according prediction capability shown in the bar chart. It shows that dQ_min exhibits the highest prediction capability (0.79) and adaptability (0.62). IntTemp and dQ_var rank second and fourth with adaptability at 0.35 and 0.24, respectively. Intercept ranks third, while the prediction capability is not satisfying, which could attribute to the fact that only the first 320 cycles are used for evaluation, while most batteries have not yet exhibited noticeable capacity diving. The internal resistance feature class and charging time feature class exhibit similar adaptability around 0.2, which can be interpreted as the degradation mechanisms in the two discharging conditions being different in type or extent. The LLI is common in fast-charging conditions, influencing the electrode morphology and resistance by lowering the ionic transport speed [208]. However, the LLI can shift to the irreversible LAM under the extremely fast-charging condition, then resistance increases by losing electrical contact, and charging time is influenced as a consequence [63]. In this regard, the internal resistance and charging time classes exhibit poor adaptability.

Figure 5. 18b presents the cycle-wise adaptability of the features. The dact and IR_dif are excluded from evaluation because they are secondary features. There exhibits an increased adaptability in the first 120 cycles, followed by an adaptability saturation until 240 cycles. Then the adaptability dives down after the $240^{th}$ cycle. This phenomenon indicates that the battery kinetic ($LiFePO_4$-graphite system) needs around 120 cycles to establish a stable degradation pattern, which holds for around another 120 cycles under the fast-charging conditions in this work. The adaptability diving can be related to the primary degradation mechanism shift of some batteries with an extremely short lifetime. Specifically, there are two batteries with lifetimes of 300 and 335; the cycle at 240 accounts for a late degradation stage (at least 70% of their total lifetime). For these two batteries, the internal resistance at the end of life (EOL) is 30% larger than the average value. Given the internal resistance changes within 5% over 85% of the lifetime as a general case in the whole data space, such an internal resistance increase indicates a shift in the underlying degradation pattern.

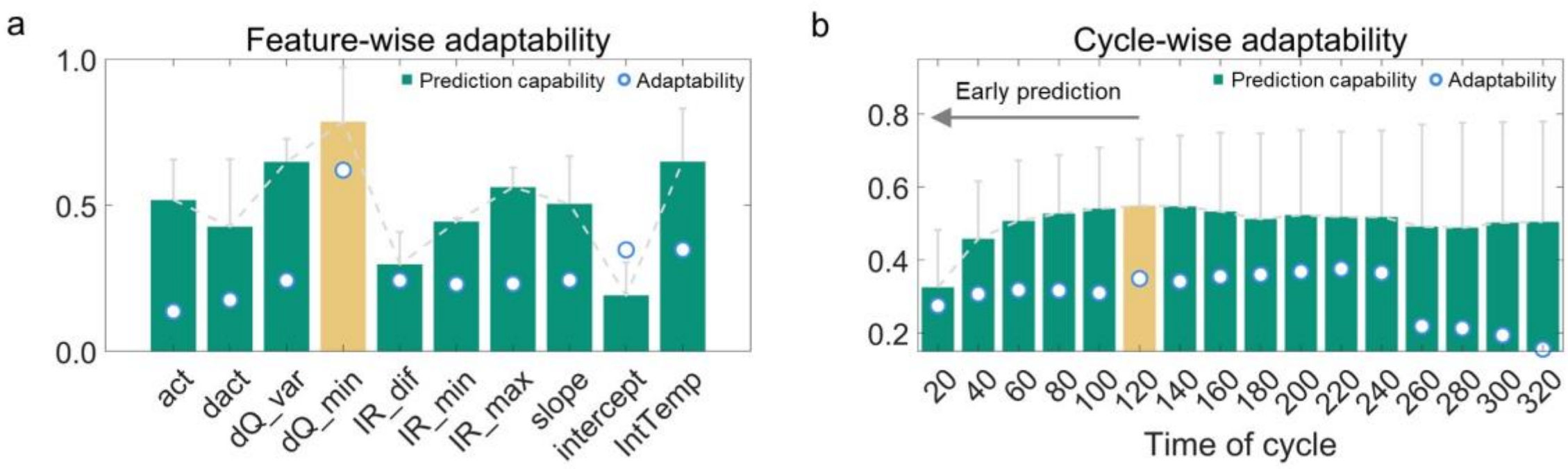

Figure 5. 18 Adaptability evaluation.

### 5.4.6 Rational Feature and Cycle Selection

Adaptability can guide the feature and cycle selection without training and testing the prediction model by trial and error. In Figure 5. 19, the prediction accuracy exhibits a significant linear positive Pearson correlation (0.66 with FC data and 0.60 with BC data, respectively) with the adaptability of the selected feature. Since this linear correlation is produced by all the five feature classes (ten features in total), such correlation is of generality. It is worth noticing that using dQ_min at FC (the first 320 cycles), and BC (the 120th cycle) produces similar prediction accuracy, while the former requires 200 cycles more data. The prediction accuracy exhibits a significant linear positive Pearson correlation with the cycle-wise adaptability, indicating that the cycle selection depicts the establishment time of the primary degradation pattern. However, it is also noticed that feature selection contributes more compared to cycle selection since the former accounts for over 6% prediction accuracy variation. The cycle selection can be regarded as an optimization based on selected features, given a 1% prediction accuracy variation. Nevertheless, the cycle selection significantly improves the computational efficiency by removing redundant degradation information without sacrificing the prediction capability. Specifically, eliminating such redundancy saves up to 99.4% of the computation time while improving prediction accuracy by selecting adaptable features.

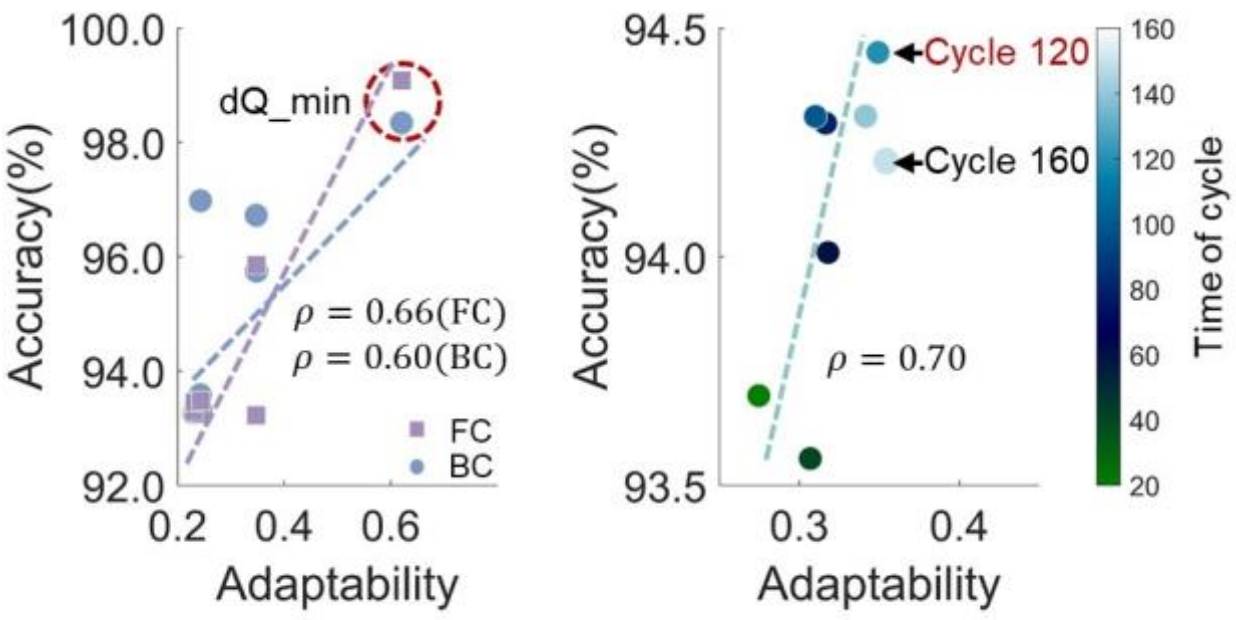

Figure 5. 19 Correlation between feature- (left) and cycle-wise (right) adaptability.

The prediction capability-adaptability matrix is shown in Figure 5. 20 to facilitate the precise feature selection for cross-operation-condition lifetime prediction. It is easy to interpret that good features for cross-condition lifetime prediction should land in the first quartile, where the prediction capability and adaptability are better than other quartiles. As such, dQ_min and IntTemp, in this case, are good features since they attempt prediction by capturing the general aging degradation mechanism across different fast-charging conditions. Overall, the quantitative adaptability evaluation method facilitates the feature engineering process by evaluating the cross-condition similarities. In this way, repeated testing and model tuning is avoided, thus saving computational cost and testing time. The selected highly adaptable features can be regarded as trustworthy, further expediting the commercial use of related lifetime prediction algorithms.

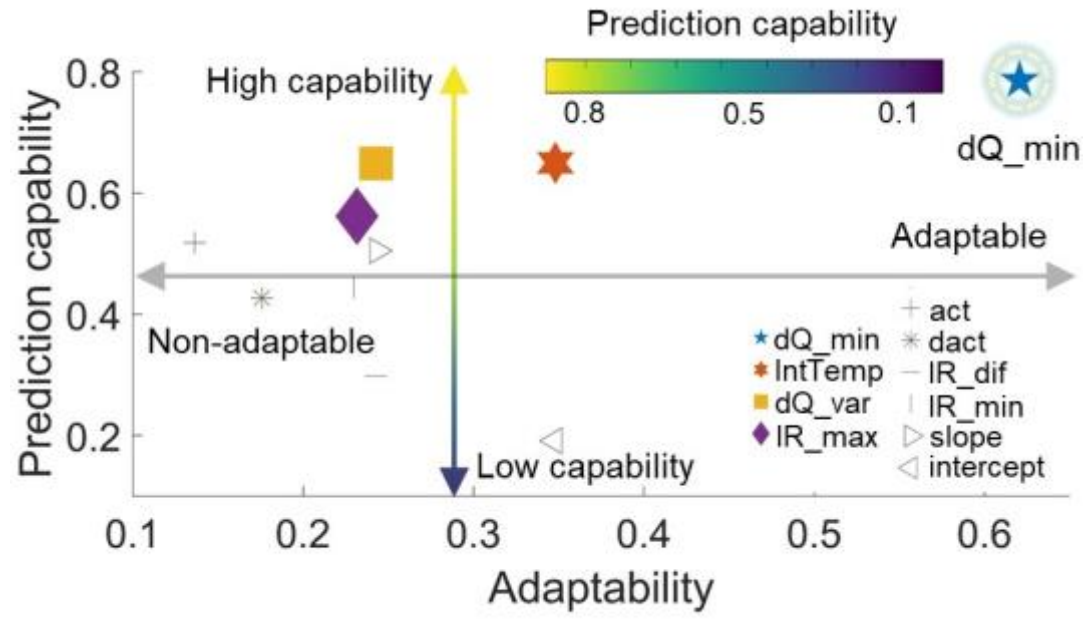

Figure 5. 20 Prediction capability-adaptability matrix.

## 5.5 CORAL For Neural Network Loss Integration

So far, the advantages of the CORAL method in feature and cycle selection have been demonstrated. However, it is still only effective in the feature engineering stage, because this requires all the data in the source and target domains to be analyzed. Such data pairs are difficult to obtain in real scenarios. In this section, we will show how to expand the CORAL method into a loss function for a neural network, so that this method can transfer knowledge with minimal target domain data. To this end, we have verified it on two tasks: charge estimation and residual value estimation. In addition, the form of physical testing has also changed from long cycle data to pulse test data. We aim to demonstrate the generalization ability of the CORAL method through different task types and physical test types.

The CORAL net aims to correct these domain differences from physical formats and capacity designs. This idea was directly inspired by the cross-operation-condition lifetime prediction of in-service LIBs[65]. However, essential difference lands on the integration of CORAL metrics into the loss function of the designed deep neural network, instead of aligning features in the feature engineering stage, where the name of CORAL net comes from. The CORAL loss aligns the distributions of source and target features by minimizing the Frobenius norm of the difference between the covariance matrices:

$$l_{\mathrm{CORAL}} = \frac{1}{4k^2}\|C_s - C_t\|_F^2 \tag{5.7}$$

where, $C_s$ and $C_t$ are the covariance matrices of source domain and target domain features $\widetilde{\boldsymbol{U}}_{\boldsymbol{s}}$ and $\widetilde{\boldsymbol{U}}_{\boldsymbol{t}}$ extracted from the pulse test (see APPENDIX U), respectively. $\|\cdot\|_F^2$ is the squared Frobenius norm. $k$ is the dimensionality of the features, and thus, $k = 21$. The factor $\frac{1}{4k^2}$ normalizes the loss concerning the dimensionality of features, ensuring that the loss values remain comparable even for different feature dimensions.

The loss function of the CORAL net is designed as:

$$L_{CORAL_net} = \boldsymbol{\alpha}^{\boldsymbol{T}} \cdot (l_{MSE}^{SOC_{src}} + l_{MSE}^{SOC_{tgt}} + l_{MSE}^{RRC_{src}} + l_{MSE}^{RRC_{tgt}}) + \beta \cdot l_{\mathrm{CORAL}} \tag{5.8}$$

where, $\boldsymbol{\alpha}$ is a weighting vector for source domain and target domain tasks, including SOC prediction and relative remaining capacity prediction. $\beta$ is a scalar value for the CORAL loss as a regularization that penalizes the model to converge to directions with large domain divergence. $l_{MSE}$ refers to the mean square error. $\boldsymbol{\alpha} = 0.075 \times [2.5 \quad 1.5 \quad 3.0 \quad 2.5]$, and $\beta = 1$ were set. The training and testing ratios were set as 0.8 and 0.2, respectively. Note that such data split is for the union set of originally tested and generated data. The random state was set as 42 in the Python 3.8.15 environment to ensure reproducibility. The training epoch for the CORAL net was 5000.

The pulse data augmentation and neural network implementation for the charge and remaining capacity prediction are detailed in APPENDIX V. Three domains are considered in this section, i.e., the 2.1Ah cylinder cells (Cylind21), 3.1Ah pouch cells (Pouch31) and 5.2Ah pouch cells (Pouch52). These cells are NMC (nickel manganese cobalt oxide) with different degradation status.

### 5.5.1 Charge Status Prediction from Pulse Test

Different from RUL prediction prognostics task presented in the previous sections, in Figure 5. 21a, predictive performance of SOC prediction cross domains is presented as a distinct new diagnostic task. Data collected from Cylind21 is the source domain, assuming the data are available due to accumulative data collection over time. The SOC prediction net can immediately infer SOC levels from the voltage dynamics without any capacity test. A 9.1% MAPE for SOC prediction of Pouch52 second-life batteries is observed using full size of Cylind21data and only 2% of the Pouch52 data, equivalent to 42 second-life battery samples collected at the test field. The Pearson correlation of 0.96 indicates a high linear correlation between the predicted and actual SOC values, and thus, the prediction is highly reliable. More importantly, the SOC net does not compromise the source domain predictive capability, evidenced by a 3.6% MAPE and 0.96 Pearson correlation. In Figure 5. 21b, for another task with more domain divergence, as shown in the voltage dynamics differences between Cylind21 and Pouch31 in APPENDIX W, the SOC net still performs robustly. Still, with only 2% of the Pouch31 data, the SOC net can immediately infer SOC levels with a 6.4% MAPE and a 0.97 Pearson correlation. We observe that the SOC prediction in the higher SOC region of the Clind21->Pouch31 is slightly divergent. This observation can be rationalized by increased domain divergence resulted from the accelerated ageing patterns in Pouch31 (see APPENDIX X), making the CORAL aided transfer between Clind21 and Pouch31 harder.

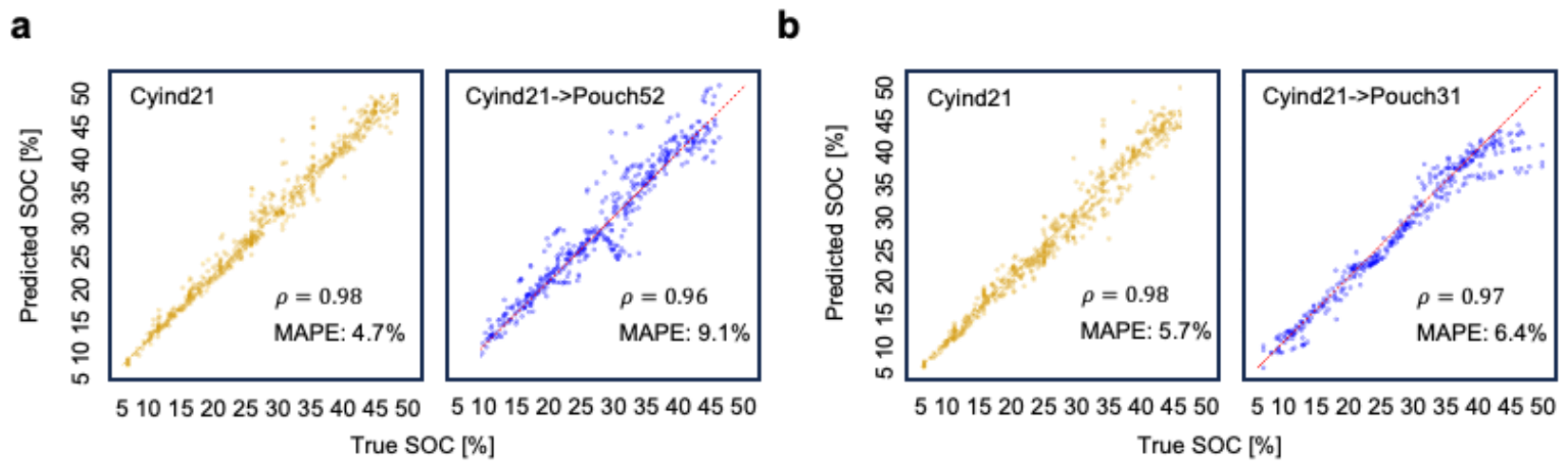

Figure 5. 21 Charge prediction performance.

## 5.5.2 Residual Value Prediction from Pulse Test

Another diagnostics task, i.e., RRC prediction, is further evaluated to demonstrate the model's flexibility. Having access to the SOC information predicted by the SOC net, the voltage dynamics are useful to predict the RRC of the second-life batteries, otherwise, the representability of voltage dynamics to battery degradation is complicated by the mixed SOC distributions, as shown in APPENDIX Y.

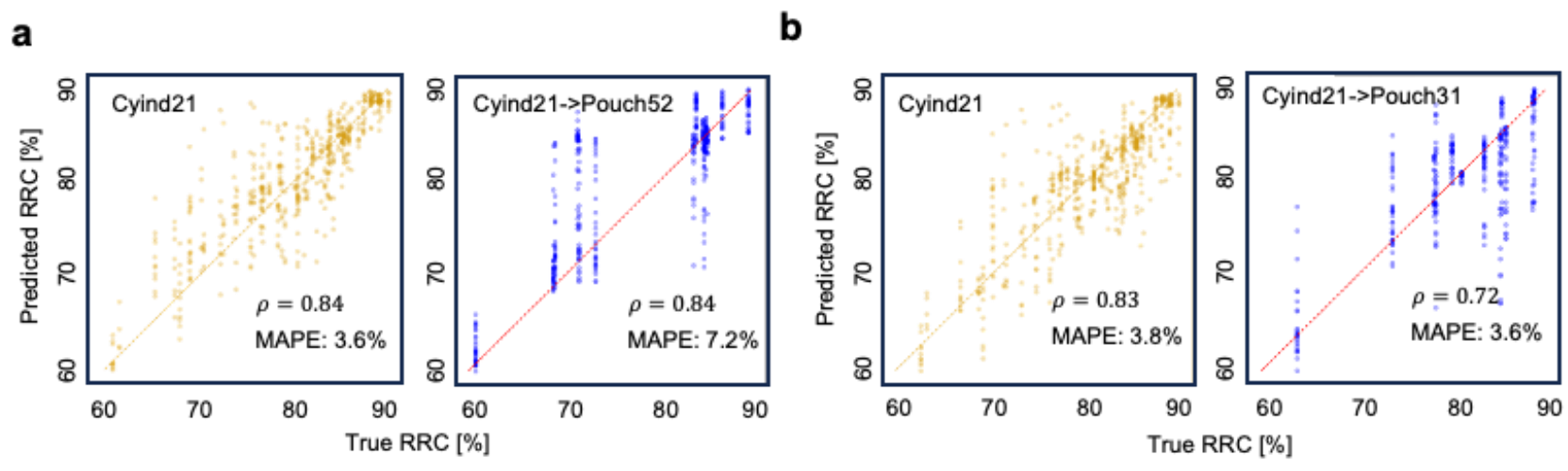

Figure 5. 22 Relative remaining capacity prediction performance.

The predicted SOC condition is appended to the voltage dynamics values that are immediately obtained from the test field for RRC prediction. Figure 5. 22a shows an RRC prediction with a 7.2% MAPE and a 0.84 Pearson correlation, only requiring 2% of the data availability. The precited RRC points show a noticeable concentration in the perfect prediction line (the red diagonal line) regardless of mixed SOC levels, demonstrating that CORAL successfully transforms heterogeneities resulting from the physical formats and capacity designs into a shared statistical space. In Figure 5. 22b, we also observe a robust result for Pouch52 second-life batteries, with a 3.6% MAPE and a 0.72 Pearson correlation. Since the batteries with same RRC are tested multiple times across SOC levels, we note there are some prediction points located far from the diagonal line, however, these sample number is small, thus without significantly impacting the overall RRC prediction error.

### 5.5.3 Generalization Risk Analysis

We perform absolute error frequency analysis to examine prediction uncertainties and risks since second-life batteries are more safety concerned. In Figure 5. 23a, a concentrated absolute error distribution can be observed, where error frequency is calculated by the ratio of the number of instances that error distributed in a certain error range and the total number. When considering a 95% accumulated error frequency, the target domain prediction is robust, with a 4.6% MAPE. In Figure 5. 23b, when transferring to a challenging condition that has a larger domain difference, the prediction result is still stable with a 3.6% MAPE.

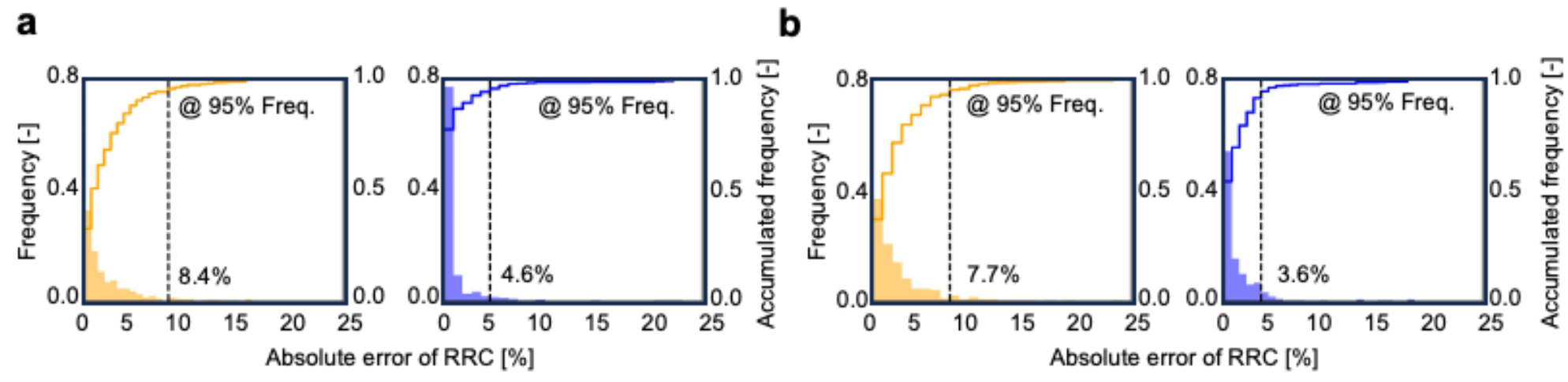

Figure 5. 23 Generalization risk analysis.

We summarize the error values reported in Table 5.4 for a clear presentation and discussion. Both the backward transfer and forward transfer are evaluated. The − and + symbol refer to backward transfer and forward transfer, respectively. The backward transfer means that the transferred model was used to make predictions in the domain where it was originally transferred from, i.e., the source domain. While the forward transfer means that the transferred model was used to make predictions in target domain that guided the transfer. The backward transfer and forward transfer evaluates model's preservability of established knowledge and the transferability to the new knowledge, respectively. It can be seen from the Table 5. 5 that both SOC predictions and RRC predictions demonstrate good knowledge preservability in the original domain, even though the difficulties vary upon different transfer scenarios. Thus, we conclude that the proposed CORAL transformation finds a domain-invariant representation of both source (Cylind21) and target (Pouch52, Pouch31) domain tasks. It is noted that the maximum prediction risk also decreases when transferring knowledge from source domain to target domain, which is preferable to second-life applications since they can be significantly relevant to improved estimation reliability, and robust second-life battery warranty strategy designs. The observed maximum remaining capacity estimation risk given a 95% confidence level is reduced by 49% averagely (45.23% and 53.25% reduction for Cylind21 -> Pouch52, and Cylind21 -> Pouch31 scenarios, respectively).

Table 5. 5 Charge and relative remaining capacity prediction performance.

| Transfer Scenario | SOC Prediction | | RRC Prediction | | |
|---|---|---|---|---|---|
| | $\rho$ | MAPE (%) | $\rho$ | MAPE (%) | Max Risk (%) |
| Cylind21 -> Pouch52 (−) | 0.98 | 4.7 | 0.84 | 3.6 | 8.4 |
| Cylind21 -> Pouch52 (+) | 0.96 | 9.1 | 0.84 | 7.2 | 4.6 |
| Cylind21 -> Pouch31 (−) | 0.98 | 5.7 | 0.83 | 3.8 | 7.7 |
| Cylind21 -> Pouch31 (+) | 0.97 | 6.4 | 0.72 | 3.6 | 3.6 |

### 5.5.4 Data Availability and Benchmarking

For the practical usage of the proposed CORAL aided transfer learning approach, one may be concerned about the impact of the field data availability that is critical to guide the model to generalize, which has significant relevance to time and cost feasibility. The proposed CORAL aided transfer learning approach is compared to the state-of-the-art machine learning models under varied field data availability scenarios. See the APPENDIX Z for detailed experimental settings. The setting is target domain only, which means that all models only have access to data in the target domain. The prediction models are trained and tested only with target domain data, except that the proposed CORAL aided transfer learning approach can use existing data from the source domain, even though data can be significantly different due to the data heterogeneities.

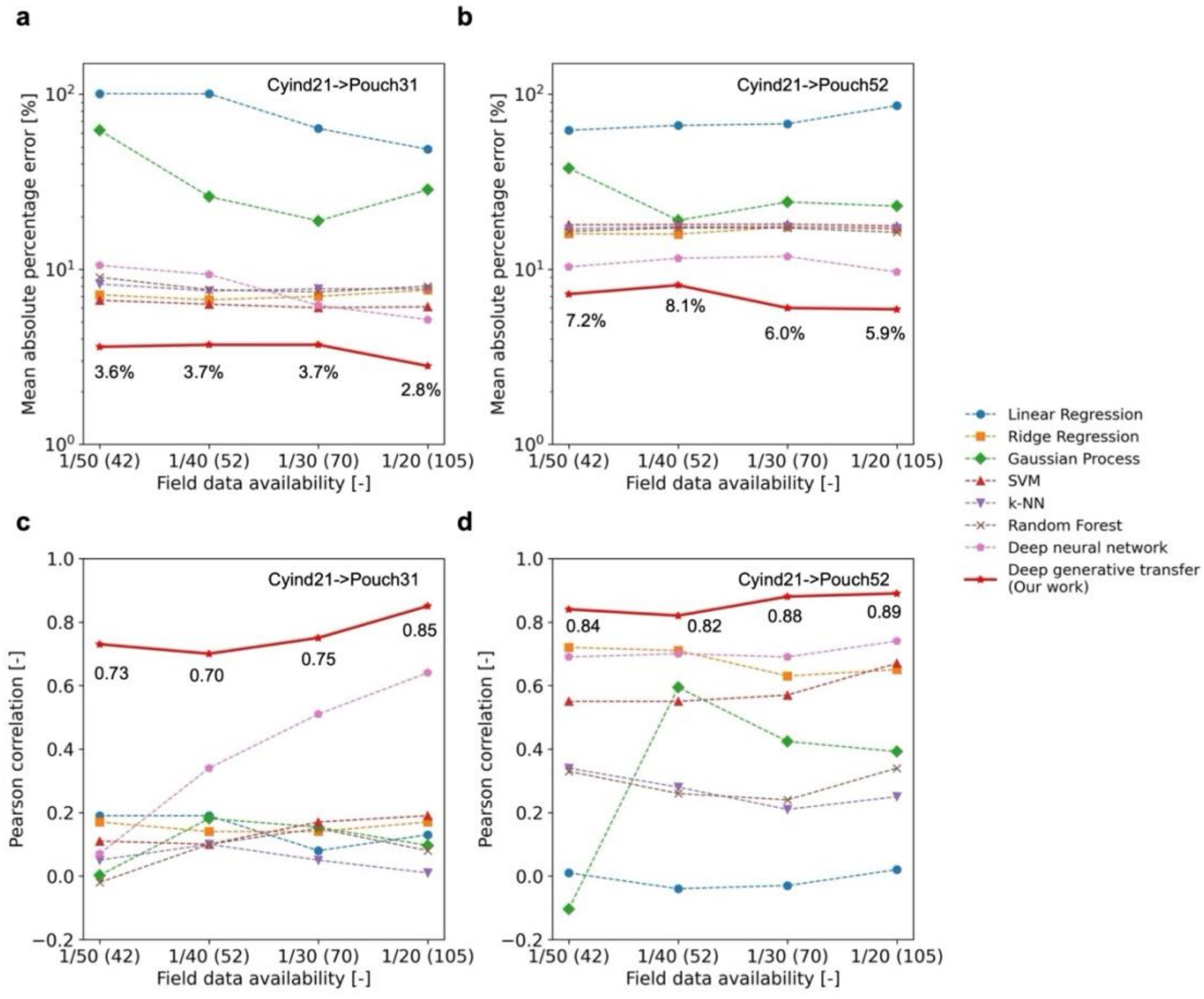

Figure 5. 24 Benchmarking performance.

In Figure 5. 24a and Figure 5. 24b, the CORAL aided transfer learning approach shows consistently lower MAPE across different field data availability and knowledge transfer tasks. In the linear regression, the prediction presents two orders of higher errors than our methods due to insufficiency in handling data scarcity and heterogeneity. When the field data availability increases from 2% to 5%, the deep neural network method approaches the CORAL aided transfer learning approach in an accelerated manner. Figure 5. 24c shows the deep neural network lags behind the CORAL aided transfer learning approach by 0.2 and 0.15 Pearson correlation values for Pouch31 and Pouch52, respectively. This phenomenon suggests that data size is critical in making decent predictions but is still challenged by data heterogeneity. Figure 5. 24d shows a better linearity of prediction when transferring from Cylind21 to Pouch52, which is understandable that the transferability between Cylind21 and Pouch52 is better than that between Cylind21 and Pouch31, see Appendix W. This demonstrates a delicate trade-off between the necessity of knowledge transfer to correct for domain divergences and the extent to which divergences are transferable. Unfortunately, it is a critical challenge that we do not always have sufficient data access for new domains, driving us to generate high-fidelity data from existing data and transfer the established model knowledge from old domains into new ones. This paper establishes a pulse test-based data curation that is fast, non-intrusive, generalizable, and transferable from already measured voltage features within one domain to target domains. The data volume requirement of such target domains is demonstrated to be as little as 2% as compared to that in the established domain. Compared with other state-of-the-art machine learning techniques, our approach stands apart as a stable method with low prediction error and minimized field data requirements.

## 5.6 Summary

To summarize, adaptive battery diagnostics and prognostics are challenging due to feature divergence under different operation conditions. To alleviate such an issue, an adaptive machine-learning framework is created to accelerate the feature engineering procedure across operation conditions. A positive linear correlation between the adaptability of the battery features and the according prediction accuracy is first reported. Using CORAL-aided MLP, we obtain a prediction accuracy of 93.3% using only one feature at the 20th cycle, while saving 99.4% of the computation time compared with the full feature full cycle model. A cross-dataset validation is performed under an unseen operation condition with a 90.3% prediction accuracy, suggesting our framework is of generality. The success of the framework is rationalized by quantifying the feature-wise and cycle-wise degradation information with the help of the adaptability evaluation. However, the data used in this paper is for multi-step fast-charging usage. Considering the difference between the aging path of the battery under multi-step fast-charging and the regular constant current (constant voltage) or dynamic cycling aging path, the numerical distribution of adaptability between different feature classes may change; thus, the 'best feature' may vary upon battery usage. We also noticed that the adaptability analysis is for the $LiFeO_4$-graphite technology. Considering that even for the same manually extracted features, there could be differences in different battery technologies. For example, dQ_min was found to be the strongest feature of adaptability in this paper, underpinned by the discharge plateau of the LiFeO4-graphite technology. For NMC batteries, however, the discharge plateau is not apparent, and the validity of dQ_min needs further verification. Nevertheless, we highlight adapting easily accessible features to inaccessible ones by studying the adaptability metric.

This CORAL method can be extended into battery management systems for cross-operation-condition predictions with distinct tasks (i.e., SOC estimation and RRC prediction), physical measurement (i.e., long cycling test and pulse test) by incorporating the proposed CORAL loss term into a neural network. The CORAL aided model achieves reliable SOC predictions with an MAPE of 9.1% for Pouch52 and 6.4% for Pouch31. Using predicted SOC as conditional information, the remaining capacity estimation obtained an MAPE of 7.2% for Pouch52 and 3.6% for Pouch31. The results were using only 2% of the field available data, and thus, considerable data curation and cost could be saved.

In summary, this method also shows high compatibility with different battery testing methods, such as long cycle testing and fast pulse testing, and achieves model adaptability and method consistency in the entire process from battery data collection, data processing, model construction to status assessment. Distinct from previous models that struggle with domain divergence or require extensive historical data curation, the proposed approach generalizes effectively with minimal field data requirements. This presents significant advances for battery reuse and recycling decision-making, where data availability is often limited. The proposed method inspires that one could preferentially generalize non-OOD data (already measured) and then adapt to OOD conditions (test field site) under the guidance of available field data since data curation time and cost are major challenges for many energy-related control and management tasks.

# CHAPTER 6 CONCLUSION AND OUTLOOK

## 6.1 Conclusion

This dissertation systematically investigates the core scientific challenges of data scarcity and data heterogeneity in LIB diagnostics and prognostics, aiming to enable sustainable remanufacturing, reuse, and recycling. These two challenges emerge consistently across all battery lifecycle stages, from lack of early-cycle data in prototype validation, to sparsity and variability of degradation signals in retired batteries, privacy-constrained fragmentation of recycling-relevant data and adaptive lifetime management. Using physics-informed understanding and machine learning, this dissertation proposes a unified framework that addresses the data scarcity and data heterogeneity challenges, offering robust and interpretable solutions for battery diagnostics and prognostics under real-world constraints. Specifically,

CHAPTER 2 introduces a physics-informed prototype verification framework for as-manufactured batteries using early-cycle data. To tackle the challenge of limited temporal information, a long-term and multi-step fast charging protocol is designed to amplify degradation signatures within short time. Feature extraction rooted in thermodynamic and kinetic loss mechanisms enables interpretable feature engineering representations. A physics-inspired transferability metric based on Arrhenius law bridges cross-temperature heterogeneity, guiding uni- and multi-domain adaptation. The proposed battery degradation trajectory prediction model shows high accuracy using just 4% of total cycle data, supporting rapid, non-destructive quality screening and informed material recycling of defective prototypes, thus addressing critical industrial sustainability challenges of defective battery prototype remanufacturing before massive production.

CHAPTER 3 focuses on residual value estimation under unknown SOC conditions, where retired batteries lack historical usage information. A rapid and generalizable pulse test protocol is designed to gather degradation indicators with reduced time and energy cost. To mitigate data heterogeneity and scarcity under random retirement conditions, a generative neural network with cross-attention mechanisms is proposed to synthesize high-quality pulse voltage signals to enable reliable residual estimation even for unseen battery samples. These synthetic signals successfully support downstream regression models for residual estimation, outperforming traditional capacity calibration methods and the state-of-the-art learning methods that currently ignore random SOC conditions. The cost and carbon footprint analysis further demonstrates the practical viability of the proposed generative learning methods in battery reusing with rapid residual evaluation.

CHAPTER 4 addresses the mixed cathode material challenges in recycling stage, where direct recycling routes are highly material-dependent. In this context, battery data is often fragmented and privacy-sensitive, preventing efficient data-driven material identification. A FL framework is proposed to enable collaborative model training across distributed data owners while preserving data confidentiality. By incorporating PBs, client simulation, and ensemble voting strategies, the model robustly adapts to data distribution shifts and incomplete information scenarios, significantly improving cathode-type classification accuracy. By utilizing the features extracted from long charge-discharge cycle (only one cycle at EOL), the proposed model exhibits 1% and 3% cathode sorting errors under homogeneous and heterogeneous recycling settings, respectively. The results demonstrate that collaborative learning can unlock otherwise isolated battery knowledge using distributed decision-making, and thus optimizing recycling decisions in a privacy-aware and scalable manner.

CHAPTER 5 extends the data-centric methods established in previous chapters by introducing an adaptive, unified and consistent framework for battery diagnostics and prognostics under diverse and dynamic operation conditions. To address cross-condition generalization, a CORAL strategy is proposed to align feature distributions across domains by minimizing the statistical distribution differences. This framework is validated on three critical tasks, i.e., charge status estimation, residual value estimation and RUL prediction, compatible with two distinct measurements, i.e., long-term cycling and short pulse test, respectively. CORAL demonstrates strong compatibility with these tasks and physical measurement methodologies, enhancing model robustness under shifting battery data availability and heterogeneity conditions. Moreover, CORAL is proven to be effective both at feature preprocessing stage and within the neural network loss function, enabling interpretable, transferable, and condition-aware machine learning. Feature-cycle sensitivity analysis informs rational selection of adaptive and predictive features. As a continuation and generalization of prior chapters, this chapter consolidates earlier insights into data scarcity, heterogeneity, delivering a unified solution for adaptive and consistent battery diagnostics and prognostics across the lifecycle for the maximized sustainability.

## 6.2 Limitation

While this dissertation validates the effectiveness of addressing data scarcity and heterogeneity through data collaboration, data generation, and domain adaptation, several natural extensions remain open for limitation discussion. Specifically,

(1) Data Collection. Data collection is a prerequisite for the machine learning model, as training cannot proceed without sufficient data. In the battery community, the available dataset size has been a major concern. Current publicly available battery test datasets cover only about 1,300 physical batteries [209, 210]. For retired batteries, the data scarcity issue becomes more pronounced, compounded by the challenge of data heterogeneity. However, battery data is not truly scarce. Major battery manufacturers conduct extensive battery testing, but due to trade secrets, these datasets remain inaccessible. The potential for multi-entity data collaboration exists, yet the challenge of privacy protection persists [18]. Another avenue is leveraging existing test data while applying generative AI to simulate and expand datasets. It has brought attention to the researchers in many battery-related tasks, such as thermal runaway prediction [116], energy consumption estimation [211], state of energy estimation [212], SOC estimation [213], RUL prediction [214], degradation path prediction [215]. However, this approach is still primarily focused on first-life batteries, with generative techniques for retired batteries being in their early stages.

(2) Feature engineering. Although end-to-end machine learning models can make powerful predictions without expert knowledge, retired batteries require interpretable results for second-life decision-making. Feature engineering, utilizing techniques such as incremental capacity, differential voltage curves, and their variations, provides domain-specific insights into degradation patterns, favorable to in-depth degradation insights [66, 194, 216, 217]. However, retired batteries provide incomplete data that lacks continuity across the full SOC range, making it difficult to analyze degradation behaviors over time. Due to heterogeneities, data collected vary across diversified retired batteries, complicating the extraction of domain-invariant and meaningful features from disparate datasets, making it challenging to generalize the feature engineering results. Thus, automated, yet interpretable, and generalized feature engineering are still highly underdeveloped in the field of retired batteries.

(3) Model interpretability. Model interpretability is critical for trustworthy machine learning, particularly in applications like second-life battery management. While pre-model interpretability can be achieved from feature engineering, interpretability during and after model construction remains a challenge. Methods like attention mechanisms, interpretable decision trees, and rule-based models progressed in enhancing interpretability during the modeling process, but more theoretical breakthroughs are needed for reliable prediction [218-221]. The post-hoc interpretability methods, such as Shapley additive explanation, and local interpretable model-agnostic explanation are the most practical since they do not influence model construction [222-224]. Both pre- and post-hoc interpretability are essential to ensure the safety and reliability of machine learning driven decisions for second-life battery management.

(4) Model adaptability. The adaptability of machine learning models is crucial for handling the heterogeneous data of retired batteries. One major challenge is OOD data, where the model encounters new, previously unseen conditions. This requires models to be adaptable to shifting data distributions, making transfer learning a promising solution [65, 66, 175, 194, 225]. Transfer learning allows models to apply knowledge learned from one domain to another. For instance, in reuse batteries, where operating conditions constantly change, models trained on historical data struggle to generalize to new scenarios defined by new data distributions. However, transfer learning alone may not be sufficient due to the improved task performance in the new domain by sacrificing the old, referred to as catastrophic forgetting [226]. Continual learning ensures that the built model can adapt to new data distributions without sacrificing knowledge on previously learned domains, maintaining a balance between old and new tasks [227-229]. Both transfer and continual learning are particularly relevant for retired batteries due to the constantly evolving, diverse applications and their accumulated data streams. However, research into applying these techniques to second-life battery scenarios is still in its early stages and presents significant opportunities for further development.

(5) Model uncertainties. Managing the model uncertainty is critical for ensuring the commercial feasibility of retired batteries, particularly in warranty design. A key challenge is that simply predicting the expected lifetime of a battery is not enough since understanding the uncertainty of predictions is essential [41, 230, 231]. If the uncertainty range is too wide, warranties can become economically unviable, as batteries may perform worse than expected average performance. Moreover, overestimating the usable life due to underrepresented uncertainty could also pose safety risks. Despite the existence of methods for measuring uncertainty in battery performance, their application to retired batteries is in its infancy. This is due to two key challenges: data scarcity, which leads to an unstable sample space, and data heterogeneity, which causes shifts in the sample space. Retired batteries often have incomplete data and highly variable histories, making it difficult to train models with a stable and representative dataset, making the uncertainty quantification an important but challenging task.

## 6.3 Outlook

This dissertation lays a solid foundation for intelligent, interpretable, and scalable battery health management throughout the entire lifecycle, yet its full potential unfolds in broader and more dynamic application contexts. Specifically,

(1) Physics Knowledge. Incorporating physics knowledge into AI is a powerful way to enhance interpretability and robustness. One method is to embed physical principles directly into the loss function, ensuring the model adheres to known physical laws while learning from the data [232-237]. This technique, known as physics-informed neural networks, has been used to integrate differential equations and physics-based constraints into the learning process, improving model accuracy and stability. However, applying this approach becomes less practical for retired batteries, where the leading degradation mechanisms are complex and not easily defined. These mechanisms are often unknown or vary significantly between different battery types and usage histories, making it difficult to apply a one-size-fits-all approach. Future research should explore more non-invasive characterizations for prior knowledge about battery degradation, thus enabling effective integration of physics knowledge into AI models.

(2) Data Sufficiency. It is important to approach the expansion of datasets with caution. While the volume of available datasets is considered considerably smaller compared to other AI domains, simply increasing the number of battery tests in a single domain does not necessarily lead to greater data diversity. To effectively expand battery datasets, more tests should focus on a wide range of battery materials, physical formats, capacity designs, usage histories, and aging mechanisms tested under varied conditions. This is of relevance to reuse that retired batteries are intended to serve new applications, but datasets with separate usages yet physically identical retired batteries are unavailable. Future research must focus on evaluating the informativeness of existing datasets to ensure meaningful data expansion with careful consideration of data sufficiency.

(3) Model Cost. While AI models are increasingly applied in battery diagnosis and prognosis, the cost of models is often overlooked, yet it is a critical barrier to real-world deployment. The cost includes data acquisition, model training, and model deployment [238, 239]. Data acquisition is particularly expensive, and by incorporating physics-based knowledge and data sufficiency analysis, unnecessary data collection can be reduced [240]. Similarly, training costs are high due to the complexity of aging mechanisms, which drives the need for advanced deep learning models, where techniques like transfer learning, continual learning, and model optimization can help mitigate costs [65, 66, 241-243]. For deployment, model interpretability can aid in trimming redundant knowledge, ensuring a balance between efficiency and performance, especially in hardware-constrained environments. Therefore, a techno-economic analysis of AI models in the retired battery field is urgently needed to determine their scalability and commercial viability, making it essential to assess these cost factors for broader industry adoption.

(4) LLM potentials. Large Language Models (LLMs) present exciting opportunities for expanding data utilization in battery research, particularly beyond traditional physical testing data [244]. Current studies emphasize the potential of LLMs to extract valuable insights from textual data sources such as academic papers, patents, technical reports, and conference proceedings, offering a new avenue for expanding usable data landscape for AI models [245, 246]. This approach can streamline literature reviews, synthesis, and extraction of technical knowledge. However, LLMs for retired batteries require careful evaluation. Key concerns include ensuring that the extracted insights apply to the highly heterogeneous nature of retired batteries and that the model is equipped to handle domain-specific degradation mechanisms. Additionally, addressing data privacy and intellectual property concerns is crucial, as textual data can be proprietary or contain false information.

(5) Model standardization. Model standardization is essential as AI applications for the sustainable use of retired batteries are still in the very initial stages [247]. Standardization efforts must focus on data collection, model construction, and model evaluation, but currently in vain. In terms of data collection, while pulse injection techniques have been proposed for fast battery diagnostics, the parameters, such as pulse width, pulse intensity, pulse combinations, and SOC ranges for pulse injection, need further generalization to ensure consistent feature engineering [40, 248-250]. For model construction, the multi-task nature of reuse, recycling, and remanufacturing calls for standardized frameworks that can generalize data for different AI tasks, without relying on individualized models [238, 251-253]. Lastly, model evaluation must follow industry-standard benchmarks that define acceptable error margins. Establishing these standards will enhance the scalability and commercial viability of AI-driven sustainable retired battery management.

In summary, as battery technologies diversify and lifecycle demands become increasingly complex, future research is encouraged to deepen the fusion of physics knowledge and experimental data, enabling models that are not only accurate but theoretically grounded and trustworthy. Expanding the representativeness and structure of datasets, rather than scale alone, will be key to enhancing generalizability across chemistries, formats, and usage scenarios. In parallel, efforts to reduce sensing costs and computational overhead will support model deployment in resource-constrained environments. Progressions toward continual, adaptive, and uncertainty-aware learning further holds promise for real-time applications where battery behaviors evolve. Together, these directions point to a future where data-driven intelligence is seamlessly integrated into sustainable battery ecosystems.

It is imperative for both industry and academia to work collaboratively toward establishing standardized scenarios for the utilization of batteries for maximized lifecycle sustainability. This involves setting clear guidelines and frameworks and ensuring the interoperability of different methods across various sectors. A unified standard would address existing inconsistencies and promote decision-making regarding whether, when, and how to put into practice the remanufacturing, reuse, and recycling of the batteries. Additionally, as machine learning increasingly plays a central role in optimizing these processes, it is essential to ensure that the clean nature of data-driven battery utilization is maintained. This can be achieved by focusing on the development of energy-efficient and cost-effective machine learning solutions and ensuring that their application aligns with sustainability development goals.

# APPENDIX A FEATURE ENGINEERING: DEFINITION

Table A.1 Feature engineering.

| ID | Taxonomy | Name | Description | Physical meaning |
|---|---|---|---|---|
| 1 | - | T | Operation temperature | - |
| 2 | Prior-cycle | U1 | Cut-off voltage value when assigned SOC is hit at each charging step | Charge acceptance at each charging step (SOC region) [254, 255] |
| 3 | | U2 | | |
| 4 | | U3 | | |
| 5 | | U4 | | |
| 6 | | U5 | | |
| 7 | | U6 | | |
| 8 | | U7 | | |
| 9 | | U8 | | |
| 10 | | U9 | | |
| 11 | In-cycle (inter-step) | VC89 | Voltage changes from the end of step 8 to the start of step 9 | Ohmic and electrochemical polarization, linked to SEI growth (pseudo relaxation) [256, 257] |
| 12 | | VD9 | Voltage drops from the start of step 9 to the minimum of step 9 | Concentration polarization (pseudo relaxation) [154] |
| 13 | | tVD9 | Time needed for VD9 | Recovery time of concentration polarization (pseudo relaxation) [154] |
| 14 | | ReVC | Voltage changes from the end of step 9 to the start of the rest | Ohmic and electrochemical polarization, linked to SEI growth (relaxation) [256, 257] |
| 15 | | ReVD | Voltage drops from the start of the rest to the minimum of the rest | Concentration polarization (relaxation) [154] |
| 16 | | tReVD | Time needed for ReVD | Recovery time of concentration polarization (relaxation) [154] |
| 17 | In-cycle (intra-step) | Vg1 | Mean value of voltage gradient at each charging step | Polarization speed at each charging step (SOC region) [258] |
| 18 | | Vg2 | | |
| 19 | | Vg3 | | |
| 20 | | Vg4 | | |
| 21 | | Vg5 | | |
| 22 | | Vg6 | | |
| 23 | | Vg7 | | |
| 24 | | Vg8 | | |
| 25 | | Vg9 | | |

Continued Table A.1 Feature engineering.

| ID | Taxonomy | Name | Description | Physical meaning |
|---|---|---|---|---|
| 26 | In-cycle (inter-step) | RVg | Ratio of Vg2 and Vg1 | |
| 27 | In-cycle (intra-step) | Q1 | Charging capacity value when assigned SOC is hit at each charging step | Charge acceptance at each charging step (SOC region) [65] |
| 28 | | Q2 | | |
| 29 | | Q3 | | |
| 30 | | Q4 | | |
| 31 | | Q5 | | |
| 32 | | Q6 | | |
| 33 | | Q7 | | |
| 34 | | Q8 | | |
| 35 | | Q9 | | |
| 36 | In-cycle (intra-step) | RL1 | Ratio of voltage and charging current at each charging step | Merged representation of ohmic, electrochemical, and concentration resistance at each charging step (SOC region) [259] |
| 37 | | RL2 | | |
| 38 | | RL3 | | |
| 39 | | RL4 | | |
| 40 | | RL5 | | |
| 41 | | RL6 | | |
| 42 | | RL7 | | |
| 43 | | RL8 | | |
| 44 | | RL9 | | |
| 45 | In-cycle (inter-step) | RO1 | Ratio of voltage change and current change at switching points between steps | Ohmic resistance from relaxation behaviors [256, 257] |
| 46 | | RO2 | | |
| 47 | | RO3 | | |
| 48 | | RO4 | | |
| 49 | | RO5 | | |
| 50 | | RO6 | | |
| 51 | | RO7 | | |
| 52 | | RO8 | | |

# APPENDIX B FEATURE ENGINEERING: CALCULATION

For cut-off voltage features in the prior-cycling stage:

$$U_i = V(soc = soc_i) \tag{B.1}$$

where, $soc_i = \{8, 20, 30, 40, 50, 61.1, 71.1, 81.1, 97\} \times 100\%$, $i = \{1,2,..,9\}$, $V$ is battery voltage, $soc$ is the state of charge, and $soc_i$ is the accumulated state of charge at the $ith$ charging step.

For inter-step voltage transient features and pseudo relaxation features in the in-cycling stage:

$$VC89 = V_9(start) - V_8(end) \tag{B.2}$$

where, $V_9$ and $V_8$ are the voltage vector in the ninth and eighth charging stage, respectively. $start$ and $end$ stand for the first and last voltage values in the vector, respectively.

$$VD9 = V_9(start) - \min(V_9) \tag{B.3}$$

where, $V_9$ is the voltage vector in the ninth charging stage. $min$ is the minimum operator.

$$tVD9 = t(V = V_9(start)) - t(V = \min(V_9)) \tag{B.4}$$

where, $V_9$ is the voltage vector in the ninth charging stage, $min$ is the minimum operator, $t$ is time and $V$ is the battery voltage.

For inter-step relaxation features in the in-cycling stage:

$$ReVC = V_9(end) - V_{re}(start) \tag{B.5}$$

where, $V_9$ and $V_{re}$ are the voltage vectors in the ninth charging stage rest stage, respectively. $start$ and $end$ stand for the first and last voltage values in the vector, respectively.

$$ReVD = V_{re}(start) - \min(V_{re}) \tag{B.6}$$

where, $V_{re}$ is the voltage vector in the rest stage. $min$ is the minimum operator.

$$tReVD = t(V = V_{re}(start)) - t(V = \min(V_{re})) \tag{B.7}$$

where, $V_{re}$ is the voltage vector in the rest stage. $min$ is the minimum operator. $t$ is time and $V$ is the battery voltage. For better time-sensitivities, $\min(V_{re})$ is taken when the 80% of the maximum voltage drop is hit.

For intra-step voltage gradient features in the in-cycling stage:

$$Vg_i = mean(G(V_i)) \tag{B.8}$$

where, $i = \{1,2,..,9\}$, $V_i$ is the voltage vector at the $ith$ charging step, $G$ is the gradient operator and $mean$ is the mean operator.

For intra-step capacity features in the in-cycling stage:

$$Q_i = Q(soc = soc_i) \tag{B.9}$$

where, $soc_i = \{8, 20, 30, 40, 50, 61.1, 71.1, 81.1, 97\} \times 100\%$, $i = \{1,2,..,9\}$, $Q$ is battery charging capacity, $soc$ is the state of charge, and $soc_i$ is the accumulated state of charge at the $ith$ charging step.

For lumped resistance features in the in-cycling stage:

$$RL_i = \frac{V_i(end) - V_i(start)}{I_i} \tag{B.10}$$

where, $i = \{1,2,..,9\}$, $V_i$ and $I_i$ are the voltage vector and current value at the $ith$ charging step, respectively. $start$ and $end$ stand for the first and last voltage values in the vector, respectively.

For ohmic resistance features in the in-cycling stage:

$$RO_i = \frac{V_{i+1}(start) - V_i(end)}{I_{i+1} - I_i} \tag{B.11}$$

where, $i = \{1,2,..,8\}$, $V_i$ and $I_i$ are the voltage vector and current value at the $ith$ charging step, respectively. $start$ and $end$ stand for the first and last voltage values in the vector, respectively.

# APPENDIX C MODEL BENCHMARKINGS

Model 1 (LSTM):

Model 1 is established to examine challenges in the long-term prediction capability of our physics-informed machine learning strategy, incorporating knowledge fusion using guiding samples and early data of batteries to be verified. In Model 1, we transform the 42-dimensional features, into time series data with a time step of 10, which means we use a sliding window with a length of 10 to predict the next capacity point iteratively. We use data from early cycles, i.e., the first 200 cycles to predict the later cycles. For the implementation details, we use a single-layer LSTM, MSE loss, and Adam optimizer. The training epoch is set to 50, and the batch size is set to 32 for a converged loss curve for limited early data (the loss is already converged in 30 epochs).

Model 2 (No-IMV):

Model 2 is established to examine the usefulness of probed IMVs. In Model 2, we do not change the machine learning pipeline as described in the Methods section. We only remove the IIMVs from the feature matrix when performing physics-informed machine learning, i.e., the temperature transfer experiment. Therefore, the model learns from extracted features, as well as their temperature divergence, disregarding the IIMVs of the batteries. The model is expected to underperform as compared with those that include IIMVs, given the evidenced fact that IIMVs influence battery capacity during long-term cycling. The remaining settings are identical to that of the Methods sections.

Model 3 (No-transfer):

Model 3 is established to verify the necessity of physics-informed machine learning for knowledge transfer regarding different degradation patterns under continuous temperature regions. In Model 3, the physics-informed machine learning, i.e., the Arrhenius law is not adopted thus the verification of target samples can only access the source sample information, regardless of the evidenced feature divergence under different temperatures. The model is expected to underperform compared with those that include a physics-informed machine learning strategy, since temperature adaptation is a necessity. The remaining settings are identical to that of the Methods sections.

Model 4 (Empirical formula):

Model 4 is established to verify the necessity of automatic temperature calibration of models as opposed to expert knowledge in a fixed temperature, which is typically not accessible in battery prototype verification as prior knowledge. In Model 4, as a common engineering practice to save test time, we use polynomial fitting to determine an empirical formula that is suitable for batteries at hand, for instance, guiding samples. The obtained empirical formula is a mapping from the cycle index to the capacity values at a fixed temperature. For verification, such empirical formula is calibrated using the distance matching between early data of batteries to be verified and guiding samples. The distance matching is implemented by a translation transformation, where the translation is an intercept shift determined by the averaged capacity differences of early data at different temperatures. Such an empirical formula makes it hard to characterize the temperature-induced degradation patterns. The cycle index of batteries to be verified is fed into the calibrated empirical formula to get the capacity values. The model is expected to underperform and is not robust since it heavily relies on the temperature variation of battery performance (which is not prior) and involves no chemical process evolution insights.

# APPENDIX D FINITE ELEMENT ANALYSIS

We visualized the evolution of various physical fields within the battery during the multi-step charging process as well as throughout the battery degradation cycle. The software environment is the Comsol Multiphysics 6.1 platform. The entire simulation process consists of two procedures: the first step focuses on the reproduction of the multi-step charging process, and the second step is modeling the capacity fade of the battery, informed by degradation insights provided by machine learning. In the simulation of the multi-step charging process, we first determined the cut-off voltage according to the real charging condition, and used it as the cut-off voltage for the subsequent charging and discharging process. In the simulation of the battery aging process, according to insights gained from machine learning, the dominant contribution to battery capacity loss is thermodynamic loss, while the polarization contributing to kinetic loss is primarily driven by concentration polarization. Therefore, our simulation involves modeling the formation of SEI on the anode, along with consumption of electrolyte and LLI due to SEI formation which represents the thermodynamic loss, concurrently contributing to increased battery impedance, reflecting kinetic loss. By adjusting the stoichiometric coefficient of LLI in the side reaction of SEI generation and the conductivity after SEI generation, we achieve control of the proportion of thermodynamic and kinetic loss, respectively. Simultaneously, the thickening of the SEI layer and the consumption of the electrolyte inherently affect the concentration polarization in the battery, thus aligning with the insights derived from machine learning. We established two sets of models, the three-dimensional model to qualitatively visualize the state of each physical quantity inside the battery, and the one-dimensional numerical model to quantitatively analyze the capacity loss.

The first part is about the basic charge transfer and mass transport processes in the battery.

Based on the pseudo-two-dimensional model framework for battery simulations [260, 261], the following are the principal governing equations:

In the electrolyte, the transport of ions is governed by the Nernst−Planck equation:

$$N_i = -D_{e,i}\left(\nabla c_{e,i} - \frac{z_i F c_{e,i}}{RT}\nabla \Phi\right) \tag{D.1}$$

where, $N_i$ is flux, $D_{e,i}$, $z_i$ and $c_{e,i}$ is the diffusion coefficient in the electrolyte, charge and concentration of species $i$, respectively. $F$ is the Faraday's constant, $R$ is the ideal gas constant, $T$ is the Kelvin temperature and $\Phi$ is the electrolyte potential.

The ions present within the electrolyte adhere to the principles of both mass conservation and charge conservation, which can be represented as:

$$\frac{\partial c_{e,i}}{\partial t} + \nabla \times N_i = 0 \tag{D.2}$$

$$\sum_i z_i\, c_{e,i} = 0 \tag{D.3}$$

where, $z_i$ is the valence of each species in the electrolyte.

At the interface of the electrolyte and the electrode, the electron transfer between $Li^+$ and Li atoms can be expressed by the following simplified reaction:

$$Li^+ + e^- \leftrightarrow Li \tag{D.4}$$

This reaction could be quantified by the Butler-Volmer equation:

$$i_{loc} = i_{ex}\left[\exp\left(\frac{\alpha_a F\eta}{RT}\right) - \exp\left(\frac{-\alpha_c F\eta}{RT}\right)\right] \tag{D.5}$$

where, $i_{loc}$ is the local current density, which could be used to quantify the local reaction rate. $\eta$ is overpotential, $\alpha_a$ and $\alpha_c$ are the anodic and cathodic charge transfer coefficients, respectively, and $i_{ex}$ is exchange current density.

The overpotential can be calculated as:

$$\eta = \phi_s - \phi_e - U_{eq} \tag{D.6}$$

where, $\phi_s$ and $\phi_l$ is the solid phase and liquid phase potential, respectively, $U_{eq}$ is the equilibrium potential of the reaction.

In the cathode/anode particles, Li atoms diffuse into/out the inner/outer particles due to the concentration gradient, and could be expressed by the Fick’s second law:

$$\frac{\partial c_s}{\partial t} = D_s\left(\frac{\partial^2 c_s}{\partial r^2} + \frac{2}{r}\frac{\partial c_s}{\partial r}\right) \tag{D.7}$$

where, $D_s$ is the diffusion coefficient of Li atoms in the cathode particles, r is the radius of the particle, $c_s$ is the Li atom concentration.

The open-circuit potential $U$ of NCM811 and graphite particles can be calculated according to the Nernst equation:

$$U = U_{eq} + \frac{RT}{nF} ln\left(\frac{c_{e,Li}}{c_s}\right) \tag{D.8}$$

where, $n$ is the number of electrons transferred.

The second part is about the simulation details of the aging process.

We choose to consider the growth of the SEI on the anode for simulating the battery aging process [262, 263], and in addition to the primary graphite lithium intercalation reactions occurring on the anode, we also analyze the following side reaction:

$$S + nLi^{+} + ne^{-} \rightarrow P_{SEI} \tag{D.9}$$

where, $S$ represents the solvent, $P_{SEI}$ denotes the products formed during the reaction, and $n$ is the number of lithium ions consumed. The generation of $P_{SEI}$ leads to the LLI within the battery, causing an increase in the resistance of the SEI, as well as a decrease in the electrolyte volume fraction within the graphite anode.

The kinetics of this side reaction can be expressed by:

$$i_{loc,SEI} = -(1 + \mathrm{HK})\frac{i_{loc}}{\exp\left(\frac{\alpha_a F\eta}{RT}\right) + \frac{q_{SEI} f J}{i_{loc}}} \tag{D.10}$$

where, $i_{loc}$ is the local current density as mentioned above, HK is a dimensionless number representing the graphite expansion factor, which depends on the graphite's state of charge. HK is zero during the lithiation process. $J$ is a dimensionless number representing the exchange current density for parasitic reactions. $q_{SEI}$ signifies the local cumulative charge caused by the formation of the SEI. $f$ is a lumped dimensionless parameter based on the properties of the SEI film.

The concentration $c_{SEI}$ of SEI can be used to calculate the SEI thickness as follows:

$$\frac{\partial c_{SEI}}{\partial t} = -\frac{\gamma_{SEI} i_{loc,SEI}}{nF} \tag{D.11}$$

where, $\gamma_{SEI}$ is the stoichiometric coefficient of SEI.

The $q_{SEI}$ above is directly proportional to the $c_{SEI}$:

$$q_{SEI} = -\frac{F c_{SEI}}{A_v} \tag{D.12}$$

where $A_v$ is the area of the electrode surface.

Then the thickness of SEI layer $\delta_{SEI}$ can be calculated:

$$\delta_{SEI} = -\frac{M_P c_{SEI}}{A_v \rho_P} \tag{D.13}$$

where, $M_P$ and $\rho_P$ are the molar mass and density of SEI, respectively.

Generally, it takes multiple cycles for a battery to exhibit noticeable capacity loss, hence it is often assumed that the incremental differences between each cycle during cycling are very small. In our model, each simulated charge-discharge cycle is considered to represent the average aging characteristics over a large number of actual cycles $\tau$. Moreover, assuming that all lithium captured in the SEI layer after a full charge-discharge cycle can be attributed to the anode, the accelerated capacity loss can be represented by re-writing the stoichiometry of the SEI formation reaction as follows:

$$(\tau - 1)S + n(\tau - 1)Li \rightarrow (\tau - 1)P_{SEI} \tag{D.14}$$

Combining with the equation (D.9), we can get:

$$\tau S + nLi^{+} + n(\tau - 1)Li \rightarrow \tau P_{SEI} \tag{D.15}$$

# APPENDIX E　FEATURE IMPORTANCE

The feature importance in different lifetime stages, i.e., the early, middle, and late 10% cycles of the entire lifetime under 25℃. Colorbar maps the cycle value in different stages, respectively. Zero importance is indicated with dashed lines, respectively.

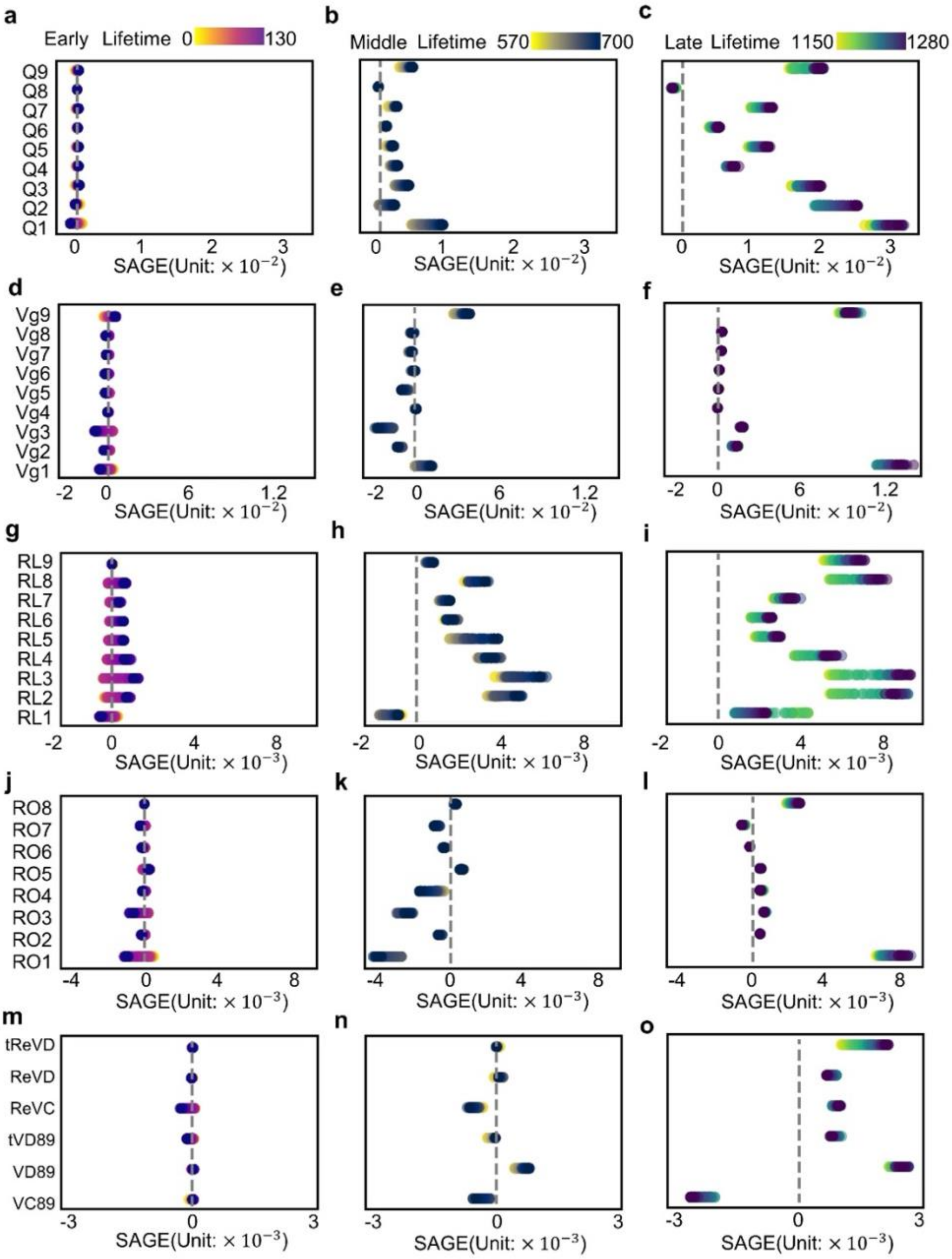

Figure E.1 Feature importance under 25℃.

The feature importance in different lifetime stages, i.e., the early, middle, and late 10% cycles of the entire lifetime under 35℃. Colorbar maps the cycle value in different stages, respectively. Zero importance is indicated with dashed lines, respectively.

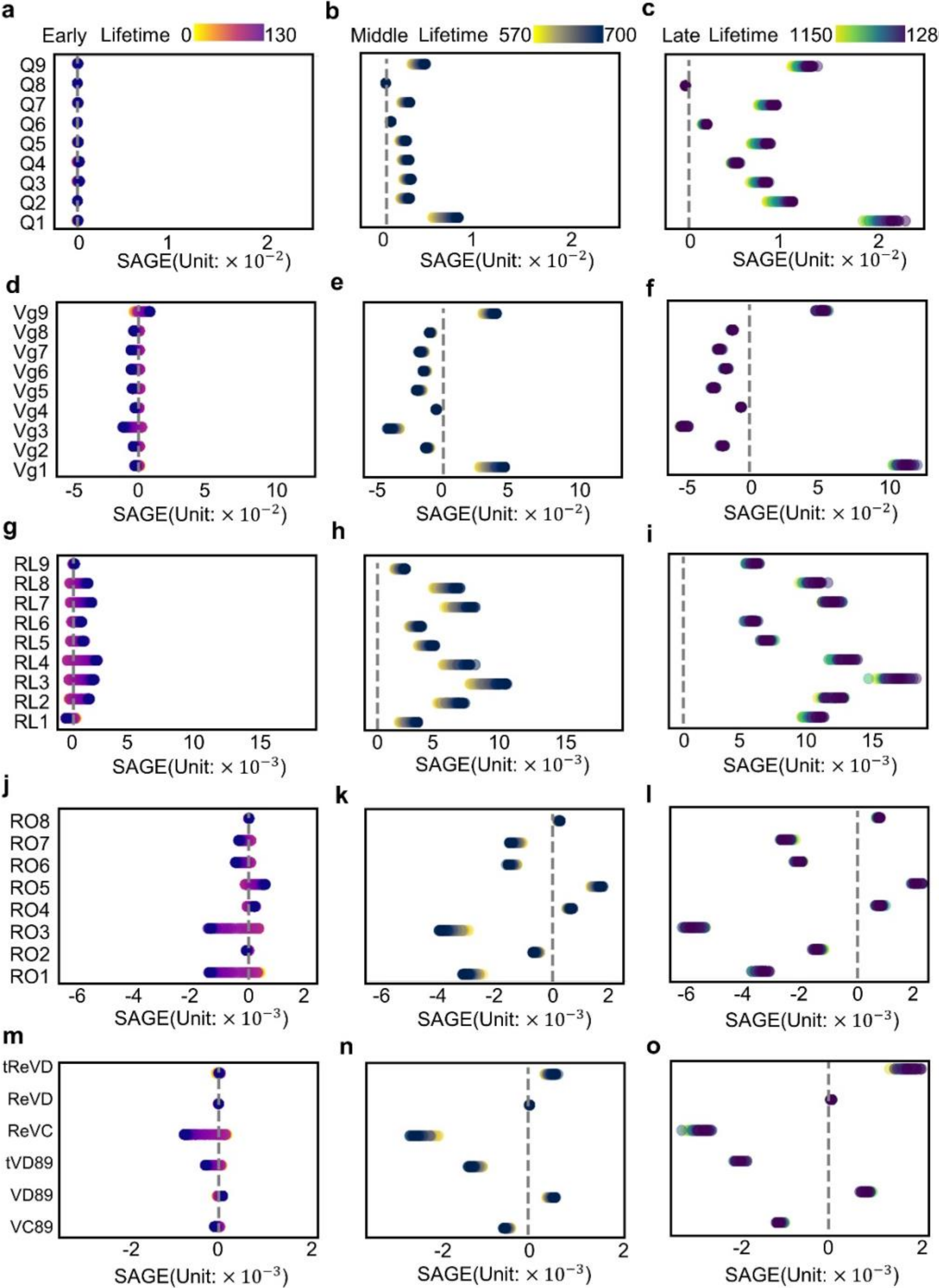

Figure E.2 Feature importance under 35℃.

The feature importance in different lifetime stages, i.e., the early, middle, and late 10% cycles of the entire lifetime under 45℃. Colorbar maps the cycle value in different stages, respectively. Zero importance is indicated with dashed lines, respectively.

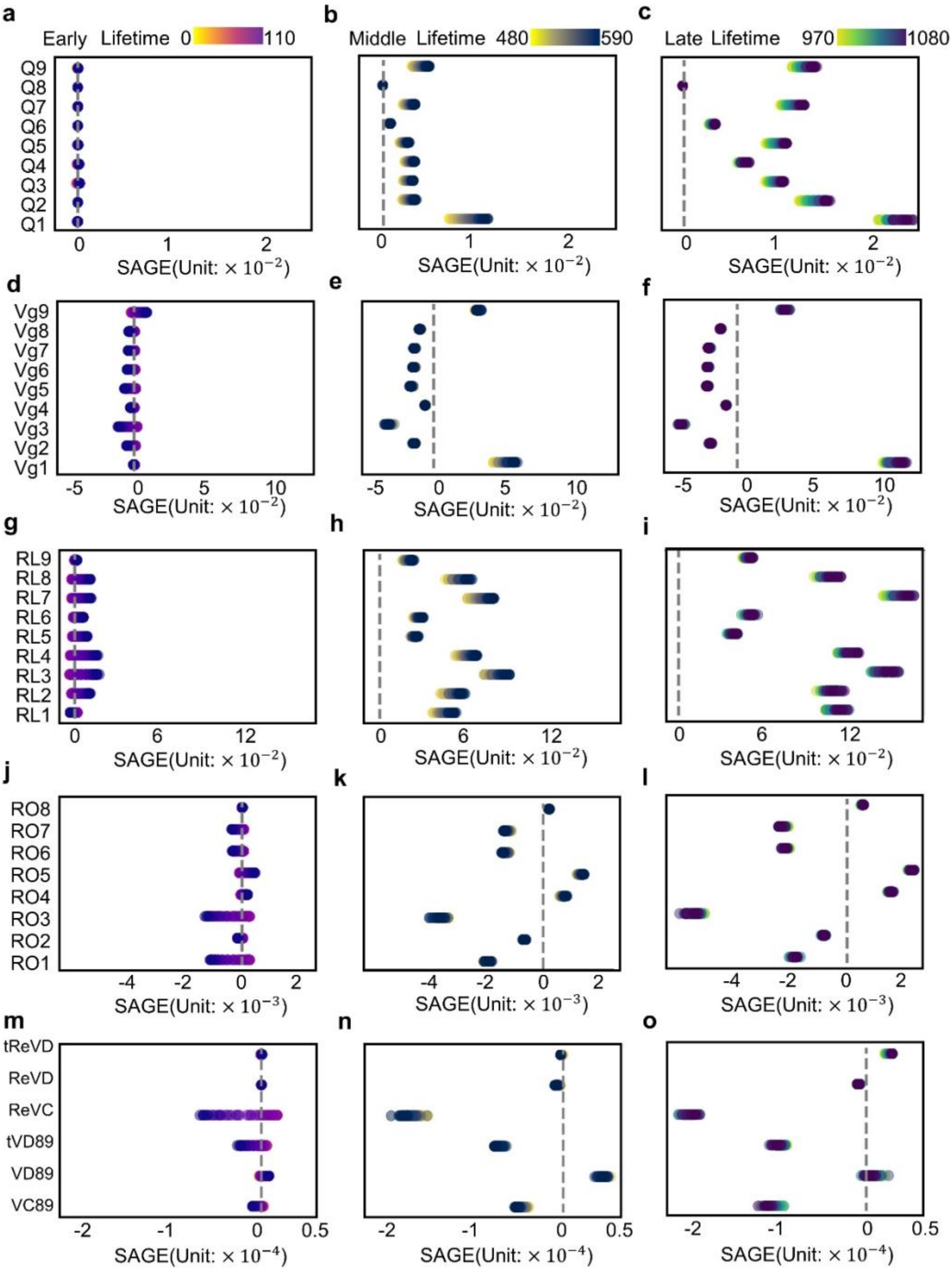

Figure E.3 Feature importance under 45℃.

The feature importance in different lifetime stages, i.e., the early, middle, and late 10% cycles of the entire lifetime under 55℃. Colorbar maps the cycle value in different stages, respectively. Zero importance is indicated with dashed lines, respectively.

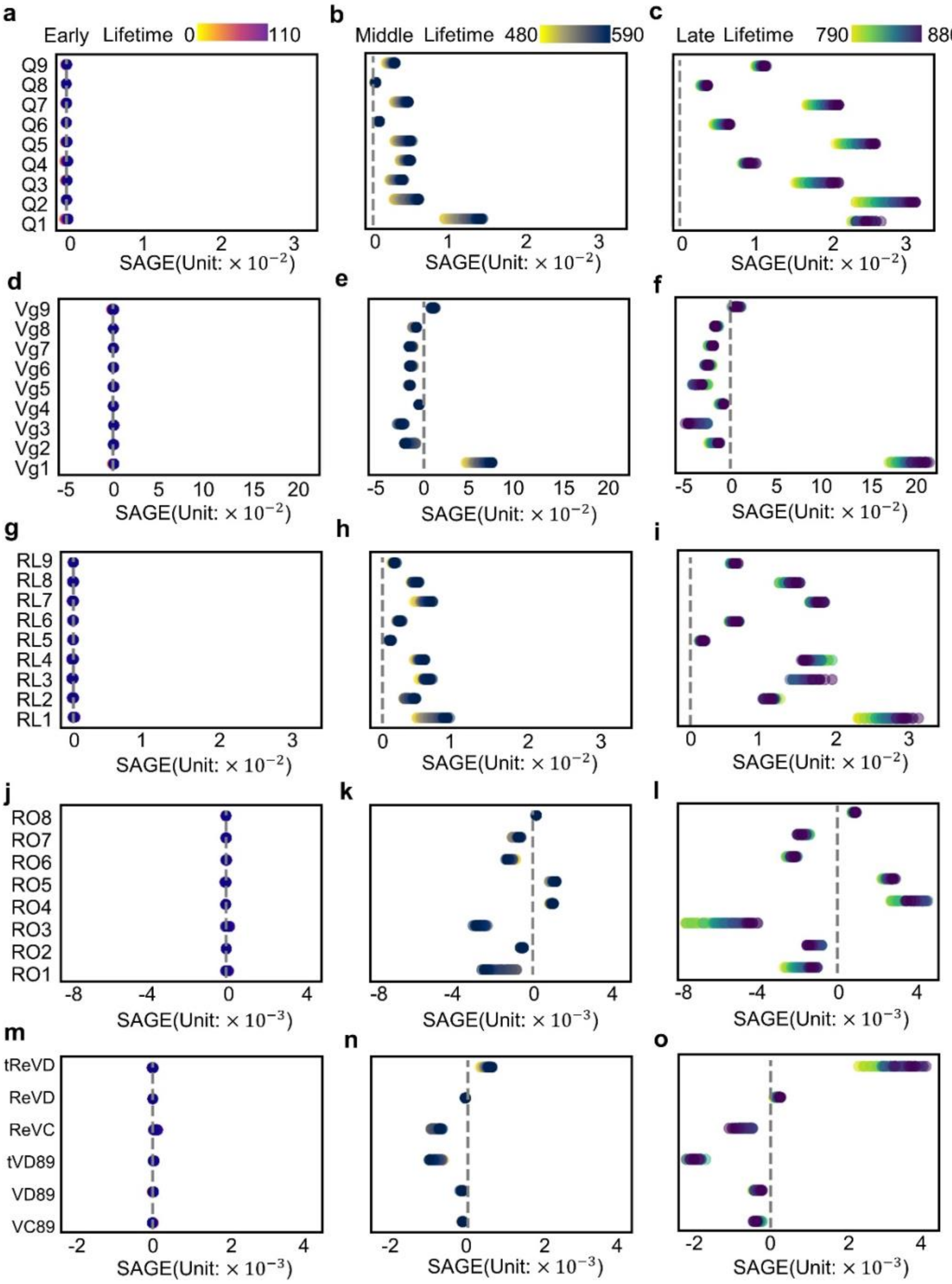

Figure E.4 Feature importance under 55℃.

# APPENDIX F SUSTAINABILITY ANALYSIS

Here we present the techno-economic assessment methodology of four battery recycling methods, i.e., refined direct, direct, hydrometallurgy, and pyrometallurgy recycling, each differentiated by their process and efficiency. The methodology begins by assessing the SOH of the prototype battery, leading to a detailed comparison of the recycling methods by decomposing the physical material treatment processes and quantifying their associated input-output relationship.

The refined direct recycling method is highlighted for its efficiency, avoiding the need to dismantle the battery structure. We note that such efficiency is enabled by our non-destructive characterization method using machine learning-inspired degradation pattern insights. This approach utilizes Lithium naphthalenide (Li-Naph) as a supplementary lithium source, directly enhancing the battery capacity without the need for preprocessing steps. The method significantly lowers costs related to materials, labor, and equipment by skipping the dismantling phase, showcasing its potential for efficient recovery of cells[183].

Contrastingly, the remaining conventional recycling methods still require extensive preprocessing, including disassembly, separation, and a blend of chemical and physical treatments. Initially, these processes involve shredding the cells into powder and removing non-recyclable battery components, yielding a 'black mass' rich in valuable materials like lithium, nickel, and cobalt. The subsequent steps differ among the methods, with some emphasizing converting the black mass into metallic salts, while others aim to rejuvenate the cathode active materials via structural repair.

The direct recycling employs lithium hydroxide as a supplementary lithium source, employing specific chemical reactions to restore the active cathode materials for both NMC811 and LFP cells, respectively:

$$\begin{aligned}&\mathrm{Li}_x(\mathrm{Ni}_{0.8}\mathrm{Co}_{0.1}\mathrm{Mn}_{0.1})\mathrm{O}_2+(1-x)\mathrm{LiOH}\\&\rightarrow\mathrm{Li}(\mathrm{Ni}_{0.8}\mathrm{Co}_{0.1}\mathrm{Mn}_{0.1})\mathrm{O}_2+\frac{1-x}{4}\mathrm{O}_2+\frac{1-x}{2}\mathrm{H}_2\mathrm{O}\end{aligned}\tag{F.1}$$

$$\begin{aligned} &\mathrm{Li}_x\mathrm{FePO}_4 + (1-x)\mathrm{LiOH} \\ &\rightarrow \mathrm{LiFePO}_4 + (1-x)/4\ \mathrm{O}_2 + (1-x)/2\ \mathrm{H}_2\mathrm{O} \end{aligned} \tag{F.2}$$

The hydrometallurgy recycling involves leaching active materials into solvents using sulfuric acid and hydrogen peroxide, extracting and precipitating critical metals like nickel, cobalt, and manganese as sulfates, and using soda ash to recover lithium carbonate:

$$\begin{aligned} &\mathrm{Li}_x(\mathrm{Ni}_{0.8}\mathrm{Co}_{0.1}\mathrm{Mn}_{0.1})\mathrm{O}_2 + (x+2)/2\ \mathrm{H}_2\mathrm{SO}_4 + (2-x)/2\ \mathrm{H}_2\mathrm{O}_2 \\ &\rightarrow \frac{x}{2}\mathrm{Li}_2\mathrm{SO}_4 + \frac{(2-x)}{2}\mathrm{O}_2 + 0.8\mathrm{NiSO}_4 + 0.1\mathrm{CoSO}_4 \\ &+0.1\mathrm{MnSO}_4 + 2\mathrm{H}_2\mathrm{O} \end{aligned} \tag{F.3}$$

$$\mathrm{Li}_x\mathrm{FePO}_4 + x/2\ \mathrm{H}_2\mathrm{SO}_4 + x/2\ \mathrm{H}_2\mathrm{O}_2 \rightarrow x/2\ \mathrm{Li}_2\mathrm{SO}_4 + \mathrm{FePO}_4 + x\mathrm{H}_2\mathrm{O} \tag{F.4}$$

$$\mathrm{Li}_2\mathrm{SO}_4 + \mathrm{Na}_2\mathrm{CO}_3 \rightarrow \mathrm{Li}_2\mathrm{CO}_3 + \mathrm{Na}_2\mathrm{SO}_4 \tag{F.5}$$

The pyrometallurgy recycling incinerates the black mass to obtain a matter of nickel and cobalt, from which lithium is eventually recovered as carbonate:

$$\mathrm{Li}_2\mathrm{SO}_4 + \mathrm{Na}_2\mathrm{CO}_3 \rightarrow \mathrm{Li}_2\mathrm{CO}_3 + \mathrm{Na}_2\mathrm{SO}_4 \tag{F.6}$$

The technology-economic assessment methodology incorporates detailed data encompassing the composition of feedstock by weight percentage, the fate and rates of components, and equipment parameters across various recycling processes. It further includes the economic and environmental aspects by detailing the prices of consumed and recovered materials, alongside cost information specific to recycling practices in China.

Here we detail the feedstock composition of different recycling methods. It is delineated by weight percentage, component fate, and recovery rates. The data is presented in the Table below. Given the assumption that cells are encased in aluminum shells, materials such as iron or steel are not considered in the composition. The pyrometallurgy and hydrometallurgy recycling methods, aim to transform black mass into lithium carbonate, nickel sulfate, cobalt sulfate, and manganese sulfate, each achieving different recovery rates. Conversely, methods like refined direct recycling and direct recycling aim to restore the active cathode materials directly. The associated data was sourced from the EverBatt model, an open-source model developed by Argonne National Laboratory, which evaluates recycling costs and environmental impacts[264].

Table F.1 The feedstock composition of different recycling methods.

| **Refined direct recycling** | | | |
| --- | --- | --- | --- |
| | LFP | NMC811 | Fate and recovery rates |
| Defective Cells | 100% | 100% | 100% recovery |
| **Preprocessing** | | | |
| | LFP | NMC811 | Fate and efficiency |
| Cathode materials | 45.7% | 42.9% | 95% to black mass |
| Graphite | 22.5% | 29.1% | 95% to black mass |
| Carbon black | 1.0% | 0.9% | 95% to black mass |
| Binder: PVDF | 1.0% | 0.9% | 5% to black mass |
| Binder: anode | 0.5% | 0.6% | 5% to black mass |
| Copper | 9.6% | 7.8% | 90% to copper powder, 5% to black mass |
| Aluminum | 5.9% | 5.2% | 90% to aluminum powder, 5% to black mass |
| Electrolyte: $LiPF_6$ | 1.9% | 1.7% | Burn for energy |
| Electrolyte: EC | 5.8% | 5.4% | Burn for energy |
| Electrolyte: DMC | 4.7% | 4.3% | Burn for energy |
| Plastic: PP | 1.1% | 0.8% | Burn for energy |
| Plastic: PE | 0.2% | 0.2% | Burn for energy |
| Plastic: PET | 0.3% | 0.3% | Burn for energy |
| **Direct recycling** | | | |
| | LFP | NMC811 | Fate and efficiency |
| Cathode materials | 65.2% | 58.2% | 90% recovery |
| Graphite | 32.2% | 39.5% | 90% recovery |
| Carbon black | 1.4% | 1.2% | 90% recovery |
| Binder: PVDF | 0.1% | 0.1% | Landfill after treatment |
| Binder: anode | 0.03% | 0.04% | Landfill after treatment |
| Copper | 0.7% | 0.6% | Discharge after treatment |
| Aluminum | 0.4% | 0.4% | Discharge after treatment |

Continued Table F.1 The feedstock composition of different recycling methods.

| **Hydrometallurgy recycling** | | | |
|---|---|---|---|
| | LFP | NMC811 | Fate and efficiency |
| Cathode materials | 65.2% | 58.2% | Li 95% to $Li_2CO_3$<br>Ni 99% to NiSO4<br>Co 99% to CoSO4<br>Mn 99% to $MnSO_4$ |
| Graphite | 32.2% | 39.5% | 90% recovery |
| Carbon black | 1.4% | 1.2% | 90% recovery |
| Binder: PVDF | 0.1% | 0.1% | Landfill after treatment |
| Binder: anode | 0.03% | 0.04% | Landfill after treatment |
| Copper | 0.7% | 0.6% | Discharge after treatment |
| Aluminum | 0.4% | 0.4% | Discharge after treatment |
| **Pyrometallurgy recycling** | | | |
| | LFP | NMC811 | Fate and efficiency |
| Cathode materials | 65.2% | 58.2% | Li 85% to $Li_2CO_3$<br>Ni 90% to NiSO4<br>Co 90% to CoSO4<br>Mn 90% to $MnSO_4$ |
| Graphite | 32.2% | 39.5% | Burn for energy |
| Carbon black | 1.4% | 1.2% | Burn for energy |
| Binder: PVDF | 0.1% | 0.1% | Burn for energy |
| Binder: anode | 0.03% | 0.04% | Burn for energy |
| Copper | 0.7% | 0.6% | Discharge after treatment |
| Aluminum | 0.4% | 0.4% | Discharge after treatment |

Here we detail the equipment parameters for the recycling methods. It is posited that the operation would handle 100,000 tonnes of rejected cells annually, operating 20 hours a day for 320 days a year. The variables such as electrical power, labor requirements, and equipment costs are subject to changes according to the volume of processed materials and the assortment of utilized machinery. Notably, the wheel loader, which runs on diesel, has its electrical power consumption listed as zero. These parameters in the Table below were also extracted from the EverBatt model [264].

Table F.2 The equipment parameters of different recycling methods.

| **Refined direct recycling** | | | | |
|---|---|---|---|---|
| Equipment | # | Electrical power (kW) | Labor requirements (person-hours/day) | Equipment Cost ($) |
| Conveyor | 1 | 29.8 | 2 | 36,067 |
| Infusion machine | 1 | 29.8 | 24 | 8,603 |
| Wheel loader | 1 | 0.0 | 20 | 183,000 |
| **Preprocessing** | | | | |
| Equipment | # | Electrical power (kW) | Labor requirements (person-hours/day) | Equipment Cost ($) |
| Hopper | 1 | 29.8 | 6 | 68,088 |
| Conveyor | 4 | 119.3 | 8 | 144,324 |
| Crusher | 1 | 149.1 | 6 | 72,061 |
| Screen, vibrating | 1 | 352.0 | 6 | 390,300 |
| Heat treatment furnace | 1 | 3897.0 | 12 | 3,363,972 |
| Cyclone | 1 | 149.1 | 6 | 1,749,237 |
| Eddy current separator | 1 | 149.1 | 0 | 411,291 |
| Air classifier | 1 | 149.1 | 6 | 326,822 |
| Gas treatment | 1 | 664.9 | 12 | 610,000 |
| Wheel loader | 1 | 0.0 | 20 | 183,000 |
| **Refined direct recycling** | | | | |
| Equipment | # | Electrical power (kW) | Labor requirements (person-hours/day) | Equipment Cost ($) |
| Conveyor | 1 | 29.8 | 2 | 36,067 |
| Infusion machine | 1 | 29.8 | 24 | 8,603 |

Continued Table F.2 The equipment parameters of different recycling methods.

| Equipment | # | Electrical power (kW) | Labor requirements (person-hours/day) | Equipment Cost ($) |
|---|---|---|---|---|
| Wheel loader | 1 | 0.0 | 20 | 183,000 |
| **Preprocessing** | | | | |
| Equipment | # | Electrical power (kW) | Labor requirements (person-hours/day) | Equipment Cost ($) |
| Hopper | 1 | 29.8 | 6 | 68,088 |
| Conveyor | 4 | 119.3 | 8 | 144,324 |
| Crusher | 1 | 149.1 | 6 | 72,061 |
| Screen, vibrating | 1 | 352.0 | 6 | 390,300 |
| Heat treatment furnace | 1 | 3897.0 | 12 | 3,363,972 |
| Cyclone | 1 | 149.1 | 6 | 1,749,237 |
| Eddy current separator | 1 | 149.1 | 0 | 411,291 |
| Air classifier | 1 | 149.1 | 6 | 326,822 |
| Gas treatment | 1 | 664.9 | 12 | 610,000 |
| Wheel loader | 1 | 0.0 | 20 | 183,000 |
| **Direct recycling** | | | | |
| Equipment | # | Electrical power (kW) | Labor requirements (person-hours/day) | Equipment Cost ($) |
| Conveyor | 1 | 29.8 | 2 | 36,067 |
| Froth flotation cell | 1 | 29.1 | 6 | 1,116,857 |
| Filter press | 2 | 19.5 | 12 | 304,372 |
| Dryer | 1 | 1975.3 | 6 | 1,737,927 |
| Ball mill | 1 | 143.4 | 6 | 274,459 |
| Heat treatment furnace | 1 | 3893.8 | 12 | 33,521,387 |

Continued Table F.2 The equipment parameters of different recycling methods.

| Equipment | # | Electrical power (kW) | Labor requirements (person-hours/day) | Equipment Cost ($) |
|---|---|---|---|---|
| Water treatment | 1 | 666.4 | 12 | 610,000 |
| Wheel loader | 1 | 0.0 | 20 | 183,000 |
| **Hydrometallurgy recycling** | | | | |
| Equipment | # | Electrical power (kW) | Labor requirements (person-hours/day) | Equipment Cost ($) |
| Conveyor | 1 | 29.8 | 2 | 36,067 |
| Leaching tank | 1 | 71.6 | 12 | 1,234,271 |
| Mixing tank | 1 | 71.6 | 6 | 1,234,271 |
| Filter press | 1 | 9.9 | 6 | 152,186 |
| Solvent extraction unit | 3 | 698.0 | 36 | 6,279,234 |
| Evaporator | 1 | 149.1 | 6 | 879,688 |
| Precipitation tank | 1 | 232.7 | 12 | 2,093,078 |
| Centrifuge | 1 | 149.1 | 6 | 469,813 |
| Dryer | 1 | 4591.1 | 6 | 2,248,876 |
| Water treatment | 1 | 666.4 | 12 | 610,000 |
| Wheel loader | 1 | 0.0 | 20 | 183,000 |
| **Pyrometallurgy recycling** | | | | |
| Equipment | # | Electrical power (kW) | Labor requirements (person-hours/day) | Equipment Cost ($) |
| Hopper | 1 | 29.8 | 6 | 68,088 |
| Conveyor | 2 | 59.7 | 4 | 72,134 |
| Smelter | 1 | 11722.2 | 24 | 28,946,046 |
| Gas treatment | 1 | 666.4 | 12 | 610,000 |
| Granulator | 1 | 20.1 | 6 | 207,097 |
| Leaching tank | 2 | 3.6 | 24 | 594,248 |

Continued Table F.2 The equipment parameters of different recycling methods.

| Equipment | # | Electrical power (kW) | Labor requirements (person-hours/day) | Equipment Cost ($) |
|---|---|---|---|---|
| Solvent extraction unit | 3 | 6.7 | 36 | 891,372 |
| Filter press | 1 | 8.2 | 6 | 152,186 |
| Precipitation tank | 1 | 8.9 | 12 | 486,111 |
| Centrifuge | 1 | 149.1 | 6 | 469,813 |
| Dryer | 1 | 96.0 | 6 | 283,778 |
| Water treatment | 1 | 666.4 | 12 | 610,000 |
| Wheel loader | 1 | 0.0 | 20 | 183,000 |

Here we detail the material prices. The analysis includes prices for both consumed and recovered materials, considering the average prices from March 20, 2023, to March 20, 2024, or the latest available data up to March 20, 2024. The pricing for chemicals is adjusted to reflect their anhydrous forms or 100% concentration levels, with specific adjustments for compounds like Li-Naph, which is derived from Naphthalene[183], and divalent metal cations prices (e.g., $Ni^{2+}$ in Ni salt/oxide) standardized to 100% concentration based on sulfate comparisons. The price for crude lithium carbonate is pegged at 50% of the battery-grade chemical price.

Table F.3 The price of consumed and recovered materials.

| **Consumed materials** | | |
|---|---|---|
| Materials | Prices ($/kg) | Source |
| Defective Cells (LFP) | 1.81 | https://data-pro.smm.cn/ |
| Defective Cells (NMC811) | 5.44 | https://data-pro.smm.cn/ |
| Hydrogen Peroxide | 0.53 | https://www.100ppi.com/ |
| Lime | 0.07 | https://www.100ppi.com/ |
| Limestone | 0.07 | https://www.100ppi.com/ |
| Lithium Carbonate | 27.39 | https://data-pro.smm.cn/ |
| Lithium Hydroxide | 47.30 | https://data-pro.smm.cn/ |

Continued Table F.3 The price of consumed and recovered materials.

| Materials | Prices ($/kg) | Source |
|---|---|---|
| Li-Naph | 1.13 | https://www.100ppi.com/ |
| Nitrogen | 0.07 | https://www.100ppi.com/ |
| Sand | 0.05 | https://www.100ppi.com/ |
| Soda Ash | 0.38 | https://www.100ppi.com/ |
| Sodium Hydroxide | 0.49 | https://www.100ppi.com/ |
| Sulfuric Acid | 0.03 | https://www.100ppi.com/ |
| **Recovered materials** | | |
| Materials | Prices ($/kg) | Source |
| Aluminum | 2.65 | https://data-pro.smm.cn/ |
| Copper | 9.65 | https://data-pro.smm.cn/ |
| $Co^{2+}$ in Co salt/oxide | 24.93 | https://data-pro.smm.cn/ |
| Graphite | 1.41 | https://data-pro.smm.cn/ |
| LFP powder | 9.75 | https://data-pro.smm.cn/ |
| Lithium Carbonate (crude) | 13.70 | https://data-pro.smm.cn/ |
| $Mn^{2+}$ in Mn salt/oxide | 2.47 | https://data-pro.smm.cn/ |
| $Ni^{2+}$ in Ni salt/oxide | 20.02 | https://data-pro.smm.cn/ |
| NMC811 powder | 31.17 | https://data-pro.smm.cn/ |
| LFP cell | 15.64 | https://data-pro.smm.cn/ |
| NMC811 cell | 32.22 | https://data-pro.smm.cn/ |

The recycling cost information pertinent to China was also gathered from EverBatt [264]. Adjustments were made to tailor the model to the geographic nuances of China, ensuring an accurate reflection of the recycling landscape within the region. The data is shown in the Table below.

Table F.4 The cost information for recycling in China.

| | Cost | Source |
|---|---|---|
| Direct labor ($/hour) | 4.60 | https://www.kanzhun.com/ |
| Electricity cost ($/kWh) | 0.10 | https://d.qianzhan.com/ |
| Natural gas cost ($/MMBTU) | 16.11 | https://data-pro.smm.cn/ |
| Water cost ($/gallon) | 0.002 | https://d.qianzhan.com/ |
| Wastewater discharge cost ($/gallon) | 13.70 | https://d.qianzhan.com/ |
| Landfill cost ($/ton) | 10.00 | https://d.qianzhan.com/ |
| Wastewater discharge ($/gallon) | 13.70 | https://data-pro.smm.cn/ |

Economic outcomes can be derived by entering the aforementioned parameters into EverBatt.

The life-cycle environmental impact and emission categories evaluated in EverBatt include total energy use, water consumption, air pollutant emissions, and greenhouse gas (GHG) emissions. The total energy use can be broken down into fossil fuel use and non-fossil fuel use, and the fossil fuel use can be further broken down into coal, natural gas, and petroleum. Air pollutant emissions modeled in EverBatt include volatile organic compound (VOC), carbon monoxide (CO), nitrogen oxides ($NO_x$), sulfur oxides (SOx), particulate matter with diameters of 10 micrometers and smaller (PM10), particulate matter with diameters of 2.5 micrometers and smaller (PM2.5), black carbon (BC), and organic carbon (OC). GHGs include carbon dioxide ($CO_2$), methane ($CH_4$), and nitrous oxide ($N_2O$). These environmental impact and emission categories are output attributes of the GREET LCA model[265].

The life-cycle environmental impacts of each process in EverBatt are calculated based on the materials and energy flows through the process, and the environmental impact intensities of each raw material and energy input obtained from the GREET model, by the following equation:

$$EI_k = \sum_i m_i \times ei_{i,k} + \sum_j q_j \times ei_{j,k} + P_k \tag{F.7}$$

Where, $EI_k$ denotes the life-cycle environmental impact/emission category $k$ for the process (for clarity's sake, let's assume the environmental impact/emission category $k$ is GHG emissions hereinafter, but it could be any of the environmental impact/emission categories listed above); $m_i$ denotes the mass (in kg) of material $i$ consumed in the process; $ei_{i,k}$ denotes the GHG emissions for 1kg of material $i$ in GREET; $q_j$ denotes the quantity (in MJ) of energy type $j$ consumed in the process; $ei_{j,k}$ denotes the GHG emissions for 1 MJ of energy type $j$ in GREET; and $P_k$ denotes GHG emissions from the process as a result of combustion or thermal decomposition of the raw materials (e.g., combustion of graphite in the pyrometallurgical recycling process, thermal decomposition of $Li_2CO_3$ in the NMC cathode powder production process).

# APPENDIX G PULSE TEST WITHOUT AGEING

The physical view of batteries for the test. (a) retired LFP batteries (prismatic 35Ah). (b) retired LMO batteries (pouch 10Ah). (c) retired NMC batteries (pouch 21Ah). (d) The illustration of the test site. (e) accelerated aging NMC batteries (cylindric 2.1Ah).

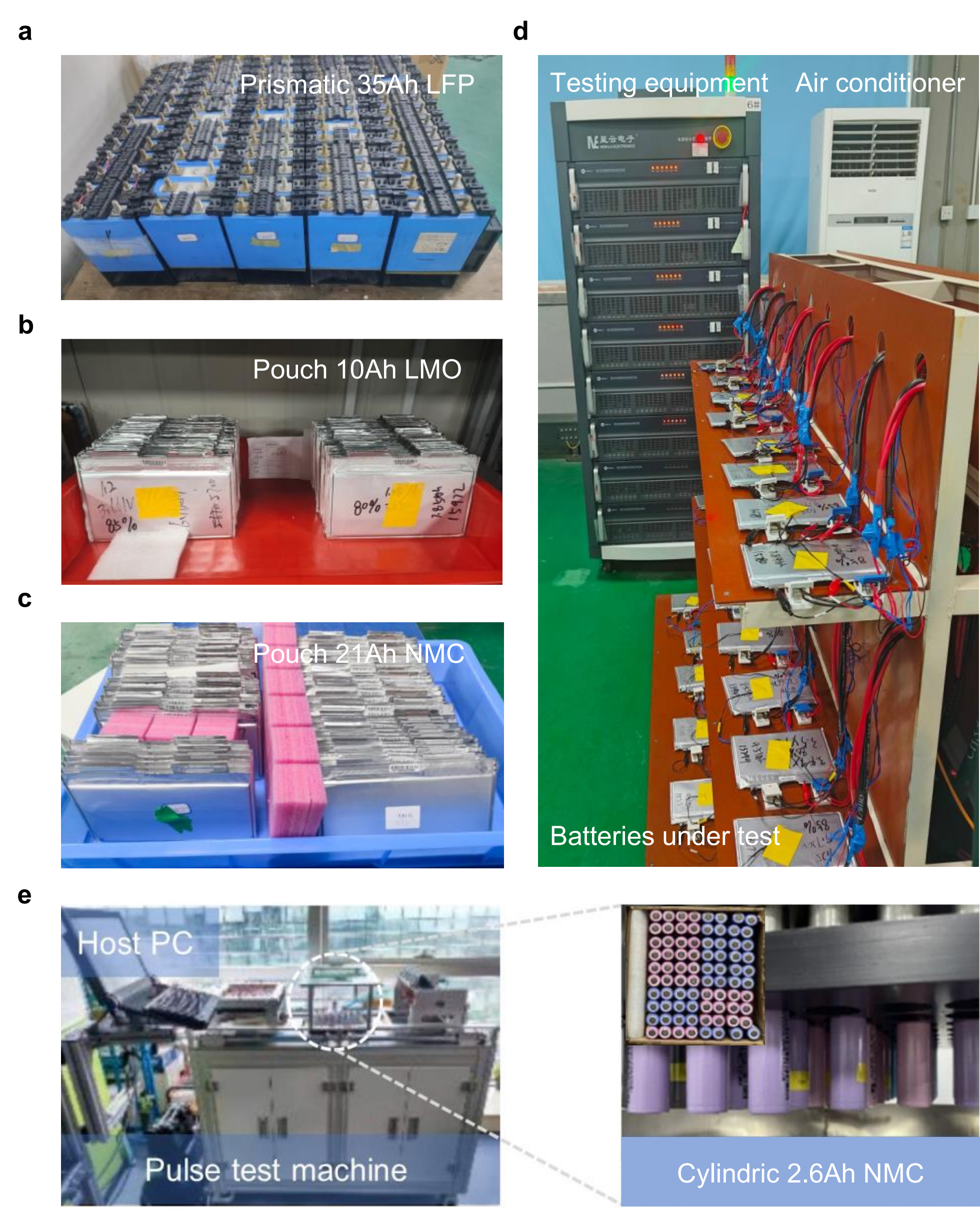

Figure G.1 The experimental setup.

# APPENDIX H　RECONSTRUCTION ERROR

The pulse data (from U1 to U21) reconstruction error of retired 21Ah NMC, 10Ah LMO and 10Ah LFP (from the top to bottom) batteries under selected state of charge (SOC) regions, respectively.

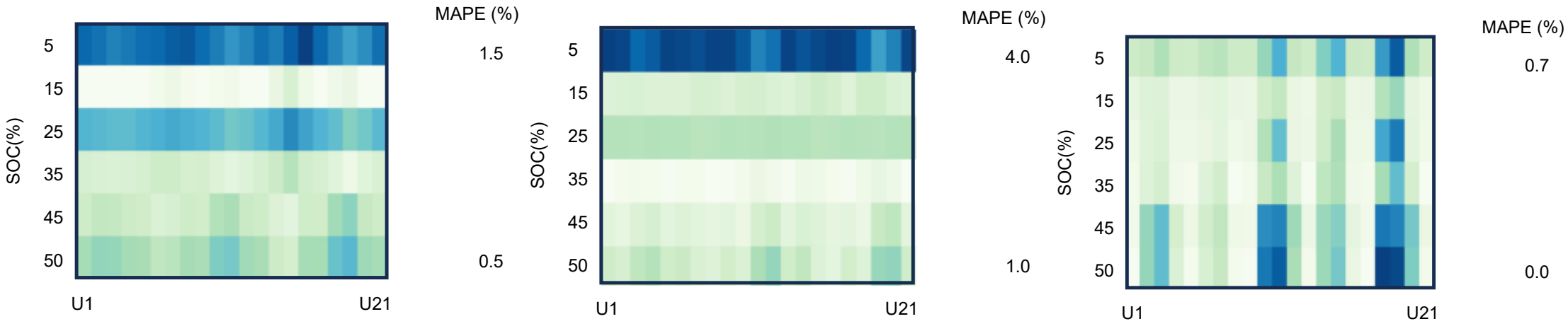

Figure H.1 The reconstruction error of different batteries.

# APPENDIX I KULLBACK-LEIBLER DIVERGENCES

Kullback-Leibler (KL) divergences evaluation of the data generation models for each feature dimension under case 0 to case 7, displayed in panels a to h. The calculation is measuring the KL divergence of real and generated pulse voltage response vectors for each feature at each testing SOC levels. The results are for retired NMC 2.1Ah batteries.

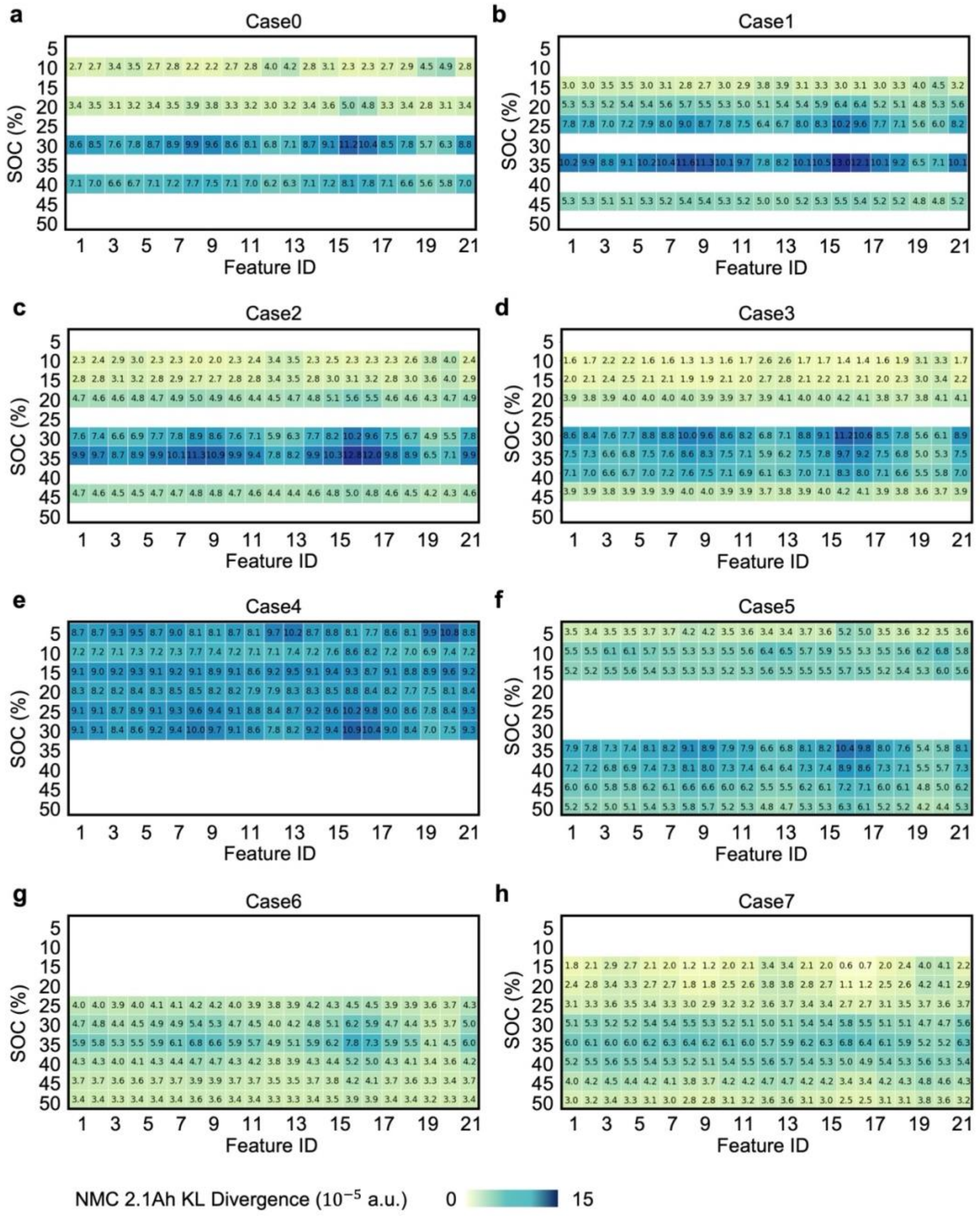

Figure I.1 The data generation performance of NMC 2.1Ah batteries.

Kullback-Leibler (KL) divergences evaluation of the data generation models for each feature dimension under case 0 to case 7, displayed in panels a to h. The calculation is measuring the KL divergence of real and generated pulse voltage response vectors for each feature at each testing SOC levels. The results are for retired LFP 35Ah batteries.

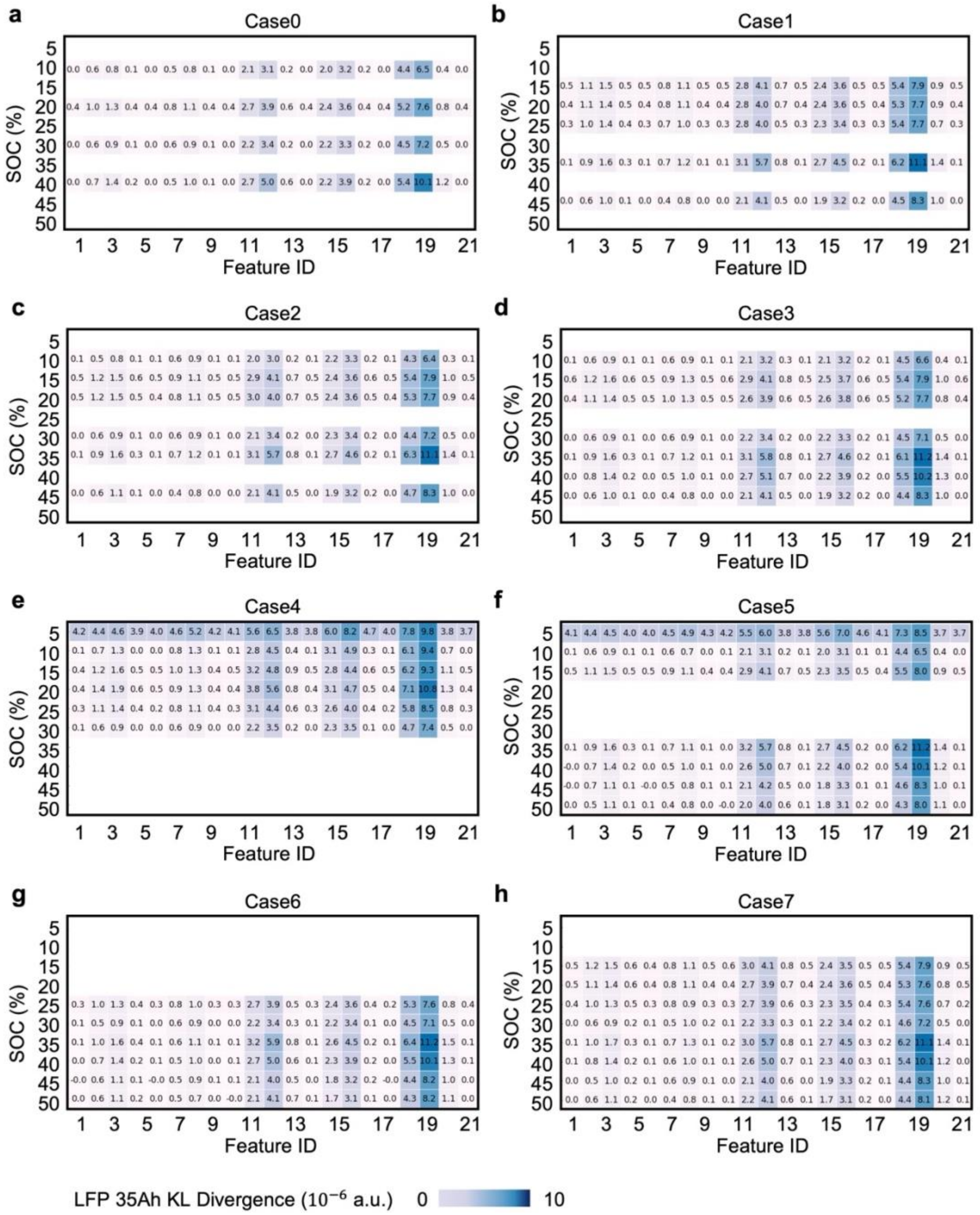

Figure I.2 The data generation performance of LFP 35Ah batteries.

Kullback-Leibler (KL) divergences evaluation of the data generation models for each feature dimension under case 0 to case 7, displayed in panels a to h. The calculation is measuring the KL divergence of real and generated pulse voltage response vectors for each feature at each testing SOC levels. The results are for retired LMO 10Ah batteries.

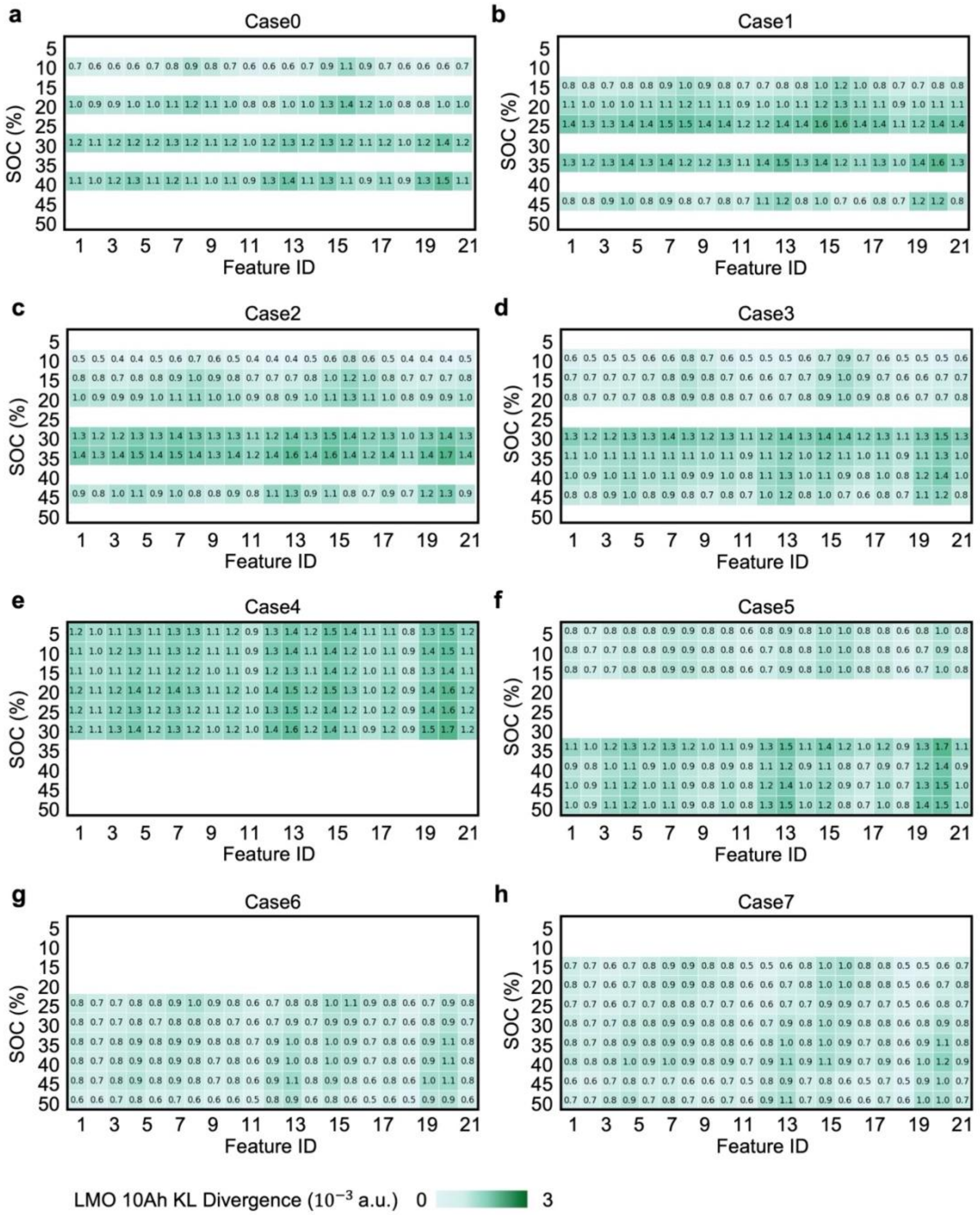

Figure I.3 The data generation performance of LMO 10Ah batteries.

Kullback-Leibler (KL) divergences evaluation of the data generation models for each feature dimension under case 0 to case 7, displayed in panels a to h. The calculation is measuring the KL divergence of real and generated pulse voltage response vectors for each feature at each testing SOC levels. The results are for retired NMC 21Ah batteries.

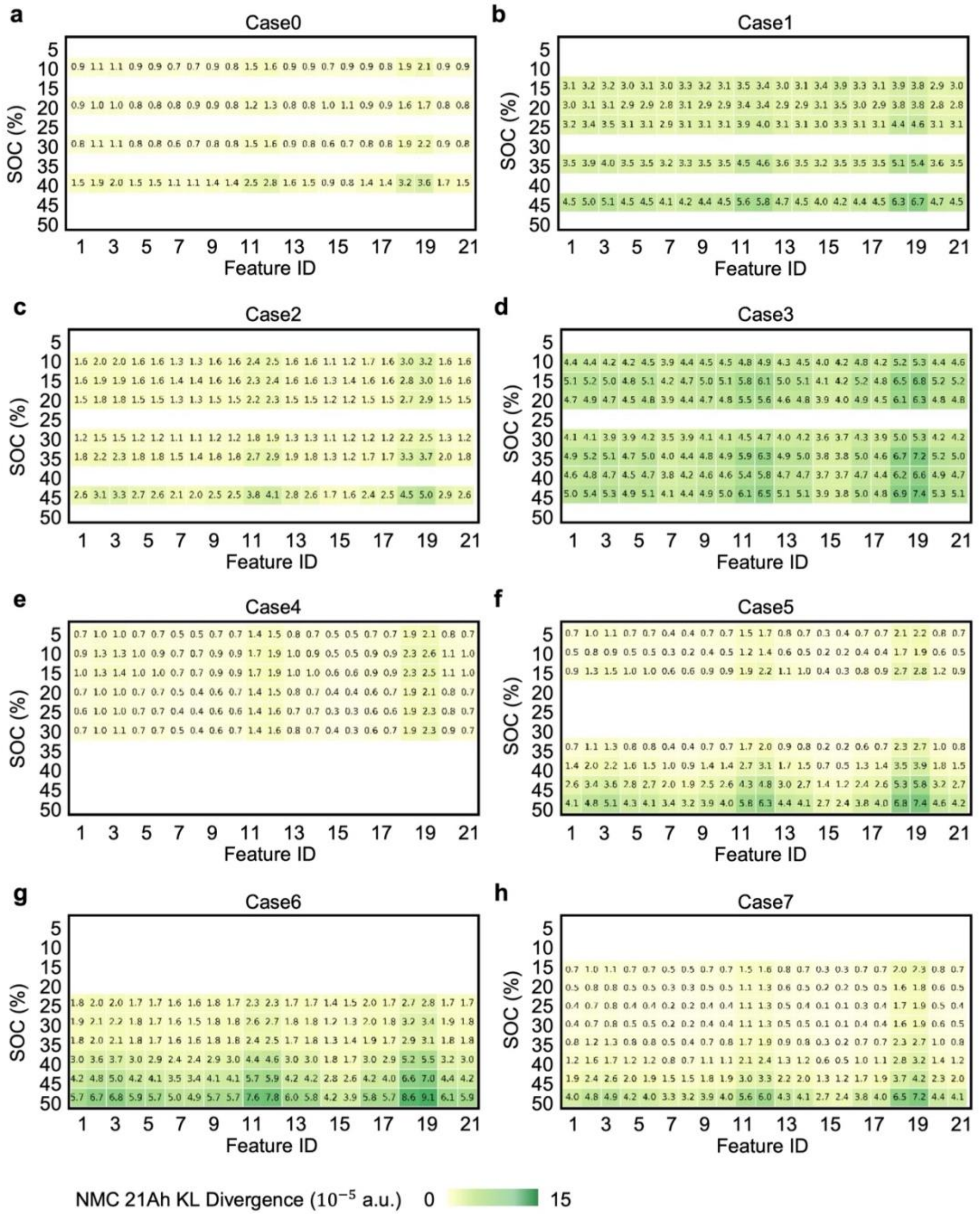

Figure I.4 The data generation performance of NMC 21Ah batteries.

# APPENDIX J MODEL TRAINGING ITERATION

The training losses, i.e., summation of reconstruction loss and KL divergence loss for different case settings and cathode material types. (a) NMC 2.1 Ah, (b) NMC 21Ah, (c) LFP 35 Ah, and (d) LMO 10 Ah.

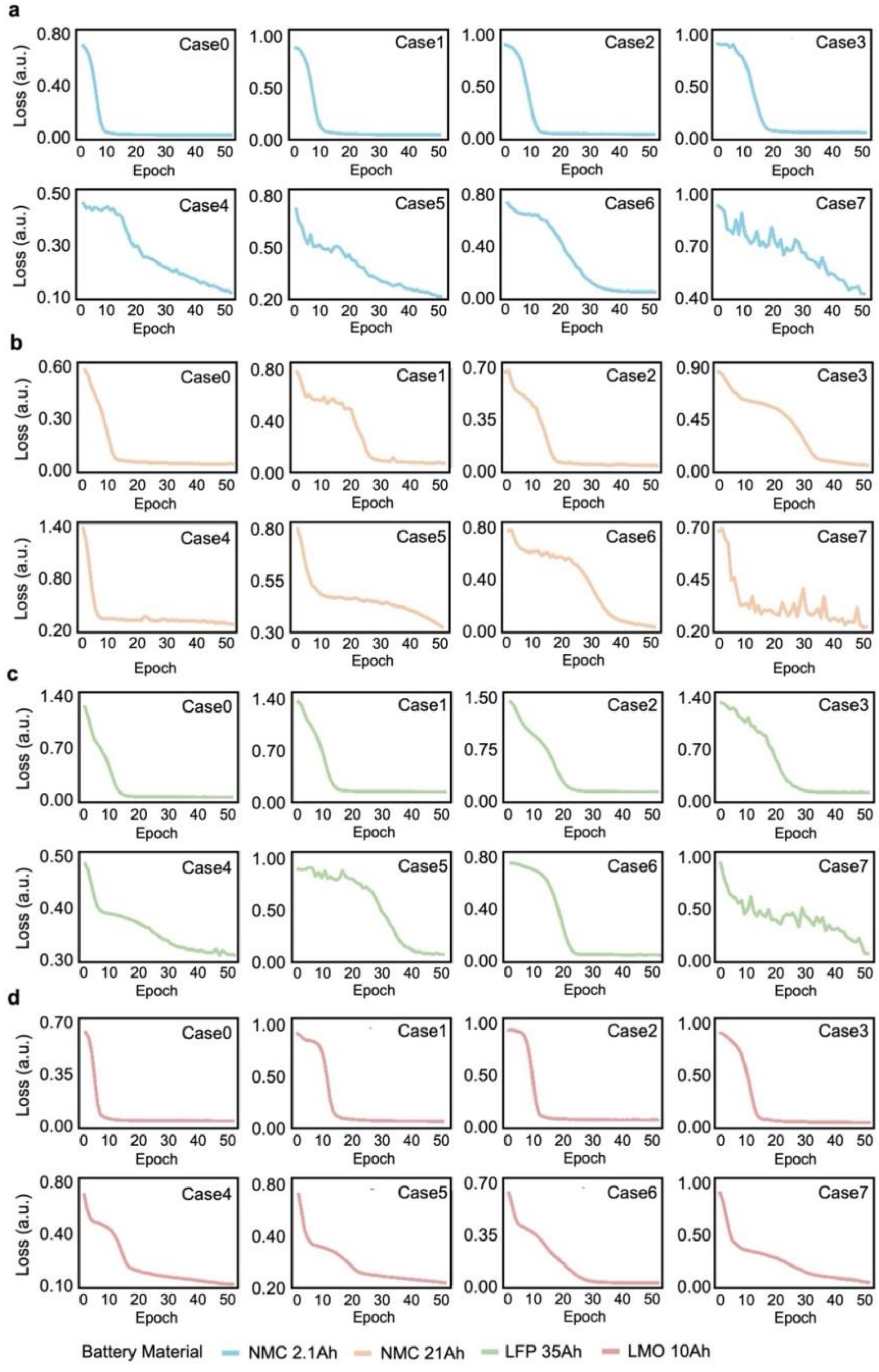

Figure J.1 The model training loss over iterations.

# APPENDIX K  CARBON AND COST REDUCTION

Economic analysis of CCCV, pulse, and generation pretreatment methods of NMC (top) and LFP (down) battery. (a) Economic performance assessment considering both pretreatment and recycling processes using different pretreatment methods in pyro-, hydro-, and direct recycling. (b) Comparative analysis of electricity carbon emission again battery retirement scale using direct recycling at 80% SOH, with pulse electricity in post-training phase excluded.

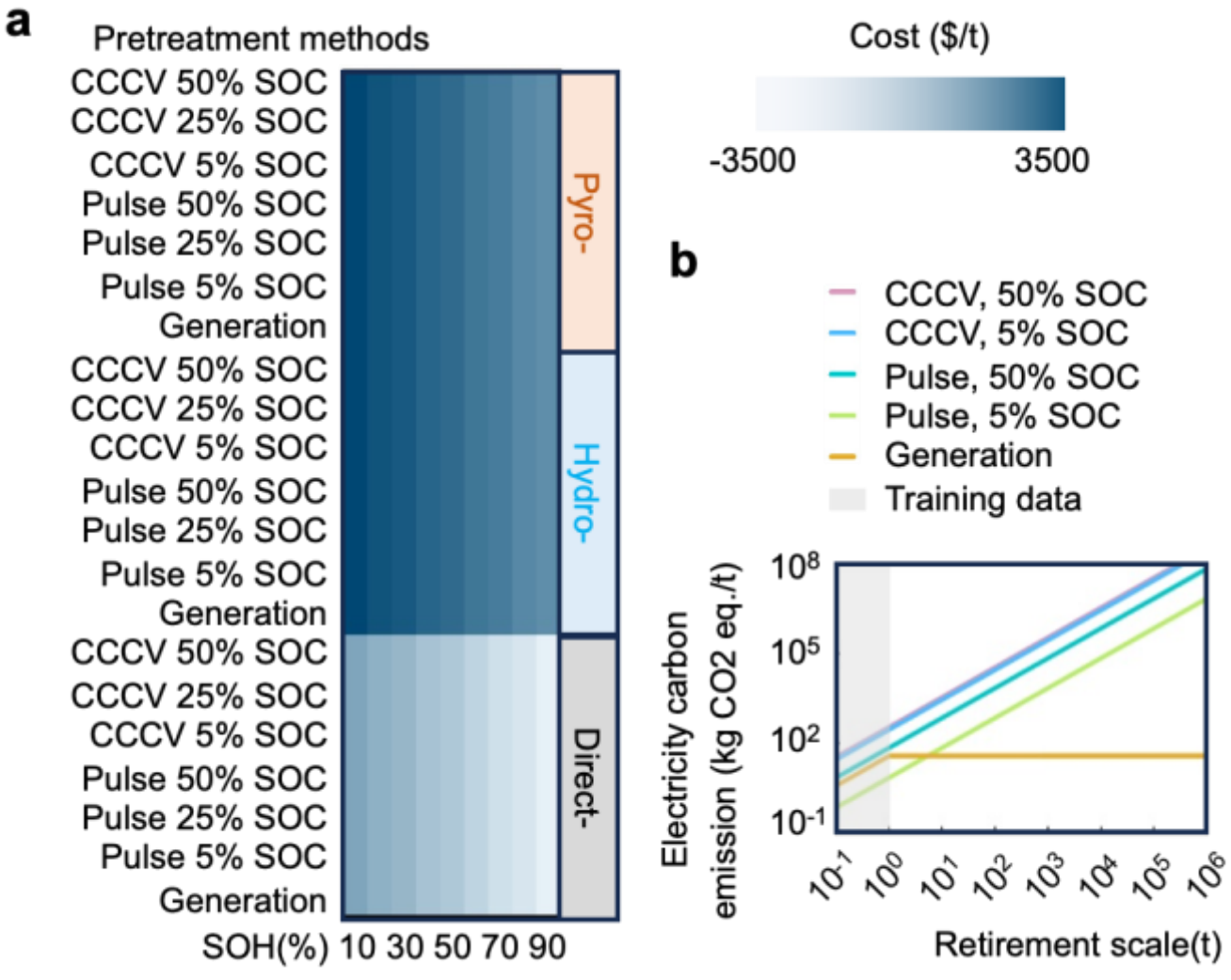

Figure K.1 The carbon and cost reduction of NMC batteries.

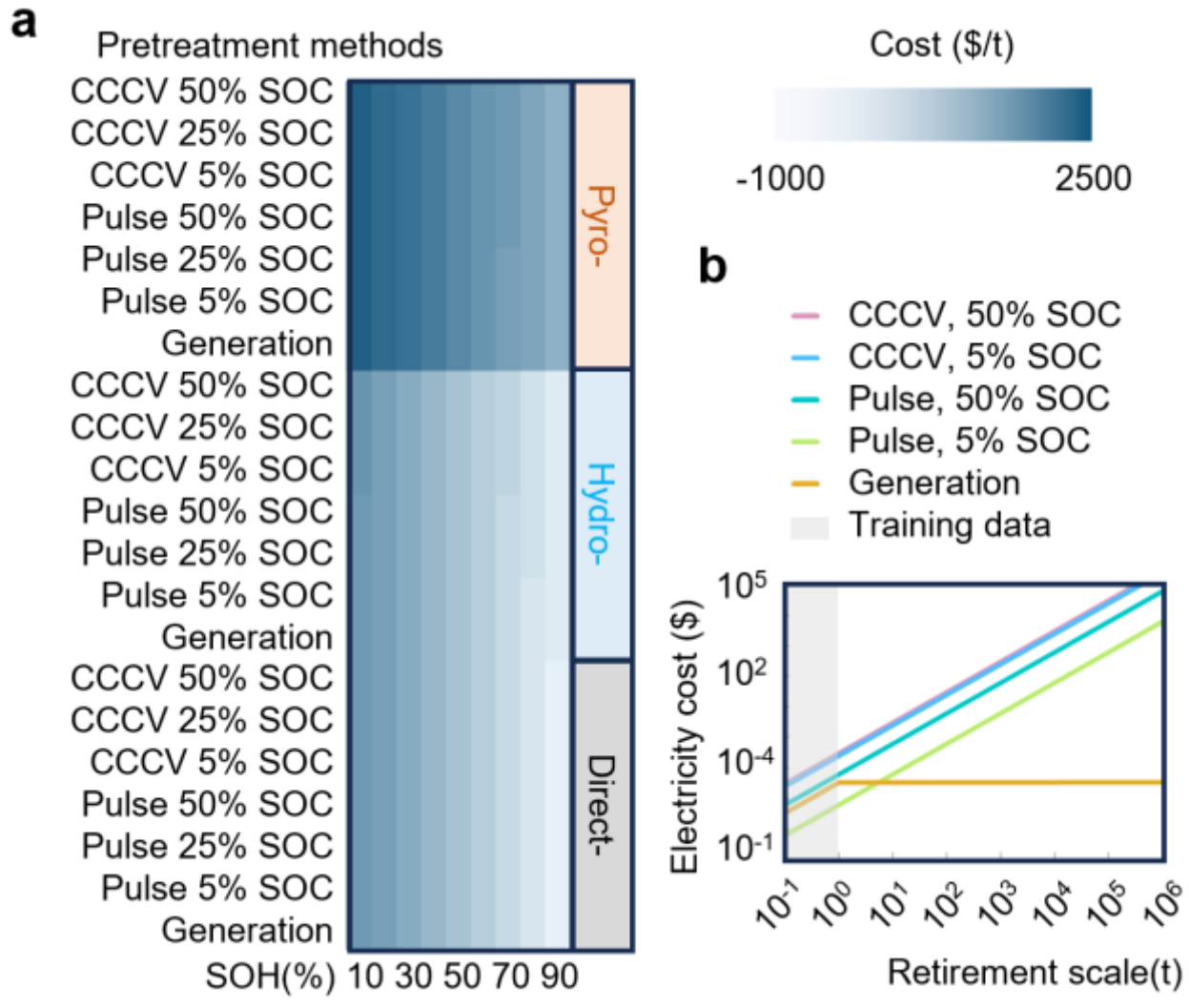

Figure K.2 The carbon and cost reduction of LFP batteries.

# APPENDIX L DATA AVAILABILITY ANALYSIS

Down-sampling of the generated data, with the proportion being 10% to 100% and testing sample size being indicated for each case. The Down-sampling is uniform across all SOC levels. The downstream SOH estimation performance is presented for each case, displayed from panel a to h. The blank SOC region is where the physical measurements are performed. The data retrieved from these SOC regions are used for training the data generation model. The results are for retired NMC 2.1Ah batteries.

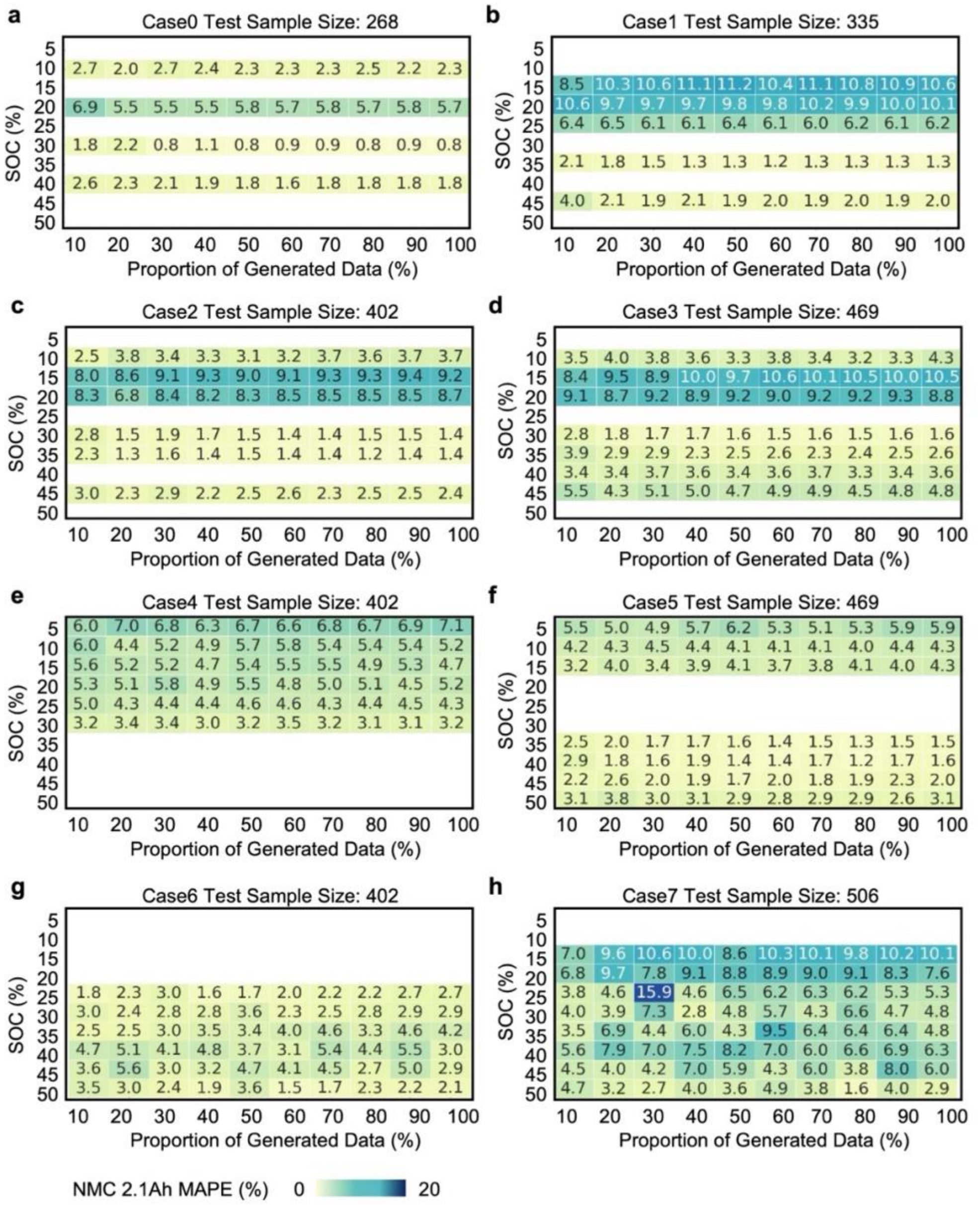

Figure L.1 The data availability analysis of NMC 2.1Ah batteries.

Down-sampling of the generated data, with the proportion being 10% to 100% and testing sample size being indicated for each case. The Down-sampling is uniform across all SOC levels. The downstream SOH estimation performance is presented for each case, displayed from panel a to h. The blank SOC region is where the physical measurements are performed. The data retrieved from these SOC regions are used for training the data generation model. The results are for retired LFP 35Ah batteries.

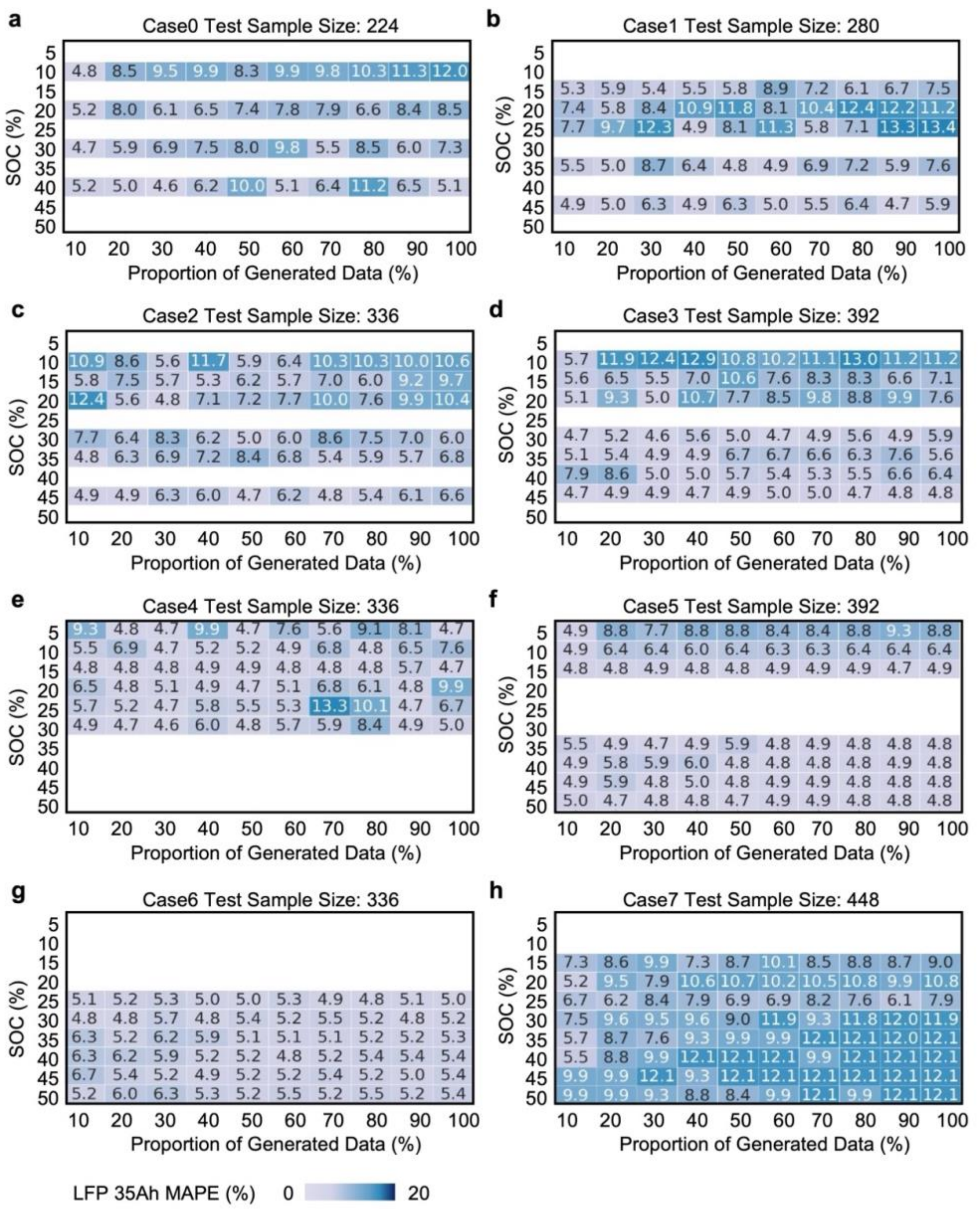

Figure L.2 The data availability analysis of LFP 35Ah batteries.

Down-sampling of the generated data, with the proportion being 10% to 100% and testing sample size being indicated for each case. The Down-sampling is uniform across all SOC levels. The downstream SOH estimation performance is presented for each case, displayed from panel a to h. The blank SOC region is where the physical measurements are performed. The data retrieved from these SOC regions are used for training the data generation model. The results are for retired LMO 10Ah batteries.

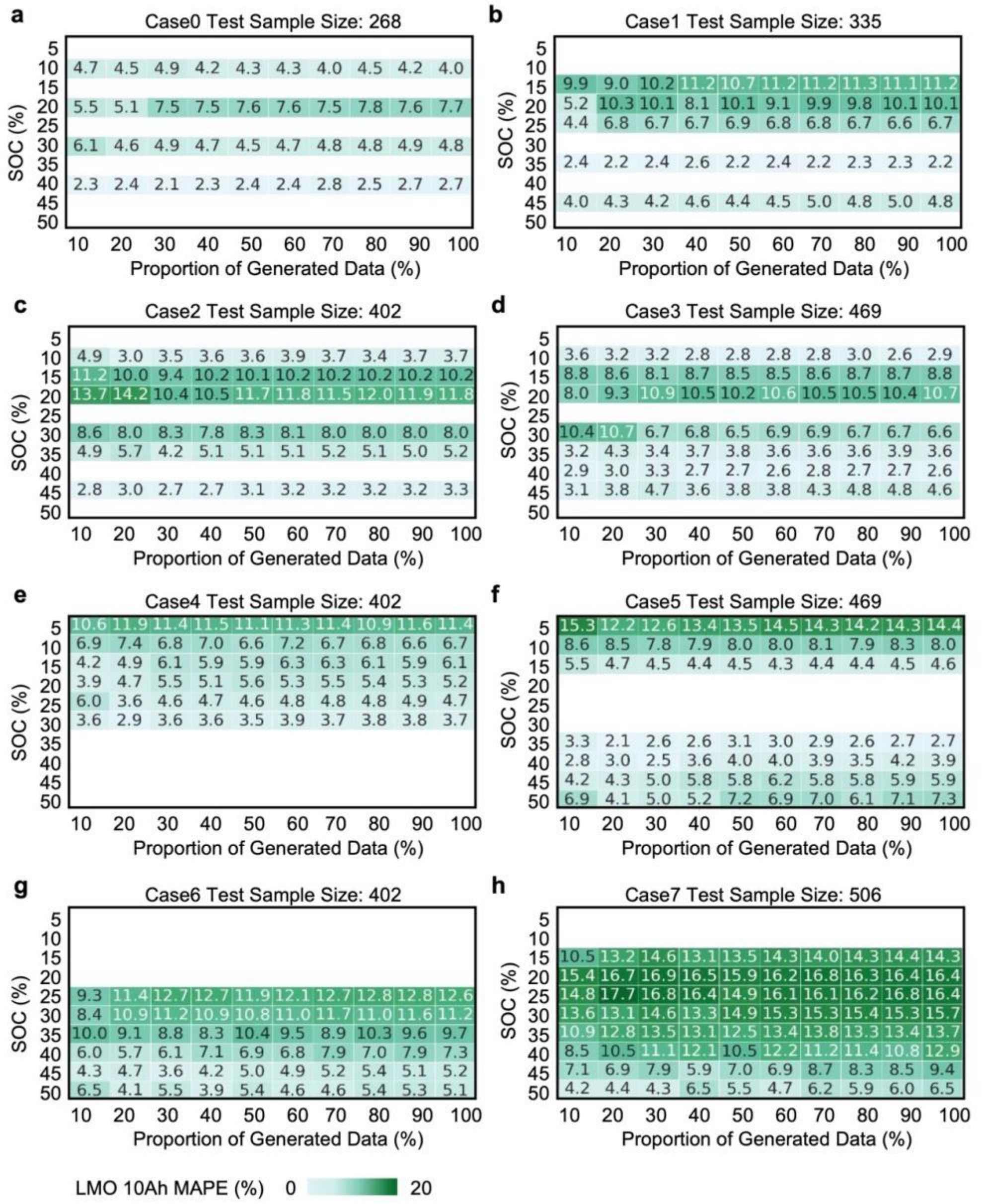

Figure L.3 The data availability analysis of LMO 10Ah batteries.

Down-sampling of the generated data, with the proportion being 10% to 100% and testing sample size being indicated for each case. The Down-sampling is uniform across all SOC levels. The downstream SOH estimation performance is presented for each case, displayed from panel a to h. The blank SOC region is where the physical measurements are performed. The data retrieved from these SOC regions are used for training the data generation model. The results are for retired NMC 21Ah batteries.

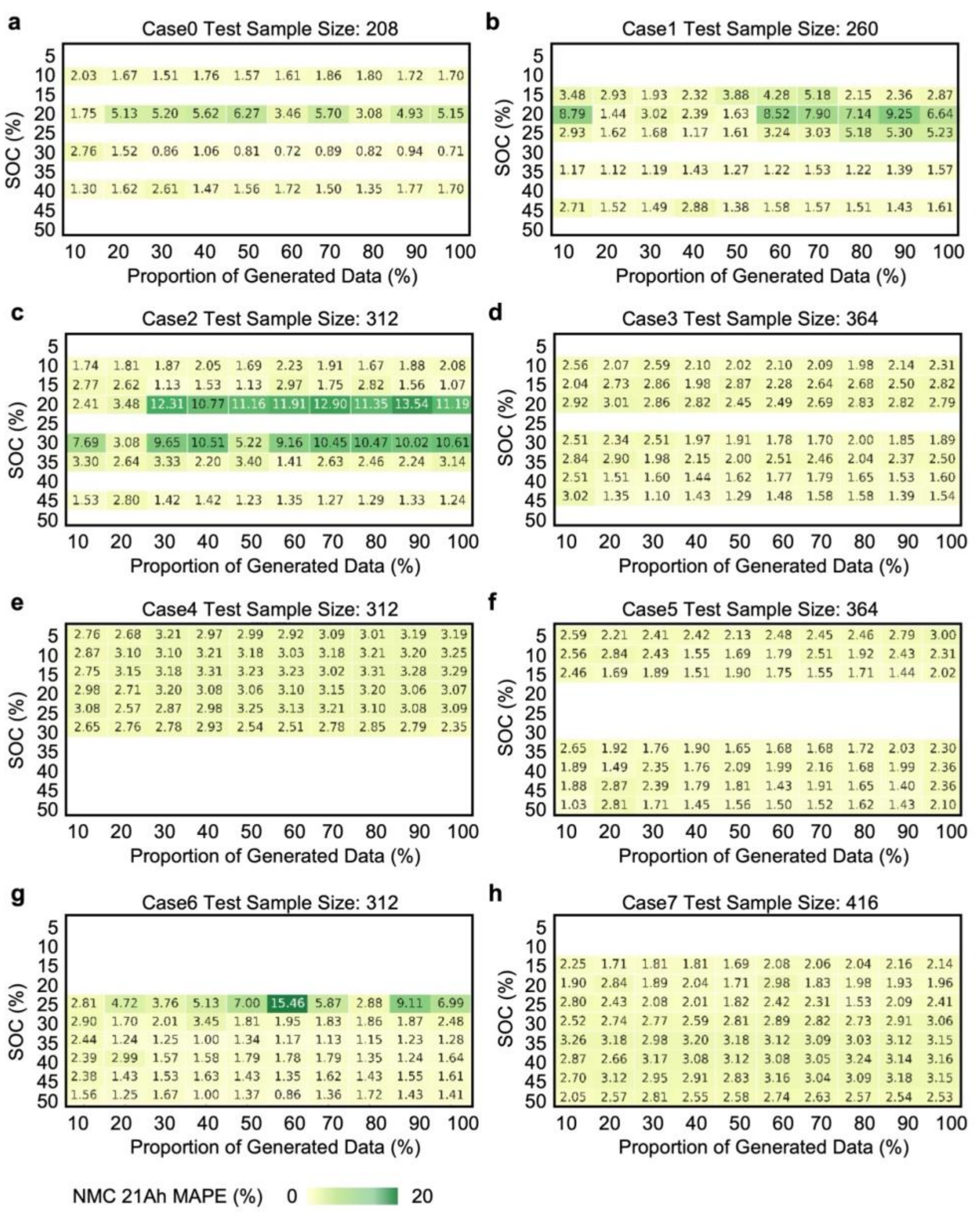

Figure L.4 The data availability analysis of NMC 21Ah batteries.

# APPENDIX M CHARGING AND DISCHARGING DATA

The voltage-capacity curve in the charging cycle for the batteries. The unit for voltage is in volts, and the unit for capacity is in Ah. Different colors in each subplot represent unique batteries. The subplot title follows a format of A: B, which indicates the batteries originate from manufacturer A with cathode material B.

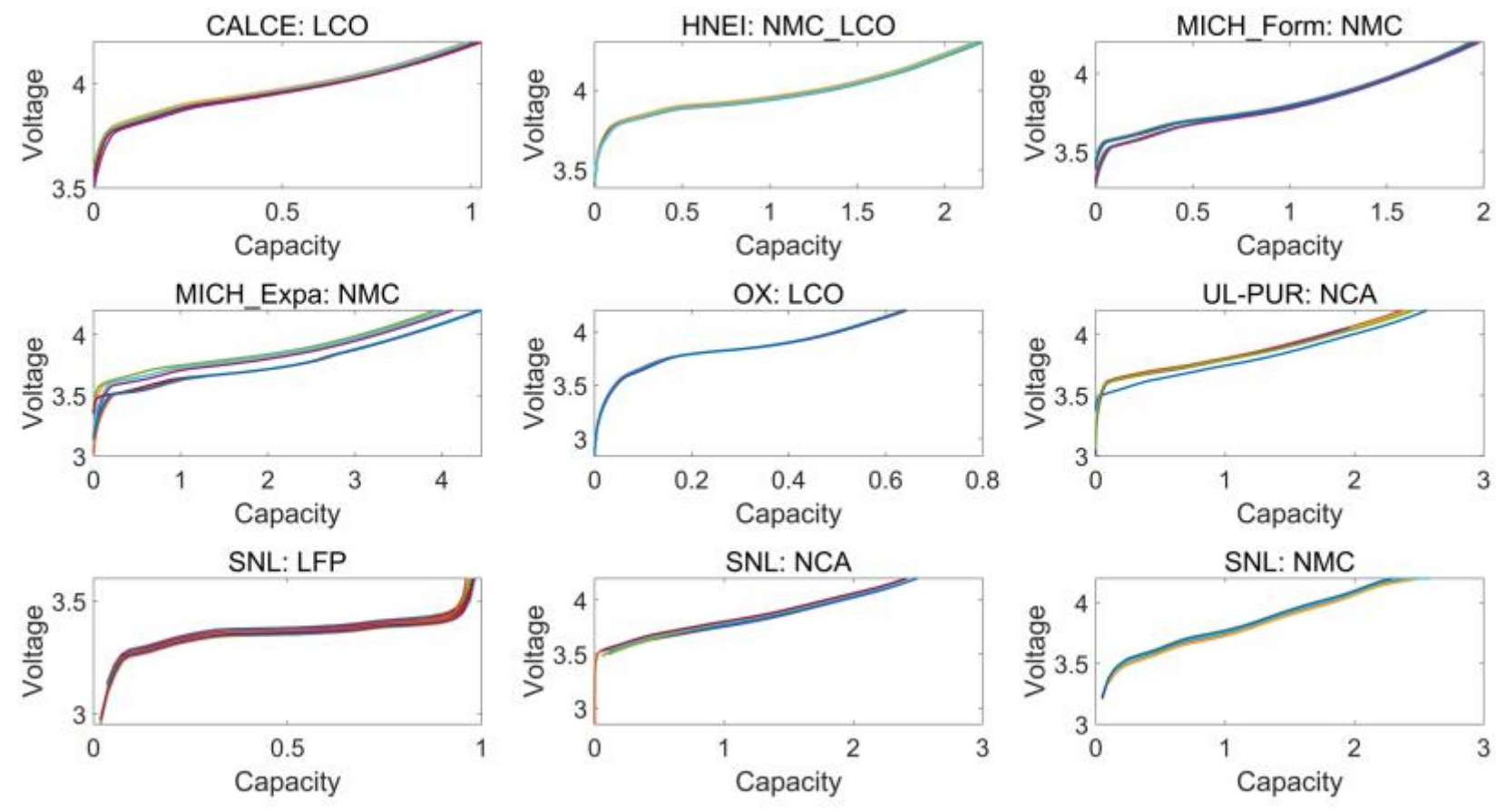

Figure M.1 The charging data.

The voltage-capacity curve in the discharging cycle for the batteries. The unit for voltage is in volts, and the unit for capacity is in Ah. Different colors in each subplot represent unique batteries. The subplot title follows a format of A: B, which indicates the batteries originate from manufacture A with cathode material B.

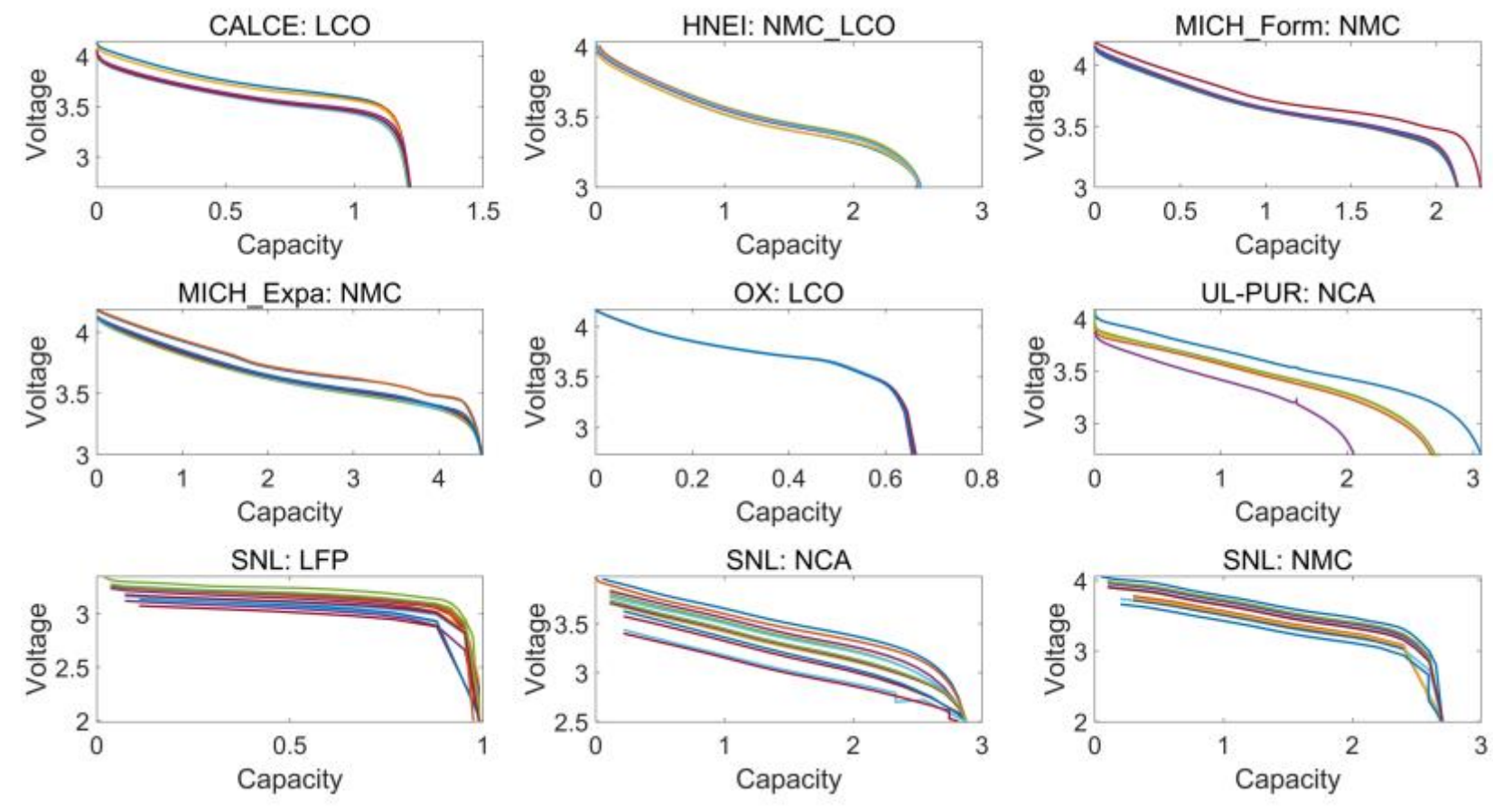

Figure M.2 The discharging data.

The dQ/dV curve in the charging cycle for the batteries. The unit for voltage is in volts, and the unit for dQ/dV is in Ah/ volts. Different colors in each subplot represent unique batteries. The subplot title follows a format of A: B, which indicates the batteries originate from manufacture A with cathode material B.

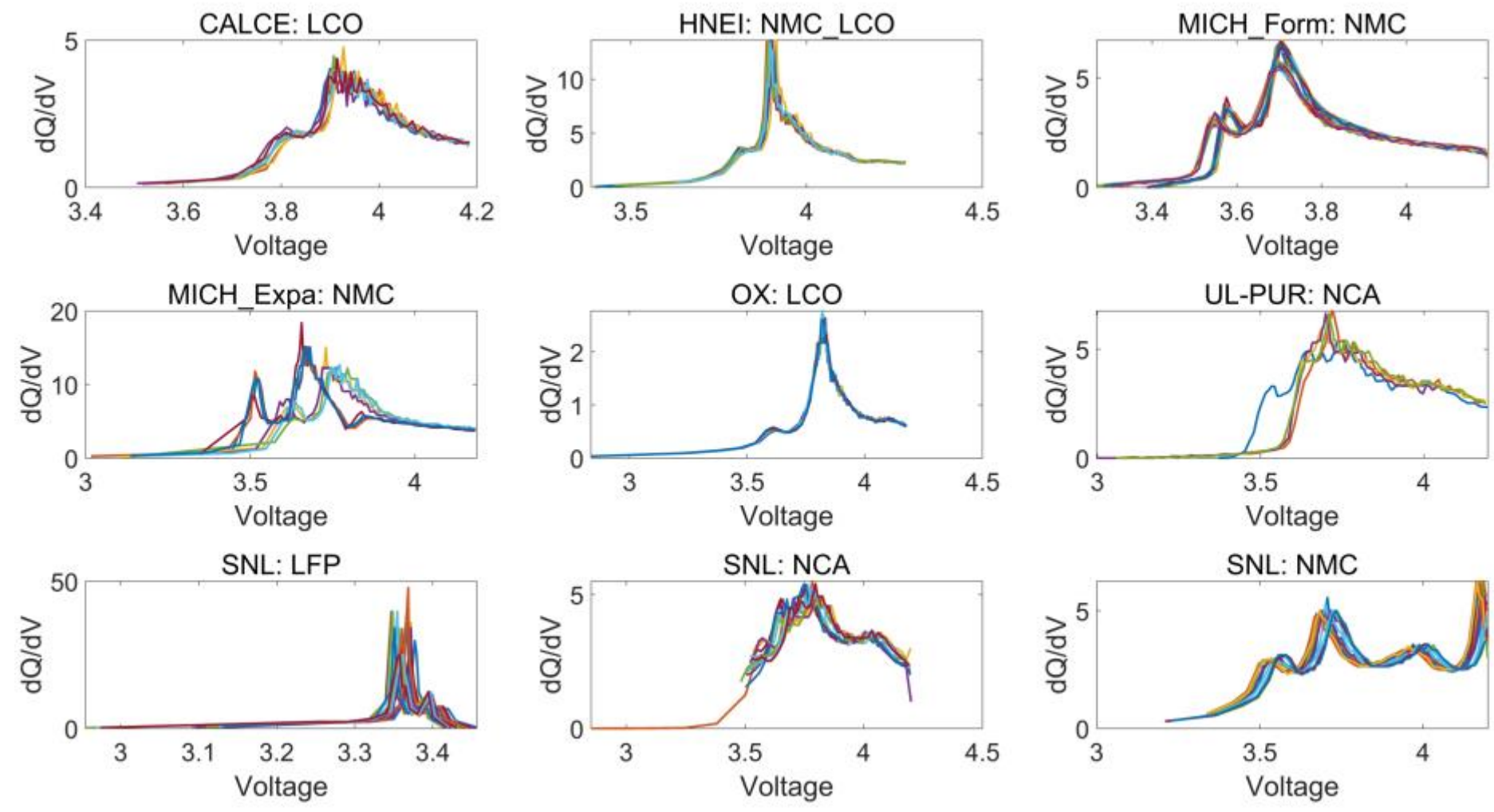

Figure M.3 The dQ/dV transformation of charging data.

The dQ/dV curve in the discharging cycle for the batteries. The unit for voltage is in volts, and the unit for dQ/dV is in Ah/ volts. Different colors in each subplot represent unique batteries. The subplot title follows a format of A: B, which indicates the batteries originate from manufacturer A with cathode material B.

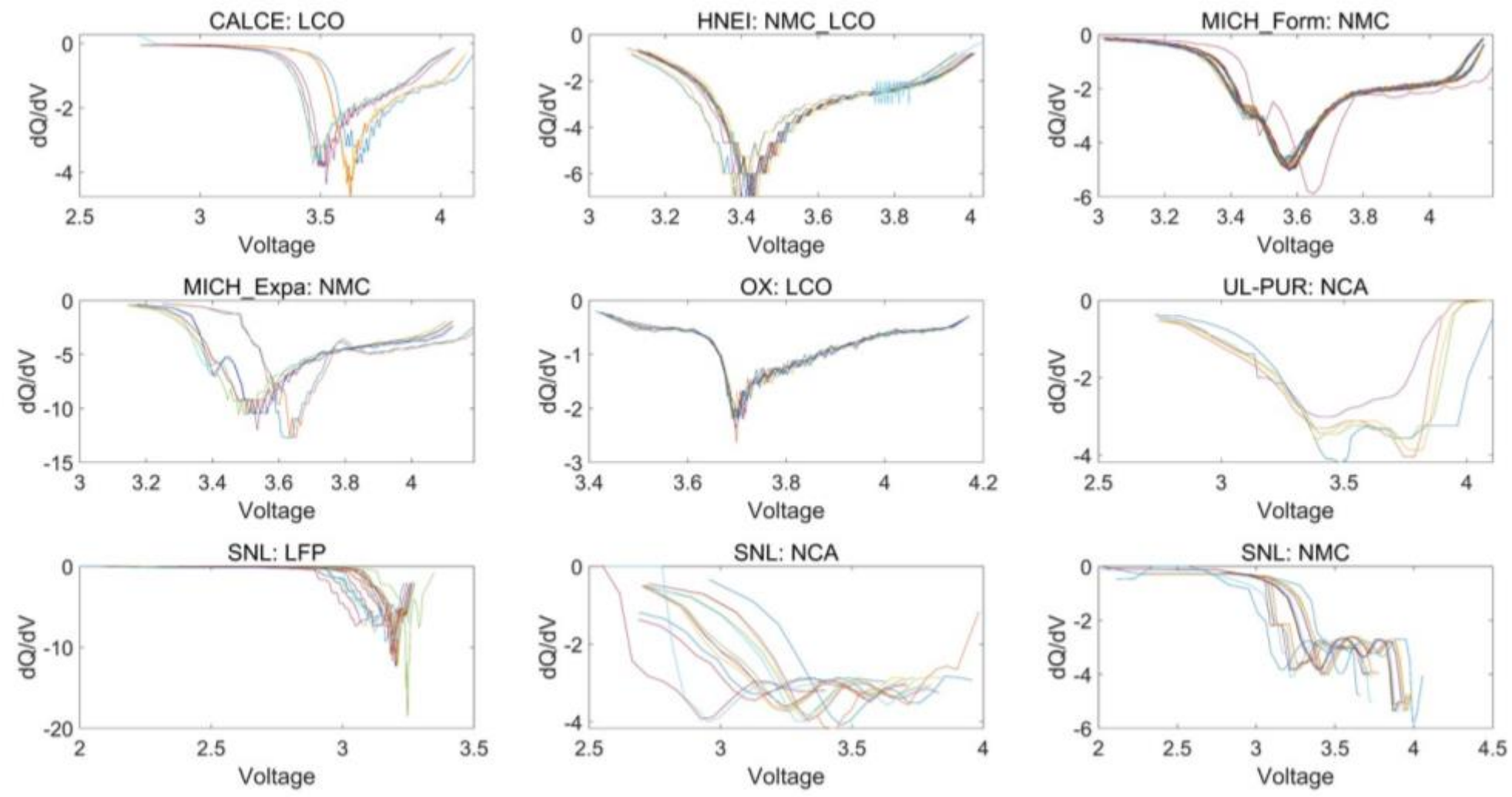

Figure M.3 The dQ/dV transformation of discharging data.

# APPENDIX N FEATURE ENGINEERING DESIGN

Table N.1 The feature engineering design.

| Feature | Definition |
|---|---|
| F1(16) | The peak intensity of the dV/dQ curve |
| F2(17) | The time when the peak intensity of the dV/dQ curve arrives |
| F3(18) | The voltage where the peak intensity of the dV/dQ curve arrives |
| F4(19) | The kurtosis statistics of the dV/dQ curve |
| F5(20) | The skewness statistics of the dV/dQ curve |
| F6(21) | Q1 quantile of the voltage |
| F7(22) | Q2 quantile of the voltage |
| F8(23) | Q3 quantile of the voltage |
| F9(24) | Q1 quantile of the capacity |
| F10(25) | Q2 quantile of the capacity |
| F11(26) | Q3 quantile of the capacity |
| F12(27) | The kurtosis statistics of the voltage |
| F13(28) | The skewness statistics of the voltage |
| F14(29) | The kurtosis statistics of the capacity |
| F15(30) | The skewness statistics of the capacity |

Note: The format of the extracted features follows Fa (Fb), which denotes that Fa and Fb are the features extracted from the charging and discharging process, respectively. The definition of the quantiles:

The Q1 quantile splits off the lowest 25% of data from the highest 75%.

The Q2 quantile cuts the data set in half.

The Q3 quantile splits off the highest 25% of data from the lowest 75%.

Kurtosis is a measure of how outlier-prone a distribution is. The kurtosis of the normal distribution equals 3. When a distribution is more outlier-prone than the normal distribution, the kurtosis is larger than 3. When a distribution is less outlier-prone than the normal distribution, the kurtosis is smaller than 3. The definition of kurtosis:

$$k = \frac{\frac{1}{n}\sum_{i=1}^{n} (x_i - \bar{x})^4}{\left(\frac{1}{n}\sum_{i=1}^{n} (x_i - \bar{x})^2\right)^2} \tag{N.1}$$

where $x_i$ is the $i$-th feature point, $\bar{x}$ is the mean value of the feature vector, and $n$ is the number of points in the feature vector.

Skewness is a measure of the asymmetry of the data around the sample mean. If skewness is negative, the data spread out more to the left of the mean than to the right. If skewness is positive, the data spread out more to the right. The definition of skewness:

$$s = \frac{\frac{1}{n}\sum_{i=1}^{n} (x_i - \bar{x})^3}{\left(\sqrt{\frac{1}{n}\sum_{i=1}^{n} (x_i - \bar{x})^2}\right)^3} \tag{N.2}$$

where $x_i$ is the $i$-th feature point, $\bar{x}$ is the mean value of the feature vector, and $n$ is the number of points in the feature vector.

# APPENDIX O NOISE INJECTION

The classification accuracy of the MV and WDV methods at selected random noise levels (NSR = 2%, 6%, and 10%), respectively.

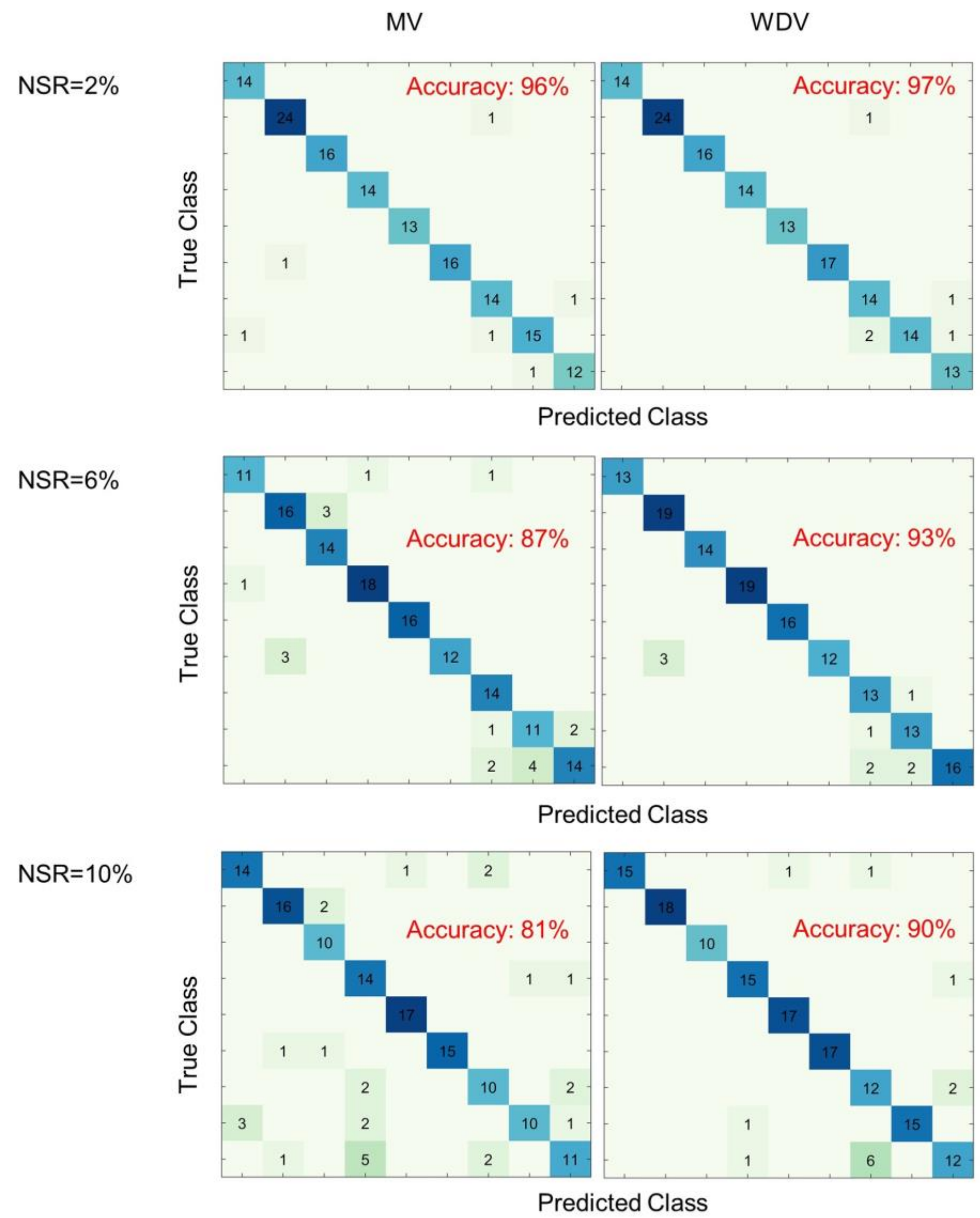

Figure O.3 The confusion matrix after noise injection.

# APPENDIX P FEATURE SPACE

The two-dimensional feature space spanned by the non-salient features. Each class is indicated as unique colors in the legend.

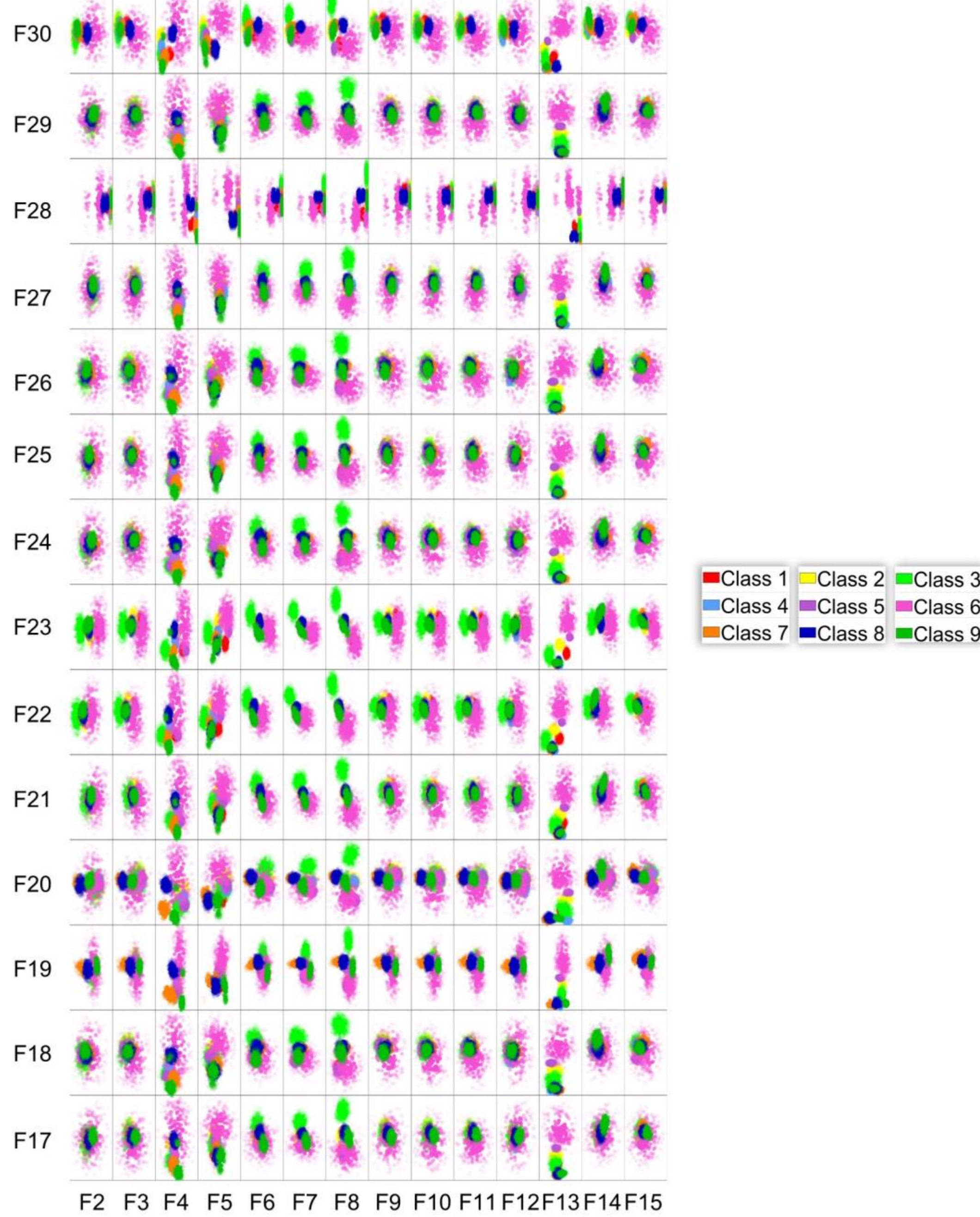

Figure P.3 The feature space.

# APPENDIX Q MATERIAL SORTING SUSTAINABILITY

Table Q.1 Recovery rate of LFP and NMC batteries using different methods.

| Method | Pyro (in %) | | Hydro (in %) | | ML-direct (in %) | |
|---|---|---|---|---|---|---|
| Type | LFP | NMC | LFP | NMC | LFP | NMC |
| Cathode materials | 94 | 94 | 98 | 94 | 90 | 90 |
| Anode materials | 94 | 94 | 95 | 94 | 90 | 90 |
| Cathode current collector | 93 | 93 | 93 | 93 | 90 | 90 |
| Anode current collector | 93 | 93 | 93 | 93 | 90 | 90 |
| Separator | 96 | 93 | 93 | 93 | 95 | 95 |
| Shell | 98 | 98 | 98 | 98 | 98 | 98 |
| Note: The data is collected from the literature[88]. | | | | | | |

Table Q.2 Partial cost of different recycling methods (¥).

| Method | Pyro | Hydro | ML-direct |
|---|---|---|---|
| Battery disassembly | 900 | 850 | 1000 |
| Sewage treatment | 800 | 990 | 700 |
| Equipment depreciation | 390 | 365 | 400 |
| Transportation | 500 | 500 | 500 |
| Average labor | 2000 | 2150 | 1564 |
| Note: The data is collected from the literature[88]. | | | |

Table Q.3 List of material prices.

| Material | Price (¥) | Source |
|---|---|---|
| $Li_2CO_3$ | 273000 | https://dianchizhijia.com/ |
| $NiSO_4$ | 32700 | https://dianchizhijia.com/ |
| $CoSO_4$ | 36300 | https://dianchizhijia.com/ |
| $MnSO_4$ | 6100 | https://dianchizhijia.com/ |
| Graphite | 25800 | https://dianchizhijia.com/ |
| Al | 18400 | https://dianchizhijia.com/ |
| Cu | 65500 | https://dianchizhijia.com/ |
| $LiFePO_4$ | 81500 | https://dianchizhijia.com/ |
| $FePO_4$ | 13000 | https://dianchizhijia.com/ |
| $Li(Ni_{0.5}C_{0.2}M_{0.3})O_2$ | 230000 | https://dianchizhijia.com/ |
| Coke | 2074 | http://quote.eastmoney.com/center/futures.html |
| $H_2SO_4$ | 550 | www.100ppi.com |
| NaOH | 2000 | www.100ppi.com |
| $N_2$ | 800 | www.100ppi.com |
| Liquid ammonia | 3800 | www.100ppi.com |
| Degraded LFP battery | 20000 | https://dianchizhijia.com/ |
| Degraded LFP cathode powder | 38750 | https://dianchizhijia.com/ |

Continued Table Q.3 List of material prices.

| Material | Price (¥) | Source |
|---|---|---|
| Degraded NMC battery | 39000 | https://dianchizhijia.com/ |
| Degraded NMC cathode powder | 115000 | https://dianchizhijia.com/ |
| Note: The update time is May 22, 2023. | | |

Table Q.4 Composition ratio of LFP and NMC batteries.

| Battery Type | NMC | LFP |
|---|---|---|
| Cathode | 0.26 | 0.25 |
| Anode | 0.15 | 0.13 |
| Cathode current collector | 0.07 | 0.06 |
| Anode current collector | 0.17 | 0.1 |
| Electrolyte | 0.1 | 0.16 |
| Separator | 0.22 | 0.27 |
| Shell | 0.03 | 0.03 |
| Note: The data is collected from the literature [90]. | | |

Table Q.5 Cost analysis of LFP and NMC batteries using different methods (individual).

| Method | Pyro (¥) | | Hydro (¥) | | ML-direct (¥) | |
|---|---|---|---|---|---|---|
| Type | LFP | NMC | LFP | NMC | LFP | NMC |
| Raw material | 9687.50 | 29900.00 | 9687.50 | 29900.00 | 9687.50 | 29900.00 |
| Reagent | 637.49 | 951.35 | 326.26 | 468.35 | 3196.19 | 5410.20 |
| Average labor | 3400.00 | 3400.00 | 3500.00 | 3500.00 | 3064.00 | 3064.00 |
| Electricity & Water | 1360.00 | 1360.00 | 1960.00 | 1960.00 | 856.00 | 856.00 |
| Equipment depreciation | 390.00 | 390.00 | 365.00 | 365.00 | 400.00 | 400.00 |
| Sewage treatment | 800.00 | 800.00 | 990.00 | 990.00 | 700.00 | 700.00 |
| Total | 16274.99 | 36801.35 | 16828.76 | 37183.35 | 17903.69 | 40330.20 |
| Note: Individual means there is only one type of battery in one recycling process, and the battery cathode type was determined by historical information assumed. The data were collected from the literatures [88, 266]. | | | | | | |

Table Q.6 Revenue analysis of LFP and NMC batteries using different methods (individual).

| Method | Pyro (¥) | | Hydro (¥) | | ML-direct (¥) | |
|---|---|---|---|---|---|---|
| Type | LFP | NMC | LFP | NMC | LFP | NMC |
| $Li_2CO_3$ | 12017.6 | 20425.6 | 12529.1 | 20425.6 | \ | \ |
| $NiSO_4$ | \ | 6403.3 | \ | 6403.3 | \ | \ |

Continued Table Q.6 Revenue analysis of LFP and NMC batteries using different methods (individual).

| Method | Pyro (¥) | | Hydro (¥) | | ML-direct (¥) | |
|---|---|---|---|---|---|---|
| Type | LFP | NMC | LFP | NMC | LFP | NMC |
| $CoSO_4$ | \ | 2847.7 | \ | 2847.7 | \ | \ |
| $MnSO_4$ | \ | 699.3 | \ | 699.3 | \ | \ |
| $Li(Ni_{0.5}C_{0.2}M_{0.3})O_2$ | \ | \ | \ | \ | \ | 53831.1 |
| $FePO_4$ | \ | \ | 3044.9 | \ | \ | \ |
| $LiFePO_4$ | \ | \ | \ | \ | 18337.5 | \ |
| Graphite | \ | \ | 4421.3 | 5047.8 | 4188.6 | 5262.6 |
| Al | 1026.7 | 1197.8 | 1026.7 | 1197.9 | 993.6 | 1159.2 |
| Cu | 6091.5 | 10355.6 | 6091.5 | 10355.6 | 7663.5 | 10021.5 |
| Total | 19135.9 | 41929.3 | 27113.5 | 46977.2 | 31183.2 | 70274.5 |
| Note: Individual means there is only one type of battery in one recycling process, and the battery cathode type was determined by historical information assumed. | | | | | | |

Table Q.7 Cost, revenue, and profit of retired battery recycling (individual).

| Methods | Type | Cost (¥) | Revenue (¥) | Profit (¥) |
|---|---|---|---|---|
| Pyro | LFP | 16274.99 | 19135.90 | 2860.91 |
| | NMC | 36801.35 | 41929.37 | 5128.02 |
| Hydro | LFP | 16828.76 | 27113.48 | 10284.72 |
| | NMC | 37183.35 | 46977.17 | 9793.82 |
| ML-direct | LFP | 17903.69 | 31183.20 | 13279.51 |
| | NMC | 40330.20 | 70274.45 | 29944.25 |
| Note: Individual means there is only one type of battery in one recycling process, and the battery cathode type was determined by historical information assumed. profit = revenue - cost | | | | |

Table Q.8 The influence of predict accuracy towards the production of ML-direct recycling (hybrid).

| Ideal production | Accuracy (%) | Actual production | Equivalent price (¥) |
|---|---|---|---|
| $Li\ (Ni_{0.5}C_{0.2}M_{0.3})\ O_2$ | Fed-WDV (97) | $Li\ (Ni_{0.5}C_{0.2}M_{0.3})\ O_2$ | 230000 |
| | Fed-MV (71) | 71 %<br>$Li(Ni_{0.5}C_{0.2}M_{0.3})O_2$<br>+<br>29 % $LiFePO_4$ | 92887.5 |
| | IL (55) | 55 %<br>$Li(Ni_{0.5}C_{0.2}M_{0.3})O_2$<br>+<br>45 % $LiFePO_4$ | 80687.5 |

Continued Table Q.8 The influence of predict accuracy towards the production of ML-direct recycling (hybrid).

| Ideal production | Accuracy (%) | Actual production | Equivalent price (¥) |
| --- | --- | --- | --- |
| $LiFePO_4$ | Fed-WDV (97) | $LiFePO_4$ | 81500 |
| | Fed-MV (71) | 70 % $LiFePO_4$ + 30 % $Li(Ni_{0.5}C_{0.2}M_{0.3})O_2$ | 60862.5 |
| | IL (55) | 55 % $LiFePO_4$ + 45 % $Li(Ni_{0.5}C_{0.2}M_{0.3})O_2$ | 73062.5 |

Note: Hybrid means there are multiple types of battery in one recycling process, and the battery cathode types were determined by our machine learning method, leveraging field information. Fed-WDV, Fed-MV, and IL stand for FL using the MDV method (our work), FL using the MV method, and non-federated IL. The impure product is regarded as degraded cathode material, which needs to be treated by hydrometallurgy again. We assumed the equivalent price is calculated by the sum of the product of accuracy and retired cathode powder price. For example, for the accuracy of 0.71 for NMC, the equivalent price = 0.71*115000+0.29*38750=92887.5.

Table Q.9 The influence of prediction accuracy on the revenue of ML-direct recycling (hybrid).

| Accuracy (%) | MR | Cathode (¥) | Others (¥) | Revenue (¥) |
| --- | --- | --- | --- | --- |
| Fed-WDV (97) | 0.00 | 53831.15 | 16443.30 | 70274.45 |
| Fed-WDV (97) | 0.33 | | | 57244.03 |
| Fed-WDV (97) | 0.50 | | | 50728.83 |
| Fed-WDV (97) | 0.67 | | | 44213.62 |
| Fed-WDV (97) | 1 | 18337.50 | 12845.70 | 31183.20 |
| Fed-MV (71) | 0.00 | 21634.54 | 15400.00 | 37034.53 |
| Fed-MV (71) | 0.33 | | | 33984.09 |
| Fed-MV (71) | 0.50 | | | 32458.88 |
| Fed-MV (71) | 0.67 | | | 30933.66 |
| Fed-MV (71) | 1 | 13994.21 | 13889.00 | 27883.22 |
| IL (55) | 0.00 | 18723.94 | 14824.38 | 33548.32 |
| IL (55) | 0.33 | | | 32822.02 |
| IL (55) | 0.50 | | | 32458.88 |
| IL (55) | 0.67 | | | 32095.73 |
| IL (55) | 1 | 16904.81 | 14464.62 | 31369.43 |

Note: Hybrid means there are multiple types of battery in one recycling process, and the battery cathode types were determined by our machine learning method, leveraging field information.
Mixed ratio, MR = LFP/(LFP+NMC)
Fed-WDV, Fed-MV, and IL stand for FL using the MDV method (our work), FL using the MV method, and non-federated IL. We assumed the revenue is calculated by the sum of the product of the cathode (weight, accuracy, and retired cathode powder price) and others (anode and current collector). For example, for the accuracy of 0.71 for the LFP/(LFP+NMC) ratio of 0.5,
the revenue =
0.90*0.5*(0.25*38750.00+0.26*115000.00)+0.50*(4188.60+993.60+7663.50)+0.50*(5262.60+1159.20+10021.0) = 32458.88.

Table Q.11 Cost, revenue, and profit of LFP and NMC batteries using different methods (hybrid).

| Method | MR | Accuracy (%) | Cost (¥) | Revenue (¥) | Profit (¥) |
|---|---|---|---|---|---|
| Pyro | 0.33 | \ | 29959.23 | 34331.54 | 4372.32 |
| | 0.5 | \ | 26538.17 | 30532.63 | 3994.46 |
| | 0.66 | \ | 23117.11 | 26733.72 | 3616.61 |
| Hydro | 0.33 | \ | 30398.48 | 40355.94 | 9957.45 |
| | 0.5 | \ | 27006.05 | 37045.32 | 10039.27 |
| | 0.66 | \ | 23613.62 | 33734.71 | 10121.09 |
| ML-direct | 0.33 | Fed-WDV (97) | 32854.70 | 57244.03 | 24389.33 |
| | | Fed-MV (71) | 30686.80 | 33984.09 | 3297.29 |
| | | IL (55) | 29490.72 | 32822.02 | 3331.30 |
| | 0.5 | Fed-WDV (97) | 29116.95 | 50728.83 | 21611.88 |
| | | Fed-MV (71) | 29116.95 | 32458.88 | 3341.93 |
| | | IL (55) | 29116.95 | 32458.88 | 3341.93 |
| | 0.66 | Fed-WDV (97) | 25379.19 | 44213.62 | 18834.42 |
| | | Fed-MV (71) | 27547.09 | 30933.66 | 3386.56 |
| | | IL (55) | 28743.17 | 32095.73 | 3352.56 |

Note: Hybrid means there are multiple types of battery in one recycling process, and the battery cathode types were determined by our machine learning method, leveraging field information.
Mixed ratio, MR = LFP/(LFP+NMC)
Fed-WDV, Fed-MV, and IL stand for FL using the MDV method (our work), FL using the MV method, and non-federated IL.
We assumed the cost is calculated by the sum of the product of individual cost and battery ratio. For example, for the LFP/(LFP+NMC) ratio of 0.5 using ML-direct recycling, the cost = 0.5*17903.69+0.5*40330.20=29916.95.

# APPENDIX R LIFETIME PREDICTION COMPARISIONS

Table R.1 The lifetime prediction comparisons.

| Ref. | Feature numbers | Cycles | Metrics | Dataset |
|---|---|---|---|---|
| Ref[63] | 5 classes (9 in total) | 100 | Err≈9.1% | D1 |
| Ref [267] | 5 classes (20 in total) | 100 | MAPE≈5.2% | D1 |
| Ref [268] | 6 classes (7 in total) | 100 | Err =3.3% | D1 |
| Ref [269] | 4 classes (10 in total) | 5/30/60 | MAPE≈15% | D1, D2 |
| Ref [217] | 11 classes (35 in total) | 100 | MAE=117cycles | D3 |
| Ref [168] | 5 classes (12 total) | 100 | RMSE=57cycles | D1 |
| Ref [270] | 7 classes (42 in total) | 100 | RMSE=115 cycles | D1 |
| Ref [271] | 1 class (10 in total) | 100 | RMSE≈32.1cycles | D1 |
| Ref [272] | 2 classes (12 in total) | 40 | Err≈22.73% | D4 |
| Ref [130] | 3 classes (6 in total) | 3 | MAPE≈10.4% | D1, D2, D5 |
| Ref [225] | 2 classes (4 in total) | 30 | RMSE=98 cycles | D1, D2, D6 |
| Ref [273] | 3 classes (4 in total) | 80 | Err≈6.3% | D1 |
| Ref [274] | 5 classes (11 in total) | 100 | RMSE≈75.8 cycles | D1 |
| Ref [275] | 3 classes (18 in total) | 100/150/200/250 | MAPE≈7% | D1 |
| Ref [276] | 5 classes (9 in total) | 100 | MAPE=10.6% | D1 |

Note:

D1: https://www.nature.com/articles/s41560-019-0356-8

D2: https://www.nature.com/articles/s41586-020-1994-5

D3: Argonne National Laboratory, including 1000 pouch cells over 40 different builds, with C-rate ranging from 1/5 to 2 (none open source)

D4: Custom-designed test on 104 commercial 18650 NCA LIBs (none open source)

D5: https://doi.org/10.1149/1945-7111/abae37

D6: Custom-designed test on 77 18650 LFP LIBs (none open source)

# APPENDIX S CYCLE-WISE PREDICTION ERROR

The prediction error distribution against the cycle number.

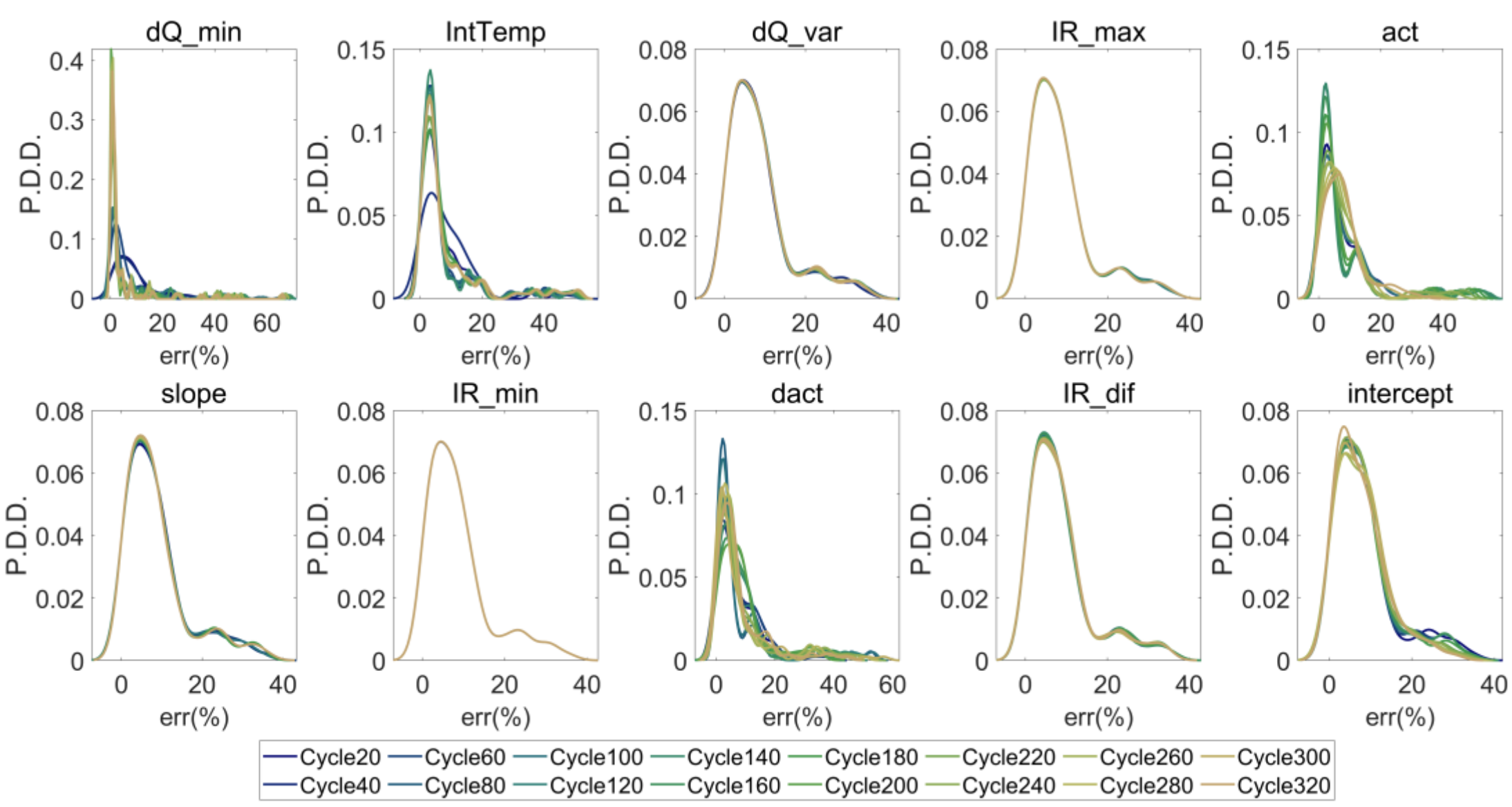

Figure S.1 The cycle-wise prediction error.

# APPENDIX T SHANNON ENTROPY CACULATION

The fidelity of the features and the computation efficiency could be sensitive to the interval selection between the cycles. Specifically, large intervals lead to information loss of the features, while small intervals result in information redundancy. Moreover, in the battery system, the feature exhibits no significant variation trends if the intervals are selected as small ones, and the feature trends can even vanish into the noise from the measurement. Herein, 8 discrete interval numbers are used to justify the interval selection, i.e., 5, 10, 15, 20, 25, 30, 35, and 40. Since each group (fast-charging and extremely fast-charging conditions) exhibits similar cycle-wise feature variations, one typical battery from each group is randomly selected for demonstration. The information entropy, i.e., the Shannon entropy, is used to evaluate the information fidelity by changing the intervals. The Shannon entropy is given by $H = -\sum_{i=1}^{n} p_i \log p_i$, where, $p_i$ is the occurrence probability of the $i^{\text{th}}$ sample points of the feature and $\sum_{i=1}^{n} p_i = 1$. The overall Shannon entropy is summed up by all Shannon entropies of each feature. The Shannon entropy exhibits slight variations with 5, 10, 15, and 20 intervals. However, when the interval is selected larger than 20, significant information loss exists, i.e., Shannon entropy decrease. Specifically, the maximum value of Shannon entropy at the interval of 15 is 1.86, while it is 1.73 at the interval of 20. Thus, 93% of the original information is preserved. When the interval increases to 25, only 74% of the original information is preserved. As a result, the interval value is selected 20 in this work to provide information fidelity. For instance, 16 sample points for each feature are given by a uniform sampling with a sampling interval of 20 in the first 320 cycles. The cycle value of 320 is determined by the shortest-lifetime cell with barcode el150800460518 in the target domain in Dataset1.

# APPENDIX U PULSE TEST WITH AGEING

There are concepts such as state of health (SOH) for residual value evaluation, but the term health could indicate multiple meanings, e.g., energy, power, and resistance. To avoid distractions, this study focuses on the relative remaining capacity (RRC) of second-life batteries since it is the most direct indicator to evaluate the residual value, which serves as the label of the estimation problem. The RRC is defined as:

$$RRC = \frac{Q_{real}}{Q_{nominal}} \tag{U.1}$$

where, the $Q_{real}$ is the remaining capacity of the second-life batteries and $Q_{nominal}$ is the nominal capacity rated by the manufacturer.

$Q_{real}$ is determined by a standard CCCV charging and discharging procedure. First, the second-life batteries are discharged to a lower cut-off voltage (2.7V) using a 1C constant current. Second, the second-life batteries are charged to the upper cut-off voltage (4.2V) using a 1C constant current, then charged using constant voltage until the current drops to 0.05 C. Third, the second-life batteries are then discharged to the lower cut-off voltage (2.7V) using a 1C constant current. C stands for charge (discharge) rate when a 1 hour of charge (discharge) is performed, which is a dimensionless scalar number.

The pulse data used in this section is obtained from accelerated aging tests to characterize different lifetime stages. Three different types of NMC (nickel manganese cobalt oxide) fresh cells are investigated in the study to demonstrate capacity design and physical format diversities, as shown in the table in this appendix. Note that there are some RRC values larger than 1, a normal observation for batteries with overloaded materials (discharged capacity is more than rated value). There were 96 second-life batteries simulated from accelerated aging tests using 17 physically independent cells. The experiments are performed for only one time to simulate the real-world requirement that one might want to know the RRC with minimal data even the measurement uncertainty could exist, which accounts for the proposed model capability.

Table U.1 The summary statistics of the second-life batteries.

| Domain | Q (Ah) | Physicl Format | Battery Quantity (Simulated Quantity) | Data Entry | RRC Range | RRC Mean (Deviation) |
|---|---|---|---|---|---|---|
| #1 | 2.1 | Cylinder | 12 (67) | 670 | 0.61-0.92 | 0.81 (0.07) |
| #2 | 3.1 | Pouch | 2 (14) | 140 | 0.68-1.01 | 0.87 (0.08) |
| #3 | 5.2 | Pouch | 3 (15) | 150 | 0.50-1.02 | 0.83 (0.16) |

In each accelerated aging cycle, all batteries were charged with a 2C constant current followed by a constant voltage at 4.2 V with C/200 or 30-minute cut-off condition followed by a 5-minute rest. The same constant discharge current was applied with a cut-off voltage of 3V followed by a 5-minute rest. Note that C stands for rate of charge (discharge) relative to that for a 1 hour of charge (discharge). After every 100 accelerated aging cycles, the batteries rested for 30 minutes to reach a steady state, preparing for the pulse injection. The pulse injection goes to 10 SOC levels, i.e., from 5% to 50%, with an increment of 0.05, since the second-life batteries are stored with relatively low SOC before redeployments. The current used to adjust SOC was 1C, thus the time for moving across SOC levels is 3 minutes. We used identical rest time, i.e., 25s, between all injected pulses. The pulse width was 5s and the amplitude was $\pm$0.5C, $\pm$1C, and $+$2C. The $+$2C pulse is the final pulse of each experiment, then the SOC adjustment is continued until the 50% SOC level is hit. The air ambient temperature was set at 25 ℃. Note that positive and negative pulses alternate to cancel the electric charge injection so that no SOC calibration is needed between different pulse injections. Since there are 10 SOC levels tested, the data entry in each domain is ten times that of the simulated battery quantity. The tested SOC are taken as the true SOC label in this paper. Specifically, data entry numbers for domain 1-3 are 670, 140, and 150, respectively, thus a total of 960 data entries were generated. The alternating setting of the accelerated ageing test and the pulse test demonstrate that safety of the technique within the wide observed RRC ranges, i.e., from 0.5 to fresh cells, even under a 2C harsh cycling conditions.

The feature engineering process is easy to implement, where features are extracted from the turning points of the voltage curve, i.e., the points with zero second-order derivative of the voltage curve[40]. Thus, the 21 feature points, from U1 to U21, are extracted. Mathematically, the feature set $U_k$ can be formulated as:

$$U_k = \underset{U}{\operatorname{argmin}} \left\| \frac{d^2 V}{dt^2} \right\|^2 \tag{U.2}$$

where, $U_k$ is the feature set, $V$ is the voltage response signal subject to the pulse current injection, $t$ is the continuous experiment time, $\|\cdot\|$ is the norm operator. Since there are 5 pulses injected, $k = 1, 2, \cdots, 21,$ and the feature dimension is 21. The feature set $U_k$ was then utilized to supervise a cross-domain model mapping to the problem label $RRC$.

# APPENDIX V NEURAL NETWORK IMPLEMENTATION

**Variational autoencoder (VAE) net for pulse voltage signal generation across SOC**

The VAE net is designed to resolve the data scarcity challenge since the data testing are expensive, which has three components, i.e., encoder, sampling layer, and decoder. The encoder is implemented by a fully connected layer with 128 neurons and ReLU activation. The encoder network receives the voltage response dynamics and SOC pairs, i.e., $[\boldsymbol{U}^{m\times21},\boldsymbol{SOC}^{m\times1}]$ as input, where $m$ is the number of observations (data entries) and 21 stands for the dimension of features. The VAE net is trained by each unique relative remaining capacity value due to considerable complexity resulting from the impact of battery degradation on $[\boldsymbol{U}^{m\times21},\boldsymbol{SOC}^{m\times1}]$ pairs.

Two separate parameters, i.e., the mean ($\widetilde{\boldsymbol{SOC}}_{\boldsymbol{mean}}$) and log-variance ($\widetilde{\boldsymbol{SOC}}_{\log_var}$) are produced from the encoder. The sampling layer is defined as:

$$\widetilde{\boldsymbol{SOC}} = \boldsymbol{\mu} + \boldsymbol{\sigma}\cdot\boldsymbol{\epsilon} \tag{V.1}$$

where, $\boldsymbol{\mu} = \widetilde{\boldsymbol{SOC}}_{\boldsymbol{mean}}$, $\boldsymbol{\sigma} = \exp\left(\mathbf{0.5}\cdot\widetilde{\boldsymbol{SOC}}_{\log_var}\right)$, and $\boldsymbol{\epsilon}$ is a random sample from a standard normal distribution, i.e., $\boldsymbol{\epsilon}\sim\boldsymbol{N}(\mathbf{0},\mathbf{1})$, and it is where generative capability of the VAE net comes from. The decoder is also implemented by a fully connected layer with 128 neurons and ReLU activation. Thus, the VAE net describes the following relationship:

$$\widetilde{\boldsymbol{U}}^{m\times21} = \text{decoder}(\widetilde{\boldsymbol{SOC}}) \tag{V.2}$$

where, $\widetilde{\boldsymbol{U}}^{m\times21}$ is the generated voltage dynamics given randomly generated SOC values described by $\widetilde{\boldsymbol{SOC}}$. The loss function of the VAE net constitutes two parts, i.e., reconstruction loss and the Kullback-Leibler (KL) divergence loss. The reconstruction loss is set as mean square error (MSE), which calculates the MSE between originally tested voltage response $U_k$ and generated voltage response $\widetilde{U}_k$, i.e., $l_{MSE}^{xent}$.

KL divergence loss $Loss_{KL}$, i.e., the KL divergence between originally tested and generated data, is given by:

$$l_{KL} = -\frac{1}{2}\sum_{i=1}^{m}\left(1 + \widetilde{\boldsymbol{SOC}}_{log_var\ i} - \widetilde{\boldsymbol{SOC}}^{2}_{mean\ i} - e^{\widetilde{\boldsymbol{SOC}}_{log_var\ i}}\right) \tag{V.3}$$

The total loss is the linear combination of $Loss_{MSE}$ and $Loss_{KL}$:

$$Loss_{VAE_net} = \omega_{xent} \cdot l^{xent}_{MSE} + \omega_{KL} \cdot l_{KL} \tag{V.4}$$

where, $\omega_{xent}$ and $\omega_{KL}$ are set to 0.5 to achieve a balance between the generation accuracy and the diversity, respectively[192].

The training epoch is 1000, the batch size is 32, the latent dimension (size of sampling layer) is 2, and the sampling intensity is 10. Sampling intensity refers to the ratio of the number of generated and originally tested observations, which is implemented by regulating the number of points taken from $\boldsymbol{\epsilon} \sim \boldsymbol{N}(\boldsymbol{0}, \boldsymbol{1})$.

**SOC net for charge level prediction from immediate available pulse voltage signal**

The SOC net is designed to learn a map from $\widetilde{\boldsymbol{U}}$ to $\widetilde{\boldsymbol{SOC}}$. The SOC net is a 10-layer deep neural network, with 512, 512, 256, 256, 128, 128, 64, 64, 32, and 32 neurons in each layer. The activation functions are all set as ReLU function. The input of the SOC net is generated, appended with the originally tested voltage dynamics signals $[\widetilde{\boldsymbol{U}}; \boldsymbol{U}]$ while the output is the predicted SOC value $\widetilde{\boldsymbol{SOC}}'$.

**Regression net for relative remaining capacity prediction**

The regression net is designed to predict the relative remaining capacity. The regression net is a 5-layer neural network, with 256, 256, 128, 128, and 64 neurons in each layer. The activation functions are all set as ReLU function. The input of the regression net is the originally tested voltage dynamics signals appended with the predicted SOC values, which is $[\boldsymbol{U}; \widetilde{\boldsymbol{SOC}}']$ while the output is the predicted $\widetilde{\boldsymbol{RRC}}$ value. In this way, immediate remaining capacity can be achieved from pulse tested voltage signals given the $\widetilde{\boldsymbol{SOC}}'$ is generated from learning the latent distribution between the $[\boldsymbol{U}^{m\times 21}, \boldsymbol{SOC}^{m\times 1}]$ pairs, thus $\widetilde{\boldsymbol{SOC}}'$ is free of physical testing cost. In a practical sense, this method enables a quick knowledge of SOC value by learning from voltage signals, and thus enabling remaining capacity prediction from voltage signals with the help of the prediction SOC condition.

# APPENDIX W VOLTAGE UNDER PULSE TEST

The data visualization. **(a)** The illustration of pulse injection, voltage dynamics (i.e., the voltage response values that subject to pulse current injections), and feature points after the feature engineering. **(b)** The heterogeneity of voltage dynamics versus feature dimensions in different domains, with Cylind21, Pouch31, and Pouch52 defined as one domain, respectively. The voltage dynamics distribution of one specific feature (U1-U21) is measured over all tested SOCs (from 5% to 50%, with a 5% increment).

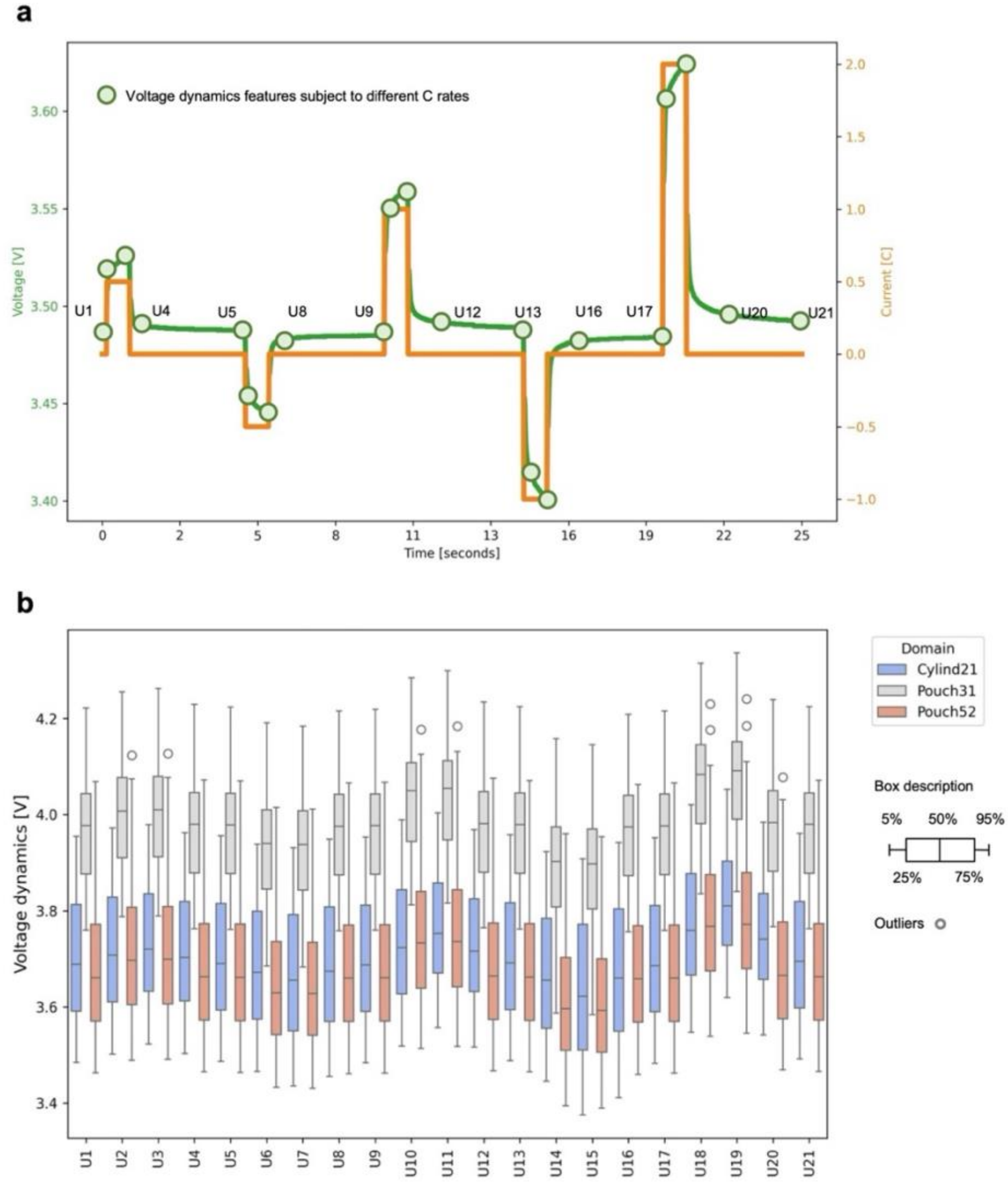

Figure W.1 The voltage test and voltage response.

# APPENDIX X DATA AUGMENTATION

Data generation under unseen (untested) SOC levels. **(a)** The direct comparison of the tested and generated voltage dynamics data in the U1 feature dimension, with other feature dimensions being presented in panel **c**. **(b)** Data distribution comparison between tested and generated voltage dynamics data using KL-divergence, i.e., KLD, (with U1 illustrated). **(c)** KLD between tested and generated data for all voltage dynamics dimensions. The color maps the maximum and minimum values across Pouch52, Pouch31, and Cylind21 second-life battery types.

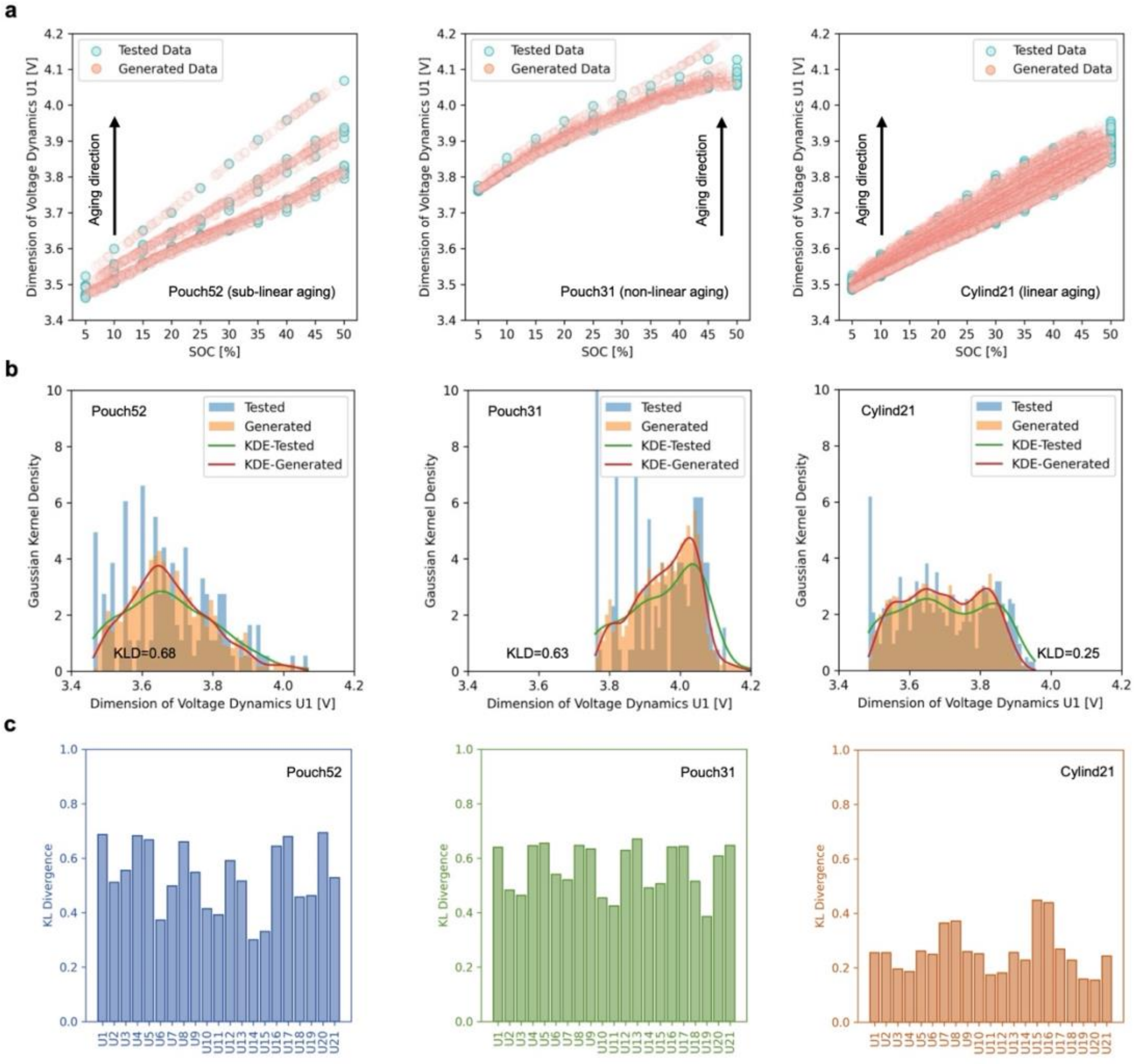

Figure X.1 The data augmentation of all feature dimensions.

# APPENDIX Y IMPACT OF CHARGE CONDITION

The effectiveness of degradation representation capability of pulse injection, using features from **(a)** given SOC information, and **(b)** with mixed SOC information (i.e., the Pearson correlation is calculated without sorting SOC value.). The correlation is calculated between the RRC and the voltage dynamics value.

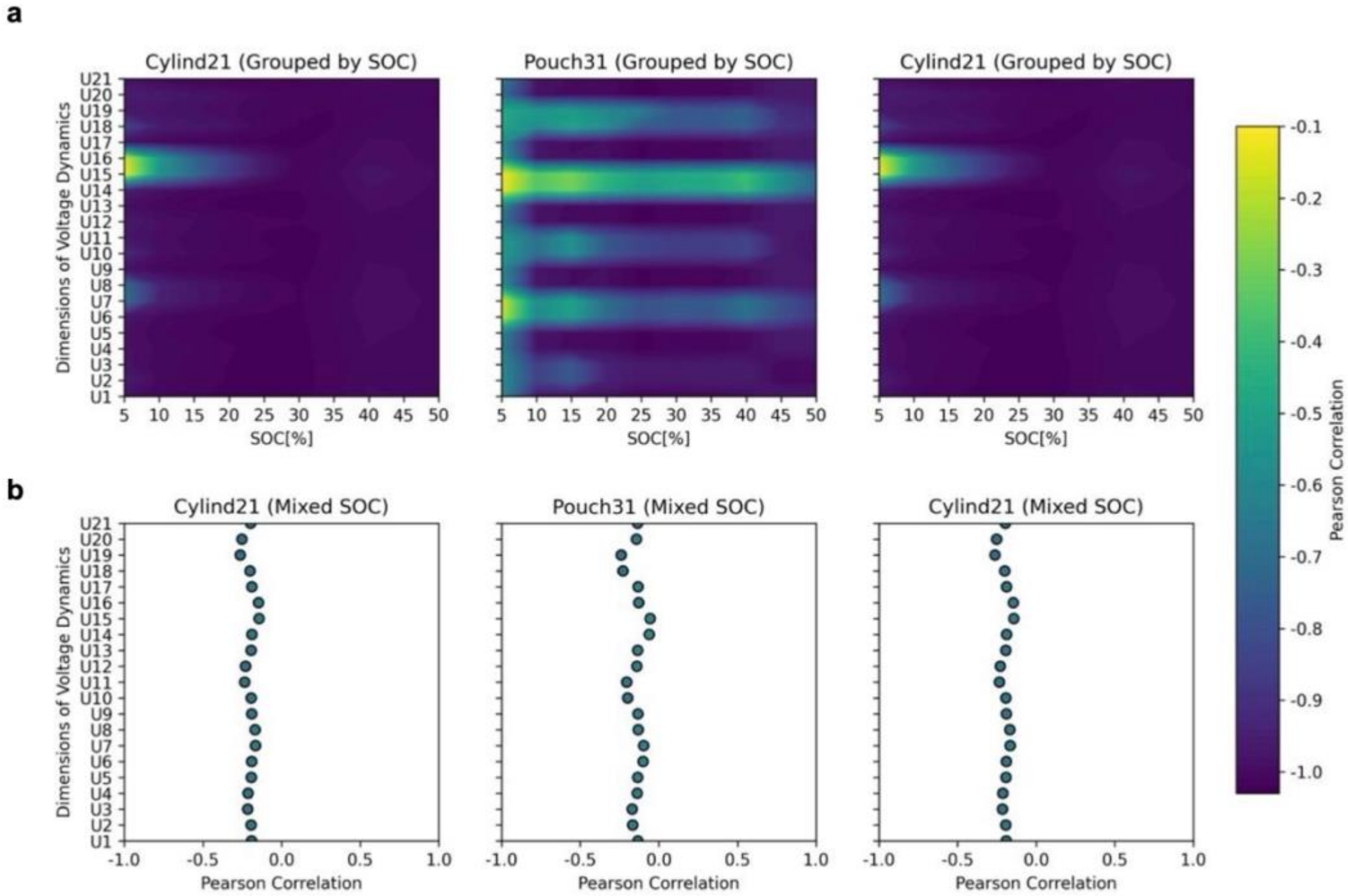

Figure Y.1 The impact of charge condition on feature effectiveness.

# APPENDIX Z BENCHMARKING MODEL SETTINGS

Linear regression, ridge regression, Gaussian process regression, support vector machine, k-nearest neighbor, and random forest model are adopted from the sci-kit learn package (version 1.3.2) with default settings instead of finely tuning the hyperparameters of these models. The deep neural network has 256, 256, 128, 128, 64, and 1 neuron in each layer with ReLU activation functions in each layer. The training epoch was 200, and after this epoch, the deep neural network performance did not increase significantly.

# ACKNOWLEDGEMENTS

I, Shengyu Tao, would like to express my deep gratitude to my supervisor Professor Xuan Zhang for the patient one-on-one talks, rigorous scientific research training, open-minded scientific research attitude and free academic pursuit. In this team, I can conduct any research that I deem meaningful, and not only can I learn, but I can also collaborate with external researchers and even teach as a senior doctoral candidate.

I would like to thank Professor Guangmin Zhou for his interdisciplinary training as a co-supervising professor. All my research topics directly come from scientific research or related industrial issues of Zhou's group, and I am sure that I have a positive impact on the sustainable use of retired batteries. Professor Zhou's guidance to do the best research has a profound impact on me, so that I understand what is good research, how to do good research, and how to present them.

I also want to thank Professor Scott J. Moura, who is an educator. When I visited the eCAL, he asked me to expand my knowledge boundaries, which could be from batteries to power grids, or from data-driven to model (physical equation) driven. His influence on me is to break boundaries of knowledge and integrate them. I feel respected and valued in Professor Scott J. Moura's research group. I will remember that the output of eCAL is two things: knowledge and people, and this will be my motto as I continue my scientific research.

I am deeply grateful to everyone who has helped me during my research journey, and I dedicate this dissertation to all of them.

# RESUME

Shengyu Tao was born on 31st Dec., 1996 in Nantong, Jiangsu Province, China. He began his bachelor's study in the School of Engineering, Shanghai Ocean University in September 2015. He got the Bachelor of Engineering degree in June 2019. He began his master's study in the School of Information Science and Technology, Fudan University in September 2019. He got the Master of Science degree in June 2022. He has started to pursue his doctor's degree in Electrical Engineering in Tsinghua University since September 2022. He was visiting the Energy, Controls, and Applications Lab (eCAL) at UC Berkeley, from September 2024 to September 2025.

First author papers

First author dataset

## Co-first author papers

## Co-author papers

# COMMENTS FROM THESIS SUPERVISOR

This dissertation addresses a timely and societally impactful topic at the intersection of energy storage, artificial intelligence, and sustainable energy: how to enable intelligent remanufacturing, reusing, and recycling of lithium-ion batteries in the presence of data scarcity and heterogeneity. With the growing volume of retired electric vehicle batteries and increasing emphasis on circular economy principles, the research provides a novel and holistic machine learning framework that contributes meaningfully to both academic scholarship and industrial practice. Its relevance spans battery engineering, sustainable systems design, and data-driven diagnostics, and is expected to attract broad interest from researchers, practitioners, and policymakers concerned with low-carbon technologies and resource efficiency.

The core contributions of the dissertation are fourfold. First, it introduces a physics-informed machine learning method for early quality control of battery prototypes, capable of predicting long-term degradation trajectories using only a small fraction of early-cycle data. This allows for efficient, non-invasive screening of manufacturing quality, with significant implications for production scalability and waste minimization. Second, the dissertation proposes a rapid residual value assessment strategy for retired batteries that combines pulse-based testing with generative modeling. This approach enables accurate estimation of battery health under unknown field conditions, while significantly reducing test time, energy use, and environmental impact. Third, a collaborative material sorting framework based on federated learning is developed, allowing high-accuracy cathode material classification using privacy-preserved data across multiple stakeholders. This innovation addresses the urgent need for secure data collaboration in direct recycling processes. Finally, the dissertation presents an adaptive diagnostics and prognostics framework that generalizes across battery states, tasks, and test protocols, leveraging correlation alignment to ensure consistent performance across heterogeneous scenarios.

The impact of this research is substantial. It advances the methodological frontier by integrating domain adaptation, privacy-aware learning, and synthetic data generation into a unified lifecycle-aware framework. Practically, it enables faster, more accurate, and more scalable assessment of battery health and value, thereby facilitating safer and more cost-effective second-life battery applications.